\pdfoutput=1
\documentclass[10pt]{aart}

\usepackage{arxiv-setting}
\usepackage{booktabs}
\usepackage{placeins}
\usepackage{tabularx}
\usepackage{manuscript-presentation}
\usepackage{authblk}
\cslet{blx@noerroretextools}\empty
\usepackage{autonum}

\renewcommand{\epsilon}{\varepsilon}
\newcommand{\clean}{{\vs}}
\newcommand{\noise}{{\vz}}
\newcommand{\manifold}{\mathcal{M}}
\newcommand{\metric}{\mathrm{g}}
\newcommand{\SNR}{\mathrm{SNR}}

\newcommand{\IOK}{\mathcal{K}} % Integral operator for the kernel K
\newcommand{\kf}{\mathrm{W}} % kernel function
\newcommand{\mPsi}{{\mathbf{\Psi}}} % map onto estimated signal coordinates
\newcommand{\foldmap}{\widetilde{\mPsi}} % fold map of Algorithm~\ref{alg:three-fold-denoising}
\newcommand{\admset}{\mathcal{A}} % admissible neighbor counts \admset_N(\zeta)
\newcommand{\radband}{\mathcal{H}} % radius band \radband_N(k)
\newcommand{\radwin}{\mathcal{R}} % valid radius window \radwin_N(\zeta)
\newcommand{\locrad}{R_{\star}} % locality radius of the clean law
\newcommand{\dgap}{\mathfrak{g}} % deterministic eigengap constant
\newcommand{\ncdf}{\Phi_{\mathrm N}} % standard normal distribution function
\newcommand{\covscale}{\bar\theta} % common covariance scale of the minimax class
\newcommand{\mismatch}{\eta}

\title{Revealing Geometry under Strong Noise: Spectral Embeddings and Clustering via Random Geometric Graphs}
\author[1]{
Hai-Xiao Wang \orcidlink{0000-0003-2730-1439}\thanks{Email: \texttt{haixwang@uw.edu}. HXW is partially supported by NSF DMS-2605423.}}
\author[2]{Xiucai Ding\thanks{Email: \texttt{xcading@ucdavis.edu}. XCD is partially supported by NSF DMS-2605423.}
}

\affil[1]{Department of Applied Mathematics, University of Washington}
\affil[2]{Department of Statistics and Data Science, University of California, Davis}

\date{}

\graphicspath{{figures/}}
\renewcommand{\abstractname}{Abstract}
\begin{document}

\maketitle

\begin{abstract}
{\centering\bfseries\abstractname\par}
Modern massive datasets often contain useful low-dimensional nonlinear structures that are hidden by noise. Spectral embedding algorithms rely on meaningful local relationships between observations, but noise accumulated across many coordinates can make the observed distances unreliable. This also complicates clustering when groups have curved or concentric shapes and cannot be distinguished by their centers. In this paper, we develop a new method that denoises pairwise distances before using them to construct spectral embeddings and identify communities. Our approach uses cross-fitted projections to estimate distances between the underlying signals, selects a neighborhood radius from the observed sample, and builds a random geometric graph whose spectral coordinates represent the latent geometric structure. By connecting nearby points, such a geometric graph can distinguish different groups even when their centers coincide. A novel recursive procedure is then proposed to extract clusters from the graph using its spectral embeddings when such clusters exist.

On the theoretical side, under a manifold model, we provide guarantees for our spectral embedding algorithm, which can be understood through the eigendecomposition of an integral operator associated with the underlying geometry. We show that the algorithm is robust to noise when the total noise energy is comparable to the total signal energy or, under additional conditions, exceeds it. To our knowledge, this is among the first spectral embedding algorithms that can handle such strong noise in datasets with nonlinear structures. In addition, our proposed clustering algorithm can recover the underlying clusters even when their centers coincide. Numerical studies demonstrate the superior performance of the proposed method in mitigating noise and revealing geometric structures for both spectral embedding and clustering.  
\end{abstract}

\section{Introduction}\label{sec:introduction}
\begin{figure}[!ht]
    \centering
    \includegraphics[width=\linewidth]{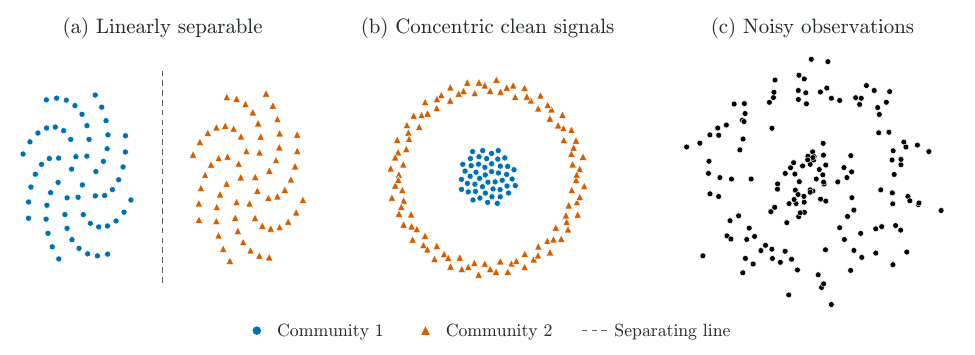}
    \caption{Conceptual linear and geometric separation. Blue circles and
    orange triangles identify the two communities in (a) and (b). The
    concentric supports in (b) are disjoint, although their centers
    coincide and no hyperplane separates them. In (c), black dots show the
    same signals after noise brings points from different communities
    close together.}
    \label{fig:GMM-Non-Linear-Separable}
\end{figure}

Clustering~\cite{abbe2018community} is a fundamental task in data analysis: it reveals hidden groups, or \emph{communities}, summarizes complex datasets, and guides further investigation, for example, identifying cell types and tissue regions in transcriptomics~\cite{li2022cell,long2023spatially}. However, identifying these groups can be challenging even the underlying signals are strong. Two difficulties stand in its way. First, different communities may have curved or concentric shapes, sharing the same geometric center rather than being linearly separable, as contrasted between Figure~\ref{fig:GMM-Non-Linear-Separable} (a) and (b). Center-based methods such as \(K\)-means~\cite{macqueen1967methods} can struggle with these groups because distances to cluster centers do not capture their geometric organization. A simple yet effective observation is that, even with overlapping geometric centers, different communities can still be distinguished by their disjoint supports. Therefore, distinguishing those communities requires information about how observations are arranged and connected locally. 
Second, measurement noise makes this task harder by moving observations away from their underlying structures and blurring the geometric distance between groups, as in Figure~\ref{fig:GMM-Non-Linear-Separable}(c). In high dimensions, noise that accumulates across many coordinates can further distort pairwise distances, making meaningful local relationships difficult to identify. As in~\cite{elkaroui2010information}, we model the observations as
\begin{align}
 \vx_j=\clean_j+\noise_j\in\R^p,\qquad j\in[N],
 \label{eqn:signal-plus-noise}
\end{align}
where \(\noise_j\) is sub-Gaussian noise with covariance \(\mGamma\),
independent across observations and of the clean signal \(\clean_j\).
We write \(\sigma^2\coloneqq\|\mGamma\|\) for the largest noise variance.
The supports lie in an unknown \(m\)-dimensional subspace, with \(m\)
typically much smaller than \(p\). We measure the strength of the signal
relative to the noise by the \emph{signal-to-noise ratio}
\begin{align}
 \SNR_N\coloneqq\frac{\E\|\clean-\clean'\|^2}{\E\|\noise-\noise'\|^2},
 \label{eqn:signal-to-noise-ratio}
\end{align}
where \(\clean,\clean'\) and \(\noise,\noise'\) are independent copies of
a clean signal and of the noise. Let \(\Delta\) denote the minimum pairwise distance among all the supports. As shown in Figure~\ref{fig:GMM-Non-Linear-Separable}(c)
, the noise moves the observations off the supports and blurs the gap, making the pairwise distances between observations from different communities smaller than \(\Delta\). It is natural to ask how the $\Delta$ and $\SNR_N$ together determine the separation of the communities.

\FloatBarrier

Spectral embeddings provide a natural framework for representing these geometric relationships. As in kernel PCA, Laplacian eigenmaps, and diffusion
maps~\cite{scholkopf1998nonlinear,belkin2003laplacian,coifman2006diffusion}, coordinates are built from eigenvalues and eigenvectors of a graph or kernel matrix, which record local connections between observations and thus inform the representation. Their effectiveness, however, depends on whether those connections reflect the underlying signals rather than measurement noise. Existing convergence guarantees for nonlinear embeddings of high-dimensional noisy data require \(\SNR_N\to\infty\)~\cite{ding2023learning,ding2025kernel,ding2026generalized}. Under the \emph{strong noise} scenario where total noise energy is comparable to the total signal energy or even exceeds it, i.e., $\SNR_N = O(1)$, it remains unclear whether meaningful nonlinear embeddings can be reconstructed accurately.

In this paper, we propose algorithms that construct meaningful spectral embeddings under strong noise conditions. We first denoise the pairwise distances via cross-validation. We then construct a \emph{Random Geometric Graph} (RGG), where two observations are connected only if their denoised distance is within a radius $r$. The $r$ is selected via a newly proposed adaptive quantile selection method. After that, we construct spectral embeddings from the kernel matrix of the RGG. These embeddings are provably close to their population counterparts, which are characterized by the eigenfunctions of an integral operator, even when $\SNR_N = O(1)$, breaking the previous barrier $\SNR_N\to\infty$ in \cite{ding2023learning}. As a down-stream task, we construct a recursive algorithm that extracts the communities based on the separation of the embedding coordinates, which provably recovers all the labels correctly with overwhelming probability.

\subsection{Overview of our methodology}\label{sec:overview}

\begin{figure}[!h]
  \centering
  \includegraphics[width=\linewidth]{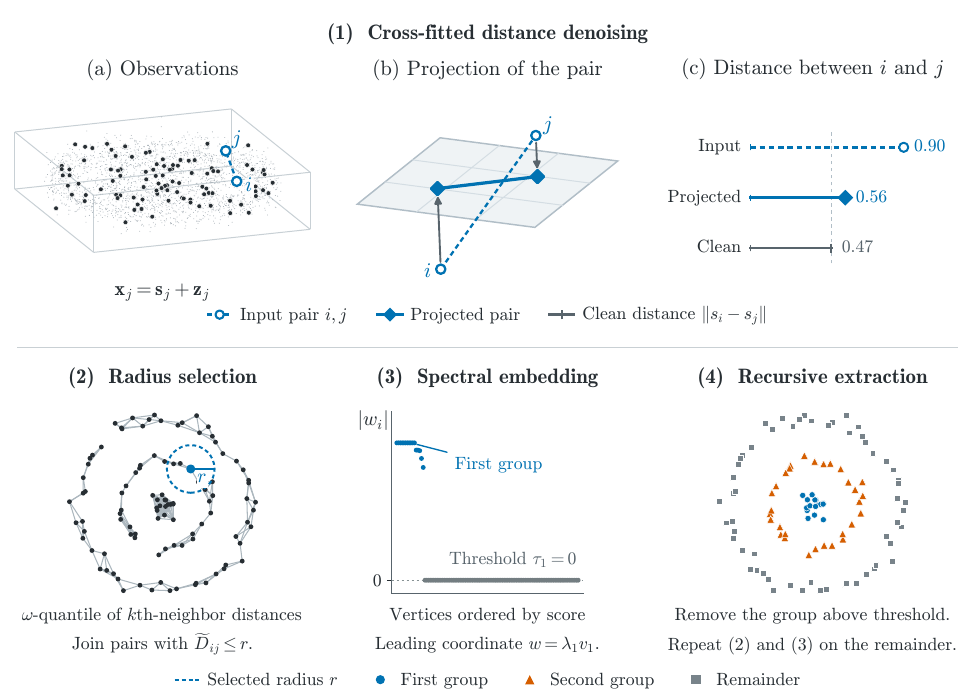}
  \caption{Four stages of the algorithm for a planar signal observed with noise
    in three coordinates. In (1), a projection fitted on a fold containing
    neither marked vertex brings their distance closer to the clean distance
    in this example. In (2), the selected radius determines which pairs of
    vertices are joined. In (3), the leading spectral coordinate identifies
    the first community. In (4), its vertices are removed and steps (2) and
    (3) are repeated on the remaining vertices.}
  \label{fig:method-overview}
\end{figure}

Our method has four steps, illustrated in
Figure~\ref{fig:method-overview}: distance denoising, radius selection,
spectral embedding, and recursive community extraction. The first step
returns a denoised distance matrix, which serves as the input for all
three later steps. The first three steps require no assumptions about
communities: their guarantees hold for a general law of the clean
signals (Section~\ref{sec:embedding-guarantees}). Only the last step
assumes that the observations form communities
(Section~\ref{sec:clustering-guarantees}).

The first step estimates the distances between the clean signals
(Algorithm~\ref{alg:three-fold-denoising}). It fits a projection onto an
estimated signal subspace on each of three folds and measures each pair
of observations after projection by a fold that contains neither of
them, as in Figure~\ref{fig:method-overview}(1). The denoised distances
are free of noise fluctuations from which the raw distances suffer (Theorem~\ref{thm:denoising-error-bound}). The output distance is close to the clean distances when \(\SNR_N\) is bounded, ensuring the validity of the subsequent steps.
The denoising algorithm is proved to be optimal in the minimax sense under the conditions of Section~\ref{sec:distance-denoising}
(Theorem~\ref{thm:coordinate-reconstruction-lower-bound}).

The second step first selects the radius from the denoised distances
(Algorithm~\ref{alg:radius-selection}), then builds a RGG with the selected radius. 
The radius is an empirical quantile of the denoised distances to their \(k\)-th nearest neighbors, as in Figure~\ref{fig:method-overview}(2). Our radius selection guarantees a hard cutoff at the selected radius, which matches the radius selected from the clean distances up to a vanishing relative error (Proposition~\ref{prop:radius-stability}).

The third step embeds the RGG at the selected radius (Algorithm~\ref{alg:RGG-spectral-embedding}). It weights each edge by a function of its length relative to the radius, and takes eigenvectors of the resulting kernel matrix, scaled by their eigenvalues, as in Figure~\ref{fig:method-overview}(3). Perturbation bounds for kernel matrices usually rely on a smooth kernel~\cite{elkaroui2010spectrum,ding2023learning}. Comparisons with the population operator through a reproducing kernel Hilbert space give errors of order \(N^{-1/2}\), which can exceed the eigenvalue scale of a local graph~\cite{ding2023learning}. Our bounds hold for the hard cutoff and vanish relative to that scale (Theorems~\ref{thm:spectral-stability} and~\ref{thm:population-eigenfunction-convergence}).

The fourth step extracts the communities one at a time (Algorithm~\ref{alg:RGG-clustering}). At each stage, it applies the second and third steps to remaining observations, assigns those whose leading coordinate exceeds a threshold to one community, and removes them, as in Figure~\ref{fig:method-overview}(4). Guarantees for spectral clustering need an eigengap and separated centers~\cite{yu2015useful,abbe2022lp,loffler2021optimality}. This step needs neither, because the leading eigenvector is supported on one connected component of the graph. It recovers every label with overwhelming probability when the input error and the sampling resolution are small compared with the separation \(\Delta\) (Theorem~\ref{thm:strong-consistency}).

\subsection{Related literature}\label{sec:related-work}
We review spectral embeddings as the noise grows, from clean samples to
\(\SNR_N\to\infty\) and then to the strong-noise regime of this paper.
We then compare our recovery guarantee with clustering by separated
supports and with random geometric graphs.

For observations sampled from the clean signals that $\vx_j=\clean_j$ in (\ref{eqn:signal-plus-noise}), spectral embeddings
are well understood. They build nonlinear coordinates from the
eigenvectors of a kernel or graph matrix. Some commonly used constructions include
kernel PCA~\cite{scholkopf1998nonlinear}, Laplacian
eigenmaps~\cite{belkin2003laplacian}, and diffusion
maps and its variants~\cite{coifman2006diffusion,singer2017spectral,singer2012vector}. As the sample
grows, graph Laplacians converge to continuum Laplace-type
operators~\cite{belkin2006convergence,singer2006graph,gine2006empirical},
with quantitative rates for their
eigenpairs~\cite{garciatrillos2020error,dunson2021spectral,wormell2021spectral}
and minimax rates~\cite{trillos2025minimax}. Likewise, the spectrum of an
empirical kernel matrix approximates that of its integral
operator~\cite{koltchinskii2000random,rosasco2010learning}, and the
fluctuations of its largest eigenvalue are
known~\cite{chatterjee2025fluctuation}. The graph results require the
scale of the graph (a bandwidth, a radius, or a neighbor count) to lie
in a range set by the sample size, the intrinsic dimension, and the
density. Within this range, spectral convergence attains optimal
rates~\cite{calder2022improved}, and \(k\)-nearest-neighbor graphs with
suitable \(k\) keep each cluster connected and different clusters
apart~\cite{maier2009optimal}. Instead of fixing the scale in advance,
one can adapt it to the data: through the distance to a nearby
neighbor~\cite{zelnikmanor2004selftuning}, through variable-bandwidth
kernels~\cite{berry2016variable}, whose graph Laplacians also
converge~\cite{cheng2022convergence}, or through the geometric mean of
nearest-neighbor distances~\cite{berry2019consistent}. Common heuristics
for choosing the scale are reviewed in~\cite{vonluxburg2007tutorial}.

However, in practice, the observations do not lie on the clean geometry and are usually contaminated by noise, as in the model~\eqref{eqn:signal-plus-noise}. How well an embedding of the
noisy observations reflects the clean geometry depends on the signal-to-noise
ratio~\eqref{eqn:signal-to-noise-ratio}, and existing guaranties concern the regime \(\SNR_N\to\infty\), where the signal energy
dominates the noise energy. In high dimensions, noise distorts kernel
spectra~\cite{elkaroui2010spectrum,elkaroui2010information}, but graph
connection Laplacians~\cite{elkaroui2016graph} and landmark diffusion
embeddings~\cite{shen2022scalability} are robust to noise under proper conditions. The
bandwidth and the signal-to-noise ratio jointly shape the spectra of
graph Laplacians and of kernel-based sensor fusion
algorithms~\cite{ding2022impact,ding2024fusion}. With a bandwidth set to
an empirical quantile of the distances, kernel embeddings of the noisy
observations converge to their clean counterparts and to population
eigenfunctions~\cite{ding2023learning}. This analysis extends to joint
and multi-view embeddings~\cite{ding2025kernel,ding2026generalized}, and
quantile bandwidths also give mixture
recovery~\cite{lin2026multikernel}. When each observation lies within
less than the bandwidth of its clean signal, the leading eigenvalues and
eigenspaces of a graph Laplacian change
little~\cite{ageev2026spectral}. These convergence guarantees require
\(\SNR_N\to\infty\) (Remark~\ref{rem:ding-ma-comparison}), failing to cover the strong noise regime.

In the strong noise regime \(\SNR_N=O(1)\), an embedding built
from the raw distances can fail (Section~\ref{sec:denoising-necessity}),
hence the noise must be removed first. Existing denoising methods include
PCA, with a uniform, rate-optimal error bound near a low-dimensional
subspace~\cite{tong2024uniform}; eigenvalue shrinkage for
multidimensional scaling~\cite{peterfreund2021multidimensional}; and
local regularization of the point cloud~\cite{garciatrillos2019local}.
A graph Laplacian built on denoised observations converges at the clean
rate when the remaining errors are small compared with the
bandwidth~\cite{ageev2026spectral}. A data-driven projection followed by a kernel spectral embedding also works at \(\SNR_N=O(1)\), requiring sparse structures on signal vectors~\cite{hu2026adaptive}. 

Our first contribution gives guaranties from the noisy observations to the embedding for all \(\SNR_N=O(1)\) without further constraints. The cross-fitted denoising has an error that matches a minimax lower bound
(Section~\ref{sec:distance-denoising}). The radius selected from the
denoised distances, and the hard-cutoff embedding at that radius, stay
close to their clean counterparts, and the embedding converges to the
eigenpairs of a population operator
(Sections~\ref{sec:stability-of-data-driven-selection}
and~\ref{sec:spectral-convergence}). These guaranties require i.i.d. sub-Gaussian noise that is weak in every
single direction compared with the smallest signal variance, although its total energy may exceed that of the signal; the clean law is
centered with a well-conditioned covariance.

Our second contribution demonstrates how the recovery of the
labels, and not only of the embedding, depends on the separation of the
communities relative to the noise. The communities are described by one geometric quantity, the separation
\(\Delta\) between their supports, and the noise by its largest variance
\(\sigma^2\) in a single direction. The recursive algorithm recovers
every label when
\(\Delta/\sigma\gg\xi_N^2\sqrt{(1+\rho_N)(m+\log N)}\) and the sampling
resolution is small compared with \(\Delta\)
(Corollary~\ref{cor:denoised-recovery}), where
\(\rho_N=\Tr\mGamma/(N\theta_{\min})\), with \(\theta_{\min}\) being the smallest signal variance, and \(\xi_N=\max\{1,\log\log N\}\). When \(\rho_N=O(1)\), this
threshold does not grow with the ambient dimension and does not involve
the total noise energy, so it can hold while \(\SNR_N\to0\). Raw distances can fail far above it: under isotropic Gaussian noise, their squared fluctuations are of the order \(\sigma^2\sqrt p\), and on two concentric
circles at bounded \(\SNR_N\), they fail throughout the range
\(\xi_N^2\sqrt{(1+\rho_N)(m+\log N)}\ll\Delta/\sigma\leq N^{-c}p^{1/4}\),
in which the denoised distances recover every label
(Section~\ref{sec:denoising-necessity}). When the regularity holds only
on cores, the fraction of misclassified observations is at most of the order of the mass outside the cores plus \(\log N/N\)
(Theorem~\ref{thm:weak-consistency}). To our knowledge, this is the
first such relationship for communities separated by their supports
rather than by their centers, under noise in a growing ambient
dimension.

Most existing clustering guarantees use a different separation, the
\emph{center separation}: the minimum distance between the means of two
communities. For isotropic Gaussian mixtures with a fixed number of
clusters, spectral clustering attains the minimax misclustering rate,
which is governed by the center separation relative to the noise level
\(\sigma\)~\cite{loffler2021optimality}. The consistency of spectral
clustering~\cite{von2008consistency}, eigenvector perturbation
bounds~\cite{yu2015useful,abbe2022lp,chen2021spectral}, and mixture
recovery with quantile bandwidths~\cite{lin2026multikernel} further rests on a spectral gap. Neither applies here: our communities may share a
center, so their center separation is zero while the support separation
\(\Delta\) is positive. Meanwhile, the kernel matrix of our graph may have
repeated leading eigenvalues, one from each community, so it need not
have a spectral gap.

Other work separates the communities by their supports, as we do.
Multi-manifold clustering handles nonlinear supports through
hierarchical mixtures~\cite{wang2008multiple}, probabilistic models of
noisy manifolds~\cite{canducci2022probabilistic}, affinities built from
tangent and geodesic information~\cite{wang2015multi-manifold}, and
spectral approximation for manifolds that may intersect and have
different dimensions~\cite{trillos2023large}. For suitable neighbor
counts, the connected components of a \(k\)-nearest-neighbor graph
identify the clusters of a density~\cite{maier2009optimal}, and kernel
eigenfunctions relate to mixture components, which suggests extracting
the components one at a time~\cite{shi2009data}, the strategy we adopt.
Closest to our setting, in~\cite{arias2011clustering} the observations
are sampled within a distance \(\tau\) of surfaces of possibly different
dimensions in a fixed ambient dimension, and the connected components of
a neighborhood graph, as well as spectral clustering, recover the
surfaces under separation conditions that are optimal up to logarithmic
factors. In our model, the noise displacement instead grows with the
ambient dimension and can far exceed \(\Delta\), so the observations do
not stay near their supports. The recursive extraction then needs
distances denoised to an error below \(\Delta\) and a radius, selected
from the same sample, between the sampling resolution and \(\Delta\)
(Section~\ref{sec:clustering-guarantees}).

Our work is also related to random geometric graphs (RGGs), in which two
vertices are joined when their latent points are
close~\cite{duchemin2023random}. Their detection
thresholds~\cite{devroye2011high,bubeck2016testing,liu2022testing,brennan2024threshold,bangachev2024fourier,bangachev2025sandwiching,baguley2025testing,mao2026smooth,du2026resolution},
spectra~\cite{cao2025spectra,ding2025edge}, metric
reconstruction~\cite{huang2025reconstructing}, and
entropy~\cite{baker2026entropy} have been studied, and community
recovery is characterized in the geometric
SBM~\cite{galhotra2023community,abbe2021community,gaudio2024exact,gaudio2025exact}
and in the Gaussian mixture block
model~\cite{li2023spectral,fernandez2026spectral}, with extensions
in~\cite{egidi2023clustering,peche2020robustness,avrachenkov2026spectral}.
In all these works the graph is observed, including when Gaussian noise
perturbs the latent points before the edges are
formed~\cite{brailovskaya2026spectral}. We instead build the graph from
denoised distances between the noisy observations, and the condition on
\(\Delta/\sigma\) keeps its edges from joining two communities: the
radius plus the error of the denoised distances stays below \(\Delta\)
(Section~\ref{sec:clustering-guarantees}).

\subsection{Notation}\label{sec:theory-notation}
Denote the set \([n]=\{1,\ldots,n\}\).  We write \(x\wedge y=\min\{x,y\}\) and \(x\vee y=\max\{x,y\}\). We write
\(u_N\lesssim v_N\), for nonnegative \(u_N\) and \(v_N\), if
\(u_N\leq Cv_N\) or \(u_N = O(v_N)\) for some constant \(C\) independent of \(N\); \(u_N\asymp v_N\) means \(u_N\lesssim v_N\lesssim u_N\), and \(u_N\ll v_N\), or \(v_N\gg u_N\), means \(u_N=o(v_N)\).

For a matrix \(\mH\), let \(\lambda_j(\mH)\) denote its \(j\)-th largest eigenvalue and \(\vv_j(\mH)\) denote the corresponding unit eigenvector. Let \(\lambda_{\min}(\mH)\) denote the smallest eigenvalue. Let \(\|\cdot\|\) denote the Euclidean norm for vectors and the operator norm for matrices, while \(\|\cdot\|_{\frob}\) denotes the Frobenius norm. For a matrix \(\mH\in\R^{N\times N}\), let \(\|\mH\|_{\max}\coloneqq\max_{i,j}|\emH_{ij}|\).
\begin{definition}[Overwhelming probability]
\label{def:overwhelming-probability}
Let \(\gE\) and \(\gS\) be events and \(X_N,Y_N\geq0\) random variables, all depending on \(N\).
\begin{enumerate}[label=(\alph*)]
\item \(\gS\) holds \emph{with overwhelming probability} if, for every fixed \(\nu>0\), there is a constant \(C_\nu>0\) such that \(\P(\gS^c)\leq C_\nu N^{-\nu}\) for all $N\geq1$.
\item \(\gS\) holds \emph{with overwhelming probability on} \(\gE\) if the same holds with \(\P(\gS^c)\) replaced by \(\P(\gE\setminus\gS)\).
\item \(X_N\lesssim Y_N\) \emph{with overwhelming probability} (on \(\gE\)) if, for every fixed \(\nu>0\), the event \(\{X_N\leq C_\nu Y_N\}\) satisfies (a) (or (b)) with the same constant \(C_\nu\).
\end{enumerate}
\end{definition}
In~(b), \(\P(\gS^c)\leq\P(\gE^c)+\P(\gE\setminus\gS)\), so \(\gS\) holds with overwhelming probability whenever \(\gE\) does.

\subsection{Organization}\label{sec:organization}
The rest of the paper is organized as follows. We present our algorithms in Section~\ref{sec:methodology}. Theoretical guarantees for each algorithm are provided in Section~\ref{sec:embedding-guarantees} and Section~\ref{sec:clustering-guarantees}. The performance of our algorithms is validated through numerical experiments in Section~\ref{sec:synthetic-experiments}. Proof details are deferred to the Appendices.

\section{Methodology}\label{sec:methodology}
\begingroup
\setlength{\intextsep}{1.2\baselineskip plus 2pt minus 2pt}
\setlength{\textfloatsep}{1.2\baselineskip plus 2pt minus 2pt}
We present our methodology in four steps:
denoising the pairwise distances (Section~\ref{sec:denoising-methodology}),
selecting a radius and building the RGG (Section~\ref{sec:data-driven-radius-selection}),
computing the spectral embedding (Section~\ref{sec:spectral-embedding-methodology}), and
extracting communities recursively (Section~\ref{sec:recursive-methodology}).
Throughout, let \(\mX=[\vx_1,\ldots,\vx_N]\in\R^{p\times N}\) denote the data
matrix, with one observation per column.

\subsection{Cross-fitted projection denoising}
\label{sec:denoising-methodology}
We first provide a general distance denoising algorithm, whose output will be used as input to the subsequent algorithms. Algorithm~\ref{alg:three-fold-denoising} computes the denoised distance matrix \(\widetilde\mD\). It splits the sample into three folds, fits a projection onto an estimated signal subspace on each fold, and measures each distance after projection on a fold that contains neither endpoint. The theoretical guarantee is given in Section~\ref{sec:distance-denoising}. We now describe the algorithm in detail.

\paragraph{Data splitting}
We first draw a uniformly random permutation \(\pi_{\mathrm{fold}}\) on \([N]\),
independently of the data matrix \(\mX\). We then split the $N$ observations into three folds as follows:
\begin{align}
 F_q\coloneqq\{i\in[N]:\ \pi_{\mathrm{fold}}(i)\equiv q\pmod 3\},\qquad q=1,2,3,
 \label{eqn:three-fold-split}
\end{align}
whose sizes differ by at most one. For each \(q\in\{1,2,3\}\), we denote the data matrix in the \(q\)-th fold by \(\mX_q=[\vx_i]_{i\in F_q}\in\R^{p\times N_q}\), where \(N_q=|F_q|\).

\paragraph{Estimation of signal dimension}
Let \(\mG_q=\mX_q^{\sT}\mX_q\) denote the \emph{Gram matrix} of the \(q\)-th fold. We then define the \emph{shifted Gram matrix} by
\begin{align}
 \widetilde\mG_q\coloneqq\mG_q-\lambda_{\min}(\mG_q)\mI_{N_q},
 \label{eqn:fold-corrected-gram}
\end{align}
where \(\lambda_{\min}(\mG_q)\) is the smallest eigenvalue of \(\mG_q\). Using the fold $q$, the signal dimension is estimated by
\begin{align}
 \widetilde m_q\coloneqq\bigl|\{j\in[N_q]:
   \lambda_j(\widetilde\mG_q)>\lambda_1(\widetilde\mG_q)/\log(\mathrm eN)\}\bigr|.
 \label{eqn:fold-rank}
\end{align}
\paragraph{Fold map} With the estimated signal dimension \(\widetilde m_q\), the fold map is defined as
 \begin{align}
 \foldmap_q\coloneqq\mX_q\widetilde\mU_q\widetilde\mLambda_q^{-1/2}
 \in\R^{p\times\widetilde m_q},
 \label{eqn:fold-map}
\end{align}
where \(\widetilde\mLambda_q=\operatorname{diag}(\lambda_1(\widetilde\mG_q),\ldots,\lambda_{\widetilde m_q}(\widetilde\mG_q)) \in \R^{\widetilde m_q\times\widetilde m_q}\) and \(\widetilde\mU_q=[\vv_1(\widetilde\mG_q),\ldots,\vv_{\widetilde m_q}(\widetilde\mG_q)] \in \R^{N_q\times\widetilde m_q}\) contain the leading \(\widetilde m_q\) eigenvalues of the shifted Gram matrix and the corresponding eigenvectors, respectively.

\paragraph{Evaluation on a held-out fold}
Given a pair \(i\neq j\), we assign it to the first fold that contains neither endpoint, i.e.,
\begin{align}
 q(i,j)\coloneqq\min\{q\in\{1,2,3\}:\ i\notin F_q,\ j\notin F_q\}.
 \label{eqn:three-fold-pair-assignment}
\end{align}
With \(q=q(i,j)\), the denoised distance is given by
\begin{align}
 \widetilde D_{ij}\coloneqq\|\foldmap_q^{\sT}(\vx_i-\vx_j)\|.
 \label{eqn:denoised-distance}
\end{align}
Note that we set \(\widetilde D_{ii}=0\) for every \(i\in[N]\).

\paragraph{Optional debiasing}
The algorithm has an optional debiasing step, which is turned off by
default. When debiasing is requested, the first step is to estimate the noise variance on fold \(q\), for nearly isotropic noise, by
\begin{align}
 \widehat\sigma_q^2\coloneqq\frac{\lambda_{\min}(\mG_q)}{p-\widetilde m_q}.
 \label{eqn:fold-noise-variance-estimator}
\end{align}
As will be demonstrated later, the debiasing procedure is provably beneficial when the denominator is large. We therefore debias fold \(q\) only if \(p-\widetilde m_q\geq20N\), and then call it \emph{debiased}. For a pair \(i\neq j\) whose fold \(q=q(i,j)\) is debiased, we replace \eqref{eqn:denoised-distance} by the debiased distance:
\begin{align}
 \widetilde D_{ij}\coloneqq
 \bigl[\|\foldmap_q^{\sT}(\vx_i-\vx_j)\|^2
       -2\widehat\sigma_q^2\Tr(\foldmap_q^{\sT}\foldmap_q)\bigr]_+^{1/2},
 \label{eqn:debiased-distance}
\end{align}
where \([z]_+=\max\{z,0\}\). Algorithm~\ref{alg:three-fold-denoising}
summarizes these steps.

\begin{algorithm}[h]
  \caption{Cross-fitted projection denoising}
  \label{alg:three-fold-denoising}
  \begin{algorithmic}[1]
  \Require Data \(\mX\in\R^{p\times N}\); folds \(\{F_q\}_{q=1}^3\) from \eqref{eqn:three-fold-split}; option \(\mathtt{debias}\in\{\mathtt{false},\mathtt{true}\}\), by default \(\mathtt{false}\).
  \For{\(q=1,2,3\)} \Comment{fit one map per fold}
      \State Compute \(\widetilde\mG_q\) by \eqref{eqn:fold-corrected-gram} and its eigenpairs
      \State \(\widetilde m_q\gets\) \eqref{eqn:fold-rank} and \(\foldmap_q\gets\) \eqref{eqn:fold-map}
      \If{\(\mathtt{debias}\) and \(p-\widetilde m_q\geq20N\)}
        \State \(\widehat\sigma_q^2\gets\) \eqref{eqn:fold-noise-variance-estimator} and mark fold \(q\) debiased
      \EndIf
  \EndFor
  \For{\(1\leq i<j\leq N\)} \Comment{evaluate on a held-out fold}
      \State \(q\gets q(i,j)\) by \eqref{eqn:three-fold-pair-assignment}
      \State \(\widetilde D_{ij}\gets\) \eqref{eqn:debiased-distance} if fold \(q\) is debiased, and \eqref{eqn:denoised-distance} otherwise; \(\widetilde D_{ji}\gets\widetilde D_{ij}\)
  \EndFor
  \State \(\widetilde D_{ii}\gets0\) for every \(i\in[N]\)
  \Ensure The denoised distance matrix \(\widetilde\mD\in\R^{N\times N}\).
  \end{algorithmic}
\end{algorithm}

\subsection{Data-driven radius selection and construction of RGG}
\label{sec:data-driven-radius-selection}
We now present a data-driven radius selection algorithm, whose output will be used as input for constructing the random geometric graph later. Algorithm~\ref{alg:radius-selection} takes a distance matrix. It then computes the nearest-neighbor distances and returns the radius at which a prescribed fraction of the active observations have at least \(k\) neighbors. Theoretical guarantees are given in Section~\ref{sec:stability-of-data-driven-selection}. We now describe the algorithm in detail.

\paragraph{Input distances}
Let \(\mD\in\R^{N\times N}\) be a symmetric and entrywise nonnegative matrix with zero diagonal. One can use the raw distance matrix \(\mD^{(\mathrm{x})}=[D^{(\mathrm{x})}_{ij}]_{i,j\in[N]}\), where \(D^{(\mathrm{x})}_{ij}=\|\vx_i-\vx_j\|\), or the denoised distance matrix \(\widetilde\mD\) from Algorithm~\ref{alg:three-fold-denoising}, with or without debiasing. The following procedures accept any such matrix.

\paragraph{Nearest-neighbor distances}
Let \(\sI\subseteq[N]\) be the nonempty set of active observations, and let
\(1\leq k<|\sI|\) be the neighbor count. For each \(i\in \sI\), we denote by
\(r_i(\sI;k)\) the \(k\)-th smallest value in
\(\{D_{ij}:j\in \sI\setminus\{i\}\}\), that is, the distance from \(i\) to its
\(k\)-th nearest active neighbor. We then sort these values as
\(r_{(1)}(\sI;k)\leq\cdots\leq r_{(|\sI|)}(\sI;k)\).

\paragraph{Selected radius}
Given a quantile level \(\omega\in(0,1)\), the selected radius is the empirical \(\omega\)-quantile of the nearest-neighbor distances, formally defined as
\begin{align}
 r(\sI;k,\omega)\coloneqq r_{(\lceil\omega |\sI|\rceil)}(\sI;k).
 \label{eqn:selected-radius}
\end{align}
Hence, at least \(\lceil\omega |\sI|\rceil\) of the active observations have at least $k$ neighbors within distance $r(\sI;k,\omega)$ from themselves. We now present the algorithm block below.

\begin{algorithm}[h]
  \caption{Data-driven radius selection}
  \label{alg:radius-selection}
  \begin{algorithmic}[1]
  \Require Distance matrix \(\mD\), active set \(\sI\subseteq[N]\), neighbor count \(1\leq k<|\sI|\), quantile level \(\omega\in(0,1)\).
  \For{\(i\in \sI\)}
      \State \(r_i(\sI;k)\gets\) the \(k\)-th smallest value in
        \(\{D_{ij}:j\in \sI\setminus\{i\}\}\)
  \EndFor
  \State Sort these values as \(r_{(1)}(\sI;k)\leq\cdots\leq r_{(|\sI|)}(\sI;k)\)
  \State Compute \(r(\sI;k,\omega)\) by \eqref{eqn:selected-radius}
  \Ensure Selected radius \(r(\sI;k,\omega)\).
  \end{algorithmic}
\end{algorithm}

\subsection{Spectral embedding procedure}
\label{sec:spectral-embedding-methodology}
Given a radius, we can build a \emph{Random Geometric Graph} (RGG) on the active
observations with given kernel function, resulting in a kernel matrix. We then obtain the eigenvalue-weighted embedding by the spectral embedding algorithm. Algorithm~\ref{alg:RGG-spectral-embedding} takes a distance matrix, a radius and an index set, and returns the embedding matrix. Theoretical guarantees are given in Section~\ref{sec:spectral-convergence}. We now describe the algorithm in detail.

\paragraph{Kernel matrix}
Let \(\sI\subseteq[N]\) be the set of active observations and let \(r>0\) be
a radius. The \emph{random geometric graph} (RGG) on \(\sI\) with radius
\(r\) is the graph with vertex set \(\sI\) in which two distinct vertices
\(i,j\in \sI\) are adjacent if and only if \(D_{ij}\leq r\). Its adjacency
matrix \(\mA(\sI;r)\in\{0,1\}^{|\sI|\times|\sI|}\) is given by
\begin{align}
 [\mA(\sI;r)]_{ij}\coloneqq\indi{i\ne j}\indi{D_{ij}\leq r},
 \qquad i,j\in \sI.
 \label{eqn:RGG-adjacency}
\end{align}
Given a kernel function \(\kf:[0,\infty)\to[0,\infty)\), we weight each edge by the function at the relative distance of its endpoints and normalize by the number of active observations. The kernel matrix is given by
\begin{align}
  [\mK(\sI;r)]_{ij}\coloneqq \frac{1}{|\sI|} [\mA(\sI;r)]_{ij} \cdot \kf\bigl({D_{ij}}/{r}\bigr),
  \qquad i,j\in \sI.
 \label{eqn:RGG-kernel-def}
\end{align}
In particular, \(\mK(\sI;r)=|\sI|^{-1}\mA(\sI;r)\) for the constant kernel function \(\kf\equiv1\).

\paragraph{Eigenvalue-weighted embedding}
Let \(\lambda_1\geq\cdots\geq\lambda_{|\sI|}\) be the eigenvalues of \(\mK(\sI;r)\), and let \(\vv_1,\ldots,\vv_{|\sI|}\) be corresponding orthonormal eigenvectors, as in Section~\ref{sec:theory-notation}. For a nonempty index set \(\sJ\subseteq[|\sI|]\), let \(\mV_{\sJ}\) be the matrix with columns \(\{\vv_j \}_{j\in \sJ}\), and let \(\mLambda_{\sJ}\) be the diagonal matrix of the corresponding eigenvalues. The eigenvalue-weighted embedding scales each eigenvector by its eigenvalue, formally,
\begin{align}
 \mZ_{\sJ}\coloneqq\mV_{\sJ}\mLambda_{\sJ}=\mK(\sI;r)\mV_{\sJ}\in\R^{|\sI|\times|\sJ|}.
 \label{eqn:method-weighted-embedding}
\end{align}
The embedding of observation \(i\) is given by the \(i\)-th row of \(\mZ_{\sJ}\). We summarize these steps in the algorithm block below.

\begin{algorithm}[h]
  \caption{Spectral embedding of the random geometric graph}
  \label{alg:RGG-spectral-embedding}
  \begin{algorithmic}[1]
  \Require Distance matrix \(\mD\), radius \(r>0\), kernel function \(\kf\), active set \(\sI\subseteq[N]\), nonempty set of spectral indices \(\sJ\subseteq[|\sI|]\).
  \State Compute \(\mK(\sI;r)\) by \eqref{eqn:RGG-kernel-def}
  \State Compute its eigenpairs \((\lambda_j,\vv_j)\)
  \State Compute \(\mZ_{\sJ}\) by \eqref{eqn:method-weighted-embedding}
    \Comment{row \(i\) embeds observation \(i\)}
  \Ensure Embedding matrix \(\mZ_{\sJ}\in\R^{|\sI|\times|\sJ|}\).
  \end{algorithmic}
\end{algorithm}

\subsection{Downstream task: recursive spectral clustering}
\label{sec:recursive-methodology}
Finally, we combine the previous steps into a recursive spectral clustering algorithm. Extracting one community at a time follows the suggestion of Shi, Belkin, and Yu~\cite{shi2009data}, who relate the leading eigenfunctions of a kernel operator to the components of a mixture. Algorithm~\ref{alg:RGG-clustering} takes a distance matrix and the number of communities \(B\), and returns the estimated community labels. Theoretical guarantees are given in Section~\ref{sec:recursive-clustering-guarantee}. We now describe the algorithm in detail.

\paragraph{Tuning parameters}
The algorithm runs in \(B-1\) stages. The neighbor count \(k=k_N\) is
chosen once from the original sample size \(N\) and is kept fixed
throughout the recursion, while each stage \(\ell\in[B-1]\) has its own
quantile level \(\omega_\ell\in(0,1)\) and threshold \(\tau_\ell\geq0\).

\paragraph{Stage quantities}
Stage \(\ell\) works on the active set \(\sI_\ell\), of size
\(N_\ell=|\sI_\ell|\), starting from \(\sI_1=[N]\). It selects the radius by
Algorithm~\ref{alg:radius-selection}, forms the kernel matrix at that
radius, and computes the leading coordinate of the eigenvalue-weighted
embedding by Algorithm~\ref{alg:RGG-spectral-embedding} with
\(\sJ=\{1\}\):
\begin{align}
 r_\ell&=r(\sI_\ell;k,\omega_\ell),
 &\mK_\ell&=\mK(\sI_\ell;r_\ell),
 &\vw_\ell&=\lambda_1(\mK_\ell)\vv_1(\mK_\ell),
 \label{eqn:stage-quantities}
\end{align}
where \(\vw_\ell\in\R^{\sI_\ell}\) is indexed by the active observations.
In case that the leading eigenvalue \(\lambda_1(\mK_\ell)\) is not simple, we fix the leading eigenvector \(\vv_1(\mK_\ell)\) as follows. Let \(\mP_\ell\) be the orthogonal projector onto the eigenspace \(\{\vv\in\R^{\sI_\ell}:\mK_\ell\vv=\lambda_1(\mK_\ell)\vv\}\), and let
\((\mP_\ell)_{\cdot i}\) denote its column indexed by \(i\in\sI_\ell\). Since
\(\mP_\ell\ne\bzero\), the index
\(i_\star\coloneqq\min\{i\in\sI_\ell:(\mP_\ell)_{\cdot i}\ne\bzero\}\)
exists, and we define the leading eigenvector as
\[
 \vv_1(\mK_\ell)\coloneqq
 \frac{(\mP_\ell)_{\cdot i_\star}}{\|(\mP_\ell)_{\cdot i_\star}\|}.
\]
 When \(\lambda_1(\mK_\ell)\) is simple, \(\mP_\ell=\vv\vv^{\sT}\) with the unit eigenvector \(\vv=\vv_1(\mK_\ell)\) up to a universal sign flip.

\paragraph{Extraction}
Stage \(\ell\) then extracts the active observations whose leading coordinate exceeds the threshold $\tau_\ell$ in absolute value. Formally, we assign them label \(\ell\) to the following set:
\begin{align}
 \widehat{\gC}_\ell\coloneqq\{j\in \sI_\ell:|[\vw_\ell]_j|>\tau_\ell\}.
 \label{eqn:method-recursive-extraction}
\end{align}
We then remove them from the current active set, so that the next stage starts
from \(\sI_{\ell+1}=\sI_\ell\setminus\widehat{\gC}_\ell\). After \(B-1\)
stages, the remaining observations \(\sI_B\) form the last community and
receive label \(B\). Together with Algorithm~\ref{alg:three-fold-denoising},
the whole method is, schematically,
\[
 \mX
 \;\longrightarrow\;\widetilde\mD
 \;\longrightarrow\;(r_\ell,\mK_\ell)
 \;\longrightarrow\;\vw_\ell
 \;\longrightarrow\;\widehat{\gC}_\ell,
 \qquad \ell=1,\ldots,B-1.
\]
We present the algorithm in the block below.

\begin{algorithm}[h]
  \caption{Recursive spectral clustering from a distance matrix}
  \label{alg:RGG-clustering}
  \begin{algorithmic}[1]
  \Require Distance matrix \(\mD\) on \([N]\), number of communities
    \(B\geq2\), kernel function \(\kf\), neighbor count \(1\leq k<N\), quantile levels \(\omega_\ell\in(0,1)\) and thresholds \(\tau_\ell\geq0\) for each stage \(\ell\in[B-1]\).
  \State \(\sI_1\gets[N]\)
  \For{\(\ell=1,\ldots,B-1\)} \Comment{stop early if a step is undefined}
      \State Compute \(r_\ell\) by Algorithm~\ref{alg:radius-selection}
      \State Compute \(\mK_\ell\) and \(\vw_\ell\) by
        \eqref{eqn:stage-quantities}
      \State Compute \(\widehat{\gC}_\ell\) by
        \eqref{eqn:method-recursive-extraction}; set
        \(\widehat{\evy}_j\gets\ell\) for \(j\in\widehat{\gC}_\ell\)
      \State \(\sI_{\ell+1}\gets \sI_\ell\setminus\widehat{\gC}_\ell\)
  \EndFor
  \State \(\widehat{\evy}_j\gets B\) for \(j\in \sI_B\)
  \Ensure Estimated label vector \(\widehat{\vy}\in[B]^N\).
  \end{algorithmic}
\end{algorithm}

\paragraph{Early stopping}
Two steps of a stage can be undefined. First, radius selection requires
\(k<N_\ell\), so that every active observation has a \(k\)-th nearest
neighbor. Second, the kernel \eqref{eqn:RGG-kernel-def} requires
\(r_\ell>0\). If either requirement fails at stage \(\ell\), we stop early:
the labels assigned at stages \(1,\ldots,\ell-1\) are kept, and every
observation still active, in \(\sI_\ell\), receives label \(\ell\).
Otherwise the run \emph{completes}. With this rule, \(\widehat\vy\) is
defined on every realization; a run that stops at the first stage labels
every observation \(1\).

\FloatBarrier
\endgroup

\section{Guarantees for denoising, radius selection, and spectral embedding}\label{sec:embedding-guarantees}
We provide theoretical guarantees for Algorithms~\ref{alg:three-fold-denoising}--\ref{alg:RGG-spectral-embedding} in this section. We state assumptions in Section~\ref{sec:setting}. Performance of the denoising algorithm is analyzed in Section~\ref{sec:distance-denoising}. We show that the selected radius and the spectral embedding are close to those computed from the clean distances in Sections~\ref{sec:stability-of-data-driven-selection} and~\ref{sec:spectral-convergence}.

\subsection{Setting and assumptions}
\label{sec:setting}
We introduce assumptions on clean signals and noisy vectors, separately.
\paragraph{Clean signals}
In the model \eqref{eqn:signal-plus-noise}, the clean signals
\(\clean_1,\ldots,\clean_N\) are i.i.d.\ with law \(\P_{\mathrm{s}}\), the
\emph{clean law}. Its support is compact and lies in the \emph{signal
subspace}, the column space of a matrix \(\mJ\in\R^{p\times m}\) with
orthonormal columns, \(p\geq m\). The mean of the clean signals is
\begin{align}
    \rvmu_{\mathrm{s}}\coloneqq\int_{\R^p}\vs\,\P_{\mathrm{s}}(\odif{\vs}).
    \label{eqn:global-signal-centering}
\end{align}
Let \(\mSigma\) denote the covariance of the signal coordinates
\(\mJ^{\sT}\clean_j\in\R^m\), with spectral decomposition
\begin{align}
    \mSigma&\coloneqq\Cov(\mJ^{\sT}\clean_j)=\mR\mTheta\mR^{\sT}
      \in\R^{m\times m},
      &\quad
    \mTheta&=\operatorname{diag}(\theta_1,\ldots,\theta_m),
    \label{eqn:signal-covariance-definition}
\end{align}
where \(\theta_{\max}=\theta_1\geq\cdots\geq\theta_m=\theta_{\min}\geq0\) and
\(\mR\in\R^{m\times m}\) is orthogonal. Let \(\clean_j^\circ\coloneqq\mR^{\sT}\mJ^{\sT}\clean_j\in\R^m\) denote the \emph{reduced signal coordinates}, which have covariance \(\mTheta\) and the same pairwise distances as the clean
signals.
\begin{assumption}[Centered mean and well-conditioned signal covariance]
\label{ass:growing-covariance}
The clean law is centered, \(\rvmu_{\mathrm{s}}=\bzero\) in \eqref{eqn:global-signal-centering}. The smallest eigenvalue \(\theta_{\min}\) of \(\mSigma\) is positive, and, for a
constant \(C>0\),
\[
 {\theta_{\max}}/{\theta_{\min}}=o(\log N),
 \qquad
 \|\mSigma^{-1/2}\mJ^{\sT}\clean_j\|^2\leq Cm\quad\text{almost surely}.
\]
\end{assumption}

\paragraph{Local mass regularity}
Sections~\ref{sec:stability-of-data-driven-selection}
and~\ref{sec:spectral-convergence} use the clean law only through the mass
that it gives to small balls and thin shells around its support points.
\begin{definition}[Local mass regularity]
\label{def:local-mass-regularity}
Let \(d\geq1\) be an integer, and let \(\ell>0\), \(R>0\), and
\(0<c_{\mathrm{g}}\leq C_{\mathrm{g}}<\infty\). A law \(\mathds{Q}\) on
\(\R^p\) has \emph{local mass regularity} with dimension \(d\), length
scale \(\ell\), locality radius \(R\), and constants
\((c_{\mathrm g},C_{\mathrm g})\) if, for every support point
\(\vs\in\operatorname{supp}(\mathds{Q})\), every radius \(0<r\leq R\), and
every relative shell width \(0<u\leq1/2\), the following hold:
\begin{align}
    c_{\mathrm{g}}(r/\ell)^d
    \leq\mathds{Q}\bigl(\{\vt:\|\vt-\vs\|\leq r\}\bigr)
    &\leq C_{\mathrm{g}}(r/\ell)^d,
    \label{eqn:uniform-clean-small-ball}\\
    \mathds{Q}\bigl(\{\vt:(1-u)r<\|\vt-\vs\|\leq(1+u)r\}\bigr)
    &\leq C_{\mathrm{g}}u
       (r/\ell)^d.
    \label{eqn:uniform-clean-shell}
\end{align}
\end{definition}
\begin{assumption}[Regular clean law]
\label{ass:regular-clean-law}
Let \(\P_{\mathrm{s}}\) denote the law for clean signals. There are a \emph{length scale} \(L>0\) and a \emph{locality radius}
\(\locrad>0\) such that \(\P_{\mathrm{s}}\) has local mass
regularity with dimension \(d\), length scale \(L\), locality radius
\(\locrad\), and constants \((c_{\mathrm g},C_{\mathrm g})\).
\end{assumption}
A ball of radius \(r\leq\locrad\) around a support point thus has mass of
order \((r/L)^d\). One example that we use in Section~\ref{sec:multi-community-manifolds} is a compact \(d\)-dimensional
manifold \(\manifold\) with a bounded density, \emph{dilated} by \(L\), that is, a law on \(L\manifold\); then \eqref{eqn:uniform-clean-small-ball} is the volume growth of small balls in \(\manifold\), and \(\locrad\lesssim L\).

\paragraph{Noise vectors}
The noise vectors in \eqref{eqn:signal-plus-noise} satisfy the following.
\begin{assumption}[Sub-Gaussian noise]
    \label{ass:noise-variance-structure}
   Suppose that \(\noise_j=\mGamma^{1/2}\vg_j\) for \(j\in[N]\), where
    \(\mGamma\in\R^{p\times p}\) is positive semidefinite and
    \(\vg_1,\ldots,\vg_N\in\R^p\) are i.i.d.\ and independent of the clean
    signals. The coordinates of \(\vg_j\) are independent, with mean zero,
    unit variance, and sub-Gaussian norm at most a constant \(C_{\mathrm{sg}}\).
\end{assumption}
Then \(\Cov(\noise_j)=\mGamma\). We write
\(\sigma^2\coloneqq\|\mGamma\|\) for the largest noise variance;
quantities normalized by \(\sigma\) are defined for \(\sigma>0\).
Constants in the results below may depend on \(C_{\mathrm{sg}}\).

\paragraph{Signal-to-noise ratio}
For independent copies \(\clean,\clean'\) of a clean signal and
\(\noise,\noise'\) of the noise, \(\E\|\clean-\clean'\|^2=2\Tr\mTheta\) and
\(\E\|\noise-\noise'\|^2=2\Tr\mGamma\), so the signal-to-noise ratio
\eqref{eqn:signal-to-noise-ratio} is
\(\SNR_N=\Tr\mTheta/\Tr\mGamma\).

\paragraph{Input distances}
Let \(D^{(\mathrm{s})}_{ij}=\|\clean_i-\clean_j\|\) denote the clean distances.
For any distance matrix \(\mD\), we measure its closeness to the clean distances as
follows.
\begin{definition}[Affine event]
\label{def:affine-event}
For a distance matrix \(\mD\in\R^{N\times N}\) and \(a,b\geq0\), the
\emph{affine event} is
\begin{align}
 \mathcal E_{\mathrm D}(a,b)\coloneqq\bigl\{|D_{ij}-D^{(\mathrm{s})}_{ij}|
 \leq aD^{(\mathrm{s})}_{ij}+b\ \text{for all }1\leq i<j\leq N\bigr\},
 \label{eqn:affine-distance-error}
\end{align}
with \emph{multiplicative error} \(a\) and \emph{additive error} \(b\).
\end{definition}
Let
\(\Dsig\coloneqq\sup\{\|\vs-\vt\|:\vs,\vt\in\operatorname{supp}(\P_{\mathrm{s}})\}\)
be the diameter of the clean support. Since \(D^{(\mathrm{s})}_{ij}\leq\Dsig\),
on \(\mathcal E_{\mathrm D}(a,b)\) we have
\(\|\mD-\mD^{(\mathrm{s})}\|_{\max}\leq a\Dsig+b\).
The multiplicative error \(a\) must vanish, and the additive error \(b\)
limits the usable radii
(Section~\ref{sec:stability-of-data-driven-selection}). For the denoised
distances and every \(\nu>0\), the event
\(\mathcal E_{\mathrm D}(C_\nu\widetilde a_N,C_\nu\widetilde b_N)\) holds
with probability at least \(1-C_\nu N^{-\nu}\) for a constant \(C_\nu\)
(Theorem~\ref{thm:denoising-error-bound}).

\paragraph{Dependence on \(N\)}
The dimension \(d\) of Definition~\ref{def:local-mass-regularity} and all
constants are independent of \(N\); every other quantity, such as the clean
law, the dimensions \(p\) and \(m\), and the noise covariance, may depend
on \(N\).

\subsection{Cross-fitted projection denoising}
\label{sec:distance-denoising}
We bound the error of the denoised distances of
Algorithm~\ref{alg:three-fold-denoising} and show that it attains the
minimax rate when the noise floor \(\rho_N\) is bounded. Proofs are in
Appendix~\ref{app:denoising}.

\paragraph{Error rates}
Let \(\mGamma_\perp\coloneqq\mP_\perp\mGamma\mP_\perp\), with
\(\mP_\perp\coloneqq\mI_p-\mJ\mJ^{\sT}\), be the noise covariance outside
the signal subspace, and let
\begin{align}
 \widetilde a_N&\coloneqq\frac{1}{\theta_{\min}}\Bigl(\sigma^2+\frac{\|\mGamma_\perp\|_{\frob}}{\sqrt N}\Bigr),
 &\rho_N&\coloneqq\frac{\Tr\mGamma}{N\theta_{\min}}.
 \label{eqn:reconstruction-rate-parameters}
\end{align}
We call debiasing \emph{active} if \(\mathtt{debias}=\mathtt{true}\) and
\(p-m\geq20N\). Debiasing estimates the average noise variance outside the
signal subspace, \(\bar\sigma^2\coloneqq\Tr\mGamma_\perp/(p-m)\), and its
error depends on the anisotropy
\(\varepsilon_{\mathrm{iso}}\coloneqq\|\mGamma-\bar\sigma^2\mI_p\|/\sigma^2\).
The additive error is
\begin{align}
 \widetilde b_N\coloneqq\sigma\sqrt{1+\rho_N}\,\times
 \begin{cases}
  \sqrt m+\sqrt{\log N}, & \text{if debiasing is not active},\\[2pt]
  (m\log N)^{1/4}+\sqrt{\log N}+\sqrt m\,\bigl(\tfrac{N}{p-m}\bigr)^{1/4}
  +\sqrt{m\varepsilon_{\mathrm{iso}}},
   & \text{if debiasing is active}.
 \end{cases}
 \label{eqn:denoising-additive-error}
\end{align}
\begin{theorem}[Denoising error bound]
\label{thm:denoising-error-bound}
Suppose Assumptions~\ref{ass:noise-variance-structure}
and~\ref{ass:growing-covariance} hold, \(m\log N=o(N)\), and
\(\widetilde a_N\log N\to0\). Then the following hold for
Algorithm~\ref{alg:three-fold-denoising}.
\begin{enumerate}[label=(\alph*)]
\item With overwhelming probability, every fold selects the signal
dimension, \(\widetilde m_q=m\), for \(q=1,2,3\).
\item Suppose that \(\varepsilon_{\mathrm{iso}}\leq1/2\) if debiasing is
active. For every \(\nu>0\), there is a constant \(C_\nu\) such that, for
\(\mD=\widetilde\mD\),
\(\P\{\mathcal E_{\mathrm D}(C_\nu\widetilde a_N,C_\nu\widetilde b_N)\}
\geq1-C_\nu N^{-\nu}\) for all sufficiently large \(N\). In particular,
\begin{align}
 \|\widetilde\mD-\mD^{(\mathrm{s})}\|_{\max}\lesssim
 \widetilde a_N\Dsig+\widetilde b_N.
 \label{eqn:local-affine-bound}
\end{align}
\end{enumerate}
\end{theorem}
\paragraph{Dimension selection and the signal-to-noise ratio}
Since
\(\|\mGamma_\perp\|_{\frob}^2\leq\|\mGamma_\perp\|\Tr\mGamma_\perp\leq\sigma^2\Tr\mGamma\),
\begin{align}
 \widetilde a_N\leq\frac{\sigma^2}{\theta_{\min}}
   \Bigl(1+\sqrt{\frac{\Tr\mGamma}{\sigma^2N}}\Bigr),
\end{align}
and, with \(\Tr\mGamma=\Tr\mTheta/\SNR_N\), the hypothesis
\(\widetilde a_N\log N\to0\) holds if
\begin{align}
 \SNR_N\gg\frac{\Tr\mTheta}{m\theta_{\min}}\,
   m\log N\Bigl(\frac{\sigma^2}{\Tr\mGamma}+\frac{\sigma}{\sqrt{N\Tr\mGamma}}\Bigr).
 \label{eqn:selection-snr}
\end{align}
The right side tends to zero when \(\Tr\mTheta\asymp m\theta_{\min}\)
and \(\Tr\mGamma\gg\sigma^2m\log N\), since \(m\log N=o(N)\), so
\(\SNR_N\) may tend to zero.

\begin{remark}[Comparison with~\cite{ding2023learning}]
\label{rem:ding-ma-comparison}
Theorem~2 of~\cite{ding2023learning}, for signals with independent
coordinates and a full-dimensional density, requires
\(p\sigma^2=o(\Tr\mTheta)\), that is, \(\SNR_N\to\infty\), and the
duo-landmark joint embeddings of~\cite{ding2025kernel} and the multi-view
embeddings of~\cite{ding2026generalized} also assume that the signal
energy dominates the noise energy.
\end{remark}

\paragraph{Optimality}
Fix \(d\geq1\). For each \(N\), fix a matrix \(\mJ\in\R^{p\times m}\) with
\(\max\{2,d\}\leq m\leq p\), a noise level \(\sigma\geq0\), and a
\emph{covariance scale} \(\covscale_N>0\). Let \(\gP_N\) be the class of
laws of the observations in the setting of Section~\ref{sec:setting},
with isotropic Gaussian noise of level \(\sigma\), whose clean law
satisfies the following for some length scale \(L>0\) and some locality
radius \(\locrad>0\).
\begin{enumerate}[label=(\roman*)]
\item \(\rvmu_{\mathrm s}=\bzero\),
\(\tfrac12\covscale_N\leq\theta_{\min}\leq\theta_{\max}\leq2\covscale_N\),
and \(\|\clean\|\leq C_*\sqrt{m\covscale_N}\) almost surely, where
\(C_*\geq3\) is a fixed constant.
\item Assumption~\ref{ass:regular-clean-law} holds with dimension \(d\),
length scale \(L\), locality radius \(\locrad\), and constants
\((c_{\mathrm g},C_{\mathrm g})\).
\end{enumerate}
\begin{theorem}[Minimax lower bound]
\label{thm:coordinate-reconstruction-lower-bound}
There are constants \(0<c_0\leq1\leq C_0<\infty\), depending only on \(d\),
with the following property. Suppose
\(\covscale_N\gg\sigma^2\log N\), and that the constants in~(ii) satisfy
\(c_{\mathrm g}\leq c_0\) and \(C_{\mathrm g}\geq C_0\).
Then
\begin{align}
 \inf_{\widehat\mD}\sup_{P\in\gP_N}
    \P_P\bigl\{\|\widehat\mD-\mD^{(\mathrm{s})}\|_{\max}
       \geq\psi_N/64\bigr\}\longrightarrow1,
 \qquad
 \psi_N\coloneqq\sigma\bigl\{(m\log N)^{1/4}+\sqrt{\log N}\bigr\}.
 \label{eqn:coordinate-minimax-rate}
\end{align}
where the infimum is over all estimators \(\widehat\mD\) computed from
\(\vx_1,\ldots,\vx_N\).
\end{theorem}
By Theorem~\ref{thm:denoising-error-bound}, \(\widetilde b_N\asymp\psi_N\)
if \(\rho_N=O(1)\) and either \(m\lesssim\log N\), or debiasing is active and
\begin{align}
 \sqrt m\,\Bigl(\frac{N}{p-m}\Bigr)^{1/4}+\sqrt{m\varepsilon_{\mathrm{iso}}}
 \lesssim(m\log N)^{1/4}+\sqrt{\log N},
 \label{eqn:debiasing-gain-condition}
\end{align}
which holds, for instance, when \(p-m\gtrsim Nm/\log N\) and
\(\varepsilon_{\mathrm{iso}}\lesssim(\log N/m)^{1/2}\). Hence
Algorithm~\ref{alg:three-fold-denoising} attains the rate
\eqref{eqn:coordinate-minimax-rate} whenever
\(\widetilde a_N\Dsig\lesssim\widetilde b_N\), without knowing \(\sigma\)
or \(m\). The lower bound is for distance estimation over a class in which
\(L\) and \(\locrad\) are free, so its hard laws are regular only at radii
below the nearest-neighbor distance of the clean signals;
Remark~\ref{rem:minimax-communities} discusses the multi-community model.

\subsection{Radius selection and neighborhood stability}
\label{sec:stability-of-data-driven-selection}
We show that, on the affine event, the radius selected by
Algorithm~\ref{alg:radius-selection} is close to the radius selected from
the clean distances, and that the two radius graphs have nearly the same
neighborhoods. Proofs are in Appendix~\ref{app:radius-neighborhood-proofs}.

\paragraph{Input accuracy}
In
Sections~\ref{sec:stability-of-data-driven-selection}--\ref{sec:clustering-guarantees},
we fix deterministic sequences \(a_N\to0\) and \(b_N\geq0\) and a constant
\(C\geq1\), write
\(\mathcal E_{\mathrm D}\coloneqq\mathcal E_{\mathrm D}(Ca_N,Cb_N)\), and
allow constants to depend on \(C\). Until
Section~\ref{sec:below-separation}, which treats the recursion, the active
set is the full sample \([N]\).

\paragraph{Valid radii}
For \(k>0\), the \emph{local scale} is
\begin{align}
    h_N(k)\coloneqq L\left(k/N\right)^{1/d},
    \label{eqn:local-knn-scale}
\end{align}
the radius of a clean ball of mass of order \(k/N\) when
\(h_N(k)\leq\locrad\). We call \(h_N(\log N)\) the \emph{sampling
resolution}, the order of the radius at which a random geometric graph on
\(N\) points in dimension \(d\) becomes connected~\cite{penrose2003random}.
\begin{definition}[Valid radius window]
\label{def:valid-radii}
For an error level \(\zeta\geq0\), the
\emph{valid radius window} is
\begin{align}
 \radwin_N(\zeta)\coloneqq\Bigl[\xi_N\max\bigl\{\zeta,\ h_N(\log N)\bigr\},\
 {\locrad}/{\xi_N}\Bigr],
 \label{eqn:radius-window}
\end{align}
where we choose the \emph{slack} \(\xi_N\coloneqq\max\{1,\log\log N\}\).
\end{definition}
The upper end keeps valid radii in the range of
Assumption~\ref{ass:regular-clean-law}.

\paragraph{Admissible neighbor counts}
The \emph{radius band} at neighbor count \(k\) is
\begin{align}
 \radband_N(k)\coloneqq\bigl[C_{\mathrm h}^{-1/d}h_N(k),\
 C_{\mathrm h}^{1/d}h_N(k)\bigr], \qquad C_{\mathrm h}\coloneqq\max\Bigl\{4\mathrm e\,C_{\mathrm g},\,\, {16}/{c_{\mathrm g}}\Bigr\},
 \label{eqn:radius-band}
\end{align}
where \(\mathrm e\) is Euler's number. The band contains the clean selected
radius (Proposition~\ref{prop:radius-stability}).
\begin{definition}[Admissible neighbor counts]
\label{def:admissible-ranks}
For \(\zeta\geq0\), the \emph{admissible neighbor counts} are
\begin{align}
 \admset_N(\zeta)\coloneqq\bigl\{k\in\{1,\ldots,N-1\}:\
 \radband_N(k)\subseteq\radwin_N(\zeta)\bigr\}.
 \label{eqn:admissible-ranks}
\end{align}
\end{definition}
Since \((h_N(k)/L)^d=k/N\), a count \(k\in\{1,\ldots,N-1\}\) is admissible
if and only if
\begin{align}
 C_{\mathrm h}N\biggl(\frac{\xi_N\max\{\zeta,\,h_N(\log N)\}}{L}\biggr)^d
 \leq k\leq
 \frac{N}{C_{\mathrm h}}\biggl(\frac{\locrad}{\xi_NL}\biggr)^d,
 \label{eqn:admissible-range}
\end{align}
so the admissible counts are consecutive integers, and
\(\admset_N(\zeta)\subseteq\admset_N(0)\).

\begin{proposition}[Existence of admissible neighbor counts]
\label{prop:admissible-ranks}
Suppose Assumption~\ref{ass:regular-clean-law} holds, and let
\(\zeta_N\geq0\). Then \(\admset_N(\zeta_N)\) is nonempty for all
sufficiently large \(N\) if
\begin{align}
 \xi_N^2\max\bigl\{\zeta_N,\ h_N(\log N)\bigr\}=o(\locrad),
 \label{eqn:admissible-existence}
\end{align}
and only if
\(\xi_N^2\max\{\zeta_N,h_N(\log N)\}\leq C_{\mathrm h}^{-2/d}\locrad\).
\end{proposition}
Condition \eqref{eqn:admissible-existence} asks the window
\(\radwin_N(\zeta_N)\) to be wide: the ratio of its ends,
\(\locrad/\{\xi_N^2\max\{\zeta_N,h_N(\log N)\}\}\), must exceed the ratio
\(C_{\mathrm h}^{2/d}\) of the ends of a radius band.

\paragraph{Radius stability}
Let \(\admset_N\coloneqq\admset_N(b_N)\) be the admissible counts for the
additive error of \(\mathcal E_{\mathrm D}\), which we assume nonempty
(Proposition~\ref{prop:admissible-ranks}).
Let \(r(k,\omega)\) be the radius \eqref{eqn:selected-radius} selected by
Algorithm~\ref{alg:radius-selection} from \(\mD\), with the active set
\(\sI=[N]\) suppressed: the smallest radius at which at least
\(\lceil\omega N\rceil\) observations have at least \(k\) neighbors. Let
\(r_{\mathrm s}(k,\omega)\) be the radius selected from
\(\mD^{(\mathrm s)}\).
\begin{proposition}[Stability of the selected radius]
\label{prop:radius-stability}
Suppose Assumption~\ref{ass:regular-clean-law} holds. With overwhelming
probability, for every \(k\in\admset_N\) and every \(\omega\in(0,1)\),
\begin{align}
 r_{\mathrm{s}}(k,\omega)\in\radband_N(k),
 \label{eqn:clean-percentile-radius-scale}
\end{align}
and, with overwhelming probability on \(\mathcal E_{\mathrm D}\),
\begin{align}
 \left|\frac{r(k,\omega)}{r_{\mathrm{s}}(k,\omega)}-1\right|
 \lesssim a_N+\frac{b_N}{h_N(k)}.
 \label{eqn:radius-relative-stability}
\end{align}
\end{proposition}

\paragraph{Neighborhood stability}
For a radius \(r\geq0\), let
\begin{align}
 \gN_i(r)\coloneqq\{j\in[N]\setminus\{i\}:D_{ij}\leq r\}
 \label{eqn:clean-noisy-neighborhoods}
\end{align}
be the neighborhood of observation \(i\), and let
\(\gN_i^{(\mathrm s)}(r)\) be the one computed from \(\mD^{(\mathrm s)}\).
We compare the neighborhoods at the two selected radii
\(r=r(k,\omega)\) and \(r_{\mathrm s}=r_{\mathrm s}(k,\omega)\) by the
average Jaccard similarity
\begin{align}
 T_N(k,\omega)\coloneqq\frac1N\sum_{i=1}^N
   J\bigl(\gN_i(r),\gN_i^{(\mathrm s)}(r_{\mathrm s})\bigr),
 \qquad
 J(A,B)\coloneqq\frac{|A\cap B|}{|A\cup B|},\quad J(\emptyset,\emptyset)\coloneqq1.
 \label{eqn:average-jaccard}
\end{align}
Since \(1-J(A,B)=|A\triangle B|/|A\cup B|\), the dissimilarity
\(1-T_N(k,\omega)\) is the average fraction of neighbors that differ
between the two graphs. The \emph{local error rate} is
\begin{align}
 \delta_N(k)\coloneqq a_N+\frac{b_N}{h_N(k)}+\frac{\log N}{k},
 \qquad k\in\admset_N,
 \label{eqn:local-error-rate}
\end{align}
the sum of the multiplicative error, the additive error relative to the
local scale, and the sampling error. It decreases in \(k\), so
\(\max_{k\in\admset_N}\delta_N(k)=\delta_N(\min\admset_N)\to0\) by
\eqref{eqn:admissible-range}.
The dissimilarity is at most of the order of the local error rate.
\begin{proposition}[Neighborhood preservation]
\label{prop:neighborhood-preservation}
Suppose Assumption~\ref{ass:regular-clean-law} holds. With overwhelming
probability on \(\mathcal E_{\mathrm D}\), for
every \(k\in\admset_N\) and every \(\omega\in(0,1)\),
\begin{align}
 1-T_N(k,\omega)
 &\lesssim\delta_N(k).
 \label{eqn:uniform-neighborhood-preservation-rate}
\end{align}
\end{proposition}

\subsection{Convergence of the spectral embedding}
\label{sec:spectral-convergence}
The spectral embedding of Algorithm~\ref{alg:RGG-spectral-embedding}
computed from the input distances converges to the one computed from the
clean distances (Theorem~\ref{thm:spectral-stability}), and both converge
to the eigenpairs of a population integral operator
(Theorem~\ref{thm:population-eigenfunction-convergence}). Proofs are in
Appendix~\ref{app:spectral-convergence-proofs}.

\paragraph{Kernel functions and kernel matrices}
\begin{definition}[Regular kernel functions]
\label{def:admissible-kernels}
A kernel function \(\kf:[0,\infty)\to[0,\infty)\) is \emph{regular} if
\begin{align}
 c_{\kf}\leq\kf(t)\leq M_{\kf},
 &\qquad t\in[0,1],
 \label{eqn:profile-positive-bounds}\\
 |\kf(t)-\kf(t')|\leq\Lambda_{\kf}|t^2-t'^2|,
 &\qquad t,t'\in[0,1],
 \label{eqn:profile-lipschitz-bound}
\end{align}
for some constants \(c_{\kf}>0\), \(M_{\kf}<\infty\), and \(\Lambda_{\kf}<\infty\).
\end{definition}
Examples include \(\kf\equiv1\), the Gaussian \(\kf(t)=\exp(-t^2)\), and the rational quadratic \(\kf(t)=(1+t^2/c)^{-a}\) with \(a,c>0\).

For \(k\in\admset_N\) and \(\omega\in(0,1)\), let \(\mK\) be the kernel
matrix \eqref{eqn:RGG-kernel-def} of the input \(\mD\) at its selected
radius \(r=r(k,\omega)\), and \(\mK^{(\mathrm s)}\) that of
\(\mD^{(\mathrm s)}\) at \(r_{\mathrm s}=r_{\mathrm s}(k,\omega)\); we
suppress \(k\) and \(\omega\). Let \((\lambda_j,\vv_j)\) and
\((\lambda_j^{(\mathrm s)},\vv_j^{(\mathrm s)})\) denote their eigenpairs,
with eigenvalues in decreasing order and orthonormal eigenvectors.

\paragraph{Population operator}
For a radius \(r>0\), define
\begin{align}
 (\IOK_r f)(\vs)\coloneqq\int\indi{\|\vs-\vt\|\leq r}\,
       \kf\bigl(\|\vs-\vt\|/r\bigr)f(\vt)\,
       \P_{\mathrm{s}}(\odif{\vt}),
 \qquad f\in L^2(\P_{\mathrm{s}}).
 \label{eqn:convolution-operator}
\end{align}
The operator is compact and self-adjoint. Let
\(\gamma_{1,r}\geq\gamma_{2,r}\geq\cdots\) list its positive
eigenvalues with multiplicity, padded by zeros, and for each positive
\(\gamma_{j,r}\) let \(\phi_{j,r}\) be a corresponding unit
eigenfunction. The \emph{population eigengap} of the \(j\)-th eigenvalue
at \(r\) is
\begin{align}
 \operatorname{gap}_j\bigl(\IOK_r\bigr)\coloneqq
 \inf_{j'\ne j}\bigl|\gamma_{j,r}-\gamma_{j',r}\bigr|.
 \label{eqn:population-eigengap}
\end{align}

\paragraph{Convergence to the clean embedding}
Besides the local error rate \(\delta_N(k)\), the bounds involve the
\emph{population error rate}
\begin{align}
 \beta_N(k)\coloneqq\Bigl(\frac{\log N}{k}\Bigr)^{1/(d+2)},
 \qquad k\in\admset_N,
 \label{eqn:population-error-rate}
\end{align}
at which the clean kernel approaches the population operator. It dominates
the sampling term \(\log N/k\) of \(\delta_N(k)\), and
\(\max_{k\in\admset_N}\beta_N(k)=\beta_N(\min\admset_N)\to0\).
In the conclusions of the two theorems below, the population quantities
are evaluated at the clean selected radius, and we write
\(\IOK\coloneqq\IOK_{r_{\mathrm s}}\),
\(\gamma_j\coloneqq\gamma_{j,r_{\mathrm s}}\), and
\(\phi_j\coloneqq\phi_{j,r_{\mathrm s}}\).
\begin{theorem}[Convergence to the clean embedding]
\label{thm:spectral-stability}
Suppose Assumption~\ref{ass:regular-clean-law} holds, and let \(\kf\) be
a regular kernel function.
\begin{enumerate}[label=(\alph*)]
\item \emph{(Eigenvalue convergence)} With overwhelming probability on
\(\mathcal E_{\mathrm D}\), for
every \(k\in\admset_N\) and every \(\omega\in(0,1)\),
\begin{align}
 \max_{j}|\lambda_j-\lambda_j^{(\mathrm s)}|
 \leq\|\mK-\mK^{(\mathrm s)}\|
 \lesssim\frac kN\,\delta_N(k).
 \label{eqn:array-kernel-transfer}
\end{align}

\item \emph{(Eigenvector and embedding convergence)} Fix \(j\geq1\) and a
sequence \(k=k_N\in\admset_N\), and suppose that, uniformly over
\(r\in\radband_N(k)\),
\begin{align}
 \frac kN\bigl\{\delta_N(k)+\beta_N(k)\bigr\}
 =o\bigl(\operatorname{gap}_j(\IOK_r)\bigr).
 \label{eqn:spectral-gap-condition}
\end{align}
Then, with overwhelming probability on \(\mathcal E_{\mathrm D}\), for every \(\omega\in(0,1)\) there is
a sign \(\chi_j\in\{-1,1\}\) such that
\begin{align}
 \|\vv_j-\chi_j\vv_j^{(\mathrm s)}\|
 &\lesssim\frac kN\,\frac{\delta_N(k)}{\operatorname{gap}_j(\IOK)},
 \label{eqn:reconstructed-singleton-eigenvector}\\
 \|\lambda_j\vv_j-\chi_j\lambda_j^{(\mathrm s)}\vv_j^{(\mathrm s)}\|
 &\lesssim\Bigl(\frac kN\Bigr)^2
   \frac{\delta_N(k)}{\operatorname{gap}_j(\IOK)}.
 \label{eqn:reconstructed-weighted-eigenvector}
\end{align}
\end{enumerate}
\end{theorem}
Both bounds are relative to the eigenvalue scale \(k/N\). The gap
condition \eqref{eqn:spectral-gap-condition} is imposed over the band
\(\radband_N(k)\), which contains the random radius \(r_{\mathrm s}\), and
it contains \(\beta_N\) so that the clean kernel inherits the population
eigengap. Applying
part~(b) to each \(j\) in a block \(\sJ\) bounds the embedding
\(\mZ_{\sJ}\) of Algorithm~\ref{alg:RGG-spectral-embedding} up to column
signs.

\paragraph{Population limit}
Let \(\lambda_j^+\coloneqq\max\{\lambda_j,0\}\) for \(j\leq N\) and
\(\lambda_j^+\coloneqq0\) for \(j>N\). To compare vectors with functions, we
use a linear isometry \(U_N:\R^N\to L^2(\P_{\mathrm{s}})\), determined by
the clean sample and \(k\), that maps a vector to a step function on a partition of
the clean support (Lemma~\ref{lem:common-space-spectral-lift} in
Appendix~\ref{app:spectral-convergence-proofs}).
\begin{theorem}[Convergence to the population operator]
\label{thm:population-eigenfunction-convergence}
Suppose the assumptions of Theorem~\ref{thm:spectral-stability} hold, and
let \(k=k_N\in\admset_N\). With overwhelming probability on
\(\mathcal E_{\mathrm D}\), the following hold for every \(\omega\in(0,1)\), with
\(r_{\mathrm s}=r_{\mathrm s}(k,\omega)\).
\begin{enumerate}[label=(\alph*)]
\item \emph{(Eigenvalue convergence)}
\begin{align}
 \sup_{j\geq1}\bigl|\lambda_j^+-\gamma_j\bigr|
 \lesssim\frac kN\bigl\{\delta_N(k)+\beta_N(k)\bigr\}.
 \label{eqn:reconstructed-population-eigenvalues}
\end{align}
\item \emph{(Eigenfunction convergence)} Fix \(j\geq1\), and suppose that
\eqref{eqn:spectral-gap-condition} holds uniformly over
\(r\in\radband_N(k)\). Then there is a sign \(\chi_j'\in\{-1,1\}\) such
that
\begin{align}
 \|U_N\vv_j-\chi_j'\phi_j\|_{L^2(\P_{\mathrm{s}})}
 \lesssim\frac kN\,\frac{\delta_N(k)+\beta_N(k)}{\operatorname{gap}_j(\IOK)}.
 \label{eqn:affine-population-lift}
\end{align}
\end{enumerate}
\end{theorem}
\begin{remark}[Size of population eigengaps]
\label{rem:circle-leading-mode-limitation}
When the population eigengap is small compared with the eigenvalue scale
\((r/L)^d\), the gap condition needs a large neighbor count. For two separated uniform circles of equal radius
\(L\), with unequal weights and \(\kf\equiv1\), the leading gap is
\(\operatorname{gap}_1(\IOK_r)\asymp(r/L)^3\) for small \(r/L\), and
\eqref{eqn:spectral-gap-condition} with \(a_N=b_N=0\) requires
\(k\gg N^{6/7}(\log N)^{1/7}\)
(Appendix~\ref{app:eigengap-remark-proof}).
\end{remark}

\section{Recovery guarantees under the multi-community model}\label{sec:clustering-guarantees}
We prove that Algorithm~\ref{alg:RGG-clustering} recovers the
communities. Section~\ref{sec:multi-community-manifolds} states the model,
Section~\ref{sec:below-separation} applies the results of
Section~\ref{sec:embedding-guarantees} on the active sets of the
recursion, Section~\ref{sec:recursive-clustering-guarantee} proves
consistency for every input that is close enough to the clean distances,
and Section~\ref{sec:denoising-necessity} gives a model in which the
denoised distances recover every label and the raw distances fail.

\subsection{Multi-community manifolds}
\label{sec:multi-community-manifolds}
The clean signal \(\clean_j\) is now generated in three steps: we first
draw its community label, then draw a point on the reference manifold of
that community, and finally embed this point into the signal subspace of
Section~\ref{sec:setting}.

\paragraph{Community labels}
There are \(B\geq1\) communities. The \(j\)-th observation is assigned a label
\(\rY_j\in[B]\) with \(\P(\rY_j=b)=\alpha_b\). We assume that no community is vanishingly small:
\begin{align}
 \alpha_{\min}\coloneqq\min_{b\in[B]}\alpha_b\geq c_{\alpha}
 \label{eqn:min-mixture-weight}
\end{align}
for some constant \(c_\alpha>0\). We write
\(\vy=(y_1,\ldots,y_N)^{\sT}\) for the realized labels and
\(\gC_b=\{j\in[N]:y_j=b\}\) for the \(b\)-th community.

\paragraph{Reference manifolds}
Following \cite{wang2015multi-manifold}, each community is described by a Riemannian manifold in \(\R^m\) and a law on it.
\begin{definition}[Multi-community Riemannian manifolds]
\label{def:reference-riemannian-manifolds}
Let \(m,d\) be integers with \(1\leq d\leq m\). For each \(b\in[B]\), let
\(\manifold_b\subset\R^m\) be a connected compact \(d\)-dimensional
\(C^2\)-embedded submanifold, possibly with boundary. The Riemannian
manifold \((\manifold_b, \metric)\) carries the metric induced by the
Euclidean inner product:
\[
 \metric_{\vu}(\xi,\eta)\coloneqq\langle\xi,\eta\rangle_{\R^m},
 \qquad\xi,\eta\in T_{\vu}\manifold_b,
\]
where \(T_{\vu}\manifold_b\) is the tangent space at \(\vu\in\manifold_b\),
with intrinsic dimension \(\dim(T_{\vu}\manifold_b)=d\). Let
\(\diam(\manifold_b)\) denote the diameter of \(\manifold_b\), with
\(\diam(A)\coloneqq\sup_{\vu,\vv\in A}\|\vu-\vv\|\) for a nonempty set
\(A\).
\end{definition}
Each reference manifold \(\manifold_b\) carries a law \(\mathds{Q}_b\), the
\emph{reference law}, which we take centered,
\(\int_{\manifold_b}\vu\,\mathds{Q}_b(\odif{\vu})=\bzero\), since the
center \(\rvmu_b\) of the embedding below absorbs any translation.

\paragraph{Embedding into the ambient space}
Let \(L>0\) be a \emph{dilation} common to all communities, and let
\(\rvmu_b\) be a point of the signal subspace, the \emph{center} of
community \(b\). The embedding
\begin{align}
    \iota_b(\vu)
    \coloneqq&\rvmu_b+L\mJ\vu, \qquad \vu\in\manifold_b,
    \label{eqn:homothetic-manifold-support}
\end{align}
places \(\manifold_b\) in the signal subspace and scales all distances by
\(L\).

\paragraph{Clean signals}
Each clean signal is the embedded point of its community:
\begin{align}
    (\vu_j\mid\rY_j=b)&\sim\mathds{Q}_b,
    & \clean_j&=\iota_{\rY_j}(\vu_j).
    \label{eqn:data-generation-clean-signal}
\end{align}
The pairs \(\{(\rY_j,\clean_j)\}_{j=1}^N\) are i.i.d.\ and independent of
the noise of Assumption~\ref{ass:noise-variance-structure}. For a nonempty
set \(\sB\subseteq[B]\), the clean signals with labels in \(\sB\) follow
the law
\begin{align}
    \P_{\mathrm{s}}^{(\sB)}
       \coloneqq\mathrm{Law}(\clean_j\mid\rY_j\in\sB)
       =\frac{1}{\alpha_\sB}\sum_{b\in\sB}\,\alpha_b\,(\iota_b)_\#\mathds{Q}_b,
    \qquad
    \alpha_\sB\coloneqq\sum_{b\in\sB}\alpha_b.
    \label{eqn:community-clean-law}
\end{align}
The law \(\P_{\mathrm{s}}^{(b)}\coloneqq\P_{\mathrm{s}}^{(\{b\})}\) of
community \(b\) has support \(\iota_b(\manifold_b)\) and mean \(\rvmu_b\).
The law \(\P_{\mathrm{s}}\coloneqq\P_{\mathrm{s}}^{([B])}\) of all clean
signals is the clean law of Section~\ref{sec:setting}: its support
\(\bigcup_{b=1}^B\iota_b(\manifold_b)\) is compact and lies in the signal
subspace, and its mean is
\(\rvmu_{\mathrm{s}}=\sum_{b=1}^B\alpha_b\rvmu_b\), so the centering of
Assumption~\ref{ass:growing-covariance} asks
\(\sum_{b=1}^B\alpha_b\rvmu_b=\bzero\).

\paragraph{Separation between communities}
The \emph{separation} is the smallest distance between two community
supports,
\begin{align}
    \Delta
    \coloneqq
    \min_{1\leq b<b'\leq B}\,
    \inf_{\substack{\vs\in\iota_b(\manifold_b)\\
                    \vt\in\iota_{b'}(\manifold_{b'})}}
    \|\vs-\vt\|,
    \label{eqn:manifold-separation}
\end{align}
with \(\Delta=+\infty\) when \(B=1\); the supports are disjoint when
\(\Delta>0\). Communities with a common center can
be separated in this sense, as the concentric circles of
Section~\ref{sec:denoising-necessity}.

\paragraph{Local mass regularity of the communities}
Assumption~\ref{ass:regular-clean-law} is replaced by one for each
community.
\begin{assumption}[Regular communities]
\label{ass:local-mass-regularity}
There is a \emph{community locality radius} \(R_{\mathrm g}>0\) such that,
for every \(b\in[B]\), the community law \(\P_{\mathrm{s}}^{(b)}\) has
local mass regularity (Definition~\ref{def:local-mass-regularity}) with
dimension \(d\), length scale \(L\), locality radius \(R_{\mathrm g}\),
and constants \((c_{\mathrm g},C_{\mathrm g})\).
\end{assumption}
\begin{assumption}[Locality radius and separation]
\label{ass:locality-separation}
There are \(B\geq2\) communities, and \(R_{\mathrm g}\geq\Delta\).
\end{assumption}
The lower bound of \eqref{eqn:uniform-clean-small-ball} at
\(r=R_{\mathrm g}\) gives \(R_{\mathrm g}\leq c_{\mathrm g}^{-1/d}L\), so
the separation is at most of order \(L\). Below, conditions on the
locality radius involve \(\Delta\) alone.

\subsection{Separation condition and stability on the active sets}
\label{sec:below-separation}
We apply Section~\ref{sec:embedding-guarantees} to the multi-community
model, at radii below half the separation and on every active set of the
recursion. Proofs are in Appendices~\ref{app:mixture-regularity-proof}
and~\ref{app:union-stability-proof}.

\paragraph{Regularity below the separation}
Below half the separation, a ball or a shell around a support point meets
only one community.
\begin{lemma}[Regularity of unions of communities]
\label{lem:mixture-regularity}
Suppose Assumption~\ref{ass:local-mass-regularity} and
\eqref{eqn:min-mixture-weight} hold, and \(\Delta>0\). Then, for every
nonempty \(\sB\subseteq[B]\), the law \(\P_{\mathrm{s}}^{(\sB)}\) of
\eqref{eqn:community-clean-law} has local mass regularity with dimension
\(d\), length scale \(L\), locality radius
\(\min\{R_{\mathrm g},\Delta/2\}\), and constants
\((c_\alpha c_{\mathrm g},C_{\mathrm g})\). In particular, the clean law
\(\P_{\mathrm{s}}\) satisfies Assumption~\ref{ass:regular-clean-law} with
\(\locrad=\min\{R_{\mathrm g},\Delta/2\}\), which equals \(\Delta/2\) under
Assumption~\ref{ass:locality-separation}.
\end{lemma}

\paragraph{Valid radii and the separation condition}
In the rest of Section~\ref{sec:clustering-guarantees}, the definitions of
Section~\ref{sec:stability-of-data-driven-selection} are used with
\(\locrad=\Delta/2\) and the constants
\((c_\alpha c_{\mathrm g},C_{\mathrm g})\) of
Lemma~\ref{lem:mixture-regularity}:
\begin{align}
 \radwin_N(\zeta)=\Bigl[\xi_N\max\bigl\{\zeta,\ h_N(\log N)\bigr\},\
 \frac{\Delta}{2\xi_N}\Bigr],
 \qquad
 C_{\mathrm h}=\max\Bigl\{4\mathrm e\,C_{\mathrm g},\
 \frac{16}{c_{\mathrm g}c_\alpha}\Bigr\}.
 \label{eqn:band-constants-communities}
\end{align}
A valid radius is thus below \(\Delta/2\), so the radius graph has no
edges between communities, and above the sampling resolution, so each
community is connected in it.

For the errors of \(\mathcal E_{\mathrm D}\), condition
\eqref{eqn:admissible-existence} becomes the \emph{separation condition}
\begin{align}
 \xi_N^2\max\bigl\{b_N,\ h_N(\log N)\bigr\}=o(\Delta),
 \label{eqn:separation-condition}
\end{align}
under which \(\admset_N\) is nonempty for all sufficiently large \(N\)
(Proposition~\ref{prop:admissible-ranks}). Up to constants and powers of
\(\xi_N\), the admissible counts run from
\(\max\{\log N,\ N(b_N/L)^d\}\) to \(N(\Delta/L)^d\).

\paragraph{Active sets of the recursion}
Algorithm~\ref{alg:RGG-clustering} applies
Algorithms~\ref{alg:radius-selection}
and~\ref{alg:RGG-spectral-embedding} to the observations that remain
after each extraction. When the extractions are correct, these form a
union of communities \(\sI_\sB\coloneqq\bigcup_{b\in \sB}\gC_b\), for a
nonempty \(\sB\subseteq[B]\). For an active set \(\sI\subseteq[N]\), we write \(r(\sI;k,\omega)\),
\(r_{\mathrm s}(\sI;k,\omega)\), \(T_N(\sI;k,\omega)\), \(\mK(\sI)\), and
\(\mK^{(\mathrm s)}(\sI)\) for the selected radii, the Jaccard similarity,
and the kernel matrices of Section~\ref{sec:embedding-guarantees} computed
on \(\sI\) in place of \([N]\); the average in \eqref{eqn:average-jaccard}
is then over \(i\in\sI\). For a nonempty \(\sB\subseteq[B]\), we
write \(\IOK_r^{(\sB)}\), \(\gamma_{j,r}^{(\sB)}\), and
\(\phi_{j,r}^{(\sB)}\) for the operator \eqref{eqn:convolution-operator}
and its eigenpairs with \(\P_{\mathrm{s}}^{(\sB)}\) in place of
\(\P_{\mathrm{s}}\), and \(\IOK_r^{(b)}=\IOK_r^{(\{b\})}\).
\begin{proposition}[Stability on the active sets]
\label{prop:union-stability}
Suppose Assumptions~\ref{ass:local-mass-regularity}
and~\ref{ass:locality-separation} hold, and let \(\kf\) be a regular kernel
function. With overwhelming probability on \(\mathcal E_{\mathrm D}\), the
following hold for every \(k\in\admset_N\), every nonempty
\(\sB\subseteq[B]\), and every \(\omega\in(0,1)\), with the active set
\(\sI=\sI_\sB\) in place of \([N]\).
\begin{enumerate}[label=(\alph*)]
\item \emph{(Radius and neighborhoods)} The selected radii
\(r(\sI_\sB;k,\omega)\) and \(r_{\mathrm s}(\sI_\sB;k,\omega)\) satisfy
\eqref{eqn:clean-percentile-radius-scale} and
\eqref{eqn:radius-relative-stability}, and
\(1-T_N(\sI_\sB;k,\omega)\lesssim\delta_N(k)\).
\item \emph{(Embedding)} Part~(a) of Theorem~\ref{thm:spectral-stability}
holds for the kernel matrices \(\mK(\sI_\sB)\) and
\(\mK^{(\mathrm s)}(\sI_\sB)\).
\item \emph{(Eigenvectors)} Fix \(j\geq1\), a sequence
\(k=k_N\in\admset_N\), and a nonempty \(\sB\subseteq[B]\) for which the
gap condition \eqref{eqn:spectral-gap-condition} holds for
\(\IOK_r^{(\sB)}\) uniformly over \(r\in\radband_N(k)\). Then part~(b) of
Theorem~\ref{thm:spectral-stability} holds on \(\sI_\sB\), with
\(\IOK_r^{(\sB)}\) in place of \(\IOK_r\).
\end{enumerate}
\end{proposition}
For \(r<\Delta\),
the kernel of \eqref{eqn:convolution-operator} vanishes between distinct
communities, so
\begin{align}
 \IOK_r^{(\sB)}=\bigoplus_{b\in \sB}\frac{\alpha_b}{\alpha_\sB}\IOK_r^{(b)}.
 \label{eqn:operator-direct-sum}
\end{align}
The spectrum of \(\IOK_r^{(\sB)}\) is thus the union of the weighted
community spectra. The gap condition \eqref{eqn:spectral-gap-condition}
asks \(\gamma_{j,r}^{(\sB)}\) to be simple and separated from the rest of
this union and from zero.

\paragraph{Separation-to-noise condition}
For the denoised distances, \(b_N=\widetilde b_N\) by
Theorem~\ref{thm:denoising-error-bound}(b), and the separation condition
\eqref{eqn:separation-condition} has two parts: the
\emph{sampling condition}
\begin{align}
 \xi_N^2h_N(\log N)=o(\Delta),
 \label{eqn:sampling-condition}
\end{align}
and the \emph{separation-to-noise condition}
\(\xi_N^2\widetilde b_N=o(\Delta)\). Since
\(\sqrt m+\sqrt{\log N}\asymp\sqrt{m+\log N}\), without debiasing the
latter reads
\begin{align}
 {\Delta}/{\sigma}\gg\xi_N^2\sqrt{(1+\rho_N)(m+\log N)},
 \label{eqn:denoised-separation-noise}
\end{align}
When \(\rho_N=O(1)\), this condition involves the noise through its
largest variance \(\sigma^2\) and the \(m\) fitted directions, whereas
\(\SNR_N\) compares total energies. Since two independent clean signals
lie in different communities with probability at least \(\alpha_{\min}\),
and then at distance at least \(\Delta\),
\(\alpha_{\min}\Delta^2\leq2\Tr\mTheta=2\SNR_N\Tr\mGamma\). Hence
\eqref{eqn:denoised-separation-noise} implies the
\emph{signal-to-noise condition}
\begin{align}
 \SNR_N\gg\frac{1}{\Tr(\mGamma)} \xi_N^4(m+\log N)\sigma^2,
 \label{eqn:snr-condition}
\end{align}
and is equivalent to it when \(\Delta^2\asymp\Tr\mTheta\) and
\(\rho_N=O(1)\).
\begin{remark}[Lower bound within the multi-community class]
\label{rem:minimax-communities}
The lower bound of Theorem~\ref{thm:coordinate-reconstruction-lower-bound}
holds within the model of Section~\ref{sec:multi-community-manifolds}. For
fixed \(B\geq1\), \(d\), and weights \(\alpha_b\), it holds over the laws of
this model that satisfy condition~(i) of the class \(\gP_N\) and
Assumption~\ref{ass:local-mass-regularity}, and whose reference laws have
bounded Lipschitz densities
(Proposition~\ref{prop:minimax-communities} in
Appendix~\ref{app:minimax-proof}). When \(B\geq2\), the laws used in the
proof have a separation below the sampling resolution \(h_N(\log N)\), so
they violate the sampling condition \eqref{eqn:sampling-condition}, and the
necessity of the separation-to-noise condition for recovery remains open.
\end{remark}

\subsection{Consistency of the recursive algorithm}
\label{sec:recursive-clustering-guarantee}
Under the separation condition \eqref{eqn:separation-condition},
Algorithm~\ref{alg:RGG-clustering} recovers every label from any input
on \(\mathcal E_{\mathrm D}\)
(Theorem~\ref{thm:strong-consistency}). Its mismatch ratio vanishes when
the local mass bounds hold only on community cores that carry all but a
small fraction of the mass (Theorem~\ref{thm:weak-consistency}). Corollary~\ref{cor:denoised-recovery}
applies both results to the denoised distances. Proofs are in
Appendix~\ref{app:recursive-recovery-proofs}.

\paragraph{Algorithm settings and error measure}
Let \(\widehat\vy\) denote the labels returned by
Algorithm~\ref{alg:RGG-clustering} from an input matrix \(\mD\), run in
exact arithmetic with the true \(B\geq2\), a neighbor count \(k=k_N\), a
regular kernel function \(\kf\)
(Definition~\ref{def:admissible-kernels}), quantile levels
\(\omega_\ell\in(0,1)\), and thresholds \(\tau_\ell=0\). We measure the
label error up to a common permutation by
\begin{align}
 \mismatch_N(\vy,\widehat{\vy})
 &\coloneqq\frac1N\min_{\pi\in\mathfrak S_B}
       \sum_{i=1}^N\indi{\pi(\widehat{\evy}_i)\ne y_i}.
 \label{eqn:mismatch-ratio}
\end{align}
With \(\tau_\ell=0\), each stage extracts the connected component of its
graph with the largest Perron root, the one containing the earliest index
on a tie (Lemma~\ref{lem:selected-community-extraction}). Recovery thus
reduces to the components of every active graph separating the
communities, or their cores.
\begin{theorem}[Strong consistency]
\label{thm:strong-consistency}
Suppose Assumptions~\ref{ass:local-mass-regularity}
and~\ref{ass:locality-separation} hold and the separation condition
\eqref{eqn:separation-condition} holds. Then \(\admset_N\) is nonempty for all sufficiently large \(N\), and for
every sequence \(k=k_N\in\admset_N\), with overwhelming probability on
\(\mathcal E_{\mathrm D}\),
Algorithm~\ref{alg:RGG-clustering} completes and
\(\mismatch_N(\vy,\widehat\vy)=0\).
\end{theorem}

\paragraph{Community cores}
The local mass bounds can fail on part of a support, for instance in
sparse tails, and still hold on a large subset of it.
\begin{assumption}[Regular cores]
\label{ass:regular-cores}
There are \(B\geq2\) communities and, for each \(b\in[B]\), a compact
connected \emph{core} \(\manifold_b^\circ\subseteq\manifold_b\) with
\(\mathds{Q}_b(\manifold_b^\circ)>0\) such that the law of
\(\clean\) given \(\{\rY=b,\ \vu\in\manifold_b^\circ\}\) satisfies the ball
bounds \eqref{eqn:uniform-clean-small-ball} at every center in
\(\iota_b(\manifold_b^\circ)\), with dimension \(d\), length scale \(L\), constants
\((c_{\mathrm g},C_{\mathrm g})\), and a locality radius
\(R_{\mathrm g}\geq\Delta\). The mass outside the cores is at most
\(\alpha_{\mathrm{out}}\), that is,
\begin{align}
 \P\{\vu_i\notin\manifold_{\rY_i}^\circ\}\leq\alpha_{\mathrm{out}}.
 \label{eqn:core-conditions}
\end{align}
\end{assumption}
We call \(\gO_N\coloneqq\{i\in[N]:\vu_i\notin\manifold_{y_i}^\circ\}\) the
\emph{exceptional observations}. Under
Assumption~\ref{ass:regular-cores}, the admissible neighbor counts are
formed with the constants of the cores and the separation \(\Delta\) of
the full supports, and they must also exceed the expected number of
exceptional observations by the slack:
\begin{align}
 \admset_N^\circ\coloneqq\{k\in\admset_N:\ k\geq\xi_NN\alpha_{\mathrm{out}}\}.
 \label{eqn:core-admissible-counts}
\end{align}

\begin{theorem}[Weak consistency]
\label{thm:weak-consistency}
Suppose Assumption~\ref{ass:regular-cores} and the separation condition
\eqref{eqn:separation-condition} hold, and that the mass outside the cores
satisfies
\begin{align}
 \xi_N^{d+1}\alpha_{\mathrm{out}}=o\Bigl(\Bigl(\frac{\Delta}{L}\Bigr)^d\Bigr).
 \label{eqn:core-mass-condition}
\end{align}
Fix \(\underline\omega\in(0,1/2]\). Then \(\admset_N^\circ\) is nonempty for
all sufficiently large \(N\), and, for every sequence
\(k=k_N\in\admset_N^\circ\) and quantile levels
\(\omega_\ell\in[\underline\omega,1-\underline\omega]\), with overwhelming
probability on \(\mathcal E_{\mathrm D}\), the algorithm completes and
\begin{align}
 \mismatch_N(\vy,\widehat\vy)\leq\frac{|\gO_N|}N
 \lesssim\alpha_{\mathrm{out}}+\frac{\log N}N,
 \label{eqn:recursive-core-probabilistic-mismatch}
\end{align}
Moreover,
\(\E\bigl[\mismatch_N(\vy,\widehat\vy)\mathds{1}_{\mathcal E_{\mathrm D}}\bigr]
\leq\alpha_{\mathrm{out}}+C_\nu N^{-\nu}\) for every fixed \(\nu>0\).
\end{theorem}
Compared with Theorem~\ref{thm:strong-consistency}, only the ball bounds
are required, and only around core points, so a support may have sparse
tails. In return, \(k\geq\xi_NN\alpha_{\mathrm{out}}\) makes the exceptional
observations \(o(k)\) on every active set, and quantiles away from \(0\)
and \(1\) keep the selected radius at a core row; only the
exceptional observations can then be mislabeled. Condition
\eqref{eqn:core-mass-condition} leaves room for such a count, and it is
necessary up to constants: if \(\admset_N^\circ\) is nonempty, then \eqref{eqn:separation-condition} and
\eqref{eqn:core-mass-condition} hold with \(O\) in place of \(o\)
(Lemma~\ref{lem:admissible-ranks-floor} in
Appendix~\ref{app:recursive-recovery-proofs}).

\paragraph{Denoised distances}
For the denoised distances, \(b_N=\widetilde b_N\), and the separation
condition \eqref{eqn:separation-condition} is the sampling condition
\eqref{eqn:sampling-condition} together with the separation-to-noise
condition.
\begin{corollary}[Recovery from the denoised distances]
\label{cor:denoised-recovery}
Suppose the hypotheses of Theorem~\ref{thm:denoising-error-bound}, the
sampling condition \eqref{eqn:sampling-condition}, and the
separation-to-noise condition \(\xi_N^2\widetilde b_N=o(\Delta)\) hold.
Let Algorithm~\ref{alg:RGG-clustering} run on \(\mD=\widetilde\mD\) with
the settings above and \(\admset_N=\admset_N(\widetilde b_N)\).
\begin{enumerate}[label=(\alph*)]
\item \emph{(Exact recovery)} Under
Assumptions~\ref{ass:local-mass-regularity}
and~\ref{ass:locality-separation}, for every sequence
\(k=k_N\in\admset_N\), with overwhelming probability, the algorithm
completes and \(\mismatch_N(\vy,\widehat\vy)=0\).
\item \emph{(Vanishing mismatch)} Under
Assumption~\ref{ass:regular-cores} and \eqref{eqn:core-mass-condition},
the conclusions of Theorem~\ref{thm:weak-consistency} hold with
overwhelming probability, and
\(\E\mismatch_N(\vy,\widehat\vy)\leq\alpha_{\mathrm{out}}+C_\nu N^{-\nu}\)
for every fixed \(\nu>0\).
\end{enumerate}
\end{corollary}

\paragraph{Signal-to-noise ratio}
When \(\Tr\mTheta\asymp m\theta_{\min}\) and
\(\Delta^2\asymp\Tr\mTheta\), conditions
\eqref{eqn:denoised-separation-noise} and \eqref{eqn:selection-snr} hold
with \(\rho_N=O(1)\) if and only if
\begin{align}
 \SNR_N\gg\max\Bigl\{\frac{\xi_N^4(m+\log N)\sigma^2}{\Tr\mGamma},\
   m\log N\Bigl(\frac{\sigma^2}{\Tr\mGamma}+\frac{\sigma}{\sqrt{N\Tr\mGamma}}\Bigr)\Bigr\}
 \qquad\text{and}\qquad
 \SNR_N\gtrsim\frac mN.
 \label{eqn:snr-envelope}
\end{align}
All three lower bounds tend to zero when \(\Tr\mGamma\gg\sigma^2(m+\xi_N^4)\log N\), since
\(m\log N=o(N)\), so \(\SNR_N\) may tend to zero with a growing signal
dimension \(m\).

\subsection{Failure of the raw distances}
\label{sec:denoising-necessity}
We give a model in which the denoised distances recover every label and
the raw distances fail, at bounded signal-to-noise ratio and with two
communities that share a center. Proofs are in
Appendices~\ref{app:raw-necessity} and~\ref{app:denoising-necessity-proof}.

\paragraph{Why the raw distances fail}
For isotropic Gaussian noise, the raw squared distances equal the clean
ones plus a common offset of about \(2p\sigma^2\) and a fluctuation of
order \(\sigma^2\sqrt p\). The selected radius absorbs the offset, but not
the fluctuation. Every valid radius lies below \(\Delta/2\), so if
\(\Delta^2\leq N^{-c}\sigma^2\sqrt p\) for a constant \(c>0\), the
fluctuation exceeds the clean squared distances at every valid radius,
and a fraction \(1-\omega\) of the observations has no neighbor at the
selected radius (Proposition~\ref{prop:raw-necessity}). Denoising leaves an error of order
\(\sigma\sqrt{(1+\rho_N)(m+\log N)}\) instead, so the guarantee for the
denoised distances and the failure of the raw distances can hold together
in the range
\begin{align}
 \xi_N^2\sqrt{(1+\rho_N)(m+\log N)}\ll\frac{\Delta}{\sigma}
 \leq N^{-c}p^{1/4}.
 \label{eqn:raw-denoised-gap}
\end{align}

\paragraph{Two concentric circles}
Let the noise be isotropic Gaussian with \(\sigma=1\), and let
\(p=\lceil N^{5/2}\rceil\), \(m=2\), \(d=1\), \(B=2\),
\(\alpha_1=\alpha_2=1/2\), and \(L=\sqrt p\). In a fixed two-dimensional
subspace, community \(b\) is the uniform law on the circle of radius
\(\varrho_b\) centered at the origin, with \(\varrho_1=L/4\) and
\(\varrho_2=L/4+\Delta\), where \(\Delta\coloneqq h_N(\log^3N)=L\log^3N/N\)
is the local scale of \(\log^3N\) neighbors. Then \(\SNR_N\to1/16\),
\(\Delta/\sigma\asymp N^{1/4}\log^3N\),
\(\sqrt{(1+\rho_N)(m+\log N)}\asymp\sqrt{\log N}\), and
\(p^{1/4}\asymp N^{5/8}\), so \eqref{eqn:raw-denoised-gap} holds with any
\(c<3/8\).
\begin{proposition}[Raw and denoised distances on two concentric circles]
\label{prop:denoising-necessity}
In the model above, fix \(\omega\in(0,1)\). For all sufficiently large \(N\),
\(\admset_N(\widetilde b_N)=\admset_N(0)\); this set contains
\(\lceil\log^2N\rceil\), and all its counts are at most \(\log^3N\). For
every deterministic sequence \(k=k_N\in\admset_N(0)\), the following hold.
\begin{enumerate}[label=(\alph*)]
\item \emph{Denoised distances.} The hypotheses of
Corollary~\ref{cor:denoised-recovery}(a) hold, and so does the
signal-to-noise condition \eqref{eqn:snr-condition}. With overwhelming
probability, Algorithm~\ref{alg:RGG-clustering} run on \(\widetilde\mD\)
with the settings of Section~\ref{sec:recursive-clustering-guarantee}
completes with \(\mismatch_N(\vy,\widehat\vy)=0\).
\item \emph{Raw distances.} With probability tending to one, the
following hold for the radius \(r=r([N];k,\omega)\) selected from
\(\mD^{(\mathrm x)}\):
\begin{enumerate}[label=(\roman*)]
\item \(r^2=2p\sigma^2+O(\sigma^2\sqrt{p\log N})\), while the clean radius
is of order \(h_N(k)\);
\item at least a fraction \(1-\omega-o(1)\) of each community has no
neighbors at the radius \(r\);
\item \(T_N([N];k,\omega)\leq\omega+o(1)\);
\item if the first stage of Algorithm~\ref{alg:RGG-clustering} uses
\(\omega_1=\omega\), then
\(\mismatch_N(\vy,\widehat\vy)\geq(1-\omega)/2-o(1)\) for every threshold
\(\tau_1\geq0\).
\end{enumerate}
\end{enumerate}
\end{proposition}
The admissible counts cannot be raised to help the raw distances: that
would need
\(h_N(k)^2\gg\sigma^2\sqrt p\), that is, \(k\gg N^{3/8}\), but then
\(h_N(k)\gg\Delta\), and the radius graph joins the two circles even for
the clean distances.
\section{Numerical experiments on synthetic data}\label{sec:synthetic-experiments}
% Generated by python -m experiments.figures from results/; do not edit.
\providecommand{\NUM}[1]{\ifcsname nr@#1\endcsname\csname nr@#1\endcsname\else\PackageError{numerical-revision}{Undefined numerical macro #1}{Regenerate the checked saved-run macros.}\fi}
\expandafter\def\csname nr@I:k0\endcsname{\ensuremath{4}}
\expandafter\def\csname nr@I:k1\endcsname{\ensuremath{13}}
\expandafter\def\csname nr@I:k2\endcsname{\ensuremath{37}}
\expandafter\def\csname nr@U:condition\endcsname{\ensuremath{4}}
\expandafter\def\csname nr@U:delta-over-sigma\endcsname{\ensuremath{50}}
\expandafter\def\csname nr@U:disk-hi\endcsname{\ensuremath{0.8}}
\expandafter\def\csname nr@U:gap\endcsname{\ensuremath{2}}
\expandafter\def\csname nr@U:inner-hi\endcsname{\ensuremath{3.8}}
\expandafter\def\csname nr@U:inner-lo\endcsname{\ensuremath{2.8}}
\expandafter\def\csname nr@U:outer-hi\endcsname{\ensuremath{7.1}}
\expandafter\def\csname nr@U:outer-lo\endcsname{\ensuremath{5.8}}
\expandafter\def\csname nr@U:q\endcsname{\ensuremath{0.3}}
\expandafter\def\csname nr@U:sigma\endcsname{\ensuremath{0.04}}
\expandafter\def\csname nr@U:sigma-ratio\endcsname{\ensuremath{0.02}}
\expandafter\def\csname nr@U:stretch\endcsname{\ensuremath{2}}
\expandafter\def\csname nr@U:theta\endcsname{\ensuremath{8.914}}
\expandafter\def\csname nr@UD3:N-max\endcsname{\ensuremath{8640}}
\expandafter\def\csname nr@UD3:N-min\endcsname{\ensuremath{540}}
\expandafter\def\csname nr@UD3:p0\endcsname{\ensuremath{10^{3}}}
\expandafter\def\csname nr@UD3:p1\endcsname{\ensuremath{10^{6}}}
\expandafter\def\csname nr@UD3:p2\endcsname{\ensuremath{3.16\times10^{6}}}
\expandafter\def\csname nr@UD:N0\endcsname{\ensuremath{540}}
\expandafter\def\csname nr@UD:N1\endcsname{\ensuremath{2160}}
\expandafter\def\csname nr@UD:N2\endcsname{\ensuremath{4320}}
\expandafter\def\csname nr@UD:cross-max\endcsname{\ensuremath{6.47}}
\expandafter\def\csname nr@UD:cross-min\endcsname{\ensuremath{6}}
\expandafter\def\csname nr@UD:reps\endcsname{\ensuremath{10}}
\expandafter\def\csname nr@UF:N0\endcsname{\ensuremath{540}}
\expandafter\def\csname nr@UF:N1\endcsname{\ensuremath{1080}}
\expandafter\def\csname nr@UF:N2\endcsname{\ensuremath{2160}}
\expandafter\def\csname nr@UF:N4\endcsname{\ensuremath{8640}}
\expandafter\def\csname nr@UF:eig-low-first\endcsname{\ensuremath{0.0344}}
\expandafter\def\csname nr@UF:eig-low-last\endcsname{\ensuremath{0.0208}}
\expandafter\def\csname nr@UF:gap-high-max\endcsname{\ensuremath{1.08}}
\expandafter\def\csname nr@UF:gap-high-min\endcsname{\ensuremath{0.832}}
\expandafter\def\csname nr@UF:gap-low-dev\endcsname{\ensuremath{0.0392}}
\expandafter\def\csname nr@UF:p\endcsname{\ensuremath{10^{3}}}
\expandafter\def\csname nr@UF:path\endcsname{\ensuremath{45}}
\expandafter\def\csname nr@UF:pop-vec-N0\endcsname{\ensuremath{0.629}}
\expandafter\def\csname nr@UF:pop-vec-N1\endcsname{\ensuremath{0.402}}
\expandafter\def\csname nr@UF:pop-vec-N2\endcsname{\ensuremath{0.5}}
\expandafter\def\csname nr@UF:pop-vec-N4\endcsname{\ensuremath{0.214}}
\expandafter\def\csname nr@UF:reps\endcsname{\ensuremath{10}}
\expandafter\def\csname nr@UF:rings\endcsname{\ensuremath{8}}
\expandafter\def\csname nr@UF:rs\endcsname{\ensuremath{0.28}}
\expandafter\def\csname nr@UF:vec-high-first\endcsname{\ensuremath{0.81}}
\expandafter\def\csname nr@UF:vec-high-last\endcsname{\ensuremath{0.591}}
\expandafter\def\csname nr@UF:vec-low-first\endcsname{\ensuremath{0.255}}
\expandafter\def\csname nr@UF:vec-low-last\endcsname{\ensuremath{0.118}}
\expandafter\def\csname nr@UH:N3\endcsname{\ensuremath{4320}}
\expandafter\def\csname nr@UH:T-high-last\endcsname{\ensuremath{0.62}}
\expandafter\def\csname nr@UH:T-high-min\endcsname{\ensuremath{0.052}}
\expandafter\def\csname nr@UH:T-low-first\endcsname{\ensuremath{0.46}}
\expandafter\def\csname nr@UH:T-low-last\endcsname{\ensuremath{0.93}}
\expandafter\def\csname nr@UH:k-count\endcsname{\ensuremath{12}}
\expandafter\def\csname nr@UH:k-max\endcsname{\ensuremath{720}}
\expandafter\def\csname nr@UH:k-min\endcsname{\ensuremath{4}}
\expandafter\def\csname nr@UH:p\endcsname{\ensuremath{10^{3}}}
\expandafter\def\csname nr@UH:rel-high-last\endcsname{\ensuremath{0.305}}
\expandafter\def\csname nr@UH:rel-high-max\endcsname{\ensuremath{0.912}}
\expandafter\def\csname nr@UH:rel-high-peak-k\endcsname{\ensuremath{16}}
\expandafter\def\csname nr@UH:rel-low-last\endcsname{\ensuremath{0.00596}}
\expandafter\def\csname nr@UH:rel-low-max\endcsname{\ensuremath{0.0515}}
\expandafter\def\csname nr@UH:reps\endcsname{\ensuremath{10}}
\expandafter\def\csname nr@UK:N\endcsname{\ensuremath{540}}
\expandafter\def\csname nr@UK:cross-first\endcsname{\ensuremath{33}}
\expandafter\def\csname nr@UK:hi0\endcsname{\ensuremath{32}}
\expandafter\def\csname nr@UK:hi5\endcsname{\ensuremath{25}}
\expandafter\def\csname nr@UK:lo0\endcsname{\ensuremath{14}}
\expandafter\def\csname nr@UK:lo5\endcsname{\ensuremath{21}}
\expandafter\def\csname nr@UK:max10\endcsname{\ensuremath{89}}
\expandafter\def\csname nr@UK:max15\endcsname{\ensuremath{55}}
\expandafter\def\csname nr@UK:noise0\endcsname{\ensuremath{0}}
\expandafter\def\csname nr@UK:noise10\endcsname{\ensuremath{0.1}}
\expandafter\def\csname nr@UK:noise15\endcsname{\ensuremath{0.15}}
\expandafter\def\csname nr@UK:noise5\endcsname{\ensuremath{0.05}}
\expandafter\def\csname nr@UK:p\endcsname{\ensuremath{1000}}
\expandafter\def\csname nr@UK:peak-k10\endcsname{\ensuremath{20}}
\expandafter\def\csname nr@UK:peak-k15\endcsname{\ensuremath{14}}
\expandafter\def\csname nr@UK:reps\endcsname{\ensuremath{100}}
\expandafter\def\csname nr@UP2:N0\endcsname{\ensuremath{2160}}
\expandafter\def\csname nr@UP2:denoised-fail-p\endcsname{\ensuremath{3.16\times10^{8}}}
\expandafter\def\csname nr@UP2:denoised-last-p\endcsname{\ensuremath{10^{8}}}
\expandafter\def\csname nr@UP2:denoised-last-snr\endcsname{\ensuremath{2.79\times10^{-4}}}
\expandafter\def\csname nr@UP2:extra-decades\endcsname{\ensuremath{4}}
\expandafter\def\csname nr@UP2:k0\endcsname{\ensuremath{40}}
\expandafter\def\csname nr@UP2:p-max\endcsname{\ensuremath{10^{9}}}
\expandafter\def\csname nr@UP2:p-min\endcsname{\ensuremath{10^{3}}}
\expandafter\def\csname nr@UP2:raw-fail-p\endcsname{\ensuremath{3.16\times10^{5}}}
\expandafter\def\csname nr@UP2:raw-last-p\endcsname{\ensuremath{10^{4}}}
\expandafter\def\csname nr@UP2:raw-last-snr\endcsname{\ensuremath{2.79}}
\expandafter\def\csname nr@UP2:reps\endcsname{\ensuremath{10}}
\expandafter\def\csname nr@UP2:select-last-p\endcsname{\ensuremath{10^{8}}}
\expandafter\def\csname nr@curve:pairs\endcsname{\ensuremath{200}}

We illustrate our two contributions in Section~\ref{sec:introduction} on synthetic communities that share a center. We describe the data in Section~\ref{sec:experiments-setup} and the findings in Section~\ref{sec:experiments-findings}.
Findings~\ref{find:denoising}--\ref{find:spectrum} follow the first, from the noisy observations to the embedding (Section~\ref{sec:embedding-guarantees}), and Findings~\ref{find:window} and~\ref{find:necessity} follow the second, the recovery of the labels (Section~\ref{sec:clustering-guarantees}).

\subsection{Data}\label{sec:experiments-setup}
The clean signals lie in the plane with \(d=m=2\), forming \(B=3\) communities of \(N/3\) observations each: a disk of radius \NUM{U:disk-hi} and two annuli with radii in \([\NUM{U:inner-lo},\NUM{U:inner-hi}]\) and \([\NUM{U:outer-lo},\NUM{U:outer-hi}]\), all centered at the origin, so their center separation is zero.
Within each community, the angle is uniform and the radius is uniform in area.
The support separation is \(\Delta=\NUM{U:gap}\), and the signal covariance is \(\theta\mI_2\) with \(\theta=\NUM{U:theta}\); we call this the \emph{common design}.
The \emph{anisotropic design} of Finding~\ref{find:necessity} stretches the signals by \(\operatorname{diag}(\NUM{U:stretch},1)\), so that \(\theta_{\max}/\theta_{\min}=\NUM{U:condition}\).

A fixed matrix \(\mJ\) with orthonormal columns embeds the signals in \(\R^p\), and the noise is isotropic Gaussian with covariance \(\mGamma=\sigma^2\mI_p\), which satisfies Assumption~\ref{ass:noise-variance-structure}; hence \(\SNR_N=\Tr\mTheta/(p\sigma^2)\), which decreases as \(1/p\).
Unless a caption says otherwise, we choose \(\sigma=\NUM{U:sigma}\), hence \(\Delta/\sigma=\NUM{U:delta-over-sigma}\); the figures report the inverse, \(\sigma/\Delta\).
To reach \(p=\NUM{UP2:p-max}\), we sample the Gram matrix of the observations instead of the observations themselves: it equals the Gram matrix of the noisy signal coordinates plus \(\sigma^2\) times an independent Wishart matrix with \(p-m\) degrees of freedom, which we sample exactly.

\subsection{Findings}\label{sec:experiments-findings}
\newcounter{finding}
\newcommand{\finding}[2]{\refstepcounter{finding}\label{#1}\paragraph{Finding~\thefinding: #2}}

All runs use Algorithms~\ref{alg:three-fold-denoising}--\ref{alg:RGG-clustering} with the constant kernel function \(\kf\equiv1\), zero thresholds, quantiles \(\omega_\ell=\NUM{U:q}\), and no debiasing, so the additive error \(\widetilde b_N\) is the first case of \eqref{eqn:denoising-additive-error}.
On the denoised distances \(\widetilde\mD\), the affine event \(\mathcal E_{\mathrm D}(C\widetilde a_N,C\widetilde b_N)\) holds with overwhelming probability (Theorem~\ref{thm:denoising-error-bound}(b)), so the results of Sections~\ref{sec:stability-of-data-driven-selection}--\ref{sec:recursive-clustering-guarantee} apply with \(a_N=\widetilde a_N\) and \(b_N=\widetilde b_N\).
The figures show means or frequencies over independent runs, without error bars; labels and clean signals are used only for evaluation.

\begin{figure}[!htbp]
\centering
\includegraphics[width=\linewidth]{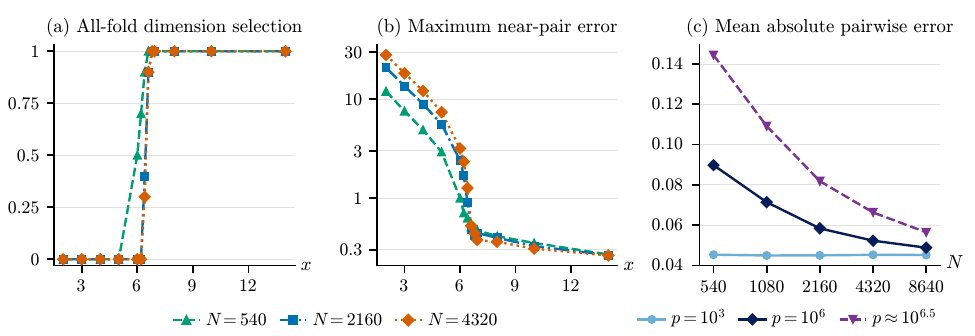}
\caption{Distance denoising on the common disk-and-annuli design at \(\sigma/\Delta=\NUM{U:sigma-ratio}\), with \NUM{UD:reps} independent runs per point.
In (a) and (b), the horizontal coordinate \(x=\theta/[\sigma^2\sqrt{p/N}\log(\mathrm eN)]\) increases as \(p\) decreases at each fixed \(N\in\{\NUM{UD:N0},\NUM{UD:N1},\NUM{UD:N2}\}\).
(a) Fraction of runs in which every fold selects the true dimension \(m=2\).
(b) Maximum absolute distance error \(|\widetilde D_{ij}-D^{(\mathrm s)}_{ij}|\) among the \NUM{curve:pairs} closest clean pairs.
(c) Mean absolute distance error over all pairs \(i<j\), as \(N\) doubles from \NUM{UD3:N-min} to \NUM{UD3:N-max} at \(p=\NUM{UD3:p0}\), \(p=\NUM{UD3:p1}\), and \(p\approx\NUM{UD3:p2}\).
Both errors are computed within each run, then averaged over runs.}
\label{fig:exp-rates-main}
\end{figure}

\finding{find:denoising}{Every fold selects the signal dimension once \(\widetilde a_N\log N\) is small, and the distance error then falls with \(N\) at large \(p\).}
Theorem~\ref{thm:denoising-error-bound}(a) predicts that every fold selects \(m\) when \(\widetilde a_N\log N\to0\).
For isotropic noise and \(p\gg N\), \(\widetilde a_N\approx\sigma^2\sqrt{p/N}/\theta\), so the horizontal coordinate \(x\) of Figure~\ref{fig:exp-rates-main}(a,b) is about \(1/\{\widetilde a_N\log(\mathrm eN)\}\).
The fraction of runs in which every fold selects \(m=2\) crosses \(1/2\) between \(x=\NUM{UD:cross-min}\) and \(x=\NUM{UD:cross-max}\), depending on \(N\) (panel (a)).
Through the same transition, the maximum error among the \NUM{curve:pairs} closest clean pairs drops sharply (panel (b)); on these pairs the multiplicative part \(\widetilde a_ND^{(\mathrm s)}_{ij}\) of the bound~\eqref{eqn:local-affine-bound} is small, so they probe its additive part \(\widetilde b_N\).
Over all pairs, the mean error falls at every doubling of \(N\) for \(p=\NUM{UD3:p1}\) and \(p\approx\NUM{UD3:p2}\), and stays nearly constant at \(p=\NUM{UD3:p0}\) (panel (c)).

\begin{figure}[!htbp]
\centering
\includegraphics[width=\linewidth]{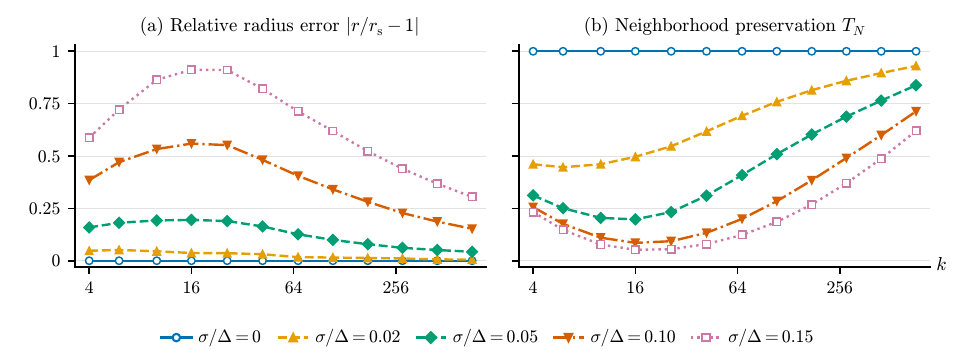}
\caption{Radius selection on the common design with \(N=\NUM{UH:N3}\), denoised input at \(p=\NUM{UH:p}\), and \(\omega=\NUM{U:q}\) on all three communities, against \NUM{UH:k-count} neighbor counts \(k\) from \NUM{UH:k-min} to \NUM{UH:k-max}, at \(\sigma/\Delta\in\{0,\NUM{U:sigma-ratio},\NUM{UK:noise5},\NUM{UK:noise10},\NUM{UK:noise15}\}\): (a) relative radius error \(|r/r_{\mathrm s}-1|\), where \(r_{\mathrm s}\) is the radius selected from the clean distances, and (b) neighborhood similarity \(T_N\) of~\eqref{eqn:average-jaccard}, each the mean over \NUM{UH:reps} runs.}
\label{fig:exp-radius}
\end{figure}

\finding{find:radius}{Once the neighbor count is admissible, the radius error and the changed neighbors fall as \(k\) grows.}
Propositions~\ref{prop:radius-stability} and~\ref{prop:neighborhood-preservation} bound the relative radius error by \(a_N+b_N/h_N(k)\) and the fraction \(1-T_N\) of changed neighbors by \(\delta_N(k)\), for admissible \(k\); both bounds decrease in \(k\).
At \(\sigma/\Delta=\NUM{U:sigma-ratio}\) (Figure~\ref{fig:exp-radius}), the radius error is at most \NUM{UH:rel-low-max} and falls to \NUM{UH:rel-low-last} at the largest \(k\), while \(T_N\) rises from \NUM{UH:T-low-first} to \NUM{UH:T-low-last}.
At \(\sigma/\Delta=\NUM{UK:noise15}\), the radius error peaks at \NUM{UH:rel-high-max} near \(k=\NUM{UH:rel-high-peak-k}\), where \(T_N\) is lowest (\NUM{UH:T-high-min}); at these counts \(b_N/h_N(k)\) is not small, so they lie below the admissible range~\eqref{eqn:admissible-range}.
Beyond them, both errors decrease, as the bounds predict, to \NUM{UH:rel-high-last} and \(T_N=\NUM{UH:T-high-last}\).

\begin{figure}[!htbp]
\centering
\includegraphics[width=\linewidth]{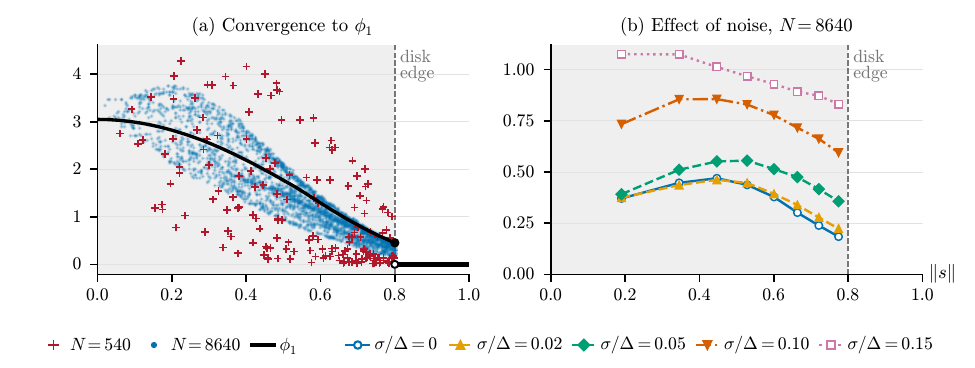}
\caption{Leading eigenvector at the first extraction stage on the common design, with \(N\) doubling from \NUM{UF:N0} to \NUM{UF:N4} and \(k=N/\NUM{UF:path}\), so that the clean radius stays near \NUM{UF:rs}, against each point's clean distance \(\|s_i\|\) from the center; the disk occupies \(\|s\|\leq\NUM{U:disk-hi}\) (shaded). Here \(\chi_1'\in\{-1,1\}\) aligns signs with \(\phi_1=\phi_{1,r_{\mathrm s}}\), normalized in \(L^2(\P_{\mathrm s})\), and \(\sqrt N\vv_1\), like \(\phi_1\), has mean square one. (a) Convergence to \(\phi_1\): \(\sqrt N\chi_1'\vv_1^{(\mathrm s)}\) for one realization at \(N=\NUM{UF:N0}\) and \(N=\NUM{UF:N4}\), and \(\phi_1\) (black); the clean eigenvector vanishes on the two annuli (\(\|s\|\geq\NUM{U:inner-lo}\)). (b) Effect of noise at \(N=\NUM{UF:N4}\), with denoised input at \(p=\NUM{UF:p}\): the disk is divided into \NUM{UF:rings} rings of equal area, and rms is the root mean square of \(\sqrt N\chi_1'(\vv_1)_i-\phi_1(\|s_i\|)\) over the points of a ring, averaged over \NUM{UF:reps} runs and plotted at the ring's mean distance; \(\sigma/\Delta=0\) is the clean input.}
\label{fig:exp-embedding}
\end{figure}

\finding{find:eigenfunction}{The leading clean eigenvector converges to a population eigenfunction that is positive exactly on the disk.}
Theorem~\ref{thm:population-eigenfunction-convergence}(b) bounds the distance between the eigenvectors and the population eigenfunctions; for clean input, \(a_N=b_N=0\), and only the population error rate \(\beta_N(k)\) remains.
Since the population eigengaps shrink with the radius (Remark~\ref{rem:circle-leading-mode-limitation}), we hold the clean radius near \NUM{UF:rs} by taking \(k=N/\NUM{UF:path}\), and examine the first extraction stage.
Its leading population eigenfunction \(\phi_1\) is positive on the disk and zero off it, and the clean eigenvector vanishes on the annuli, so one threshold extracts the disk (Figure~\ref{fig:exp-embedding}(a)).
The distance between \(\vv_1^{(\mathrm s)}\) and the normalized evaluations of \(\phi_1\) at the sample points, a sample version of \(\|U_N\vv_1^{(\mathrm s)}-\chi_1'\phi_1\|_{L^2(\P_{\mathrm s})}\), falls from \NUM{UF:pop-vec-N0} at \(N=\NUM{UF:N0}\) to \NUM{UF:pop-vec-N4} at \(N=\NUM{UF:N4}\), with one rise, from \NUM{UF:pop-vec-N1} at \(N=\NUM{UF:N1}\) to \NUM{UF:pop-vec-N2} at \(N=\NUM{UF:N2}\).

\begin{figure}[!htbp]
\centering
\includegraphics[width=\linewidth]{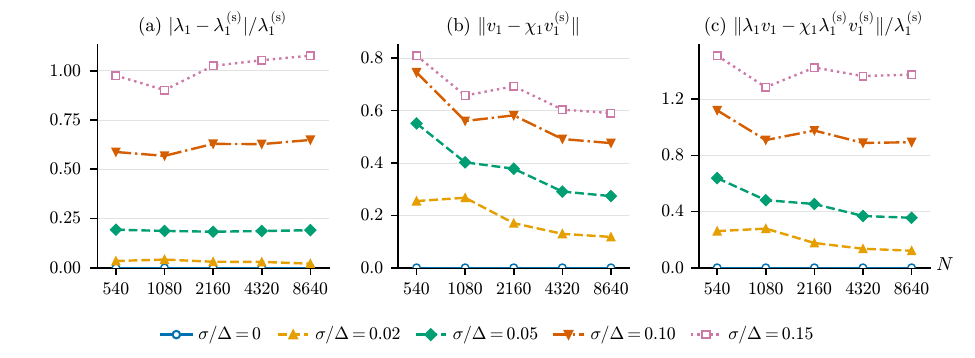}
\caption{Leading eigenpair along the path of Figure~\ref{fig:exp-embedding}, with denoised input at \(p=\NUM{UF:p}\) and \(\sigma/\Delta\in\{0,\NUM{U:sigma-ratio},\NUM{UK:noise5},\NUM{UK:noise10},\NUM{UK:noise15}\}\), against the clean kernel: (a) relative eigenvalue error \(|\lambda_1-\lambda_1^{(\mathrm s)}|/\lambda_1^{(\mathrm s)}\), (b) eigenvector error \(\|\vv_1-\chi_1\vv_1^{(\mathrm s)}\|\), and (c) error of the eigenvalue-weighted coordinate \(\|\lambda_1\vv_1-\chi_1\lambda_1^{(\mathrm s)}\vv_1^{(\mathrm s)}\|/\lambda_1^{(\mathrm s)}\), each the mean over \NUM{UF:reps} runs.}
\label{fig:exp-spectrum}
\end{figure}

\finding{find:spectrum}{At a fixed noise level, the error of the leading eigenpair levels off at a floor set by the noise inside the signal subspace.}
Theorem~\ref{thm:spectral-stability} bounds the eigenvalue error by \((k/N)\delta_N(k)\), that is, by \(\delta_N(k)\) relative to the eigenvalue scale \(k/N\), and the eigenvector errors by the same rate divided by the population eigengap.
Along the fixed-radius path of Finding~\ref{find:eigenfunction}, the local scale \(h_N(k)\) is fixed, and the term \(b_N/h_N(k)\) of \(\delta_N(k)\) does not vanish: \(\widetilde b_N\) contains \(\sigma\sqrt m\), the noise inside the signal subspace, which the projection keeps.
The errors therefore level off (Figure~\ref{fig:exp-spectrum}).
At \(\sigma/\Delta=\NUM{U:sigma-ratio}\) they still decrease over the simulated range: the eigenvalue error from \NUM{UF:eig-low-first} to \NUM{UF:eig-low-last}, and the eigenvector error from \NUM{UF:vec-low-first} to \NUM{UF:vec-low-last}.
From \(\sigma/\Delta=\NUM{UK:noise5}\) on, the eigenvalue error is flat or rising in \(N\), and at \(\sigma/\Delta=\NUM{UK:noise15}\) the eigenvector error falls only from \NUM{UF:vec-high-first} to \NUM{UF:vec-high-last}.
The error to \(\phi_1\) shows the same floor at \(N=\NUM{UF:N4}\) (Figure~\ref{fig:exp-embedding}(b)): it stays within \NUM{UF:gap-low-dev} of the clean error at \(\sigma/\Delta=\NUM{U:sigma-ratio}\), and from \(\sigma/\Delta=\NUM{UK:noise5}\) on it exceeds the clean error in every ring, with rms values between \NUM{UF:gap-high-min} and \NUM{UF:gap-high-max} at \(\sigma/\Delta=\NUM{UK:noise15}\).

\begin{figure}[!htbp]
    \centering
    \includegraphics[width=\linewidth]{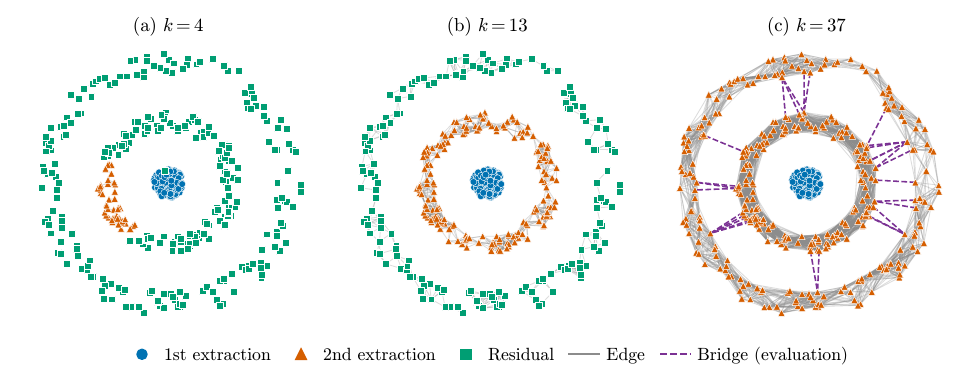}
    \caption{One clean realization of the common design with \(N=\NUM{UK:N}\) at \(k=\NUM{I:k0}\), \NUM{I:k1} and \NUM{I:k2}. Every vertex is colored by its saved membership; the residual class receives its label without another spectral extraction, need not be connected, and is empty at \(k=\NUM{I:k2}\). Grey lines are edges of the second-stage radius graph, which join active vertices only; bridges between communities (dashed) are drawn from the true labels.}
    \label{fig:numerical-2d-same-center}
\end{figure}
\begin{figure}[!htbp]
\centering
\includegraphics[width=\linewidth]{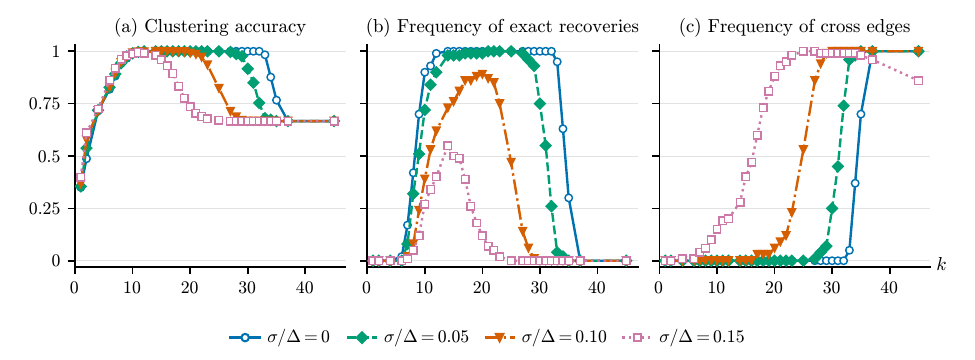}
\caption{Neighbor-count window on the common design with \(N=\NUM{UK:N}\), denoised input at \(p=\NUM{UK:p}\), and \(\sigma/\Delta\in\{\NUM{UK:noise0},\NUM{UK:noise5},\NUM{UK:noise10},\NUM{UK:noise15}\}\), with \NUM{UK:reps} realizations per neighbor count, paired across counts and noise ratios: (a) mean clustering accuracy \(1-\mismatch_N\), with \(\mismatch_N\) from~\eqref{eqn:mismatch-ratio}, (b) frequency of exact recovery \(\mismatch_N=0\), and (c) frequency of a second-stage cross-community edge, among the runs with a defined second stage.}
\label{fig:exp-k-window}
\end{figure}

\finding{find:window}{Exact recovery holds over a window of neighbor counts, which narrows as the noise grows.}

Theorem~\ref{thm:strong-consistency} gives exact recovery for every admissible neighbor count (Definition~\ref{def:admissible-ranks}); up to constants and the slack, these counts run from \(\max\{\log N,N(b_N/L)^d\}\) to \(N(\Delta/L)^d\), so the noise moves only the lower end.
We call the tested counts at which every run recovers exactly the \emph{recovery window}.
Figure~\ref{fig:numerical-2d-same-center} shows on one clean realization how it ends: at \(k=\NUM{I:k0}\) the second extraction is a fragment, and at \(k=\NUM{I:k2}\) the second-stage graph joins the two annuli.
Without noise, the window runs from \(k=\NUM{UK:lo0}\), the first count at which every extracted community is connected, through \(k=\NUM{UK:hi0}\), the last count before the first second-stage cross-community edge, at \(k=\NUM{UK:cross-first}\) (Figure~\ref{fig:exp-k-window}).
At \(\sigma/\Delta=\NUM{UK:noise5}\), it shrinks to \(k=\NUM{UK:lo5}\) through \NUM{UK:hi5}.
The lower end rises, as the theorem predicts; unlike the admissible range, the upper end also falls, because noise moves the first cross-community edges to smaller counts (panel (c)).
At the two larger noise levels the window is empty: the best counts recover exactly in \NUM{UK:max10} and \NUM{UK:max15} of \NUM{UK:reps} runs, at \(k=\NUM{UK:peak-k10}\) and \(k=\NUM{UK:peak-k15}\).

\begin{figure}[!htbp]
    \centering
    \includegraphics[width=\linewidth]{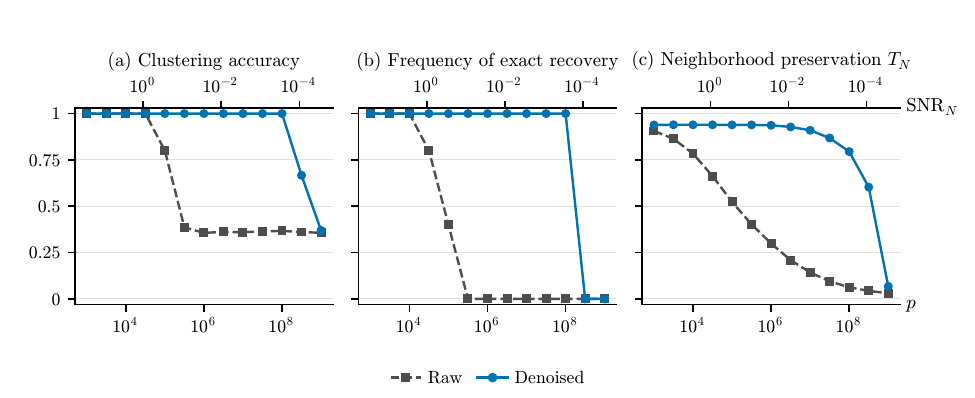}
    \caption{Raw distances \(\mD^{(\mathrm x)}\) and denoised distances \(\widetilde\mD\) on the anisotropic design with \(N=\NUM{UP2:N0}\), \(k=\NUM{UP2:k0}\), \(\sigma/\Delta=\NUM{U:sigma-ratio}\), and \(p\) from \NUM{UP2:p-min} to \NUM{UP2:p-max} in half decades, with \NUM{UP2:reps} paired runs per dimension: (a) mean clustering accuracy, (b) frequency of exact recovery, and (c) mean neighborhood similarity \(T_N\), against \(p\) (bottom) and \(\SNR_N\) (top).}
    \label{fig:exp-phase-anisotropic}
\end{figure}

\finding{find:necessity}{At a fixed separation-to-noise ratio, denoised distances recover the labels far beyond the dimension at which raw distances fail.}
Along this sweep, \(\Delta/\sigma=\NUM{U:delta-over-sigma}\) is fixed and \(\SNR_N\) falls as \(1/p\) (Figure~\ref{fig:exp-phase-anisotropic}).
Raw input recovers exactly in every run up to \(p=\NUM{UP2:raw-last-p}\), where \(\SNR_N=\NUM{UP2:raw-last-snr}\), and in no run at \(p=\NUM{UP2:raw-fail-p}\).
Denoised input recovers exactly in every run up to \(p=\NUM{UP2:denoised-last-p}\), where \(\SNR_N=\NUM{UP2:denoised-last-snr}\), \NUM{UP2:extra-decades} decades beyond raw input, and in no run at \(p=\NUM{UP2:denoised-fail-p}\).
It ends where dimension selection ends: every fold of every run selects \(m=2\) up to \(p=\NUM{UP2:select-last-p}\), as Theorem~\ref{thm:denoising-error-bound}(a) requires.
The raw distances fail by the mechanism of Section~\ref{sec:denoising-necessity}: their squared fluctuation, of order \(\sigma^2\sqrt p\), exceeds the squared local scale.
Proposition~\ref{prop:denoising-necessity} proves this failure, together with exact recovery from \(\widetilde\mD\), on two concentric circles at sample sizes far beyond those simulated here.

\section*{Acknowledgments}
Part of this research was performed when both authors were attending the workshop ``Recent Advances in Random Networks'' at the Institute for Mathematical and Statistical Innovation (IMSI) in January 2026, which is supported by the National Science Foundation (Grant No. DMS-2425650). Part of the research was conducted during HXW's visit to UC Davis in Oct 2026. Both authors acknowledge support from NSF DMS-2605423.  
%%%%%%%%%%%%%%%%%%%%%%%%%%%%%%%%%%%%%%%%%%%%%%
%%              Bibliography                %%
%%%%%%%%%%%%%%%%%%%%%%%%%%%%%%%%%%%%%%%%%%%%%%
\printbibliography
%%%%%%%%%%%%%%%%%%%%%%%%%%%%%%%%%%%%%%%%%%%%%%
%%               Appendix                   %%
%%%%%%%%%%%%%%%%%%%%%%%%%%%%%%%%%%%%%%%%%%%%%%
\newpage
\appendix
\phantomsection
\addcontentsline{toc}{section}{Appendices}
\section{Proofs in Section~\ref{sec:distance-denoising}}
\label{app:denoising}

Throughout this section, \(c\) and \(C\) denote positive constants whose
values may change between occurrences; they may depend on the constant
\(C_{\mathrm{sg}}\) of Assumption~\ref{ass:noise-variance-structure}, and \(C_\nu\) may
also depend on \(\nu\).
Appendix~\ref{app:denoising-proof} proves
Theorem~\ref{thm:denoising-error-bound},
Appendix~\ref{app:minimax-proof} proves
Theorem~\ref{thm:coordinate-reconstruction-lower-bound}, and
Appendix~\ref{app:hard-law-construction} constructs the hard laws used
there. General probability tools are collected in
Appendix~\ref{app:technical-lemmas}.

\subsection{Proof of Theorem~\ref{thm:denoising-error-bound}}
\label{app:denoising-proof}

\paragraph{Idea of the proof}
Fix a pair \(i<j\) and its fold \(q=q(i,j)\). Since
\(\vx_i-\vx_j=(\clean_i-\clean_j)+(\noise_i-\noise_j)\), the triangle
inequality applied to \eqref{eqn:denoised-distance} gives
\begin{align}
 \bigl|\widetilde D_{ij}-D^{(\mathrm s)}_{ij}\bigr|
 \leq\underbrace{\bigl|\|\foldmap_q^{\sT}(\clean_i-\clean_j)\|-D^{(\mathrm s)}_{ij}\bigr|}_{\text{signal distortion}}
 +\underbrace{\|\foldmap_q^{\sT}(\noise_i-\noise_j)\|}_{\text{held-out noise}}.
 \label{eqn:denoising-error-decomposition}
\end{align}
On an event that depends on fold~\(q\) alone, \(\foldmap_q\) is a
near-isometry on the signal subspace (Lemma~\ref{lem:fold-event}), so the
signal distortion is at most \(C_\nu\widetilde a_ND^{(\mathrm s)}_{ij}\).
The held-out noise is where cross-fitting enters: neither endpoint lies in
\(F_q\), so \(\noise_i-\noise_j\) is independent of \(\foldmap_q\), and
\(\foldmap_q^{\sT}(\noise_i-\noise_j)\) is a centered sub-Gaussian vector in
\(\R^m\) with covariance
\(2\foldmap_q^{\sT}\mGamma\foldmap_q\preceq2\sigma^2\foldmap_q^{\sT}\foldmap_q\).
Lemma~\ref{lem:held-out-pair} bounds its norm by
\(C_\nu\sigma\sqrt{1+\rho_N}\,(\sqrt m+\sqrt{\log N})\), which is
\(\widetilde b_N\) without debiasing. Its squared norm has mean
\(2\Tr(\foldmap_q^{\sT}\mGamma\foldmap_q)\leq C_\nu\sigma^2(1+\rho_N)m\);
for nearly isotropic noise, debiasing subtracts an estimate of this
trace and leaves its fluctuation and an error proportional to
\(\varepsilon_{\mathrm{iso}}\).

\paragraph{Role of the hypotheses}
The condition \(m\log N=o(N)\) makes the second-moment matrix of each
fold concentrate; \(\widetilde a_N\log N\to0\) keeps the noise eigenvalues
of every fold below the selection threshold of \eqref{eqn:fold-rank}; and
\(\varepsilon_{\mathrm{iso}}\leq1/2\) gives \(\bar\sigma^2\geq\sigma^2/2\),
so that \(\widehat\sigma_q^2\) estimates a variance of order \(\sigma^2\).

\paragraph{The fold Gram matrix}
Complete \(\mJ\) to an orthogonal matrix
\([\mJ,\mJ_\perp]\in\R^{p\times p}\), and write
\(\mS_q=[\mJ^{\sT}\clean_i]_{i\in F_q}\in\R^{m\times N_q}\) and
\(\mXi_q=[\noise_i]_{i\in F_q}\). Then
\(\mJ_\perp^{\sT}\mGamma\mJ_\perp\) has the same nonzero
eigenvalues as \(\mGamma_\perp=\mP_\perp\mGamma\mP_\perp\), so it
has trace \(\Tr\mGamma_\perp\) and Frobenius norm
\(\|\mGamma_\perp\|_{\frob}\). The clean signals lie in the range of
\(\mJ\), so \(\mX_q=\mJ\mS_q+\mXi_q\) and the Gram matrix of fold~\(q\)
splits exactly as
\begin{align}
 \mG_q=\mA_q^{\sT}\mA_q+(\Tr\mGamma_\perp)\mI_{N_q}+\mDelta_q,
 \qquad
 \mA_q=\mJ^{\sT}\mX_q,
 \qquad
 \mDelta_q=(\mJ_\perp^{\sT}\mXi_q)^{\sT}\mJ_\perp^{\sT}\mXi_q-(\Tr\mGamma_\perp)\mI_{N_q}.
 \label{eqn:fold-gram-decomposition}
\end{align}
The first summand carries the signal: it has rank \(m\), and its nonzero
eigenvalues are of order between \(N\theta_{\min}\) and \(N\theta_{\max}\).
The second is a noise floor, a multiple of the identity because the
noise vectors are iid, whatever their covariance; the shift in
\eqref{eqn:fold-corrected-gram} removes it up to \(\|\mDelta_q\|\). The third
is a fluctuation of size
\(\sigma^2N+\|\mGamma_\perp\|_{\frob}\sqrt N=N\theta_{\min}\widetilde a_N\).
After the shift, \(m\) eigenvalues lie above \(cN\theta_{\min}\) and the
others below \(C_\nu N\theta_{\min}\widetilde a_N\), and the threshold of
\eqref{eqn:fold-rank} separates the two groups. The selected directions are
partly noise, so projecting onto them shrinks clean distances, and the
fold map corrects for this.
Since
\(\mG_q\widetilde\mU_q=\widetilde\mU_q(\widetilde\mLambda_q+\lambda_{\min}(\mG_q)\mI)\),
the map \eqref{eqn:fold-map} satisfies
\begin{align}
 \foldmap_q
 =\underbrace{\mX_q\widetilde\mU_q
   \bigl(\widetilde\mLambda_q+\lambda_{\min}(\mG_q)\mI\bigr)^{-1/2}}_{\text{orthonormal columns}}
 \bigl(\mI+\lambda_{\min}(\mG_q)\widetilde\mLambda_q^{-1}\bigr)^{1/2},
 \qquad
 \foldmap_q^{\sT}\foldmap_q
 =\mI+\lambda_{\min}(\mG_q)\widetilde\mLambda_q^{-1}.
 \label{eqn:corrected-map-factorization}
\end{align}
The first factor has orthonormal columns and projects onto the estimated
signal directions; the diagonal second factor undoes the shrinkage.

For \(q=1,2,3\), let \(\mathcal F_q\) be the \(\sigma\)-field generated by
the fold permutation \(\pi_{\mathrm{fold}}\) and by the clean signals and
the noise vectors of the observations in \(F_q\). The fold map
\(\foldmap_q\) is \(\mathcal F_q\)-measurable. The next lemma collects the
properties of the fitted folds.
\begin{lemma}[Good folds]
\label{lem:fold-event}
Under the conditions of Theorem~\ref{thm:denoising-error-bound}, for every
fixed \(\nu>0\) there are events \(\mathcal E_q\in\mathcal F_q\),
\(q=1,2,3\), with \(\P(\bigcup_{q=1}^3\mathcal E_q^c)\leq C_\nu N^{-\nu-3}\)
for all sufficiently large \(N\), such that on \(\mathcal E_q\):
\begin{enumerate}[label=(\roman*)]
\item fold \(q\) selects the signal dimension, \(\widetilde m_q=m\);
\item the smallest eigenvalue of the Gram matrix is close to the noise
floor,
\(|\lambda_{\min}(\mG_q)-\Tr\mGamma_\perp|
\leq C_\nu\bigl(\|\mGamma_\perp\|_{\frob}\sqrt N+\sigma^2N\bigr)\);
\item the fold map is a near-isometry on the signal subspace, with bounded
norm:
\begin{align}
 \bigl|\|\foldmap_q^{\sT}\mJ\vv\|-\|\vv\|\bigr|
 &\leq C_\nu\widetilde a_N\|\vv\|\quad(\vv\in\R^m),
 &
 \|\foldmap_q\|^2&\leq C_\nu(1+\rho_N).
 \label{eqn:fold-event-bounds}
\end{align}
In particular, for all \(i<j\),
\begin{align}
 \big|\|\foldmap_q^{\sT}(\clean_i-\clean_j)\|-D^{(\mathrm{s})}_{ij}\big|
 &\leq C_\nu\widetilde a_N\,D^{(\mathrm{s})}_{ij}.
 \label{eqn:three-fold-clean-distance-distortion}
\end{align}
\end{enumerate}
\end{lemma}

\begin{proof}[Proof of Lemma~\ref{lem:fold-event}]
Since \(\theta_{\max}/\theta_{\min}=o(\log N)\) by
Assumption~\ref{ass:growing-covariance} and \(\widetilde a_N\log N\to0\),
\begin{align}
 \frac{\sigma^2}{\theta_{\min}}\cdot\frac{\theta_{\max}}{\theta_{\min}}\to0,
 \qquad
 \sigma^2N+\|\mGamma_\perp\|_{\frob}\sqrt N
 =N\theta_{\min}\widetilde a_N.
 \label{eqn:denoising-rate-identities}
\end{align}
The fold sizes satisfy \(N_q\geq N/4\) for \(N\geq12\), so \(m<N_q\) for
all sufficiently large \(N\), since \(m\log N=o(N)\). The permutation
defining the folds is independent of the data, so the observations in each
fold are iid from the model.

\smallskip
\noindent\emph{Step 1: clean signals on a fold.}
The columns of \(\mS_q\) in \eqref{eqn:fold-gram-decomposition} are iid
with covariance \(\mSigma\) in \eqref{eqn:signal-covariance-definition}.
Let \(\breve{\vs}_i=\mSigma^{-1/2}\mJ^{\sT}\clean_i\). Then
\(\E\breve{\vs}_i\breve{\vs}_i^{\sT}=\mI_m\), and
\(\|\breve{\vs}_i\|^2\leq Cm\) by
Assumption~\ref{ass:growing-covariance}.
Lemma~\ref{lem:isotropic-matrix-chernoff} with
\(\mX_i=\breve{\vs}_i\breve{\vs}_i^{\sT}\) and \(K=Cm\), conjugated by \(\mSigma^{1/2}\),
gives
\begin{align}
 \P\left\{\tfrac12N_q\mSigma\npreceq\mS_q\mS_q^{\sT}
        \ \text{ or }\ 
        \mS_q\mS_q^{\sT}\npreceq\tfrac32N_q\mSigma\right\}
 &\leq2m\exp(-cN_q/m).
 \label{eqn:growing-training-covariance-tail}
\end{align}
Since \(m\log N=o(N)\), this is at most \(N^{-\nu-4}\) for all
three folds and all sufficiently large \(N\). On the complementary event,
the singular values of \(\mS_q\) lie between \(c\sqrt{N\theta_{\min}}\) and
\(C\sqrt{N\theta_{\max}}\).

\smallskip
\noindent\emph{Step 2: noise on a fold.}
Lemma~\ref{lem:fold-noise-bounds} with \(t=(\nu+5)\log N\leq N_q\) gives,
with probability at least \(1-C_\nu N^{-\nu-4}\) for all three folds,
\(\|\mJ^{\sT}\mXi_q\|\leq C_\nu\sigma\sqrt N\) and
\begin{align}
 \|\mDelta_q\|
 &\leq C_\nu\bigl(\|\mGamma_\perp\|_{\frob}\sqrt N+\sigma^2N\bigr)
 \leq C_\nu N\theta_{\min}\widetilde a_N,
 \label{eqn:three-fold-wishart-bound}
\end{align}
where the last step is \eqref{eqn:denoising-rate-identities}.

Let \(\mathcal E_q\) be the event that the bounds of Steps~1 and~2 hold
for fold~\(q\). It is determined by \(\pi_{\mathrm{fold}}\) and by the
signals and the noise of fold~\(q\), so \(\mathcal E_q\in\mathcal F_q\),
and \(\P(\bigcup_q\mathcal E_q^c)\leq C_\nu N^{-\nu-3}\). The remaining
steps are deterministic on \(\mathcal E_q\), for all sufficiently large
\(N\). There, the signal-direction data
\(\mA_q=\mS_q+\mJ^{\sT}\mXi_q\) satisfy
\(\|\mA_q\mA_q^{\sT}-\mS_q\mS_q^{\sT}\|
\leq C_\nu N(\sigma\theta_{\max}^{1/2}+\sigma^2)=o(N\theta_{\min})\), by
the first relation of \eqref{eqn:denoising-rate-identities}. With Step~1,
\begin{align}
 cN\theta_{\min}
 \leq\lambda_m(\mA_q\mA_q^{\sT})
 \leq\lambda_1(\mA_q\mA_q^{\sT})
 \leq CN\theta_{\max}.
 \label{eqn:three-fold-parallel-signal-bound}
\end{align}

\smallskip
\noindent\emph{Step 3: noise floor and dimension selection.}
In the decomposition \eqref{eqn:fold-gram-decomposition}, the first
summand is positive semidefinite of rank at most \(m<N_q\). Hence
\(\lambda_{\min}(\mG_q)\geq \Tr\mGamma_\perp-\|\mDelta_q\|\), and the Rayleigh
quotient of \(\mG_q\) at a unit vector in the kernel of the first summand
gives \(\lambda_{\min}(\mG_q)\leq\Tr\mGamma_\perp+\|\mDelta_q\|\). Thus
\begin{align}
 |\lambda_{\min}(\mG_q)-\Tr\mGamma_\perp|\leq\|\mDelta_q\|,
 \label{eqn:noise-floor-sandwich}
\end{align}
which with \eqref{eqn:three-fold-wishart-bound} is~(ii).
With \(\mDelta_q'\coloneqq\mDelta_q+\{\Tr\mGamma_\perp-\lambda_{\min}(\mG_q)\}\mI_{N_q}\),
\begin{align}
 \widetilde{\mG}_q
 &=\mA_q^{\sT}\mA_q+\mDelta_q',
 &\|\mDelta_q'\|&\leq2\|\mDelta_q\|.
 \label{eqn:three-fold-empirical-shift}
\end{align}
By \eqref{eqn:three-fold-wishart-bound}--\eqref{eqn:three-fold-empirical-shift}
and \(\widetilde\mG_q\succeq\bzero\),
\begin{align}
 cN\theta_{\min}&\leq\lambda_{m}(\widetilde{\mG}_q)\leq\lambda_{1}(\widetilde{\mG}_q)\leq CN\theta_{\max},
 &0\leq\lambda_{m+1}(\widetilde{\mG}_q)&\leq C_\nu N\theta_{\min}\widetilde a_N.
 \label{eqn:three-fold-corrected-positivity}
\end{align}
Since \(\theta_{\max}/\theta_{\min}=o(\log N)\) and
\(\widetilde a_N\log(\mathrm eN)\to0\), we have
\(CN\theta_{\max}/\log(\mathrm eN)<cN\theta_{\min}\) and
\(C_\nu N\theta_{\min}\widetilde a_N<cN\theta_{\min}/\log(\mathrm eN)\) for all
sufficiently large \(N\). Hence
\begin{align}
 \lambda_{m}(\widetilde{\mG}_q)>\lambda_{1}(\widetilde{\mG}_q)/\log(\mathrm eN)>\lambda_{m+1}(\widetilde{\mG}_q),
 \label{eqn:three-fold-rank-selection-gap}
\end{align}
so exactly \(m\) eigenvalues exceed the threshold of
\eqref{eqn:fold-rank}. This is~(i).

\smallskip
\noindent\emph{Step 4: signal distortion.}
By~(i), \(\mH_q\coloneqq\foldmap_q^{\sT}\mJ\) is \(m\times m\).
Since \(\mJ^{\sT}\mX_q=\mA_q\),
\(\widetilde\mU_q^{\sT}\widetilde\mG_q\widetilde\mU_q=\widetilde\mLambda_q\), and
\(\mA_q^{\sT}\mA_q=\widetilde\mG_q-\mDelta_q'\),
\begin{align}
 \mH_q\mH_q^{\sT}
 &=\mI_m-\widetilde{\mLambda}_q^{-1/2}
   \widetilde{\mU}_q^{\sT}\mDelta_q'
   \widetilde{\mU}_q\widetilde{\mLambda}_q^{-1/2}.
 \label{eqn:three-fold-signal-distortion-identity}
\end{align}
By \eqref{eqn:corrected-map-factorization} and
\eqref{eqn:three-fold-wishart-bound}--\eqref{eqn:three-fold-corrected-positivity},
\begin{align}
 \|\mH_q\mH_q^{\sT}-\mI_m\|
 &\leq C_\nu\widetilde a_N,
 &\|\foldmap_q\|^2
 &\leq1+\frac{\Tr\mGamma_\perp+\|\mDelta_q\|}{cN\theta_{\min}}
 \leq C_\nu (1+\rho_N).
 \label{eqn:three-fold-transform-bounds}
\end{align}
Since \(\mH_q\) is square, \(\mH_q^{\sT}\mH_q\) has the same eigenvalues, so
every singular value of \(\mH_q\) is within \(C_\nu\widetilde a_N\) of one.
This is the first bound of \eqref{eqn:fold-event-bounds}; the second is
\eqref{eqn:three-fold-transform-bounds}. Finally,
\eqref{eqn:three-fold-clean-distance-distortion} is the first bound at
\(\vv=\mJ^{\sT}(\clean_i-\clean_j)\), since
\(\clean_i-\clean_j=\mJ\mJ^{\sT}(\clean_i-\clean_j)\).
\end{proof}

The second lemma treats a pair held out from a fitted fold.
\begin{lemma}[Held-out pairs]
\label{lem:held-out-pair}
Fix \(q\in\{1,2,3\}\), a pair \(i<j\), and \(t>0\). Let \(\mPsi\) be an
\(\mathcal F_q\)-measurable random matrix with \(p\) rows and
\(\widetilde k\) columns, and put
\(\vu=\mPsi^{\sT}(\clean_i-\clean_j)\),
\(\vh=\mPsi^{\sT}(\noise_i-\noise_j)\), and
\(\mV=2\mPsi^{\sT}\mGamma\mPsi\). There is a constant \(C\),
depending only on \(C_{\mathrm{sg}}\), such that, on the event
\(\{q(i,j)=q\}\), conditionally on \(\mathcal F_q\) and
\((\clean_i,\clean_j)\), \(\vh\) has mean zero and covariance \(\mV\), and
the following hold except with conditional probability at most
\(8\mathrm e^{-t}\):
\begin{enumerate}[label=(\alph*)]
\item \(\|\vh\|\leq C\|\mV\|^{1/2}(\sqrt{\widetilde k}+\sqrt t)\);
\item \(\bigl|\|\vu+\vh\|^2-\Tr\mV-\|\vu\|^2\bigr|
\leq C\bigl\{\sqrt{t\|\mV\|}\,\|\vu\|+\|\mV\|(\sqrt{\widetilde kt}+t)\bigr\}\).
\end{enumerate}
Consequently, for every event \(\mathcal E'\in\mathcal F_q\), the
probability that \(\mathcal E'\cap\{q(i,j)=q\}\) holds and (a) or (b)
fails is at most \(8\mathrm e^{-t}\).
\end{lemma}
\begin{proof}[Proof of Lemma~\ref{lem:held-out-pair}]
The event \(\{q(i,j)=q\}\) is determined by \(\pi_{\mathrm{fold}}\), so it
lies in \(\mathcal F_q\). On it, neither endpoint belongs to \(F_q\), so
\(\vw\coloneqq(\vg_i-\vg_j)/\sqrt2\) is independent of the conditioning,
and \(\vh=\mB\vw\) with the conditionally fixed matrix
\(\mB=\sqrt2\,\mPsi^{\sT}\mGamma^{1/2}\), for which
\(\mB\mB^{\sT}=\mV\). The coordinates of \(\vw\) are independent, with
mean zero, unit variance, and sub-Gaussian norm at most \(CC_{\mathrm{sg}}\).
Since \(\Tr\mV\leq\widetilde k\|\mV\|\) and
\(\|\mV\|_{\frob}^2\leq\widetilde k\|\mV\|^2\),
Lemma~\ref{lem:gaussian-quadratic-form} with \(\vu=\bzero\) gives
\(\|\vh\|^2\leq\Tr\mV+C\|\mV\|(\sqrt{\widetilde kt}+t)
\leq C\|\mV\|(\sqrt{\widetilde k}+\sqrt t)^2\), and with general \(\vu\) it
gives~(b). The last claim follows by integrating the conditional bound
over \(\mathcal E'\cap\{q(i,j)=q\}\in\mathcal F_q\).
\end{proof}

\begin{proof}[Proof of Theorem~\ref{thm:denoising-error-bound}]
Fix \(\nu>0\), let \(\mathcal E_q\) be the events of
Lemma~\ref{lem:fold-event}, and put \(t=(\nu+5)\log N\).

\smallskip
\noindent\emph{Part~(a).} This is Lemma~\ref{lem:fold-event}(i).

\smallskip
\noindent\emph{Part~(b): held-out pairs.}
Fix \(i<j\) and \(q\in\{1,2,3\}\), and put
\(\vu=\foldmap_q^{\sT}(\clean_i-\clean_j)\) and
\(\mV=2\foldmap_q^{\sT}\mGamma\foldmap_q\). On \(\mathcal E_q\), the map
\(\foldmap_q\) has \(m\) columns and
\(\|\mV\|\leq2\sigma^2\|\foldmap_q\|^2\leq C_\nu\sigma^2(1+\rho_N)\) by
\eqref{eqn:fold-event-bounds}.
Lemma~\ref{lem:held-out-pair} with \(\mPsi=\foldmap_q\) and
\(\mathcal E'=\mathcal E_q\) therefore shows that, except on an event of
probability at most \(8N^{-\nu-5}\), the following hold on
\(\mathcal E_q\cap\{q(i,j)=q\}\):
\begin{align}
 \|\foldmap_q^{\sT}(\noise_i-\noise_j)\|
 &\leq C_\nu\sigma\sqrt{1+\rho_N}\,\bigl(\sqrt m+\sqrt{\log N}\bigr),
 \label{eqn:three-fold-held-out-noise}\\
 \bigl|\|\foldmap_q^{\sT}(\vx_i-\vx_j)\|^2-\Tr\mV-\|\vu\|^2\bigr|
 &\leq \eta_1\|\vu\|+\eta_2,
 \label{eqn:debiased-squared-norm-concentration}
\end{align}
with \(\eta_1=C_\nu\sigma\sqrt{(1+\rho_N)\log N}\) and
\(\eta_2=C_\nu\sigma^2(1+\rho_N)(\sqrt{m\log N}+\log N)\). A union bound
over the three folds and the fewer than \(N^2\) pairs, with
\(\P(\bigcup_q\mathcal E_q^c)\leq C_\nu N^{-\nu-3}\), gives an event
\(\mathcal E_*\subseteq\bigcap_q\mathcal E_q\) of probability at least
\(1-C_\nu N^{-\nu}\) on which both bounds hold for every pair \(i<j\) with
\(q=q(i,j)\). On \(\mathcal E_*\), every fold selects \(\widetilde m_q=m\),
and \(|\|\vu\|-D^{(\mathrm s)}_{ij}|\leq C_\nu\widetilde a_ND^{(\mathrm s)}_{ij}\)
by \eqref{eqn:three-fold-clean-distance-distortion}.

\smallskip
\noindent\emph{Part~(b) without active debiasing.}
Let \(\mathtt{debias}=\mathtt{false}\) or \(p-m<20N\). On
\(\mathcal E_*\), no fold is debiased, and
\eqref{eqn:denoising-error-decomposition},
\eqref{eqn:three-fold-clean-distance-distortion}, and
\eqref{eqn:three-fold-held-out-noise} show that
\(\mathcal E_{\mathrm D}(C_\nu\widetilde a_N,C_\nu\widetilde b_N)\) holds for
\(\mD=\widetilde\mD\), with \(\widetilde b_N\) from the first case of
\eqref{eqn:denoising-additive-error}.

\smallskip
\noindent\emph{Part~(b) with active debiasing.}
Let \(\mathtt{debias}=\mathtt{true}\) and \(p-m\geq20N\), let
\(\varepsilon\coloneqq\varepsilon_{\mathrm{iso}}\leq1/2\), and put
\(\mE\coloneqq\mGamma-\bar\sigma^2\mI_p\), so that
\(\|\mE\|=\varepsilon\sigma^2\). Since
\(\sigma^2\leq\bar\sigma^2+\|\mE\|\), we have
\(\bar\sigma^2\geq\sigma^2/2\). Hence
\((p-m)\sigma^2\leq2\Tr\mGamma_\perp\leq2N\theta_{\min}\rho_N\) and
\(\|\mGamma_\perp\|_{\frob}\geq\Tr\mGamma_\perp/\sqrt{p-m}\geq\sigma^2\sqrt{p-m}/2\),
so that
\begin{align}
 \frac{(p-m)\sigma^2}{N\theta_{\min}}\leq2\rho_N,
 \qquad
 \frac{\sigma^2}{\theta_{\min}}\Bigl(1+\sqrt{\frac{p-m}{N}}\Bigr)\leq3\widetilde a_N.
 \label{eqn:near-isotropic-rates}
\end{align}
For \(q=1,2,3\), let \(\mathcal E_q^{\mathrm{iso}}\in\mathcal F_q\) be the
event on which the two bounds of Lemma~\ref{lem:weighted-fold-noise} hold
for the noise of fold~\(q\), with \(t=(\nu+5)\log N\) and with the
deterministic matrices \(\mB\) and \(\mW\) below. Then
\(\P(\bigcup_q(\mathcal E_q^{\mathrm{iso}})^c)\leq C N^{-\nu-5}\), and we
work on \(\mathcal E_*\cap\bigcap_q\mathcal E_q^{\mathrm{iso}}\), which has
probability at least \(1-C_\nu N^{-\nu}\).

On this event, every fold is debiased, with
\(\widehat\sigma_q^2=\lambda_{\min}(\mG_q)/(p-m)\). By
Lemma~\ref{lem:fold-event}(ii), \(\|\mGamma_\perp\|_{\frob}\leq\sigma^2\sqrt{p-m}\),
and \(N/(p-m)\leq1/20\),
\[
 |\widehat\sigma_q^2-\bar\sigma^2|
 =\frac{|\lambda_{\min}(\mG_q)-\Tr\mGamma_\perp|}{p-m}
 \leq C_\nu\sigma^2\sqrt{\frac{N}{p-m}}.
\]
The trace in \eqref{eqn:debiased-squared-norm-concentration} is
\(\Tr\mV=2\Tr(\foldmap_q^{\sT}\mGamma\foldmap_q)\), and the
debiased distance subtracts \(2\widehat\sigma_q^2\Tr(\foldmap_q^{\sT}\foldmap_q)\).
Their difference is
\[
 \Tr\mV-2\widehat\sigma_q^2\Tr(\foldmap_q^{\sT}\foldmap_q)
 =2\Tr(\foldmap_q^{\sT}\mE\foldmap_q)
  +2(\bar\sigma^2-\widehat\sigma_q^2)\Tr(\foldmap_q^{\sT}\foldmap_q).
\]
Since
\(\Tr(\foldmap_q^{\sT}\foldmap_q)\leq m\|\foldmap_q\|^2\leq C_\nu(1+\rho_N)m\),
the second term is at most
\[
 \eta_3\coloneqq2|\widehat\sigma_q^2-\bar\sigma^2|\Tr(\foldmap_q^{\sT}\foldmap_q)
 \leq C_\nu\sigma^2(1+\rho_N)\,m\sqrt{\frac{N}{p-m}}.
\]
For the first term, write
\(\foldmap_q=\mJ\mH_q^{\sT}+\mJ_\perp\foldmap_{q,\perp}\), with
\(\mH_q=\foldmap_q^{\sT}\mJ\) from the proof of Lemma~\ref{lem:fold-event}
and
\(\foldmap_{q,\perp}=\mJ_\perp^{\sT}\foldmap_q=\mJ_\perp^{\sT}\mXi_q\widetilde\mU_q\widetilde\mLambda_q^{-1/2}\),
since the clean signals have no component along \(\mJ_\perp\). With
\(\mE_\perp=\mJ_\perp^{\sT}\mE\mJ_\perp\),
\[
 \Tr(\foldmap_q^{\sT}\mE\foldmap_q)
 =\underbrace{\Tr(\mH_q\mJ^{\sT}\mE\mJ\mH_q^{\sT})}_{T_1}
 +2\underbrace{\Tr(\mH_q\mJ^{\sT}\mE\mJ_\perp\foldmap_{q,\perp})}_{T_2}
 +\underbrace{\Tr(\foldmap_{q,\perp}^{\sT}\mE_\perp\foldmap_{q,\perp})}_{T_3}.
\]
By Lemma~\ref{lem:fold-event}(iii), \(\|\mH_q\|\leq2\), so
\(|T_1|\leq m\|\mH_q\|^2\|\mE\|\leq4m\varepsilon\sigma^2\).
By \eqref{eqn:three-fold-corrected-positivity},
\(\|\widetilde\mLambda_q^{-1/2}\|\leq(cN\theta_{\min})^{-1/2}\) and
\(\Tr\widetilde\mLambda_q^{-1}\leq m/(cN\theta_{\min})\). Write
\(\mXi_q=\mGamma^{1/2}\mXi_q^{\circ}\), where \(\mXi_q^{\circ}\) collects the
vectors \(\vg_i\), \(i\in F_q\).

For \(T_2\), let
\(\mB=\mJ^{\sT}\mE\mJ_\perp\mJ_\perp^{\sT}\mGamma^{1/2}\in\R^{m\times p}\),
so that \(\|\mB\|\leq\varepsilon\sigma^3\) and
\(\|\mB\|_{\frob}\leq\sqrt m\|\mB\|\). On
\(\mathcal E_q^{\mathrm{iso}}\),
\(\|\mB\mXi_q^{\circ}\|\leq C_\nu\varepsilon\sigma^3\sqrt N\), and therefore
\[
 |T_2|\leq m\|\mH_q\|\,\|\mB\mXi_q^{\circ}\|\,\|\widetilde\mLambda_q^{-1/2}\|
 \leq C_\nu m\varepsilon\sigma^2\Bigl(\frac{\sigma^2}{\theta_{\min}}\Bigr)^{1/2}
 \leq C_\nu m\varepsilon\sigma^2\widetilde a_N^{1/2}.
\]
For \(T_3\), let
\(\mW=\mGamma^{1/2}\mJ_\perp\mE_\perp\mJ_\perp^{\sT}\mGamma^{1/2}\),
so that
\(\foldmap_{q,\perp}^{\sT}\mE_\perp\foldmap_{q,\perp}
=\widetilde\mLambda_q^{-1/2}\widetilde\mU_q^{\sT}((\mXi_q^{\circ})^{\sT}\mW\mXi_q^{\circ})\widetilde\mU_q\widetilde\mLambda_q^{-1/2}\).
Since \(\mJ_\perp^{\sT}\mGamma\mJ_\perp=\mE_\perp+\bar\sigma^2\mI\)
and \(\Tr\mE_\perp=\Tr\mGamma_\perp-(p-m)\bar\sigma^2=0\),
\[
 \Tr\mW=\Tr\{\mE_\perp(\mE_\perp+\bar\sigma^2\mI)\}=\|\mE_\perp\|_{\frob}^2
 \leq(p-m)\varepsilon^2\sigma^4,
 \qquad
 \|\mW\|\leq\varepsilon\sigma^4,
 \qquad
 \|\mW\|_{\frob}\leq\varepsilon\sigma^4\sqrt{p-m}.
\]
On \(\mathcal E_q^{\mathrm{iso}}\),
\(\|(\mXi_q^{\circ})^{\sT}\mW\mXi_q^{\circ}\|\leq(p-m)\varepsilon^2\sigma^4
+C_\nu\varepsilon\sigma^4(\sqrt{(p-m)N}+N)\). Hence, by
\eqref{eqn:near-isotropic-rates},
\[
 |T_3|\leq\Tr(\widetilde\mLambda_q^{-1})\,\|(\mXi_q^{\circ})^{\sT}\mW\mXi_q^{\circ}\|
 \leq C_\nu m\varepsilon\sigma^2\Bigl\{\frac{\varepsilon(p-m)\sigma^2}{N\theta_{\min}}
   +\frac{\sigma^2}{\theta_{\min}}\Bigl(1+\sqrt{\frac{p-m}{N}}\Bigr)\Bigr\}
 \leq C_\nu m\varepsilon\sigma^2(\rho_N+\widetilde a_N).
\]
Since \(\widetilde a_N\to0\), the first term is therefore at most
\[
 \eta_4\coloneqq2|\Tr(\foldmap_q^{\sT}\mE\foldmap_q)|
 \leq C_\nu\,m\,\varepsilon_{\mathrm{iso}}\,\sigma^2(1+\rho_N).
\]
With \eqref{eqn:debiased-squared-norm-concentration}, the squared distance
in \eqref{eqn:debiased-distance} before truncation is within
\(\eta_1\|\vu\|+\eta_2+\eta_3+\eta_4\) of \(\|\vu\|^2\).
Lemma~\ref{lem:square-root-perturbation} then gives
\[
 \bigl|\widetilde D_{ij}-\|\vu\|\bigr|
 \leq\eta_1+\sqrt{\eta_2+\eta_3+\eta_4}
 \leq C_\nu\sigma\sqrt{1+\rho_N}\,
   \Bigl\{(m\log N)^{1/4}+\sqrt{\log N}
      +\sqrt m\,\Bigl(\frac{N}{p-m}\Bigr)^{1/4}
      +\sqrt{m\varepsilon_{\mathrm{iso}}}\Bigr\},
\]
and \eqref{eqn:three-fold-clean-distance-distortion} adds at most
\(C_\nu\widetilde a_ND^{(\mathrm s)}_{ij}\). Hence
\(\mathcal E_{\mathrm D}(C_\nu\widetilde a_N,C_\nu\widetilde b_N)\) holds for
\(\mD=\widetilde\mD\), with \(\widetilde b_N\) from the second case of
\eqref{eqn:denoising-additive-error}. This completes part~(b).
\end{proof}

\paragraph{The anisotropy bias}
The term \(\sqrt{m\varepsilon_{\mathrm{iso}}}\) in
\eqref{eqn:denoising-additive-error} is a bias. The trace to be removed,
\(2\Tr(\foldmap_q^{\sT}\mGamma\foldmap_q)\), involves the noise
variance in the fitted directions, whereas \(\widehat\sigma_q^2\)
estimates \(\bar\sigma^2\). By the proof above, their difference is at most
of order \(m\varepsilon_{\mathrm{iso}}\sigma^2(1+\rho_N)\), and its leading
part, \(2\{\Tr(\mJ^{\sT}\mGamma\mJ)-m\bar\sigma^2\}\), does not depend
on the data. If, for instance,
\(\mJ^{\sT}\mGamma\mJ=(1+\varepsilon)\bar\sigma^2\mI_m\), it
shifts every debiased squared distance by about
\(2m\varepsilon\bar\sigma^2\), so the condition on
\(\varepsilon_{\mathrm{iso}}\) in \eqref{eqn:debiasing-gain-condition}
cannot be relaxed for this estimator. No correction computed from the
covariance of the observations can remove this shift, since
\(\mJ^{\sT}\Cov(\vx_j)\mJ=\mSigma+\mJ^{\sT}\mGamma\mJ\) does not
separate the two terms.

\subsection{Proof of Theorem~\ref{thm:coordinate-reconstruction-lower-bound}}
\label{app:minimax-proof}
We prove the lower bound within the multi-community model of
Section~\ref{sec:multi-community-manifolds}, for every fixed number of
communities.

\paragraph{The class of laws}
Fix \(B\geq1\), \(d\geq1\), and weights \(\alpha_b>0\) with
\(\sum_b\alpha_b=1\), and, for each \(N\), the parameters
\(\mJ\in\R^{p\times m}\) with \(\max\{2,d\}\leq m\leq p\),
\(\sigma\geq0\), and \(\covscale_N>0\) of the class \(\gP_N\). Let
\(\gP_N^{(B)}\) be the class of laws of the model of
Section~\ref{sec:multi-community-manifolds} with these parameters and
isotropic Gaussian noise of level \(\sigma\) such that
\begin{enumerate}[label=(\roman*)]
\item condition~(i) of the class \(\gP_N\) holds;
\item Assumption~\ref{ass:local-mass-regularity} holds with constants
\((c_{\mathrm g},C_{\mathrm g})\);
\item each reference law \(\mathds{Q}_b\) has a density \(f_b\) with
respect to the \(d\)-dimensional volume measure on \(\manifold_b\), with
\(c_f\leq f_b\leq C_f\) and
\(|f_b(\vu)-f_b(\vv)|\leq\Lambda_f\|\vu-\vv\|\) for all
\(\vu,\vv\in\manifold_b\), for constants \(0<c_f\leq C_f<\infty\) and
\(0\leq\Lambda_f<\infty\).
\end{enumerate}
\begin{proposition}[Lower bound within the multi-community model]
\label{prop:minimax-communities}
There are constants \(0<c_0\leq1\leq C_0<\infty\), depending only on \(d\)
and \(\alpha_{\min}\), with the following property. If
\(\covscale_N\gg\sigma^2\log N\), \(c_f\leq1\), \(c_{\mathrm g}\leq c_0\),
and \(C_f,C_{\mathrm g}\geq C_0\), then
\[
 \inf_{\widehat\mD}\sup_{P\in\gP_N^{(B)}}
    \P_P\bigl\{\|\widehat\mD-\mD^{(\mathrm{s})}\|_{\max}
       \geq\psi_N/64\bigr\}\longrightarrow1.
\]
\end{proposition}

\paragraph{Reduction to the proposition}
A law with a single community satisfies
Assumption~\ref{ass:regular-clean-law} with the constants and the locality
radius of that community, so \(\gP_N^{(1)}\subseteq\gP_N\), and
Theorem~\ref{thm:coordinate-reconstruction-lower-bound} is the case
\(B=1\) of the proposition. For \(\sigma=0\) the bound is trivial, so
assume \(\sigma>0\). It suffices to find laws \(P_N\in\gP_N^{(B)}\) with
\begin{align}
 \sup_{\widehat\mD}\P_{P_N}\bigl\{\|\widehat\mD-\mD^{(\mathrm{s})}\|_{\max}
                      <\psi_N/64\bigr\}\longrightarrow0.
 \label{eqn:coordinate-reconstruction-lower-bound}
\end{align}
The bound holds even for estimators that know \(\mJ\), \(\sigma\), and
\(P_N\); since a randomized rule is an average of deterministic ones, we
bound deterministic rules.

\paragraph{Reductions of the observations}
The signals of the laws below lie in the range of \(\mJ\), and the \(p-m\)
coordinates orthogonal to it are pure noise, which an estimator can
simulate; so it suffices to bound estimators that observe
\(\mJ^{\sT}\vx_i\in\R^m\). Distances are invariant under translation, so
the laws below need to be centered only up to a known translation.

\paragraph{Idea of the proof}
Let \(\delta_0=\frac14\sigma\max\{\sqrt{\log N},(m\log N)^{1/4}\}\), and,
for \(\covscale>0\), let \(\va\) have the law
\begin{align}
 \va=\begin{cases}
 \bzero & \text{with probability }1/4\quad\text{(null)},\\
 (\delta_0/\sqrt m)\,\boldsymbol{\varepsilon},\
   \boldsymbol{\varepsilon}\in\{-1,1\}^m
   & \text{each with probability }2^{-m}/4\quad\text{(weak)},\\
 \pm\sqrt{2m\covscale}\,(\mI_m)_{\cdot j},\ j\in[m]
   & \text{each with probability }1/(4m)\quad\text{(calibration)}.
 \end{cases}
 \label{eqn:hard-finite-law}
\end{align}
A null and a weak atom are at distance \(\delta_0\asymp\psi_N\),
but no test tells them apart from one noisy observation with error below
\(\frac14N^{-1/32}\). The two scales in \(\delta_0\) are the two
terms of \(\psi_N\): at \(\sigma\sqrt{\log N}\) a single weak atom is hard
to detect, and at \(\sigma(m\log N)^{1/4}\) the mixture over the \(2^m\)
weak atoms is. About half of the points are null or weak, so, with
probability tending to one, every estimator misclassifies one of them, and its distance to a null point then
has error at least \(\delta_0/4\). The calibration atoms
carry almost all of the variance, so that
\(\theta_{\min},\theta_{\max}\asymp\covscale\).
Lemma~\ref{lem:growing-coordinate-product-testing} makes this precise,
using the testing facts of Lemma~\ref{lem:two-point-testing} in
Appendix~\ref{app:technical-lemmas}.
Lemma~\ref{lem:finite-law-approximation} replaces the finite law by a law
of the model whose points are close to the atoms, and
Lemma~\ref{lem:growing-coordinate-gaussian-transfer} shows that the
failure persists. Appendix~\ref{app:hard-law-construction} proves
Lemma~\ref{lem:finite-law-approximation}, by smoothing the finite law into
a curve and thickening the curve into \(B\) communities of dimension
\(d\).

\begin{lemma}[A hard finite law]
\label{lem:growing-coordinate-product-testing}
Let \(m\geq2\), \(\sigma>0\), and \(\covscale>0\), let \(\va\) have the law
\eqref{eqn:hard-finite-law}, and observe \(\va_i+\sigma\vg_i\) for iid
copies \(\va_1,\ldots,\va_N\) of \(\va\) and iid
\(\vg_i\sim\Normal(\bzero,\mI_m)\). Then
\[
 \sup_{\widehat\mD}\P\Bigl\{\max_{i<j}
   \bigl|\widehat D_{ij}-\|\va_i-\va_j\|\bigr|<\delta_0/4\Bigr\}
 \leq(3/4)^{\lfloor N/2\rfloor}+\mathrm e^{-N/32}
   +\exp\{-N^{31/32}/32\}.
\]
\end{lemma}

\begin{proof}[Proof of Lemma~\ref{lem:growing-coordinate-product-testing}]
\emph{Step 1: one point.} Let \(P_0=\Normal(\bzero,\sigma^2\mI_m)\) and
let \(P_1\) be the uniform mixture, over
\(\boldsymbol{\varepsilon}\in\{-1,1\}^m\), of
\(\Normal(\delta_0 m^{-1/2}\boldsymbol{\varepsilon},\,\sigma^2\mI_m)\):
the laws of one observation of a null and of a weak point. For
independent uniform \(\boldsymbol{\varepsilon},\boldsymbol{\varepsilon}'\), a
Gaussian computation and \(\cosh y\leq\mathrm e^{y^2/2}\) give
\[
 1+\chi^2(P_1,P_0)
 =\E\exp\{\delta_0^2\langle\boldsymbol{\varepsilon},\boldsymbol{\varepsilon}'\rangle/(m\sigma^2)\}
 =\cosh(\delta_0^2/(m\sigma^2))^m
 \leq\exp\{\delta_0^4/(2m\sigma^4)\}.
\]
Each component of \(P_1\) has relative entropy
\(\delta_0^2/(2\sigma^2)\) with respect to \(P_0\), so
Lemma~\ref{lem:two-point-testing}(b), with the first term if
\(m\leq\log N\) and the second otherwise, and
Lemma~\ref{lem:two-point-testing}(a) give
\begin{align}
 \operatorname{KL}(P_1\Vert P_0)
 \leq\min\left\{\frac{\delta_0^2}{2\sigma^2},\frac{\delta_0^4}{2m\sigma^4}\right\}
 \leq\frac{\log N}{32},
 \qquad
 \beta_N\geq\tfrac14\mathrm e^{-\operatorname{KL}(P_1\Vert P_0)}
 \geq\tfrac14N^{-1/32},
 \label{eqn:growing-coordinate-testing-kl}
\end{align}
where \(\beta_N\) is the smallest balanced error of a test of \(P_0\)
against \(P_1\).

\emph{Step 2: many independent points.} Reveal the type of the first
\(n_1=\lfloor N/2\rfloor\) points and which of the remaining points are
calibration points; revealing information can only help the estimator. Let
\(\mathcal E_{\mathrm{bit}}\) be the event that some revealed point
\(i_0\) is null and that the number \(n_{\mathrm{bit}}\) of remaining null or
weak points is at least \(N/8\). Since
\(n_{\mathrm{bit}}\sim\operatorname{Bin}(N-n_1,1/2)\) has mean at least
\(N/4\), \eqref{eqn:binomial-half-mean} of Lemma~\ref{lem:binomial-tails} gives
\[
 \P(\mathcal E_{\mathrm{bit}}^c)\leq(3/4)^{n_1}+\mathrm e^{-N/32}.
\]
Given the revealed information, these \(n_{\mathrm{bit}}\) points carry iid
balanced null/weak bits, observed independently through \(P_0\) or
\(P_1\). The posterior law of the bits is therefore a product over the
points, so the rule that maximizes the probability of recovering all of
them decides each bit from its own observation, and it is correct on each
bit with probability at most \(1-\beta_N\). By
\eqref{eqn:growing-coordinate-testing-kl}, every rule thus recovers all
bits with probability at most
\((1-\beta_N)^{n_{\mathrm{bit}}}\leq\exp(-N^{31/32}/32)\) on
\(\mathcal E_{\mathrm{bit}}\). If all distance errors are below
\(\delta_0/4\), thresholding \(\widehat D_{i_0j}\) at
\(\delta_0/2\) recovers every bit, since
\(\|\va_{i_0}-\va_j\|\in\{0,\delta_0\}\). Hence the success
probability is at most
\(\P(\mathcal E_{\mathrm{bit}}^c)+\exp(-N^{31/32}/32)\).
\end{proof}

\begin{lemma}[Approximating a finite law by a law of the model]
\label{lem:finite-law-approximation}
Let \(1\leq d\leq m\) with \(m\geq2\), let \(B\geq1\), and let
\(\alpha_b>0\) with \(\sum_b\alpha_b=1\). Let \(\va\) have a finite law
supported on \(M\geq2\) distinct points of \(\R^m\) of norm at most \(R\).
For every \(\eta\in(0,1]\),
there are a random vector \(\vs_0\in\R^m\), coupled with \(\va\), and a
law of the model of Section~\ref{sec:multi-community-manifolds} in
\(\R^m\), with \(B\) communities of dimension \(d\) and weights
\(\alpha_b\), such that:
\begin{enumerate}[label=(\alph*)]
\item the clean signal of the law is \(\vs_0-\E\vs_0\);
\item \(\P\{\|\vs_0-\va\|>\eta\}\leq\eta\)
and \(\|\vs_0\|\leq R+\eta\) almost surely;
\item the reference laws are uniform, with densities \(f_b=1/\alpha_b\),
and Assumption~\ref{ass:local-mass-regularity} holds with constants
\(c_0\in(0,1]\) and \(C_0<\infty\) that depend only on \(d\) and
\(\alpha_{\min}\);
\item the locality radius \(R_{\mathrm g}\) of~(c) satisfies
\((R_{\mathrm g}/L)^d\leq C\eta\), and
\((\Delta/L)^d\leq C\eta\) when \(B\geq2\), with \(C\)
depending only on \(d\).
\end{enumerate}
\end{lemma}
The proof is in Appendix~\ref{app:hard-law-construction}.

\begin{lemma}[Transfer between coupled laws]
\label{lem:growing-coordinate-gaussian-transfer}
Let \(\sigma>0\), and let \(\va\) and \(\vs_0\) be random vectors in
\(\R^m\) coupled so that
\(\P\{\|\vs_0-\va\|>\eta\}\leq\eta\), let
\((\va_i,\vs_{0,i})\), \(i\in[N]\), be iid copies of this coupling, and
consider the experiments with observations \(\va_i+\sigma\vg_i\) and
\(\vs_{0,i}+\sigma\vg_i\). For every \(t>0\),
\[
 \sup_{\widehat\mD}\P\Bigl\{\max_{i<j}
   \bigl|\widehat D_{ij}-\|\vs_{0,i}-\vs_{0,j}\|\bigr|<t\Bigr\}
 \leq\sup_{\widehat\mD}\P\Bigl\{\max_{i<j}
   \bigl|\widehat D_{ij}-\|\va_i-\va_j\|\bigr|<t+2\eta\Bigr\}
   +2N\eta+\frac{N\eta}{2\sigma}.
\]
\end{lemma}

\begin{proof}[Proof of Lemma~\ref{lem:growing-coordinate-gaussian-transfer}]
Given \((\va_i,\vs_{0,i})\), couple the two observations of point \(i\)
as in Lemma~\ref{lem:two-point-testing}(c); they differ with probability
at most \(\|\vs_{0,i}-\va_i\|/(2\sigma)\), hence unconditionally with
probability at most \(\eta+\eta/(2\sigma)\).
Off an event of probability at most
\(2N\eta+N\eta/(2\sigma)\), all observations
agree and all target distances differ by at most \(2\eta\).
Applying the same estimator to both observation arrays proves the claim.
\end{proof}

\begin{proof}[Proof of Proposition~\ref{prop:minimax-communities}]
\emph{Step 1: finite law and approximation scales.} Let \(\va\) have the
law \eqref{eqn:hard-finite-law} with \(\covscale=\covscale_N\). It is
centered, and
\begin{align}
 \operatorname{Cov}(\va)
   =\left(\covscale_N+\frac{\delta_0^2}{4m}\right)\mI_m.
 \label{eqn:growing-coordinate-calibration-covariance}
\end{align}
Since \(\delta_0^2/m\leq\sigma^2\log N/16=o(\covscale_N)\),
all atoms lie in the ball of radius \(R=\sqrt{2m\covscale_N}\) for large
\(N\). Choose \(\eta\in(0,1]\) with
\begin{align}
 \frac{N\eta}{\sigma}+N\eta\longrightarrow0,
 \qquad\eta\leq\frac{\delta_0}{64},
 \qquad R\eta+\eta R^2=o(\covscale_N).
 \label{eqn:growing-coordinate-approximation-scales}
\end{align}

\emph{Step 2: a law in \(\gP_N^{(B)}\).} Apply
Lemma~\ref{lem:finite-law-approximation} to \(\va\) with these scales, and
let \(P_N\) be the resulting law, with clean signal
\(\vs=\vs_0-\E\vs_0\). It is centered, and by part~(c) of the lemma it
satisfies the conditions of \(\gP_N^{(B)}\) on the manifolds, the
densities, and the local mass regularity. For the rest of condition~(i),
write
\(\vs_0\vs_0^{\sT}-\va\va^{\sT}=(\vs_0-\va)\vs_0^{\sT}+\va(\vs_0-\va)^{\sT}\) on the
coupling event, and bound both outer products by
\((R+\eta)^2\) off it:
\[
 \|\E\vs_0\|\leq\eta+3\eta R,
 \qquad
 \|\operatorname{Cov}(\vs_0)-\operatorname{Cov}(\va)\|
 \leq C(R\eta+\eta R^2)=o(\covscale_N).
\]
With \eqref{eqn:growing-coordinate-calibration-covariance}, this gives
\(\frac12\covscale_N\leq\theta_{\min}\leq\theta_{\max}\leq2\covscale_N\)
for large \(N\), and
\(\|\vs\|\leq2(R+\eta)\leq3\sqrt{m\covscale_N}\). By the
reductions, it suffices to bound estimators that observe
\(\vs_{0,i}+\sigma\vg_i\in\R^m\).

\emph{Step 3: conclusion.} Since
\(\delta_0/8+2\eta<\delta_0/4\),
Lemma~\ref{lem:growing-coordinate-gaussian-transfer} and
\eqref{eqn:growing-coordinate-approximation-scales} bound the success
probability at tolerance \(\delta_0/8\) by the finite-law
success probability at tolerance \(\delta_0/4\) plus \(o(1)\).
Lemma~\ref{lem:growing-coordinate-product-testing} makes the latter tend
to zero. Since \(8\delta_0\geq\psi_N\), this proves
\eqref{eqn:coordinate-reconstruction-lower-bound}.
\end{proof}

\paragraph{Regime of the hard laws}
By part~(d) of
Lemma~\ref{lem:finite-law-approximation} and
\(N\eta\to0\), the law \(P_N\) has
\((R_{\mathrm g}/L)^d=o(1/N)\), and \((\Delta/L)^d=o(1/N)\) when
\(B\geq2\). Its locality radius and its separation are therefore below
the sampling resolution \(h_N(\log N)=L(\log N/N)^{1/d}\).

\subsection{Construction of the hard law}
\label{app:hard-law-construction}
This subsection proves Lemma~\ref{lem:finite-law-approximation}.
Lemma~\ref{lem:growing-coordinate-curve-approximation} smooths a finite
law into a curve, Lemma~\ref{lem:curve-thickening} thickens a curve into
\(B\) communities of dimension \(d\), and
Lemma~\ref{lem:finite-law-approximation} follows by composing the two.

\begin{lemma}[Smoothing the finite law into a curve]
\label{lem:growing-coordinate-curve-approximation}
Let \(m\geq2\), and let \(\va\) have a finite law supported on
\(M\geq2\) distinct points \(\va_1,\ldots,\va_M\in\R^m\), with
probabilities \(w_j>0\) and \(\|\va_j\|\leq R\). For every \(\eta>0\),
there is a compact connected \(C^\infty\) embedded curve \(\Gamma\subset\R^m\)
with two endpoints whose uniform arclength point \(\vs^\Gamma\) can be
coupled with \(\va\) so that
\[
 \P\{\|\vs^\Gamma-\va\|>\eta\}\leq\eta,
 \qquad \sup_{\vs\in\Gamma}\|\vs\|\leq R+\eta.
\]
\end{lemma}

The next lemma thickens a curve into \(B\) communities of dimension
\(d\). Write \(|\mathbb B^k|\) for the volume of the unit ball of \(\R^k\),
with \(|\mathbb B^0|=1\), \(\mathbb B(\vz,r)\) for the closed ball of
radius \(r\) around \(\vz\), and \(\operatorname{Vol}\) for
\(d\)-dimensional volume.

\begin{lemma}[Thickening a curve into communities]
\label{lem:curve-thickening}
Let \(\gamma:[0,\ell]\to\R^m\) be an arclength parameterization of a
compact connected \(C^\infty\) embedded curve \(\Gamma\) with two
endpoints. Let \(\kappa>0\) bound its
curvature, and let
\(s_\Gamma\coloneqq\inf\{\|\gamma(t)-\gamma(t')\|:|t-t'|\geq1/(40\kappa)\}\),
which is positive, and infinite if there is no such pair. Let \(1\leq d\leq m\), \(B\geq1\), and weights
\(\alpha_b>0\) with \(\sum_b\alpha_b=1\), and let \(\varepsilon,g>0\) satisfy
\begin{align}
 40\kappa\varepsilon\leq1,\qquad 4\varepsilon\leq s_\Gamma,\qquad
 B(g+2\varepsilon)\leq\ell/2,\qquad 8\varepsilon\leq\alpha_{\min}\ell.
 \label{eqn:thickening-conditions}
\end{align}
Then there are disjoint compact connected \(d\)-dimensional \(C^\infty\)
embedded submanifolds \(\mathcal T_1,\ldots,\mathcal T_B\subset\R^m\),
possibly with boundary, and \(B\) parameters in \([0,\ell)\), the
\emph{cuts}, with the following properties. Let
\(\mathcal T=\bigcup_b\mathcal T_b\), and let \(\vs^{\mathcal T}\) be uniform on
\(\mathcal T\).
\begin{enumerate}[label=(\alph*)]
\item \emph{Volumes.} \(\operatorname{Vol}(\mathcal T_b)=\alpha_b\operatorname{Vol}(\mathcal T)\)
and
\(\frac12|\mathbb B^{d-1}|\varepsilon^{d-1}\ell\leq\operatorname{Vol}(\mathcal T)\leq|\mathbb B^{d-1}|\varepsilon^{d-1}\ell\).
\item \emph{Structure.} \(\vs^{\mathcal T}=\gamma(t)+\mN(t)\vq\) for a random
parameter \(t\) and a random \(\vq\in\R^{d-1}\) with
\(\|\vq\|\leq\varepsilon\), where \(\mN(t)\in\R^{m\times(d-1)}\) has
orthonormal columns orthogonal to \(\gamma'(t)\). The law of \(t\) is
within total variation \(B(g+2\varepsilon)/\ell\) of the uniform law on
\([0,\ell]\). In particular, \(\vs^{\mathcal T}\) and the uniform arclength point
\(\vs^\Gamma\) of \(\Gamma\) can be coupled so that
\(\P\{\|\vs^{\mathcal T}-\vs^\Gamma\|>\varepsilon\}\leq B(g+2\varepsilon)/\ell\).
\item \emph{Model.} Let \(L=\operatorname{Vol}(\mathcal T)^{1/d}\), let
\(\rvmu_b\) be the mean of the uniform law on \(\mathcal T_b\), and let
\(\rvmu\) be the mean of \(\vs^{\mathcal T}\). The reference manifolds
\((\mathcal T_b-\rvmu_b)/L\) with their uniform laws, the locations
\(\rvmu_b-\rvmu\), the dilation \(L\), and the weights \(\alpha_b\) define a
law of the model with clean signal \(\vs^{\mathcal T}-\rvmu\). Its densities
are \(f_b=1/\alpha_b\), and Assumption~\ref{ass:local-mass-regularity} holds
with \(R_{\mathrm g}=c\min\{\varepsilon,s_\Gamma\}\) for an absolute
constant \(c\in(0,1/8]\) and with constants \(c_0\in(0,1]\) and \(C_0<\infty\)
depending only on \(d\) and \(\alpha_{\min}\). Moreover,
\(\Delta\geq\frac12\min\{g,s_\Gamma\}\), and \(\Delta\leq g\) when
\(B\geq2\).
\end{enumerate}
\end{lemma}

\begin{proof}[Proof of Lemma~\ref{lem:finite-law-approximation}]
Apply Lemma~\ref{lem:growing-coordinate-curve-approximation} with
\(\eta/2\) to obtain a
curve \(\Gamma\), of length \(\ell\), with a curvature bound \(\kappa\)
and the constant \(s_\Gamma\) of Lemma~\ref{lem:curve-thickening}. Then
apply Lemma~\ref{lem:curve-thickening} to \(\Gamma\) with
\[
 \varepsilon=g=\min\Bigl\{\frac{\eta}2,\
 \frac{\eta\ell}{6B},\ \frac1{40\kappa},\
 \frac{s_\Gamma}4,\ \frac{\alpha_{\min}\ell}8\Bigr\},
\]
which satisfies \eqref{eqn:thickening-conditions} since
\(\eta\leq1\). Let \(\vs_0=\vs^{\mathcal T}\), coupled with
\(\va\) by composing the coupling of \(\va\) with \(\vs^\Gamma\) and that
of \(\vs^\Gamma\) with \(\vs^{\mathcal T}\). Parts~(a) and~(c) are
part~(c) of Lemma~\ref{lem:curve-thickening}. For~(b), the two couplings
fail with probability at most
\(\eta/2+3B\varepsilon/\ell\leq\eta\), and
otherwise
\(\|\vs_0-\va\|\leq\eta/2+\varepsilon\leq\eta\);
moreover
\(\|\vs_0\|\leq\sup_{\vs\in\Gamma}\|\vs\|+\varepsilon\leq R+\eta\).
For~(d), \(R_{\mathrm g}\leq\varepsilon\), and \(\Delta\leq g=\varepsilon\)
when \(B\geq2\), by part~(c) of Lemma~\ref{lem:curve-thickening}, while its
part~(a) gives
\((\varepsilon/L)^d=\varepsilon^d/\operatorname{Vol}(\mathcal T)
\leq2\varepsilon/(|\mathbb B^{d-1}|\ell)
\leq\eta/(3B|\mathbb B^{d-1}|)\).
\end{proof}

\begin{proof}[Proof of Lemma~\ref{lem:growing-coordinate-curve-approximation}]
\emph{Polygonal arc.} Rotate so that the first coordinate \(x\) takes
distinct values at the atoms, and join consecutive atoms in the order of
\(x\) by segments. The arc is the graph of a piecewise linear map
\(x\mapsto f(x)\in\R^{m-1}\), hence embedded.

\emph{Smoothing.} Choose disjoint balls of radius at most
\(\eta\) around the atoms that meet only their incident
segments, extending the end segments slightly. Inside each ball, smooth
the corner and keep straight entrance and exit collars; call the smoothed
graph \(x\mapsto(x,f_0(x))\).

\emph{Matching the masses.} Inside ball \(j\), replace the graph by
\[
 x\longmapsto\bigl(x,f_0(x)+\varepsilon_0\,\mathfrak b(x)\sin(\mathfrak f x)\,\vu\bigr),
\]
with a fixed unit vector \(\vu\in\R^{m-1}\), a smooth bump \(\mathfrak b\)
supported in the ball's parameter interval and equal to one on a
subinterval, an amplitude \(\varepsilon_0>0\) small enough to keep the
graph in the ball, and a frequency \(\mathfrak f>0\). The length of the arc
in the ball is continuous in \(\mathfrak f\) and grows linearly as
\(\mathfrak f\to\infty\), so for every large \(\ell_\Gamma\) some
\(\mathfrak f\) makes it \(w_j\ell_\Gamma\). The result is a smooth
embedded graph \(\Gamma\) of length \(\ell_\Gamma+L_0\), where \(L_0\) is
the fixed length outside the balls; choose \(\ell_\Gamma\) with
\(L_0\leq\eta(\ell_\Gamma+L_0)\).

\emph{Coupling.} Under arclength measure, the arcs inside the balls carry
mass at least \(1-\eta\), split in the proportions
\(w_j\). Couple a point in ball \(j\) with the atom \(\va_j\), and a point
outside the balls with an independent copy of \(\va\). Then
\(\|\vs^\Gamma-\va\|\leq\eta\) unless \(\vs^\Gamma\) lies
outside the balls, and every point of \(\Gamma\) lies on a segment between
atoms or within \(\eta\) of an atom, which gives the
support bound.
\end{proof}

\begin{proof}[Proof of Lemma~\ref{lem:curve-thickening}]
\emph{Construction.} Remove from \([0,\ell]\) the open intervals of length
\(g\) centered at the cuts, placed below; one cut is at \(0\), and the
interval around it consists of \([0,g/2)\) and
\((\ell-g/2,\ell]\). Let \(\sL_b=[t_b^-,t_b^+]\), \(b\in[B]\), be the remaining
intervals. Fix a nondecreasing \(C^\infty\) function
\(\chi_0:[0,\infty)\to[0,1]\), positive on \((0,\infty)\), with
\(\chi_0(x)=x\) for \(x\leq1/4\) and \(\chi_0(x)=1\) for \(x\geq1\), and put
\(\chi_b(t)=\chi_0((t-t_b^-)/\varepsilon)\,\chi_0((t_b^+-t)/\varepsilon)\) and
\[
 \Omega_b=\bigl\{(t,\vq)\in \sL_b\times\R^{d-1}:\|\vq\|^2\leq\varepsilon^2\chi_b(t)\bigr\},
 \qquad
 F_b(t,\vq)=\gamma(t)+\mN_b(t)\vq,
 \qquad
 \mathcal T_b=F_b(\Omega_b),
\]
with \(\Omega_b=\sL_b\) when \(d=1\). Here \(\mN_b\) is a rotation-minimizing
frame on \(\sL_b\): its columns are orthonormal and orthogonal to \(\gamma'\),
and \(\mN_b'=-\gamma'\vk_b^{\sT}\) with \(\vk_b=\mN_b^{\sT}\gamma''\); it exists
since this linear equation preserves both properties. Then
\(\|\vk_b\|\leq\kappa\) and \(\|\mN_b'\|\leq\kappa\). The boundary of
\(\Omega_b\) is \(C^\infty\): where \(\chi_b>0\) this is clear, and near the
ends of \(\sL_b\) it reads \(|t-t_b^-|=\|\vq\|^2/\varepsilon\) or
\(|t_b^+-t|=\|\vq\|^2/\varepsilon\). After scaling by \(1/\varepsilon\), the
sections of \(\Omega_b\) near its ends have a fixed profile, so
\(\Omega_b\) satisfies an interior ball condition with radius
\(c'\varepsilon\) for an absolute \(c'>0\).

\emph{Local geometry.} Fix \(z_0=(t_0,\vq_0)\in\Omega_b\) and
\(\vv=F_b(z_0)\), and let \(\mF_0=\mathrm dF_b(t_0,\bzero)\), which has
orthonormal columns. On the convex window
\(W=\{(t,\vq):t\in \sL_b,\ |t-t_0|<1/(40\kappa),\ \|\vq\|\leq\varepsilon\}\),
\[
 \|\mathrm dF_b(t,\vq)-\mF_0\|
 \leq\|\gamma'(t)-\gamma'(t_0)\|+\kappa\|\vq\|+\|\mN_b(t)-\mN_b(t_0)\|
 \leq2\kappa|t-t_0|+\kappa\varepsilon\leq\tfrac1{10}.
\]
For \(z=z_0+\lambda\vw\in W\) with \(\vw\) a unit vector, the segment from
\(z_0\) to \(z\) lies in \(W\), so
\(\|F_b(z)-\vv-\lambda\mF_0\vw\|\leq\lambda/10\). Hence
\(\varphi(\lambda)=\|F_b(z_0+\lambda\vw)-\vv\|\) satisfies
\begin{align}
 \tfrac9{10}\lambda\leq\varphi(\lambda)\leq\tfrac{11}{10}\lambda,
 \qquad
 \tfrac12\leq\varphi'(\lambda)\leq\tfrac{11}{10}:
 \label{eqn:thickening-radial}
\end{align}
the derivative is
\(\langle\mathrm dF_b\vw,(F_b-\vv)/\varphi\rangle\geq\langle\mF_0\vw,(F_b-\vv)/\varphi\rangle-\frac1{10}\geq\frac{9/10}{11/10}-\frac1{10}\).
Outside \(W\), \(\|F_b(t,\vq)-\vv\|\geq\|\gamma(t)-\gamma(t_0)\|-2\varepsilon\geq s_\Gamma/2\).
Thus \(F_b\) is injective, \(\mathcal T_b\) is an embedded \(C^\infty\)
submanifold with boundary, and, for \(r\leq R_{\mathrm g}\leq s_\Gamma/8\), the
preimage of \(\mathbb B(\vv,2r)\) lies in \(W\).

\emph{Volume element.} The partial derivatives
\(\partial_tF_b=(1-\vk_b^{\sT}\vq)\gamma'\) and \(\partial_{\vq}F_b=\mN_b\) have
orthogonal ranges, so the volume element of \(F_b\) is
\(1-\vk_b^{\sT}\vq\in[39/40,41/40]\). Integrating over the ball of
\(\vq\), where the linear term vanishes,
\begin{align}
 \operatorname{Vol}(\mathcal T_b)
 =|\mathbb B^{d-1}|\varepsilon^{d-1}\int_{\sL_b}\chi_b^{(d-1)/2}\,\mathrm dt
 =|\mathbb B^{d-1}|\varepsilon^{d-1}\bigl(|\sL_b|-\lambda_d\varepsilon\bigr),
 \qquad
 \lambda_d=2\int_0^1\bigl(1-\chi_0^{(d-1)/2}\bigr)\in[0,2],
 \label{eqn:thickening-volume}
\end{align}
since \(F_b\) is injective. Choose the cuts so that
\(|\sL_b|-\lambda_d\varepsilon=\alpha_b(\ell-Bg-B\lambda_d\varepsilon)\). Then
part~(a) follows from \eqref{eqn:thickening-volume} and
\(B(g+2\varepsilon)\leq\ell/2\), and \(|\sL_b|\geq\alpha_{\min}\ell/2\geq4\varepsilon\).

\emph{Part~(b).} By the volume element, \(t\) has density proportional to
\(\chi_b(t)^{(d-1)/2}\) on \(\sL_b\), with normalizing constant at most
\(\ell\), so it is at least \(1/\ell\) wherever \(\chi_b=1\). The set where
\(\chi_b<1\) or \(t\) lies in a removed interval has length at most
\(B(g+2\varepsilon)\), which bounds the total variation. For the coupling,
couple \(t\) with a uniform parameter as in
Lemma~\ref{lem:two-point-testing}(c); when they agree,
\(\|\vs^{\mathcal T}-\gamma(t)\|=\|\vq\|\leq\varepsilon\).

\emph{Part~(c).} Let \(r\leq R_{\mathrm g}=c\min\{\varepsilon,s_\Gamma\}\)
with \(c\leq1/8\) small enough that \(r\leq c'\varepsilon\). By
\eqref{eqn:thickening-radial}, the preimage of
\(\mathcal T_b\cap\mathbb B(\vv,r)\) lies between
\(\Omega_b\cap\mathbb B(z_0,10r/11)\) and \(\mathbb B(z_0,10r/9)\), and the
interior ball condition gives
\(|\Omega_b\cap\mathbb B(z_0,10r/11)|\geq|\mathbb B^d|(5r/11)^d\).
Along each ray from \(z_0\), \eqref{eqn:thickening-radial} confines the
preimage of the shell \(\{(1-u)r<\|\vs-\vv\|\leq(1+u)r\}\) to an interval
of length at most \(4ur\) at distance at most \(5r/3\) from \(z_0\). With
the volume element and polar coordinates around \(z_0\),
\[
 \tfrac{39}{40}|\mathbb B^d|\bigl(\tfrac{5r}{11}\bigr)^d
 \leq\operatorname{Vol}\bigl(\mathcal T_b\cap\mathbb B(\vv,r)\bigr)
 \leq\tfrac{41}{40}|\mathbb B^d|\bigl(\tfrac{10r}9\bigr)^d,
 \qquad
 \operatorname{Vol}(\text{shell}\cap\mathcal T_b)
 \leq\tfrac{41}{40}d|\mathbb B^d|\,4ur\bigl(\tfrac{5r}3\bigr)^{d-1}.
\]
The reference manifold \((\mathcal T_b-\rvmu_b)/L\) has volume \(\alpha_b\)
by part~(a), so its uniform density is \(1/\alpha_b\), and
\(\P(\|\clean-\vv\|\leq r\mid\rY=b)=\operatorname{Vol}(\mathcal T_b\cap\mathbb B(\vv,r))/(\alpha_bL^d)\).
The display therefore gives Assumption~\ref{ass:local-mass-regularity} with
\(c_0=\frac{39}{40}|\mathbb B^d|(5/11)^d\leq1\) and
\(C_0=\max\{1,\frac{41}{40}|\mathbb B^d|(10/9)^d,\frac{41}{40}|\mathbb B^d|4d(5/3)^{d-1}\}/\alpha_{\min}\);
these are the constants of Proposition~\ref{prop:minimax-communities},
and \(C_0\geq1/\alpha_{\min}\geq f_b\).
For the separation, let \(\vs=F_b(t,\vq)\) and \(\vt=F_{b'}(t',\vq')\) with
\(b\ne b'\). If \(|t-t'|\geq1/(40\kappa)\), then
\(\|\vs-\vt\|\geq s_\Gamma-2\varepsilon\geq s_\Gamma/2\). Otherwise
\(|t-t'|\geq g\), \(\mN_b(t)^{\sT}\gamma'(t)=\bzero\), and
\(|\langle\mN_{b'}(t')\vq',\gamma'(t)\rangle|\leq\varepsilon\kappa|t-t'|\), so
\(|\langle\vs-\vt,\gamma'(t)\rangle|\geq|t-t'|(1-\frac1{40}-\frac1{40})\geq g/2\).
Conversely, when \(B\geq2\), two consecutive intervals \(\sL_b\) and
\(\sL_{b'}\) are separated by a removed interval of length \(g\), and
their endpoints give points \(\gamma(t_b^+)\in\mathcal T_b\) and
\(\gamma(t_{b'}^-)\in\mathcal T_{b'}\) at distance at most \(g\), so
\(\Delta\leq g\).
\end{proof}

\section{Proofs in Section~\ref{sec:stability-of-data-driven-selection}}
\label{app:radius-neighborhood-proofs}

Throughout this section, constants depend only on \(d\), \(B\),
\(c_\alpha\), \(c_{\mathrm g}\), and \(C_{\mathrm g}\), and \(C_\nu\) may
also depend on \(\nu\); their values may change between occurrences. The
setting, the affine event, and the clean event below are used by all
proofs of this appendix and of
Appendix~\ref{app:spectral-convergence-proofs};
Appendix~\ref{app:clean-event-proof} proves that the clean event holds
with overwhelming probability.

\paragraph{Setting of the proofs}
The proofs of this appendix and of
Appendix~\ref{app:spectral-convergence-proofs} use the model of
Section~\ref{sec:multi-community-manifolds} only through four properties.
\begin{enumerate}[label=(S\arabic*)]
\item The labeled clean signals \((\rY_j,\clean_j)\) are i.i.d.
\item There are \(B\geq1\) communities, with weights
\(\alpha_b\geq c_\alpha\).
\item The community laws \(\P_{\mathrm s}^{(b)}\) have compact supports at
mutual distance at least \(\Delta\), with \(\Delta=+\infty\) when \(B=1\).
\item The community laws satisfy
Assumption~\ref{ass:local-mass-regularity}.
\end{enumerate}
All statements are uniform over the unions of communities
\(\sI_\sB\coloneqq\bigcup_{b\in\sB}\gC_b\),
\(\emptyset\ne\sB\subseteq[B]\). Throughout these two appendices,
\begin{align}
 \locrad\coloneqq\min\{R_{\mathrm g},\Delta/2\},
 \qquad
 C_{\mathrm h}\coloneqq\max\Bigl\{4\mathrm e\,C_{\mathrm g},\
 \frac{16}{c_{\mathrm g}c_\alpha}\Bigr\},
 \label{eqn:proof-setting}
\end{align}
and the valid radii and the admissible neighbor counts
(Definitions~\ref{def:valid-radii} and~\ref{def:admissible-ranks}) are
formed with these two quantities. By
Lemma~\ref{lem:mixture-regularity}, which
Appendix~\ref{app:mixture-regularity-proof} proves, \(\locrad\) is a
locality radius of the clean law. This setting covers the two cases of the main text.
\begin{enumerate}[label=(\roman*)]
\item \emph{Sections~\ref{sec:stability-of-data-driven-selection}
and~\ref{sec:spectral-convergence}.} Under
Assumption~\ref{ass:regular-clean-law}, take \(B=1\), with the clean law
as the only community, its locality radius as \(R_{\mathrm g}\), and
\(c_\alpha=1\). Then \(\sI_\sB=[N]\), and \eqref{eqn:proof-setting} gives
the locality radius of Assumption~\ref{ass:regular-clean-law} and the
band constant \eqref{eqn:radius-band}.
\item \emph{Section~\ref{sec:clustering-guarantees}.} Under
Assumptions~\ref{ass:local-mass-regularity}
and~\ref{ass:locality-separation}, \(B\geq2\), \(\locrad=\Delta/2\), and
\eqref{eqn:proof-setting} is \eqref{eqn:band-constants-communities}.
\end{enumerate}

\paragraph{Admissible neighbor counts}
The band constant \(C_{\mathrm h}\) of \eqref{eqn:proof-setting} satisfies
\(C_{\mathrm h}\geq4\mathrm eC_{\mathrm g}\) and \(c_{\mathrm g}C_{\mathrm h}\geq16/c_\alpha\).
It is chosen so that, at radii up to \(\locrad\), a clean ball of radius
\(C_{\mathrm h}^{-1/d}h_N(k)\) contains in expectation at most
\(k/(4\mathrm e)\) other clean signals, and one of radius
\(C_{\mathrm h}^{1/d}h_N(k)\) at least \(4k\).
Let \(\zeta\geq0\) and \(k\in\admset_N(\zeta)\). By
Definitions~\ref{def:valid-radii} and~\ref{def:admissible-ranks},
\(C_{\mathrm h}^{-1/d}h_N(k)\geq\xi_Nh_N(\log N)\), \(C_{\mathrm h}^{-1/d}h_N(k)\geq\xi_N\zeta\), and
\(C_{\mathrm h}^{1/d}h_N(k)\leq\locrad/\xi_N\). The first inequality is
\(k\geq C_{\mathrm h}\xi_N^d\log N\). The last, with
\(\locrad\leq R_{\mathrm g}\leq c_{\mathrm g}^{-1/d}L\) and
\(c_{\mathrm g}C_{\mathrm h}\geq16/c_\alpha\), bounds \(k/N\): since
\((h_N(k)/L)^d=k/N\),
\begin{align}
 \frac kN\leq\frac1{C_{\mathrm h}}\Bigl(\frac{\locrad}{\xi_NL}\Bigr)^d
 \leq\frac1{c_{\mathrm g}C_{\mathrm h}\xi_N^d}
 \leq\frac{c_\alpha}{16\,\xi_N^d}.
 \label{eqn:admissible-count-upper-bound}
\end{align}
Hence
\begin{align}
 \frac{\log N}{k}\leq\frac1{C_{\mathrm h}\xi_N^d},\qquad
 \frac{C_{\mathrm h}^{1/d}\zeta}{h_N(k)}\leq\frac1{\xi_N},\qquad
 \frac{C_{\mathrm h}^{1/d}h_N(k)}{\locrad}\leq\frac1{\xi_N},\qquad
 \frac kN\leq\frac{c_\alpha}{16\,\xi_N^d}.
 \label{eqn:window-bounds}
\end{align}
In particular, \(C_{\mathrm h}^{1/d}h_N(k)\leq\locrad\), so the radius
band lies in the range of the local mass bounds and below half the
separation.

\paragraph{The affine event}
In the proofs,
\(\mathcal E_{\mathrm D}=\mathcal E_{\mathrm D}(Ca_N,Cb_N)\) is the affine
event of Section~\ref{sec:stability-of-data-driven-selection}. We absorb
\(C\) into \(a_N\) and \(b_N\), so constants may depend on \(C\). The
proofs use the input error only through
\begin{align}
 \sup_{k\in\admset_N}\Bigl\{a_N+\frac{C_{\mathrm h}^{1/d}b_N}{h_N(k)}\Bigr\}
 \longrightarrow0,
 \label{eqn:affine-errors-vanish}
\end{align}
which holds because \(a_N\to0\) and, by \eqref{eqn:window-bounds} with
\(\zeta=b_N\), \(C_{\mathrm h}^{1/d}b_N/h_N(k)\leq C/\xi_N\to0\) for
\(k\in\admset_N\).

\paragraph{The clean event}
For an index set \(\sI\subseteq[N]\), a point \(i\in\sI\), and a radius
\(r>0\), let
\(\gN_i(\sI;r)\coloneqq\{j\in\sI\setminus\{i\}:D_{ij}\leq r\}\) be the
neighborhood of \(i\) in \(\sI\), and let \(\gN_i^{(\mathrm s)}(\sI;r)\),
\(r_i^{(\mathrm s)}(\sI;k)\), and \(r_{\mathrm s}(\sI;k,\omega)\) be the
neighborhood, the \(k\)-th nearest-neighbor distance, and the selected
radius \eqref{eqn:selected-radius} computed from \(\mD^{(\mathrm s)}\).
Let
\(\deg_i^{(\mathrm s)}(\sI;r)\coloneqq|\gN_i^{(\mathrm s)}(\sI;r)|\) be the
clean degree of \(i\), and, for \(0\leq u\leq1/2\), let
\begin{align}
 \cleanShell_i^{(\mathrm s)}(\sI;r,u)
 \coloneqq\bigl\{j\in \sI\setminus\{i\}:|D^{(\mathrm s)}_{ij}/r-1|\leq u\bigr\}
 \label{eqn:clean-shell}
\end{align}
be the clean signals in \(\sI\) whose distance from \(i\) lies within
relative distance \(u\) of \(r\). With the constants
\begin{align}
 c_{\mathrm{deg}}\coloneqq\frac{c_{\mathrm g}c_\alpha}{8C_{\mathrm h}},
 \qquad
 C_{\mathrm{deg}}\coloneqq(\mathrm e-1)C_{\mathrm g}C_{\mathrm h},
 \qquad
 t_\nu\coloneqq(\nu+8)\log N,
 \label{eqn:clean-degree-constants}
\end{align}
let \(\admset\subseteq\{1,\ldots,N-1\}\) be a deterministic family of
neighbor counts, by default \(\admset=\admset_N\). In the intersections
below, \(b\) ranges over \([B]\), \(k\) over \(\admset\), \(\sB\) over the
nonempty subsets of \([B]\), \(\omega\) over \((0,1)\), \(i\) over
\(\sI_\sB\), and \(u\) over \([0,1/2]\), and we write \(\sI=\sI_\sB\) and
\(r_{\mathrm s}=r_{\mathrm s}(\sI_\sB;k,\omega)\). Define
\begin{alignat}{2}
 \mathcal E_{\mathrm{bal}}
 &\coloneqq\bigcap_{b}\bigl\{|\gC_b|\geq\lfloor c_\alpha N/2\rfloor\bigr\},
 &\qquad
 \mathcal E_{\mathrm{deg}}(\nu)
 &\coloneqq\smashoperator{\bigcap_{k,\sB,\omega,i}}\bigl\{c_{\mathrm{deg}}k
   \leq\deg_i^{(\mathrm s)}(\sI;r_{\mathrm s})\leq C_{\mathrm{deg}}k+t_\nu\bigr\},
 \label{eqn:clean-event-balance-degree}\\
 \mathcal E_{\mathrm{scale}}
 &\coloneqq\smashoperator{\bigcap_{k,\sB,\omega}}\bigl\{r_{\mathrm s}\in\radband_N(k)\bigr\},
 &
 \mathcal E_{\mathrm{shell}}(\nu)
 &\coloneqq\smashoperator{\bigcap_{k,\sB,\omega,i,u}}
   \bigl\{|\cleanShell_i^{(\mathrm s)}(\sI;r_{\mathrm s},u)|\leq C_{\mathrm{deg}}ku+t_\nu\bigr\}.
 \label{eqn:clean-event-scale-shell}
\end{alignat}
The \emph{clean event} is
\begin{align}
 \mathcal E_N(\nu)\coloneqq\mathcal E_{\mathrm{bal}}\cap\mathcal E_{\mathrm{scale}}
 \cap\mathcal E_{\mathrm{deg}}(\nu)\cap\mathcal E_{\mathrm{shell}}(\nu).
 \label{eqn:clean-event}
\end{align}
It depends on the clean sample only, and its four parts do the following.
\begin{itemize}
\item On \(\mathcal E_{\mathrm{bal}}\), every union \(\sI_\sB\) has
\(|\sI_\sB|\geq c_\alpha N/4\) once \(N\geq8/c_\alpha\), so
\(k<|\sI_\sB|\) for every \(k\in\admset_N\) by \eqref{eqn:window-bounds},
and the selected radii on \(\sI_\sB\) are defined.
\item On \(\mathcal E_{\mathrm{scale}}\), the clean selected radius is of
order \(h_N(k)\).
\item On \(\mathcal E_{\mathrm{deg}}(\nu)\), every clean neighborhood, and
hence every Jaccard denominator and kernel row, has order \(k\); the
additive \(t_\nu\) is the price of a Chernoff bound that is uniform over
all points (Lemma~\ref{lem:uniform-radial-interval-count}).
\item On \(\mathcal E_{\mathrm{shell}}(\nu)\), thin shells around the clean
radius are sparse; this bounds how many neighbors change when the
distances and the radius are perturbed
(Lemma~\ref{lem:changed-neighbors}).
\end{itemize}
The radius \(r_{\mathrm s}\) is selected from the sample, so each part
holds uniformly over the neighbor counts, the unions, and the quantile
levels.

\begin{lemma}[Clean event]
\label{lem:clean-event}
Suppose Assumption~\ref{ass:local-mass-regularity} holds and fix \(\nu>0\).
There are constants \(C_{\mathrm{ev},\nu}\) and \(N_0(\nu)\), depending only on \(\nu\) and
the model constants, with the following property. If \(N\geq N_0(\nu)\) and
the family satisfies \(\underline k\geq C_{\mathrm{ev},\nu}\log N\),
\(\overline k<c_\alpha N/4\), and
\begin{align}
 C_{\mathrm h}^{1/d}h_N(\overline k)\leq\min\{R_{\mathrm g},\Delta/2\},
 \label{eqn:finite-local-domain}
\end{align}
where \(\underline k=\min\admset\) and \(\overline k=\max\admset\),
then \(\P\{\mathcal E_N(\nu)^c\}\leq B\mathrm e^{-c_\alpha N/8}+2N^{-\nu}\).
For the admissible neighbor counts \(\admset_N=\admset_N(b_N)\) of
Definition~\ref{def:admissible-ranks}, these conditions hold for all
sufficiently large \(N\) by \eqref{eqn:window-bounds}.
\end{lemma}
Appendix~\ref{app:clean-event-proof} proves Lemma~\ref{lem:clean-event}.
The event \(\mathcal E_N(\nu)\) therefore fails with probability at most
\(C_\nu N^{-\nu}\) for all sufficiently large \(N\), so a property that
holds on \(\mathcal E_N(\nu)\cap\mathcal E_{\mathrm D}\) holds with
overwhelming probability on \(\mathcal E_{\mathrm D}\).

\subsection{Proof of Proposition~\ref{prop:admissible-ranks}}
\label{app:admissible-ranks-proof}
Write \(\underline r\coloneqq\xi_N\max\{\zeta_N,h_N(\log N)\}\) and
\(\overline r\coloneqq\locrad/\xi_N\) for the endpoints of
\(\radwin_N(\zeta_N)\), and let
\begin{align}
 x_N\coloneqq C_{\mathrm h}N\Bigl(\frac{\underline r}{L}\Bigr)^d,
 \qquad
 y_N\coloneqq\frac{N}{C_{\mathrm h}}\Bigl(\frac{\overline r}{L}\Bigr)^d
 \label{eqn:admissible-ends}
\end{align}
be the two ends of \eqref{eqn:admissible-range}, so that a count
\(k\in\{1,\ldots,N-1\}\) is admissible if and only if
\(x_N\leq k\leq y_N\).
\begin{proof}[Proof of Proposition~\ref{prop:admissible-ranks}]
If some count is admissible, then \(x_N\leq y_N\), that is,
\(\overline r/\underline r\geq C_{\mathrm h}^{2/d}\), which is the converse
claim. Now suppose \eqref{eqn:admissible-existence}. Then
\(\overline r/\underline r\to\infty\), so \(y_N\geq2x_N\) for all
sufficiently large \(N\). Since \(\underline r\geq\xi_Nh_N(\log N)\) and
\(N(h_N(\log N)/L)^d=\log N\), also
\(x_N\geq C_{\mathrm h}\xi_N^d\log N\geq1\) for large \(N\). The interval
\([x_N,y_N]\) then has length at least \(x_N\geq1\), so it contains an
integer \(k\geq1\). Finally \(y_N\leq c_\alpha N/(16\xi_N^d)<N-1\) for
\(N\geq2\), as in \eqref{eqn:admissible-count-upper-bound}, so \(k\) is
admissible.
\end{proof}

\subsection{Proof of Proposition~\ref{prop:radius-stability}}
\label{app:proof-radius-stability}

\paragraph{Idea of the proof}
On the affine event, the selected radius moves by at most the input
error, \(|r-r_{\mathrm s}|\leq a_Nr_{\mathrm s}+b_N\); this step is
deterministic (Lemma~\ref{lem:affine-order-statistics}). On the clean
event, \(r_{\mathrm s}\in\radband_N(k)\). Dividing by
\(r_{\mathrm s}\geq C_{\mathrm h}^{-1/d}h_N(k)\) gives the proposition,
on every union of communities
(Lemma~\ref{lem:union-radius-stability}).

\begin{lemma}[Affine order statistics]
\label{lem:affine-order-statistics}
Let \(\sI\subseteq[N]\) with \(|\sI|\geq2\), let \(\mD\) be an input matrix
with the conventions of Section~\ref{sec:data-driven-radius-selection}, and
let \(r\) and \(r_{\mathrm s}\) be the selected radii on \(\sI\) computed
from \(\mD\) and \(\mD^{(\mathrm s)}\) with the same \(1\leq k<|\sI|\) and
\(\omega\in(0,1)\). If
\(|D_{ij}-D^{(\mathrm s)}_{ij}|\leq aD^{(\mathrm s)}_{ij}+b\) for all distinct
\(i,j\in \sI\), with \(0\leq a<1\) and \(b\geq0\), then
\begin{align}
 (1-a)r_{\mathrm s}-b\leq r\leq(1+a)r_{\mathrm s}+b.
 \label{eqn:affine-radius-bound}
\end{align}
\end{lemma}
\begin{proof}[Proof of Lemma~\ref{lem:affine-order-statistics}]
We use the monotonicity of order statistics: if \(\vv,\vw\in\R^n\)
satisfy \(v_\ell\leq w_\ell\) for every \(\ell\in[n]\), then
\(v_{(\ell)}\leq w_{(\ell)}\) for every \(\ell\), since at least \(\ell\)
entries of \(\vv\) are then at most \(w_{(\ell)}\). Fix \(i\in \sI\). The
entrywise bounds
\((1-a)D^{(\mathrm s)}_{ij}-b\leq D_{ij}\leq(1+a)D^{(\mathrm s)}_{ij}+b\),
\(j\in \sI\setminus\{i\}\), and this monotonicity
give the same bounds for the \(k\)-th smallest row values, because an
increasing affine map commutes with order statistics. Applying
this monotonicity again to the outer quantile
proves \eqref{eqn:affine-radius-bound}.
\end{proof}

\begin{lemma}[Radius stability on unions]
\label{lem:union-radius-stability}
Suppose Assumption~\ref{ass:local-mass-regularity} holds. For all
sufficiently large \(N\), on
\(\mathcal E_N(\nu)\cap\mathcal E_{\mathrm D}\), for every
\(k\in\admset_N\), every nonempty \(\sB\subseteq[B]\), and every
\(\omega\in(0,1)\), the radii \(r=r(\sI_\sB;k,\omega)\) and
\(r_{\mathrm s}=r_{\mathrm s}(\sI_\sB;k,\omega)\) satisfy
\begin{align}
 r_{\mathrm s}\in\radband_N(k),
 \qquad
 \Bigl|\frac r{r_{\mathrm s}}-1\Bigr|
 \leq a_N+\frac{C_{\mathrm h}^{1/d}b_N}{h_N(k)}.
 \label{eqn:union-radius-stability}
\end{align}
\end{lemma}
\begin{proof}[Proof of Lemma~\ref{lem:union-radius-stability}]
The first claim is \(\mathcal E_{\mathrm{scale}}\). On
\(\mathcal E_{\mathrm{bal}}\), \(k<|\sI_\sB|\), so
Lemma~\ref{lem:affine-order-statistics} applies on
\(\mathcal E_{\mathrm D}\) with \(a=a_N<1\) and \(b=b_N\), and gives
\(|r-r_{\mathrm s}|\leq a_Nr_{\mathrm s}+b_N\). Dividing by
\(r_{\mathrm s}\geq C_{\mathrm h}^{-1/d}h_N(k)\) proves the second claim.
\end{proof}

\begin{proof}[Proof of Proposition~\ref{prop:radius-stability}]
In case~(i) of the setting, \(\sI_\sB=[N]\), and
Lemma~\ref{lem:union-radius-stability} gives
\eqref{eqn:clean-percentile-radius-scale} on \(\mathcal E_N(\nu)\), since
its first claim is the clean event \(\mathcal E_{\mathrm{scale}}\), and
\eqref{eqn:radius-relative-stability} on
\(\mathcal E_N(\nu)\cap\mathcal E_{\mathrm D}\). The event
\(\mathcal E_N(\nu)\) fails with probability at most \(C_\nu N^{-\nu}\).
\end{proof}

\subsection{Proof of Proposition~\ref{prop:neighborhood-preservation}}
\label{app:neighborhood-preservation-proof}

\paragraph{Idea of the proof}
On the affine event, every neighbor that changes between the two graphs
lies in a thin shell around the clean radius. On the clean event, a shell
of relative width \(u\) holds at most order \(ku+\log N\) points, and
every clean neighborhood has order \(k\) points
(Lemma~\ref{lem:changed-neighbors}). Dividing the first count by the
second gives the three terms of \(\delta_N(k)\), on every union of
communities (Lemma~\ref{lem:union-neighborhood-preservation}).

\begin{lemma}[Changed neighbors]
\label{lem:changed-neighbors}
Fix \(k\in\admset_N\), a nonempty \(\sB\subseteq[B]\), and
\(\omega\in(0,1)\); write \(\sI=\sI_\sB\), \(r=r(\sI;k,\omega)\), and
\(r_{\mathrm s}=r_{\mathrm s}(\sI;k,\omega)\). On
\(\mathcal E_N(\nu)\), suppose
\(|D_{ij}-D^{(\mathrm s)}_{ij}|\leq aD^{(\mathrm s)}_{ij}+b\) for all distinct
\(i,j\in \sI\), with \(a+C_{\mathrm h}^{1/d}b/h_N(k)\leq1/16\). Then
\(\delta_{\mathrm{rel}}\coloneqq a+b/r_{\mathrm s}\leq1/16\),
\(r\geq(1-\delta_{\mathrm{rel}})r_{\mathrm s}>0\), and
\begin{align}
 \max_{i\in \sI}
 \bigl|\gN_i(\sI;r)\triangle\gN_i^{(\mathrm{s})}(\sI;r_{\mathrm{s}})\bigr|
 \leq C_{\mathrm{deg}}k\,\frac{2\delta_{\mathrm{rel}}}{1-a}+t_\nu
 \leq C\,k\Bigl(a+\frac b{h_N(k)}\Bigr)+t_\nu.
 \label{eqn:changed-neighbor-count}
\end{align}
\end{lemma}
\begin{proof}[Proof of Lemma~\ref{lem:changed-neighbors}]
On \(\mathcal E_{\mathrm{scale}}\), \(r_{\mathrm s}\geq C_{\mathrm h}^{-1/d}h_N(k)\), so
\(\delta_{\mathrm{rel}}\leq1/16\), and \eqref{eqn:affine-radius-bound}
gives \(r\geq(1-\delta_{\mathrm{rel}})r_{\mathrm s}\). Let
\(j\in\gN_i(\sI;r)\triangle\gN_i^{(\mathrm s)}(\sI;r_{\mathrm s})\). If the
edge is lost, then \(D^{(\mathrm s)}_{ij}\leq r_{\mathrm s}\) and
\((1+a)D^{(\mathrm s)}_{ij}+b\geq D_{ij}>r\geq(1-a)r_{\mathrm s}-b\); if it
is gained, then \(D^{(\mathrm s)}_{ij}>r_{\mathrm s}\) and
\((1-a)D^{(\mathrm s)}_{ij}-b\leq D_{ij}\leq r\leq(1+a)r_{\mathrm s}+b\).
Rearranging,
\begin{align}
 r_{\mathrm s}-\frac{2(ar_{\mathrm s}+b)}{1+a}\leq D^{(\mathrm s)}_{ij}
 \leq r_{\mathrm s}+\frac{2(ar_{\mathrm s}+b)}{1-a},
 \label{eqn:affine-changed-shell}
\end{align}
so \(j\) lies in \(\cleanShell_i^{(\mathrm s)}(\sI;r_{\mathrm s},u)\) with
\(u=2\delta_{\mathrm{rel}}/(1-a)\leq2/15\), and
\(\mathcal E_{\mathrm{shell}}(\nu)\) bounds the size of this shell. Finally
\(u\leq(32/15)\{a+C_{\mathrm h}^{1/d}b/h_N(k)\}\).
\end{proof}

\begin{lemma}[Neighborhood preservation on unions]
\label{lem:union-neighborhood-preservation}
Suppose Assumption~\ref{ass:local-mass-regularity} holds. For all
sufficiently large \(N\), on
\(\mathcal E_N(\nu)\cap\mathcal E_{\mathrm D}\), for every
\(k\in\admset_N\), every nonempty \(\sB\subseteq[B]\), and every
\(\omega\in(0,1)\),
\begin{align}
 1-T_N(\sI_\sB;k,\omega)
 \leq C_\nu\Bigl\{a_N+\frac{b_N}{h_N(k)}+\frac{\log N}{k}\Bigr\}
 =C_\nu\delta_N(k).
 \label{eqn:union-neighborhood-preservation}
\end{align}
\end{lemma}
\begin{proof}[Proof of Lemma~\ref{lem:union-neighborhood-preservation}]
By \eqref{eqn:affine-errors-vanish},
\(a_N+C_{\mathrm h}^{1/d}b_N/h_N(k)\leq1/16\) for every \(k\in\admset_N\)
and all sufficiently large \(N\), so Lemma~\ref{lem:changed-neighbors}
applies to every \(k\), \(\sB\), and \(\omega\). Write \(\sI=\sI_\sB\).
Each union in the denominator of \eqref{eqn:average-jaccard} contains
\(\gN_i^{(\mathrm s)}(\sI;r_{\mathrm s})\), whose size is at least
\(c_{\mathrm{deg}}k\) on \(\mathcal E_{\mathrm{deg}}(\nu)\). Bounding the
average over \(i\in\sI\) by the maximum and using
\eqref{eqn:changed-neighbor-count} and \(t_\nu\leq C_\nu\log N\) proves
\eqref{eqn:union-neighborhood-preservation}.
\end{proof}

\begin{proof}[Proof of Proposition~\ref{prop:neighborhood-preservation}]
In case~(i) of the setting, \(\sI_\sB=[N]\), and
Lemma~\ref{lem:union-neighborhood-preservation} gives
\(1-T_N(k,\omega)\leq C_\nu\delta_N(k)\) for every \(k\in\admset_N\) and
\(\omega\in(0,1)\), on an event that fails with probability at most
\(C_\nu N^{-\nu}\), which is
\eqref{eqn:uniform-neighborhood-preservation-rate}.
\end{proof}

\subsection{The clean event holds with overwhelming probability}
\label{app:clean-event-proof}
This subsection proves Lemma~\ref{lem:clean-event}.

\paragraph{Idea of the proof}
The clean event has lower and upper parts. The lower parts, namely the
radius bracket in \(\mathcal E_{\mathrm{scale}}\) and the lower degree
bound in \(\mathcal E_{\mathrm{deg}}(\nu)\), come from binomial bounds
(Lemma~\ref{lem:local-binomial-counts}) on the \(k\)-NN distance of each point, combined by a union bound over
points and counts (Lemma~\ref{lem:finite-radius-degree-localization}).
The upper parts, namely the upper degree bound and
\(\mathcal E_{\mathrm{shell}}(\nu)\), must hold at sample-selected radii,
so they come from counts that are uniform over all intervals
(Lemma~\ref{lem:uniform-local-upper-counts}, which rests on
Lemma~\ref{lem:uniform-radial-interval-count}).

\paragraph{Notation}
Call a neighbor count
\(k\) \emph{local} if \(k<c_\alpha N/4\) and
\(C_{\mathrm h}^{1/d}h_N(k)\leq\min\{R_{\mathrm g},\Delta/2\}\). Since \(h_N\) increases
with \(k\), every count of a family with \(\overline k<c_\alpha N/4\) and
\eqref{eqn:finite-local-domain} is local.

For each \(i\in[N]\), write \(b=\rY_i\), \(n_i=|\gC_b|-1\), and
\(\mathcal F_i=\sigma((\rY_\ell)_{\ell\in[N]},\clean_i)\). For a closed
interval \(\sL\subseteq\R\), let
\[
 M_i(\sL)\coloneqq
 \#\{j\in\gC_b\setminus\{i\}:D^{(\mathrm{s})}_{ij}\in \sL\},
 \qquad
 \P_i(\sL)\coloneqq
 \P_{\mathrm s}^{(b)}\bigl(\{\vs:\|\vs-\clean_i\|\in \sL\}\bigr),
\]
so that \(\P_i\) is the conditional law of the distance from \(\clean_i\)
to an independent clean draw from community \(b\). Define
\begin{align}
 \mathcal E_{\mathrm{up}}(\nu)\coloneqq
 \bigcap_{i\in[N]}\ \bigcap_{\sL\subseteq\R\,\text{closed interval}}
 \left\{M_i(\sL)\leq(\mathrm e-1)n_i\P_i(\sL)+t_\nu\right\}.
 \label{eqn:uniform-local-upper-event}
\end{align}

\begin{lemma}[Uniform upper counts]
\label{lem:uniform-local-upper-counts}
Suppose Assumption~\ref{ass:local-mass-regularity} holds and \(N\geq4\).
Then \(\P(\mathcal E_{\mathrm{up}}(\nu)^c)\leq N^{-\nu}\), and on
\(\mathcal E_{\mathrm{up}}(\nu)\), for every nonempty \(\sB\subseteq[B]\),
\(\sI=\sI_\sB\), \(i\in \sI\), \(0<r\leq\min\{R_{\mathrm g},\Delta/2\}\), and
\(0\leq u\leq1/2\),
\begin{align}
 \deg_i^{(\mathrm{s})}(\sI;r)
 &\leq(\mathrm e-1)C_{\mathrm g}|\sI|(r/L)^d+t_\nu,
 &
 |\cleanShell_i^{(\mathrm{s})}(\sI;r,u)|
 &\leq(\mathrm e-1)C_{\mathrm g}u|\sI|(r/L)^d+t_\nu.
 \label{eqn:finite-union-uniform-upper-counts}
\end{align}
\end{lemma}

\begin{proof}[Proof of Lemma~\ref{lem:uniform-local-upper-counts}]
Conditional on \(\mathcal F_i\), the labels determine \(\gC_b\setminus\{i\}\)
and \(n_i\); enumerating this set in an order determined by the labels,
the distances from \(\clean_i\) are conditionally iid with law \(\P_i\).
Lemma~\ref{lem:uniform-radial-interval-count} with \(n_0=N\) and
\(t=t_\nu-1\), and \((N^2+1)^2\leq4N^4\), give
\(\P(\mathcal E_{\mathrm{up}}(\nu)^c)\leq4\mathrm eN^5N^{-\nu-8}\leq N^{-\nu}\).

A ball of radius \(r\leq\Delta/2\) and a shell with outer radius
\((1+u)r\leq3\Delta/4\) around \(\clean_i\) contain no point of another
community, so
\(\deg_i^{(\mathrm{s})}(\sI;r)=M_i([0,r])\) and
\(|\cleanShell_i^{(\mathrm{s})}(\sI;r,u)|=M_i([(1-u)r,(1+u)r])\).
For \(0<a\leq R_{\mathrm g}\) and \(0<v\leq1/2\), \eqref{eqn:uniform-clean-shell}
gives \(\P_i(\{a\})\leq C_{\mathrm g}v(a/L)^d\); letting \(v\downarrow0\)
shows that \(\P_i\) has no atoms at positive local radii, so the closed
shell has the mass of the half-open shell in
\eqref{eqn:uniform-clean-shell}. With \(n_i\leq|\sI|\),
\eqref{eqn:uniform-clean-small-ball} and \eqref{eqn:uniform-clean-shell}
give \(n_i\P_i([0,r])\leq C_{\mathrm g}|\sI|(r/L)^d\) and
\(n_i\P_i([(1-u)r,(1+u)r])\leq C_{\mathrm g}u|\sI|(r/L)^d\), and
\eqref{eqn:uniform-local-upper-event} applied to these two intervals
proves \eqref{eqn:finite-union-uniform-upper-counts}. The event controls
every center and closed interval at once, so the bounds hold for
sample-selected radii and widths.
\end{proof}

For a family \(\admset\) of neighbor counts, define the localization event
\begin{align}
 \mathcal E_{\mathrm{knn}}\coloneqq\mathcal E_{\mathrm{bal}}\cap
 \bigcap_{\substack{\emptyset\ne \sB\subseteq[B],\ k\in\admset:\\
 k\text{ local}}}
 \ \bigcap_{i\in \sI_\sB}
 \left\{\begin{array}{l}
 C_{\mathrm h}^{-1/d}h_N(k)\leq r_i^{(\mathrm{s})}(\sI_\sB;k)\leq C_{\mathrm h}^{1/d}h_N(k),\\
 \deg_i^{(\mathrm{s})}(\sI_\sB;C_{\mathrm h}^{-1/d}h_N(k))\geq c_{\mathrm{deg}}k
 \end{array}\right\}.
 \label{eqn:finite-union-lower-endpoint-degree}
\end{align}

\begin{lemma}[Radius and lower-degree localization]
\label{lem:finite-radius-degree-localization}
Suppose Assumption~\ref{ass:local-mass-regularity} holds and
\(N\geq8/c_\alpha\). Then
\begin{align}
 \P(\mathcal E_{\mathrm{knn}}^c)
 &\leq B \mathrm e^{-c_\alpha N/8}
       +3\cdot2^BN|\admset|\mathrm e^{-c_{\mathrm{deg}}\underline{k}/4},
 \label{eqn:finite-rank-failure}
\end{align}
with \(\underline k=\min\admset\). On \(\mathcal E_{\mathrm{knn}}\),
simultaneously for every nonempty \(\sB\subseteq[B]\), every local
\(k\in\admset\), \(\sI=\sI_\sB\), and every \(\omega\in(0,1)\),
\begin{align}
 C_{\mathrm h}^{-1/d}h_N(k)\leq r_{\mathrm{s}}(\sI;k,\omega)\leq C_{\mathrm h}^{1/d}h_N(k).
 \label{eqn:finite-union-radius-bracket}
\end{align}
\end{lemma}

\begin{proof}[Proof of Lemma~\ref{lem:finite-radius-degree-localization}]
\noindent\emph{Step 1: balance and one row.}
By Lemma~\ref{lem:class-count-concentration},
\(\P(\mathcal E_{\mathrm{bal}}^c)\leq B\mathrm e^{-c_\alpha N/8}\).
Condition on a label vector in \(\mathcal E_{\mathrm{bal}}\), and fix a
nonempty \(\sB\), a local \(k\in\admset\), \(\sI=\sI_\sB\), a row
\(i\in\gC_b\subseteq \sI\), and its clean center; then \(k<c_\alpha N/4\leq|\sI|\).
The other \(n_b=|\gC_b|-1\) members of its community are
conditionally iid, with \(c_\alpha N/4\leq n_b\leq N\).
For \(v\in\{C_{\mathrm h}^{-1/d},C_{\mathrm h}^{1/d}\}\), the radius
\(vh_N(k)\) is at most \(\min\{R_{\mathrm g},\Delta/2\}\), so the ball of
that radius contains only points of community \(b\), and since
\(N(h_N(k)/L)^d=k\), Lemma~\ref{lem:local-binomial-counts} applies with
conditional means
\[
 \mu_i(v)\coloneqq\E\bigl[\deg_i^{(\mathrm s)}(\sI;vh_N(k))\bigm|
 (\rY_\ell)_\ell,\clean_i\bigr]
 \in\Bigl[\frac{c_{\mathrm g}c_\alpha}{4}v^dk,\ C_{\mathrm g}v^dk\Bigr].
\]
Write \(\mu_i^\pm\coloneqq\mu_i(C_{\mathrm h}^{\pm1/d})\). Since
\(C_{\mathrm h}\geq4\mathrm eC_{\mathrm g}\) and
\(c_{\mathrm g}C_{\mathrm h}\geq16/c_\alpha\), we have
\(\mu_i^-\leq C_{\mathrm g}k/C_{\mathrm h}\leq k/(4\mathrm e)\),
\(\mu_i^+\geq c_{\mathrm g}c_\alpha C_{\mathrm h}k/4\geq4k\), and
\(\mu_i^-\geq c_{\mathrm g}c_\alpha k/(4C_{\mathrm h})=2c_{\mathrm{deg}}k\).
The \(k\)-NN distance \(r_i^{(\mathrm s)}(\sI;k)\) is below
\(C_{\mathrm h}^{-1/d}h_N(k)\) only if at least \(k\) points lie in the
smaller ball, and above \(C_{\mathrm h}^{1/d}h_N(k)\) only if fewer than
\(k\leq\mu_i^+/2\) lie in the larger one; the degree at
\(C_{\mathrm h}^{-1/d}h_N(k)\) is below \(c_{\mathrm{deg}}k\) only if it is
below \(\mu_i^-/2\). Lemma~\ref{lem:local-binomial-counts} bounds the
conditional probability that one of these occurs by
\(4^{-k}+\mathrm e^{-k/2}+\mathrm e^{-c_{\mathrm{deg}}k/4}
\leq3\mathrm e^{-c_{\mathrm{deg}}\underline k/4}\).

\smallskip
\noindent\emph{Step 2: all rows, counts, and selected radii.}
The union \(\sI_\sB\) is
determined by the labels, so averaging over \(\clean_i\)
and a union bound over at most \(2^B|\admset|\) pairs and \(N\) rows
give \eqref{eqn:finite-rank-failure}. Every selected radius is an order statistic of the \(k\)-NN distances,
so it inherits their bracket.
\end{proof}

\begin{proof}[Proof of Lemma~\ref{lem:clean-event}]
\smallskip
\noindent\emph{Step 1: probability.}
Lemmas~\ref{lem:finite-radius-degree-localization}
and~\ref{lem:uniform-local-upper-counts} bound the failure probabilities
of \(\mathcal E_{\mathrm{knn}}\) and \(\mathcal E_{\mathrm{up}}(\nu)\).
Since \(|\admset|\leq N\) and \(\underline k\geq C_{\mathrm{ev},\nu}\log N\),
the choice \(C_{\mathrm{ev},\nu}=4(\nu+3)/c_{\mathrm{deg}}\) and a large
\(N_0(\nu)\) give
\(3\cdot2^BN^2\mathrm e^{-c_{\mathrm{deg}}\underline k/4}\leq N^{-\nu}\), so
\[
 \P\{(\mathcal E_{\mathrm{knn}}\cap\mathcal E_{\mathrm{up}}(\nu))^c\}
 \leq B\mathrm e^{-c_\alpha N/8}+2N^{-\nu}.
\]
It remains to show that
\(\mathcal E_{\mathrm{knn}}\cap\mathcal E_{\mathrm{up}}(\nu)\subseteq\mathcal E_N(\nu)\).

\smallskip
\noindent\emph{Step 2: the four parts of the clean event.}
By the hypotheses on the family, every \(k\in\admset\) is local. Fix
\(k\), \(\sB\), \(\omega\), and \(i\in\sI_\sB\), and write
\(r_{\mathrm s}=r_{\mathrm s}(\sI_\sB;k,\omega)\).
\begin{itemize}
\item \(\mathcal E_{\mathrm{bal}}\) is part of \(\mathcal E_{\mathrm{knn}}\).
\item \(\mathcal E_{\mathrm{scale}}\): the bracket
\eqref{eqn:finite-union-radius-bracket} gives
\(r_{\mathrm s}\in\radband_N(k)\).
\item Lower degree: closed neighborhoods grow with the radius, so
\(\deg_i^{(\mathrm s)}(\sI_\sB;r_{\mathrm s})\geq
\deg_i^{(\mathrm s)}(\sI_\sB;C_{\mathrm h}^{-1/d}h_N(k))\geq c_{\mathrm{deg}}k\).
\item Upper degree and shells: since
\(r_{\mathrm s}\leq C_{\mathrm h}^{1/d}h_N(k)\leq\min\{R_{\mathrm g},\Delta/2\}\)
and \(|\sI_\sB|(r_{\mathrm s}/L)^d\leq N(r_{\mathrm s}/L)^d\leq C_{\mathrm h}k\),
the bounds \eqref{eqn:finite-union-uniform-upper-counts} give
\(\deg_i^{(\mathrm s)}(\sI_\sB;r_{\mathrm s})\leq C_{\mathrm{deg}}k+t_\nu\) and
\(|\cleanShell_i^{(\mathrm s)}(\sI_\sB;r_{\mathrm s},u)|\leq C_{\mathrm{deg}}ku+t_\nu\)
for every \(u\in[0,1/2]\).
\end{itemize}
The last two items give \(\mathcal E_{\mathrm{deg}}(\nu)\) and
\(\mathcal E_{\mathrm{shell}}(\nu)\).

\smallskip
\noindent\emph{Step 3: the admissible counts.}
For \(\admset=\admset_N\), \eqref{eqn:window-bounds} gives
\(\underline k\geq C_{\mathrm h}\xi_N^d\log N\),
\(\overline k\leq c_\alpha N/(16\xi_N^d)<c_\alpha N/4\), and
\(C_{\mathrm h}^{1/d}h_N(\overline k)\leq\locrad/\xi_N\leq\min\{R_{\mathrm g},\Delta/2\}\).
Since \(\xi_N\to\infty\), the conditions of the lemma hold for all
sufficiently large \(N\).
\end{proof}

\section{Proofs in Section~\ref{sec:spectral-convergence}}
\label{app:spectral-convergence-proofs}

This section keeps the setting, the events, and the conventions of
Appendix~\ref{app:radius-neighborhood-proofs}; constants may also depend
on the kernel constants \(c_{\kf}\), \(M_{\kf}\), and \(\Lambda_{\kf}\) of
Definition~\ref{def:admissible-kernels}. The comparison between the clean
kernel and the population operator stated below is used by both proofs
and proved in Appendix~\ref{app:uniform-clean-spectral-approximation}.

\paragraph{Notation for matrices}
\label{par:spectral-notation}
Eigenvalues are in decreasing order, with orthonormal eigenvectors.
For a symmetric \(\mH\in\R^{n\times n}\), write
\(\lambda_j^+(\mH)\coloneqq\max\{\lambda_j(\mH),0\}\) for \(j\leq n\) and
\(\lambda_j^+(\mH)\coloneqq0\) for \(j>n\), so that
\(\lambda_j^+=\lambda_j^+(\mK)\) in
Theorem~\ref{thm:population-eigenfunction-convergence}.
For a real symmetric \(\mH\in\R^{n\times n}\) and a nonempty
\(\sJ\subseteq[n]\), let \(\mV_{\sJ}(\mH)\) have columns \(\vv_j(\mH)\),
\(j\in \sJ\), let \(\mLambda_{\sJ}(\mH)\) be the diagonal matrix of the
corresponding eigenvalues, and put
\(\mZ_{\sJ}(\mH)=\mV_{\sJ}(\mH)\mLambda_{\sJ}(\mH)\) as in
\eqref{eqn:method-weighted-embedding}, and
\begin{align}
 \mP_{\sJ}(\mH)&=\mV_{\sJ}(\mH)\mV_{\sJ}(\mH)^{\sT},
 &\operatorname{gap}_{\sJ}(\mH)&=\min_{j\in \sJ,\ j'\notin \sJ}
            |\lambda_j(\mH)-\lambda_{j'}(\mH)|,
 \label{eqn:finite-cluster-gap}
\end{align}
with \(\operatorname{gap}_{[n]}(\mH)=+\infty\) and
\(\operatorname{gap}_j=\operatorname{gap}_{\{j\}}\). As in
Theorem~\ref{thm:spectral-stability}, \((\lambda_j,\vv_j)\) and
\((\lambda_j^{(\mathrm s)},\vv_j^{(\mathrm s)})\) are the eigenpairs of
\(\mK(\sI)\) and \(\mK^{(\mathrm s)}(\sI)\) on the active set under
discussion. For a deterministic radius \(r>0\),
\(\mK^{(\mathrm s)}(\sI;r)\) is the clean kernel matrix on \(\sI\) with
\(r\) in place of the selected radius. Diagonal-completed kernels are
\(\overline{\mK}(\sI;r)=\mK(\sI;r)+\kf(0)|\sI|^{-1}\mI_{|\sI|}\), and similarly
\(\overline{\mK}^{(\mathrm s)}(\sI;r)\); on the
full sample \(\overline{\mK}=\overline{\mK}([N];r)\) and
\(\overline{\mK}^{(\mathrm s)}=\overline{\mK}^{(\mathrm s)}([N];r_{\mathrm s})\).

\paragraph{The population operator}
When the radius is fixed we drop the subscript \(r\) from \(\IOK_r^{(\sB)}\),
\(\gamma_{j,r}^{(\sB)}\), and \(\phi_{j,r}^{(\sB)}\). The list
\(\gamma_{1,r}\geq\gamma_{2,r}\geq\cdots\) is padded by zeros or accumulates
at zero, so the population eigengap \eqref{eqn:population-eigengap}
satisfies \(\operatorname{gap}_j(\IOK_r)\leq\gamma_{j,r}\): the eigenvalue
\(\gamma_{j,r}\) is at distance at least \(\operatorname{gap}_j(\IOK_r)\)
from the rest of the spectrum of \(\IOK_r\), including its nonpositive
part.

\paragraph{The clean kernel and the population operator}
The following statements, which concern the clean sample only, are
proved in Appendix~\ref{app:uniform-clean-spectral-approximation}.

\begin{lemma}[Local spectral bounds]
\label{lem:population-hard-cutoff-operator}
Suppose Assumption~\ref{ass:local-mass-regularity} holds, and let \(\kf\)
be a regular kernel function. For every nonempty \(\sB\subseteq[B]\) and
every \(0<r\leq\locrad\), the operator \(\IOK_r^{(\sB)}\) is self-adjoint
and Hilbert--Schmidt, hence compact, and
\(\|\IOK_r^{(\sB)}\|\leq C(r/L)^d\).
\end{lemma}

\begin{lemma}[Clean spectrum on a window of radii]
\label{lem:uniform-clean-positive-spectrum}
Suppose that Assumption~\ref{ass:local-mass-regularity} holds with
\(B\geq1\) communities and that the kernel function is regular. Fix
constants \(0<w_1<w_2\) and \(\nu>0\). There are constants
\(c_\nu\in(0,1]\), \(C_\nu\), and \(N_\nu\), depending only on \(\nu\) and
on the model, kernel, and window constants, with the following property.
Let \(N\geq N_\nu\), and let \(1\leq k<N\) be an integer with
\begin{align}
 \log N\leq c_\nu k,
 \qquad
 h_N(k)\leq c_\nu\locrad,
 \label{eqn:cell-conditions}
\end{align}
with \(\locrad\) of \eqref{eqn:proof-setting}. Write \(h_N=h_N(k)\),
\(\beta_N=\beta_N(k)\) for the population error rate
\eqref{eqn:population-error-rate}, and
\(\varepsilon_N\coloneqq C_\nu(k/N)\beta_N\). There is an event
\(\mathcal E_{\mathrm{cell}}(\nu)\), determined by the clean sample, with
\(\P\{\mathcal E_{\mathrm{cell}}(\nu)^c\}\leq N^{-\nu}\), on which the
following hold for every \(r\in[w_1h_N,w_2h_N]\).
\begin{enumerate}[label=(\alph*)]
\item \emph{(Eigenvalues)} For every \(j\geq1\),
\(|\lambda_j^+(\overline{\mK}^{(\mathrm{s})}([N];r))-\gamma_{j,r}|\leq \varepsilon_N\).
\item \emph{(Eigengaps)} If \(j\geq1\) and
\(\operatorname{gap}_j(\IOK_r)\geq\dgap\) for some
\(\dgap>2\varepsilon_N\), then \(j\leq N\) and
\(\operatorname{gap}_j(\mK^{(\mathrm{s})}([N];r))\geq\dgap-2\varepsilon_N\).
\end{enumerate}
\end{lemma}

\begin{lemma}[Isometric lift]
\label{lem:common-space-spectral-lift}
In the setting of Lemma~\ref{lem:uniform-clean-positive-spectrum}, there
is a linear isometry \(U_N:\R^N\to L^2(\P_{\mathrm s})\), defined on every
realization from the clean sample and \(k\), with the following
properties.
\begin{enumerate}[label=(\alph*)]
\item \emph{(Lift)} On \(\mathcal E_{\mathrm{cell}}(\nu)\), for every
\(r\in[w_1h_N,w_2h_N]\),
\(\|U_N\overline{\mK}^{(\mathrm{s})}([N];r)U_N^*-\IOK_r\|\leq \varepsilon_N\).
\item \emph{(Eigenfunctions)} Let \(r>0\), \(j\geq1\), and \(\dgap>0\) with
\(\operatorname{gap}_j(\IOK_r)\geq\dgap\), and let \(\mH\) be a symmetric
\(N\times N\) matrix with
\(\|U_N\mH U_N^*-\IOK_r\|\leq\varepsilon\leq\dgap/4\). Then
\(\lambda_j(\mH)\) is a simple positive eigenvalue of \(\mH\), and there
is a sign \(\chi\in\{-1,1\}\) with
\(\|U_N\vv_j(\mH)-\chi\phi_{j,r}\|_{L^2(\P_{\mathrm s})}\leq2\varepsilon/\dgap\).
\end{enumerate}
\end{lemma}

\begin{corollary}[Unions of communities]
\label{cor:union-clean-positive-spectrum}
Suppose the assumptions of Lemma~\ref{lem:uniform-clean-positive-spectrum}
hold, and fix \(\nu>0\). There are constants \(c_\nu\in(0,1]\), \(C_\nu\),
and \(N_\nu\) such that, for every \(N\geq N_\nu\) and every integer
\(1\leq k<N\) that satisfies \eqref{eqn:cell-conditions} with this
\(c_\nu\), the following holds on an event of probability at least
\(1-C_\nu N^{-\nu}\): for every nonempty \(\sB\subseteq[B]\) and every
\(r\in\radband_N(k)\), parts~(a) and~(b) of
Lemma~\ref{lem:uniform-clean-positive-spectrum} hold for
\(\overline{\mK}^{(\mathrm{s})}(\sI_\sB;r)\),
\(\mK^{(\mathrm{s})}(\sI_\sB;r)\), and \(\IOK_r^{(\sB)}\), with
\(\varepsilon_N=C_\nu(k/N)\beta_N(k)\).
\end{corollary}
For a sequence \(k=k_N\in\admset_N(0)\), the conditions
\eqref{eqn:cell-conditions} hold for all sufficiently large \(N\), by
\eqref{eqn:window-bounds}.

\subsection{Proof of Theorem~\ref{thm:spectral-stability}}
\label{app:reconstructed-spectral-transfer}

\paragraph{Idea of the proof}
The affine distance error transfers to the kernel entries, which gives
the kernel bound and, by Weyl's inequality (Lemma~\ref{lem:weyl}(a)), the
eigenvalues (Lemma~\ref{lem:affine-distance-calculus}). Under the gap condition, the
clean kernel inherits the population eigengap
(Corollary~\ref{cor:union-clean-positive-spectrum}), and the Davis--Kahan
bound (Corollary~\ref{cor:singleton-symmetric-spectral-perturbation})
gives the eigenvector. Lemma~\ref{lem:union-spectral-stability}
states the result on every union of communities.

\begin{lemma}[Affine distance errors and kernel transfer]
\label{lem:affine-distance-calculus}
In the setting of Lemma~\ref{lem:affine-order-statistics}, suppose
\(r_{\mathrm s}>0\) and \(\delta_{\mathrm{rel}}=a+b/r_{\mathrm s}<1\), let
\(\kf\) be a regular kernel function, and write
\(\gN_i=\gN_i(\sI;r)\) and
\(\gN_i^{(\mathrm s)}=\gN_i^{(\mathrm s)}(\sI;r_{\mathrm s})\). Then
\(r\geq(1-\delta_{\mathrm{rel}})r_{\mathrm s}>0\), and the following hold.
\begin{enumerate}[label=(\alph*)]
\item For every \(i\in\sI\) and \(j\in\gN_i\cap\gN_i^{(\mathrm s)}\),
\(|D_{ij}^2/r^2-(D^{(\mathrm s)}_{ij})^2/r_{\mathrm s}^2|\leq4\delta_{\mathrm{rel}}/(1-\delta_{\mathrm{rel}})\).
\item For every \(i\in\sI\),
\begin{align}
 \sum_{j\in \sI}\bigl|[\mK(\sI)]_{ij}-[\mK^{(\mathrm s)}(\sI)]_{ij}\bigr|
 \leq\frac1{|\sI|}\Bigl\{\frac{4\Lambda_{\kf}\delta_{\mathrm{rel}}}{1-\delta_{\mathrm{rel}}}\,
   \deg_i^{(\mathrm s)}(\sI;r_{\mathrm s})
 +M_{\kf}\bigl|\gN_i\triangle\gN_i^{(\mathrm s)}\bigr|\Bigr\}.
 \label{eqn:two-array-kernel-row-bound}
\end{align}
\item If, in addition, the hypotheses of
Lemma~\ref{lem:changed-neighbors} hold, then
\begin{align}
 \|\mK(\sI)-\mK^{(\mathrm{s})}(\sI)\|
 &\leq C\frac{k}{|\sI|}\Bigl\{a+\frac b{h_N(k)}+\frac{t_\nu}k\Bigr\}.
 \label{eqn:affine-local-kernel}
\end{align}
\end{enumerate}
\end{lemma}
\begin{proof}[Proof of Lemma~\ref{lem:affine-distance-calculus}]
By \eqref{eqn:affine-radius-bound},
\(|r-r_{\mathrm s}|\leq\delta_{\mathrm{rel}}r_{\mathrm s}\), so
\(r\geq(1-\delta_{\mathrm{rel}})r_{\mathrm s}>0\). On a common edge,
\(D^{(\mathrm s)}_{ij}\leq r_{\mathrm s}\), so the distance error is also at
most \(\delta_{\mathrm{rel}} r_{\mathrm s}\); adding and subtracting
\(D^{(\mathrm s)}_{ij}/r\) gives
\(|D_{ij}/r-D^{(\mathrm s)}_{ij}/r_{\mathrm s}|\leq2\delta_{\mathrm{rel}}/(1-\delta_{\mathrm{rel}})\),
and (a) follows because both normalized distances lie in \([0,1]\).
For~(b), the kernels are hollow with factor \(1/|\sI|\). On a common edge,
\eqref{eqn:profile-lipschitz-bound} and~(a) bound the difference of the
two entries by
\(4\Lambda_{\kf}\delta_{\mathrm{rel}}(1-\delta_{\mathrm{rel}})^{-1}/|\sI|\),
and there are at most \(\deg_i^{(\mathrm s)}(\sI;r_{\mathrm s})\) common
edges. On a changed edge, exactly one entry vanishes, and the other is at
most \(M_{\kf}/|\sI|\) by \eqref{eqn:profile-positive-bounds}. For~(c),
under the hypotheses of Lemma~\ref{lem:changed-neighbors},
\(\delta_{\mathrm{rel}}\leq a+C_{\mathrm h}^{1/d}b/h_N(k)\leq1/16\), clean
degrees are at most \(C_{\mathrm{deg}}k+t_\nu\) on
\(\mathcal E_{\mathrm{deg}}(\nu)\), and
\eqref{eqn:changed-neighbor-count} bounds the changed neighbors. The
difference is symmetric, so its norm is at most its largest absolute row
sum (Lemma~\ref{lem:schur-test}(a)), which gives
\eqref{eqn:affine-local-kernel}.
\end{proof}

\begin{lemma}[Spectral stability on unions]
\label{lem:union-spectral-stability}
Suppose Assumption~\ref{ass:local-mass-regularity} holds, and let \(\kf\)
be a regular kernel function.
\begin{enumerate}[label=(\alph*)]
\item For all sufficiently large \(N\), on
\(\mathcal E_N(\nu)\cap\mathcal E_{\mathrm D}\), for every
\(k\in\admset_N\), every nonempty \(\sB\subseteq[B]\), and every
\(\omega\in(0,1)\),
\begin{align}
 \|\mK(\sI_\sB)-\mK^{(\mathrm s)}(\sI_\sB)\|
 \leq C_\nu\frac kN\,\delta_N(k).
 \label{eqn:union-kernel-transfer}
\end{align}
\item Fix \(j\geq1\), a sequence \(k=k_N\in\admset_N\), and a nonempty
\(\sB\subseteq[B]\) such that the gap condition
\eqref{eqn:spectral-gap-condition} holds for \(\IOK_r^{(\sB)}\) uniformly
over \(r\in\radband_N(k)\). Then, with overwhelming probability on
\(\mathcal E_{\mathrm D}\), for every \(\omega\in(0,1)\) there is a sign \(\chi_j\in\{-1,1\}\) such that the
eigenpairs of \(\mK(\sI_\sB)\) and \(\mK^{(\mathrm s)}(\sI_\sB)\) satisfy
\eqref{eqn:reconstructed-singleton-eigenvector} and
\eqref{eqn:reconstructed-weighted-eigenvector}, with
\(\IOK_{r_{\mathrm s}}^{(\sB)}\) in place of \(\IOK_{r_{\mathrm s}}\).
\end{enumerate}
\end{lemma}
\begin{proof}[Proof of Lemma~\ref{lem:union-spectral-stability}]
\noindent\emph{Part~(a).}
Fix \(k\), \(\sB\), and \(\omega\), and write \(\sI=\sI_\sB\). On
\(\mathcal E_{\mathrm{bal}}\), \(|\sI|\geq c_\alpha N/4\). By
\eqref{eqn:affine-errors-vanish}, for all sufficiently large \(N\) the
hypotheses of Lemma~\ref{lem:changed-neighbors} hold with \(a=a_N\) and
\(b=b_N\) for every \(k\in\admset_N\), and
\eqref{eqn:affine-local-kernel} with \(t_\nu\leq C_\nu\log N\) gives
\eqref{eqn:union-kernel-transfer}.

\smallskip
\noindent\emph{Part~(b).}
By \eqref{eqn:window-bounds}, the sequence \(k=k_N\) satisfies
\eqref{eqn:cell-conditions} for all sufficiently large \(N\). Intersect
\(\mathcal E_N(\nu)\cap\mathcal E_{\mathrm D}\) with the event of
Corollary~\ref{cor:union-clean-positive-spectrum} for this \(k\). Since
\(\mathcal E_N(\nu)\) and the event of the corollary fail with probability
at most \(C_\nu N^{-\nu}\), the claims below hold with overwhelming
probability on \(\mathcal E_{\mathrm D}\). Fix
\(\omega\), write \(\sI=\sI_\sB\) and
\(r_{\mathrm s}=r_{\mathrm s}(\sI;k,\omega)\in\radband_N(k)\), and put
\(\widehat g_N\coloneqq\operatorname{gap}_j(\IOK_{r_{\mathrm s}}^{(\sB)})\).
The uniform gap condition gives
\((k/N)\{\delta_N(k)+\beta_N(k)\}\leq\eta_N\widehat g_N\) for a
deterministic sequence \(\eta_N\to0\). Hence the error
\(\varepsilon_N=C_\nu(k/N)\beta_N(k)\) of the corollary is at most
\(\widehat g_N/4\) for all sufficiently large \(N\), and its part~(b)
gives
\(\operatorname{gap}_j(\mK^{(\mathrm s)}(\sI))\geq\widehat g_N/2\).
By part~(a),
\(\|\mK(\sI)-\mK^{(\mathrm s)}(\sI)\|\leq C_\nu(k/N)\delta_N(k)
<\widehat g_N/4\leq\operatorname{gap}_j(\mK^{(\mathrm s)}(\sI))/2\)
for all sufficiently large \(N\).
Corollary~\ref{cor:singleton-symmetric-spectral-perturbation} then gives a
sign \(\chi_j\) with
\[
 \|\vv_j-\chi_j\vv_j^{(\mathrm s)}\|
 \leq\frac{4\sqrt2\,\|\mK(\sI)-\mK^{(\mathrm s)}(\sI)\|}{\widehat g_N},
 \qquad
 \|\lambda_j\vv_j-\chi_j\lambda_j^{(\mathrm s)}\vv_j^{(\mathrm s)}\|
 \leq\|\mK(\sI)-\mK^{(\mathrm s)}(\sI)\|
 \Bigl(1+\frac{4\sqrt2\,|\lambda_j^{(\mathrm s)}|}{\widehat g_N}\Bigr).
\]
The first bound is \eqref{eqn:reconstructed-singleton-eigenvector}. For
the second, the clean kernel is symmetric with nonnegative entries, so
\(|\lambda_j^{(\mathrm s)}|\) is at most its largest row sum
(Lemma~\ref{lem:schur-test}(a)),
\(M_{\kf}(C_{\mathrm{deg}}k+t_\nu)/|\sI|\leq C_\nu k/N\) on
\(\mathcal E_{\mathrm{deg}}(\nu)\). Moreover,
\(\widehat g_N\leq\gamma_{j,r_{\mathrm s}}^{(\sB)}
\leq\|\IOK_{r_{\mathrm s}}^{(\sB)}\|\leq Ck/N\) by
Lemma~\ref{lem:population-hard-cutoff-operator}, so the factor in
brackets is at most \(C_\nu(k/N)/\widehat g_N\). This proves
\eqref{eqn:reconstructed-weighted-eigenvector}.
\end{proof}

\begin{proof}[Proof of Theorem~\ref{thm:spectral-stability}]
In case~(i) of the setting, \(\sI_\sB=[N]\) and
\(\IOK_r^{(\sB)}=\IOK_r\). Part~(a) is
Lemma~\ref{lem:union-spectral-stability}(a), on
\(\mathcal E_N(\nu)\cap\mathcal E_{\mathrm D}\), where \(\mathcal E_N(\nu)\)
fails with probability at most \(C_\nu N^{-\nu}\), together with Weyl's inequality
(Lemma~\ref{lem:weyl}(a)) for the eigenvalues. Part~(b) is
Lemma~\ref{lem:union-spectral-stability}(b).
\end{proof}

\subsection{Proof of Theorem~\ref{thm:population-eigenfunction-convergence}}
\label{app:population-eigenfunction-convergence}

\paragraph{Idea of the proof}
Theorem~\ref{thm:spectral-stability} compares the input kernel with the
clean kernel, and Lemmas~\ref{lem:uniform-clean-positive-spectrum}
and~\ref{lem:common-space-spectral-lift} compare the clean kernel with
the population operator, uniformly over the radius band, which contains
the selected radius. The triangle inequality combines the two.

\begin{proof}[Proof of Theorem~\ref{thm:population-eigenfunction-convergence}]
We are in case~(i) of the setting, and all population quantities are
evaluated at \(r_{\mathrm s}=r_{\mathrm s}(k,\omega)\).

\smallskip
\noindent\emph{Step 1: the isometry and the event.}
By \eqref{eqn:window-bounds}, the sequence \(k=k_N\) satisfies
\eqref{eqn:cell-conditions} for all sufficiently large \(N\). Apply
Lemmas~\ref{lem:uniform-clean-positive-spectrum}
and~\ref{lem:common-space-spectral-lift} with \(w_1=C_{\mathrm h}^{-1/d}\)
and \(w_2=C_{\mathrm h}^{1/d}\), so that the window of radii is
\(\radband_N(k)\), and let \(U_N\) be the isometry of
Lemma~\ref{lem:common-space-spectral-lift}. Fix \(\nu>0\) and work on
\(\mathcal E_N(\nu)\cap\mathcal E_{\mathrm D}\cap\mathcal E_{\mathrm{cell}}(\nu)\).
Since \(\mathcal E_N(\nu)\cap\mathcal E_{\mathrm{cell}}(\nu)\) fails with
probability at most \(C_\nu N^{-\nu}\) for all sufficiently large \(N\),
the claims below hold with overwhelming probability on
\(\mathcal E_{\mathrm D}\). On this event, \(r_{\mathrm s}\in\radband_N(k)\) for
every \(\omega\in(0,1)\), so both lemmas apply at \(r=r_{\mathrm s}\).

\smallskip
\noindent\emph{Step 2: eigenvalues.}
Adding the same scalar diagonal leaves the difference of the two
kernels unchanged, so Lemma~\ref{lem:union-spectral-stability}(a) gives
\(\|\overline{\mK}-\overline{\mK}^{(\mathrm s)}\|\leq C_\nu(k/N)\delta_N(k)\).
The positive-part eigenvalue list is one-Lipschitz in the operator norm
by Lemma~\ref{lem:weyl}(b), and part~(a) of
Lemma~\ref{lem:uniform-clean-positive-spectrum} bounds
\(\sup_j|\lambda_j^+(\overline{\mK}^{(\mathrm s)})-\gamma_j|\). Hence
\(\sup_j|\lambda_j^+(\overline{\mK})-\gamma_j|
\leq C_\nu(k/N)\{\delta_N(k)+\beta_N(k)\}\). Moreover,
\(|\lambda_j^+(\overline{\mK})-\lambda_j^+(\mK)|\leq\kf(0)/N
\leq M_{\kf}(k/N)\delta_N(k)\) for every \(j\), because
\(\overline{\mK}=\mK+\kf(0)N^{-1}\mI_N\), \(\kf(0)\leq M_{\kf}\) by
\eqref{eqn:profile-positive-bounds}, and \(\delta_N(k)\geq\log N/k\geq1/k\).
This proves \eqref{eqn:reconstructed-population-eigenvalues}.

\smallskip
\noindent\emph{Step 3: eigenfunctions.}
By Step~2 and part~(a) of Lemma~\ref{lem:common-space-spectral-lift},
\[
 \|U_N\overline{\mK}U_N^*-\IOK\|
 \leq\|\overline{\mK}-\overline{\mK}^{(\mathrm s)}\|
     +\|U_N\overline{\mK}^{(\mathrm s)}U_N^*-\IOK\|
 \leq C_\nu\frac kN\bigl\{\delta_N(k)+\beta_N(k)\bigr\}.
\]
Write \(\widehat g_N\coloneqq\operatorname{gap}_j(\IOK_{r_{\mathrm s}})\). By
\eqref{eqn:spectral-gap-condition} at \(r_{\mathrm s}\in\radband_N(k)\),
the right side is at most \(\widehat g_N/4\) for all sufficiently large
\(N\). Part~(b) of Lemma~\ref{lem:common-space-spectral-lift}, with
\(\mH=\overline{\mK}\), whose eigenvectors are those of \(\mK\), then
gives a sign \(\chi_j'\) with
\(\|U_N\vv_j-\chi_j'\phi_j\|
\leq2C_\nu(k/N)\{\delta_N(k)+\beta_N(k)\}/\widehat g_N\), which is
\eqref{eqn:affine-population-lift}.
\end{proof}

\subsection{The clean kernel approximates the population operator}
\label{app:uniform-clean-spectral-approximation}
This subsection proves Lemmas~\ref{lem:population-hard-cutoff-operator},
\ref{lem:uniform-clean-positive-spectrum},
and~\ref{lem:common-space-spectral-lift} and
Corollary~\ref{cor:union-clean-positive-spectrum}.

\paragraph{Idea of the proofs}
We partition the support into cells of diameter of order
\(\beta_Nh_N\) and replace both the operator and the clean kernel by step
kernels that are constant on pairs of cells. The replacement costs a
relative error \(\beta_N\), from the shell of pairs whose cutoff
indicators change. The two step kernels differ only through the cell
masses, whose relative sampling error is of order
\(\sqrt{\log N/(k\beta_N^d)}\). The two errors balance at
\(\beta_N^{d+2}=\log N/k\). This replaces the RKHS comparison
of~\cite{ding2023learning}, whose absolute error of order \(N^{-1/2}\)
can exceed the local scale \(k/N\) for the hard cutoff, where the kernel
need not be positive semidefinite.
\begin{proof}[Proof of Lemma~\ref{lem:population-hard-cutoff-operator}]
The kernel of \(\IOK_r^{(\sB)}\) is real, symmetric, and bounded, hence
square-integrable with respect to
\(\P_{\mathrm{s}}^{(\sB)}\otimes\P_{\mathrm{s}}^{(\sB)}\), so the operator
is self-adjoint and Hilbert--Schmidt. The kernel is at most \(M_{\kf}\) and
vanishes beyond distance \(r\), and a ball of radius \(r\leq\locrad\)
around a support point has mass at most \(C_{\mathrm g}(r/L)^d\) under
\(\P_{\mathrm{s}}^{(\sB)}\) (Lemma~\ref{lem:mixture-regularity}). The
Lemma~\ref{lem:schur-test}(b) gives the bound.
\end{proof}

\begin{proof}[Proof of Lemmas~\ref{lem:uniform-clean-positive-spectrum}
and~\ref{lem:common-space-spectral-lift}]
The proof compares the clean kernel with the operator through four
intermediate objects built on one partition of the support into cells:
\[
 \overline{\mK}^{(\mathrm s)}([N];r)
 \overset{\text{Step 4}}{\approx}\widehat\mT
 \overset{\text{Step 5}}{\cong}\widehat\mH_{\mathrm{cell}}
 \overset{\text{Step 5}}{\approx}\mH_{\mathrm{cell}}
 \overset{\text{Step 5}}{\cong}\IOK^{\mathrm{step}}
 \overset{\text{Step 4}}{\approx}\IOK,
\]
where \(\approx\) is an operator-norm error of order \((k/N)\beta_N\) and
\(\cong\) is an identity of nonzero spectra. In words: the clean kernel
is close to its step version \(\widehat\mT\), which is constant on pairs of
cells; \(\widehat\mT\) is the sample cell matrix
\(\widehat\mH_{\mathrm{cell}}\) in disguise; the sample and population cell
matrices differ only through the cell masses; the population cell matrix
\(\mH_{\mathrm{cell}}\) is the step operator \(\IOK^{\mathrm{step}}\) in
disguise; and the step operator is close to \(\IOK\). The two cell
matrices are \(n_{\mathrm{cell}}\times n_{\mathrm{cell}}\), so the comparison
between an operator and an \(N\times N\) matrix becomes a comparison of two
matrices of the same size. The cells are defined in Step~1, the step
kernels in Step~3, and the cell matrices in Step~5. Steps~1--3 build the cells,
the event on which the empirical cell masses are accurate, and the step
kernels; Steps~6 and~7 read off the eigenvalues and the eigengaps; Steps~8
and~9 construct the isometry \(U_N\) and prove the eigenfunction bound.

\smallskip
\noindent\emph{Step 0: scales and constants.}
All constants below depend only on the model and kernel constants and on
\(w_1,w_2\), except where \(\nu\) is indicated; \(c_\nu\) is taken small
and \(N_\nu\) large in the course of the proof. Recall that
\((h_N/L)^d=k/N\) and \(\beta_N^{d+2}=\log N/k\leq c_\nu\leq1\), and put
\(\delta=\beta_Nh_N\). Every ball and shell below has radius at most
\(w_2h_N+4\delta\), and, for every \(r\in[w_1h_N,w_2h_N]\),
\[
 w_2h_N+4\delta\leq(w_2+4)h_N\leq\locrad
 \quad\text{if }c_\nu\leq\frac1{w_2+4},
 \qquad
 \frac{4\delta}r\leq\frac{4\beta_N}{w_1}\leq\frac12
 \quad\text{if }c_\nu\leq\Bigl(\frac{w_1}8\Bigr)^{d+2}.
\]

\smallskip
\noindent\emph{Step 1: the partition.}
On each compact community support, choose a finite maximal
\(\delta\)-separated set of representatives and its nearest-representative
Voronoi cells, with ties assigned by a deterministic ordering, and
enumerate all cells as \(\mathcal V_a\), \(a\in[n_{\mathrm{cell}}]\), with
representatives \(\vc_a\); \(n_{\mathrm{cell}}\) is the number of cells. By
maximality, and since every point within \(\delta/3\) of \(\vc_a\) is at
distance at least \(2\delta/3\) from every other representative,
\[
 \operatorname{supp}\cap\{\vs:\|\vs-\vc_a\|\leq\delta/3\}
 \subseteq\mathcal V_a\subseteq\{\vs:\|\vs-\vc_a\|\leq\delta\},
 \qquad\text{so}\qquad\diam(\mathcal V_a)\leq2\delta.
\]
The small-ball bounds and \eqref{eqn:min-mixture-weight} then give
\begin{align}
    c_2\frac kN\beta_N^d\leq
    \P_{\mathrm{s}}(\mathcal V_a)
    \leq C_3\frac kN\beta_N^d,
    \qquad n_{\mathrm{cell}}\leq\frac{C_3N}{k\beta_N^d}.
    \label{eqn:uniform-spectrum-cell-masses}
\end{align}

\smallskip
\noindent\emph{Step 2: the count event.}
Let \(\widehat\P_{\mathrm{s}}=N^{-1}\sum_i\delta_{\clean_i}\) be the
empirical clean law, and let
\[
 \mathcal E_{\mathrm{cell}}(\nu)=\Bigl\{
 \max_{a}\Bigl|\frac{\widehat\P_{\mathrm s}(\mathcal V_a)}{\P_{\mathrm s}(\mathcal V_a)}-1\Bigr|
 \leq C'_\nu\beta_N\Bigr\},
 \qquad c_\nu\leq(2C'_\nu)^{-(d+2)},
\]
so that \(C'_\nu\beta_N\leq1/2\). Each \(N\widehat\P_{\mathrm s}(\mathcal V_a)\)
is binomial with mean at least \(c_2k\beta_N^d\), and \(k\beta_N^d\geq1\)
gives \(n_{\mathrm{cell}}\leq C_3N\). By
\eqref{eqn:binomial-multiplicative-tail} of Lemma~\ref{lem:binomial-tails} and a union bound, for a large
\(C'_\nu\) and \(N\geq N_\nu\),
\[
 \P\{\mathcal E_{\mathrm{cell}}(\nu)^c\}
 \leq2n_{\mathrm{cell}}\exp\Bigl(-\frac{c_2C_\nu'^2k\beta_N^{d+2}}3\Bigr)
 \leq CN\exp\Bigl(-\frac{c_2C_\nu'^2\log N}3\Bigr)\leq N^{-\nu}.
\]
On \(\mathcal E_{\mathrm{cell}}(\nu)\), every cell is occupied, and every
union \(\mathcal U\) of cells satisfies
\[
 \tfrac12\P_{\mathrm s}(\mathcal U)\leq\widehat\P_{\mathrm s}(\mathcal U)
 \leq2\P_{\mathrm s}(\mathcal U).
\]
The partition and the event do not depend on \(r\); fix any
\(r\in[w_1h_N,w_2h_N]\) from now on.

\smallskip
\noindent\emph{Step 3: step kernels.}
Let \(w_r\) be the kernel of \(\IOK\), and let \(\mathrm{rep}(\vs)=\vc_a\)
for \(\vs\in\mathcal V_a\) be the representative of the cell that contains
\(\vs\). The \emph{representative kernel matrix} \(\mH_{\mathrm{rep}}\) has
one entry per pair of cells, and replacing both arguments of \(w_r\) by
their representatives gives the \emph{step operator} and the
\emph{sample step matrix}:
\begin{align}
 w_r(\vs,\vt)&=\indi{\|\vs-\vt\|\leq r}\,\kf(\|\vs-\vt\|/r),
 &[\mH_{\mathrm{rep}}]_{aa'}&=w_r(\vc_a,\vc_{a'}),\\
 (\IOK^{\mathrm{step}}f)(\vs)&=\int w_r(\mathrm{rep}(\vs),\mathrm{rep}(\vt))f(\vt)\,\P_{\mathrm s}(\odif{\vt}),
 &[\widehat{\mT}]_{ij}&=\frac1N\,w_r(\mathrm{rep}(\clean_i),\mathrm{rep}(\clean_j)).
\end{align}
Both are constant on pairs of cells. For comparison,
\([\overline{\mK}^{(\mathrm{s})}([N];r)]_{ij}=N^{-1}w_r(\clean_i,\clean_j)\).

\smallskip
\noindent\emph{Step 4: step-kernel errors.}
For \(\vs\in\mathcal V_a\) and \(\vt\in\mathcal V_{a'}\), the distances
\(v=\|\vs-\vt\|\) and \(v'=\|\vc_a-\vc_{a'}\|\) differ by at most
\(2\delta\). If both are at most \(r\), then
\eqref{eqn:profile-lipschitz-bound} gives
\(|\kf(v/r)-\kf(v'/r)|\leq\Lambda_{\kf}|v^2-(v')^2|/r^2\leq4\Lambda_{\kf}\delta/r\leq C_4\beta_N\);
if their cutoff indicators differ, then \(|v'-r|\leq2\delta\). Hence
\[
 \bigl|w_r(\vs,\vt)-w_r(\mathrm{rep}(\vs),\mathrm{rep}(\vt))\bigr|
 \leq C_4\beta_N\indi{\vt\in\mathcal U_a}
     +2M_{\kf}\indi{\vt\in\mathcal U_a^{\partial}},
 \]
 where we define
 \[
 \mathcal U_a=\smashoperator{\bigcup_{a':\|\vc_a-\vc_{a'}\|\leq r+2\delta}}\mathcal V_{a'}, \qquad \qquad, \mathcal U_a^{\partial}=\smashoperator{\bigcup_{a':|\|\vc_a-\vc_{a'}\|-r|\leq2\delta}}\mathcal V_{a'}.
\]
The union \(\mathcal U_a\) lies in the ball about \(\vc_a\) of radius
\(r+3\delta\), and \(\mathcal U_a^{\partial}\) in the shell
\(r-4\delta<\|\vt-\vc_a\|\leq r+4\delta\), of relative half-width
\(4\delta/r\leq1/2\). By \eqref{eqn:uniform-clean-small-ball} and
\eqref{eqn:uniform-clean-shell},
\[
 \P_{\mathrm{s}}(\mathcal U_a)\leq C_5\frac kN,
 \qquad
 \P_{\mathrm s}(\mathcal U_a^{\partial})\leq C_5\frac kN\beta_N,
\]
and, by Step~2, the same bounds hold for the empirical masses up to a
factor two. Lemma~\ref{lem:schur-test}(b) and~(a)
give, for every \(r\in[w_1h_N,w_2h_N]\),
\begin{align}
    \|\IOK-\IOK^{\mathrm{step}}\|\leq C_6\frac kN\beta_N,
    \qquad
    \|\overline{\mK}^{(\mathrm{s})}([N];r)-\widehat{\mT}\|\leq C_6\frac kN\beta_N.
    \label{eqn:uniform-spectrum-step-errors}
\end{align}

\smallskip
\noindent\emph{Step 5: cell matrices.}
With the diagonal matrices of the population and empirical cell masses,
\(\mM=\operatorname{diag}(\P_{\mathrm s}(\mathcal V_a))\) and
\(\widehat\mM=\operatorname{diag}(\widehat\P_{\mathrm s}(\mathcal V_a))\), the
\emph{cell matrices} are
\[
 \mH_{\mathrm{cell}}\coloneqq\mM^{1/2}\mH_{\mathrm{rep}}\mM^{1/2},
 \qquad
 \widehat\mH_{\mathrm{cell}}\coloneqq\widehat\mM^{1/2}\mH_{\mathrm{rep}}\widehat\mM^{1/2}.
\]
They are \(\IOK^{\mathrm{step}}\) and \(\widehat\mT\) written in orthonormal
bases of cell indicators,
\(\indi{\mathcal V_a}/\P_{\mathrm s}(\mathcal V_a)^{1/2}\) in
\(L^2(\P_{\mathrm s})\) and
\(\indi{\clean_i\in\mathcal V_a}/(N\widehat\P_{\mathrm s}(\mathcal V_a))^{1/2}\)
in \(\R^N\); both operators vanish on the orthogonal complements of these
bases. Hence, with multiplicity,
\[
 \operatorname{spec}(\IOK^{\mathrm{step}})\setminus\{0\}
 =\operatorname{spec}(\mH_{\mathrm{cell}})\setminus\{0\},
 \qquad
 \operatorname{spec}(\widehat\mT)\setminus\{0\}
 =\operatorname{spec}(\widehat\mH_{\mathrm{cell}})\setminus\{0\}.
\]
By Lemma~\ref{lem:population-hard-cutoff-operator} and
\eqref{eqn:uniform-spectrum-step-errors},
\(\|\mH_{\mathrm{cell}}\|=\|\IOK^{\mathrm{step}}\|\leq C_7k/N\). With the
diagonal matrix \(\mS_{\mathrm{cell}}=(\widehat\mM\mM^{-1})^{1/2}\) of
square-root ratios of empirical to population cell masses,
\[
 \widehat\mH_{\mathrm{cell}}=\mS_{\mathrm{cell}}\mH_{\mathrm{cell}}\mS_{\mathrm{cell}},
 \qquad
 \|\mS_{\mathrm{cell}}-\mI\|\leq C'_\nu\beta_N,
 \qquad
 \|\mS_{\mathrm{cell}}\|\leq2,
\]
hence
\begin{align}
    \|\widehat\mH_{\mathrm{cell}}-\mH_{\mathrm{cell}}\|
    \leq\|\mH_{\mathrm{cell}}\|\,\|\mS_{\mathrm{cell}}-\mI\|(\|\mS_{\mathrm{cell}}\|+1)
    \leq C_\nu\frac kN\beta_N.
    \label{eqn:uniform-spectrum-cell-matrix-error}
\end{align}

\smallskip
\noindent\emph{Step 6: Lemma~\ref{lem:uniform-clean-positive-spectrum}(a).}
By Lemma~\ref{lem:weyl}(b),
\[
 \sup_j\bigl|\lambda_j^+(\overline{\mK}^{(\mathrm s)}([N];r))-\lambda_j^+(\widehat\mT)\bigr|
 \leq\|\overline{\mK}^{(\mathrm s)}([N];r)-\widehat\mT\|,
 \qquad
 \sup_j\bigl|\lambda_j^+(\widehat\mH_{\mathrm{cell}})-\lambda_j^+(\mH_{\mathrm{cell}})\bigr|
 \leq\|\widehat\mH_{\mathrm{cell}}-\mH_{\mathrm{cell}}\|,
\]
and
\(\sup_j|\lambda_j^+(\IOK^{\mathrm{step}})-\gamma_{j,r}|\leq\|\IOK^{\mathrm{step}}-\IOK\|\).
The identities of Step~5 give
\(\lambda_j^+(\widehat\mT)=\lambda_j^+(\widehat\mH_{\mathrm{cell}})\) and
\(\lambda_j^+(\mH_{\mathrm{cell}})=\lambda_j^+(\IOK^{\mathrm{step}})\) for
every \(j\). Adding the three bounds, \eqref{eqn:uniform-spectrum-step-errors}
and \eqref{eqn:uniform-spectrum-cell-matrix-error} give part~(a) after
enlarging \(C_\nu\).

\smallskip
\noindent\emph{Step 7: Lemma~\ref{lem:uniform-clean-positive-spectrum}(b).}
Write \(\lambda_i=\lambda_i(\overline{\mK}^{(\mathrm{s})}([N];r))\) and
\(\mu_i=\lambda_i^+(\overline{\mK}^{(\mathrm{s})}([N];r))\), so that
\(|\mu_i-\gamma_i|\leq\varepsilon_N\) for every \(i\geq1\) by part~(a). Since
\(\gamma_j\geq\operatorname{gap}_j(\IOK_r)\geq\dgap\), we have
\(\mu_j\geq\dgap-\varepsilon_N>0\), so \(j\leq N\) and \(\mu_j=\lambda_j\).
Since \(\lambda_i=\mu_i\) for \(i<j\) and \(\lambda_i\leq\mu_i\) for all \(i\),
\[
 \lambda_i-\lambda_j\geq\gamma_i-\gamma_j-2\varepsilon_N\quad(i<j),
 \qquad
 \lambda_j-\lambda_i\geq\mu_j-\mu_i\geq\gamma_j-\gamma_i-2\varepsilon_N\quad(j<i\leq N),
\]
and both right sides are at least \(\dgap-2\varepsilon_N\). Hence
\(\operatorname{gap}_j(\overline{\mK}^{(\mathrm{s})}([N];r))\geq\dgap-2\varepsilon_N\),
and
\(\mK^{(\mathrm{s})}([N];r)=\overline{\mK}^{(\mathrm{s})}([N];r)-\kf(0)N^{-1}\mI_N\)
has the same eigengaps.

\smallskip
\noindent\emph{Step 8: the isometry and Lemma~\ref{lem:common-space-spectral-lift}(a).}
The clean law is atomless: by the upper bound of
\eqref{eqn:uniform-clean-small-ball}, a support point of a community has
mass at most \(C_{\mathrm g}(r/L)^d\) for every \(0<r\leq R_{\mathrm g}\),
hence zero. Fix a deterministic measurable rule that splits a cell, or
the whole support, into any given number of pieces of equal mass. When
all cells are occupied, split each \(\mathcal V_a\) into
\(N\widehat\P_{\mathrm s}(\mathcal V_a)\) pieces of equal mass, assigned in
increasing order to the sample indices in that cell; otherwise, an event
of probability at most \(N^{-\nu}\), split the whole support into \(N\)
pieces of mass \(1/N\), so that \(U_N\) is defined on every realization.
With \(\widetilde{\mathcal V}_i\) the piece of index \(i\), let
\[
 (U_N\vv)(\vt)=\frac{v_i}{\P_{\mathrm s}(\widetilde{\mathcal V}_i)^{1/2}}
 \quad(\vt\in\widetilde{\mathcal V}_i),
 \qquad
 [U_N^*f]_i=\P_{\mathrm s}(\widetilde{\mathcal V}_i)^{-1/2}
   \int_{\widetilde{\mathcal V}_i}f\,\odif{\P_{\mathrm s}}.
\]
The pieces cover the support up to a null set, so \(U_N\) is an isometry
on every realization, and it depends on the clean sample and \(k\) only.

When all cells are occupied, \(U_N\) maps each normalized sample-cell
vector to \(\indi{\mathcal V_a}/\P_{\mathrm s}(\mathcal V_a)^{1/2}\). Since
the pieces of a cell have equal mass and \([\widehat\mT]_{ij}\) depends on
\(i\) and \(j\) only through their cells, \(\widehat\mT U_N^*f=\bzero\)
whenever \(f\) is orthogonal to the cell indicators. So
\(U_N\widehat{\mT}U_N^*\) and \(\IOK^{\mathrm{step}}\) vanish off the span
of the cell indicators and have matrices \(\widehat\mH_{\mathrm{cell}}\) and
\(\mH_{\mathrm{cell}}\) on it, and
\[
 \|U_N\overline{\mK}^{(\mathrm s)}([N];r)U_N^*-\IOK\|
 \leq\|\overline{\mK}^{(\mathrm s)}([N];r)-\widehat\mT\|
 +\underbrace{\|U_N\widehat{\mT}U_N^*-\IOK^{\mathrm{step}}\|}_{=\|\widehat\mH_{\mathrm{cell}}-\mH_{\mathrm{cell}}\|}
 +\|\IOK^{\mathrm{step}}-\IOK\|
 \leq C_\nu\frac kN\beta_N,
\]
by \eqref{eqn:uniform-spectrum-step-errors} and
\eqref{eqn:uniform-spectrum-cell-matrix-error}. This is part~(a) of Lemma~\ref{lem:common-space-spectral-lift}.

\smallskip
\noindent\emph{Step 9: Lemma~\ref{lem:common-space-spectral-lift}(b).}
We let \(T\coloneqq U_N\mH U_N^*\), a compact self-adjoint operator. Since
\(U_N^*U_N=\mI_N\), its nonzero eigenvalues are those of \(\mH\), with
eigenvectors \(U_N\vv_i(\mH)\), and the compact-operator part,
Lemma~\ref{lem:weyl}(b), gives \(|\lambda_i^+(\mH)-\gamma_i|\leq\varepsilon\) for
every \(i\geq1\). Since \(\gamma_j\geq\dgap\),
\(\lambda_j^+(\mH)\geq\dgap-\varepsilon>0\), so \(\lambda_j(\mH)\) is a
positive eigenvalue of \(\mH\). Every other spectral value of \(T\) is at
distance at least \(\dgap-\varepsilon\) from \(\gamma_j\): a positive one is
within \(\varepsilon\) of some \(\gamma_i\) with \(i\ne j\), and a nonpositive
one is at distance at least \(\gamma_j\geq\dgap\). In particular,
\(\lambda_j(\mH)\) is simple. Let \(\widehat\Pi\) be the orthogonal
projector onto \(u=U_N\vv_j(\mH)\). The spectral theorem and
\((T-\gamma_j)\phi_j=(T-\IOK_r)\phi_j\) give
\[
 (\dgap-\varepsilon)\|(1-\widehat\Pi)\phi_j\|
 \leq\|(T-\gamma_j)\phi_j\|=\|(T-\IOK_r)\phi_j\|\leq\varepsilon,
\]
so \(\|(1-\widehat\Pi)\phi_j\|\leq4\varepsilon/(3\dgap)\), and, for the unit
vectors \(u\) and \(\phi_j\),
\[
 \min_{\chi=\pm1}\|u-\chi\phi_j\|^2=2-2|\langle u,\phi_j\rangle|
 \leq2(1-\langle u,\phi_j\rangle^2)=2\|(1-\widehat\Pi)\phi_j\|^2
 \leq\Bigl(\frac{2\varepsilon}{\dgap}\Bigr)^2.
\]
\end{proof}

\begin{proof}[Proof of Corollary~\ref{cor:union-clean-positive-spectrum}]
\smallskip
\noindent\emph{Step 1: one union.}
Fix a nonempty \(\sB\subseteq[B]\) and condition on \(\sI_\sB\), writing
\(N_\sB=|\sI_\sB|\). Given \(\sI_\sB\), the clean signals indexed by
\(\sI_\sB\) are iid from the law \(\P_{\mathrm{s}}^{(\sB)}\) of
\eqref{eqn:community-clean-law}, which has the properties that
Lemma~\ref{lem:uniform-clean-positive-spectrum} needs:
\begin{itemize}
\item its mixture weights are \(\alpha_b/\alpha_\sB\geq c_\alpha\);
\item its component laws satisfy Assumption~\ref{ass:local-mass-regularity}
with the same constants;
\item its separation \(\Delta_\sB\) is at least \(\Delta\), so
\(\min\{R_{\mathrm g},\Delta_\sB/2\}\geq\locrad\);
\item its operator is \(\IOK^{(\sB)}\).
\end{itemize}
For every index set with \(N_\sB\geq c_\alpha N/4\), apply
Lemma~\ref{lem:uniform-clean-positive-spectrum} to this sample of size
\(N_\sB\), whose local scale is \(h(\sI_\sB;k)\coloneqq L(k/N_\sB)^{1/d}\), with
\(w_1=\{c_\alpha/(4C_{\mathrm h})\}^{1/d}\), \(w_2=C_{\mathrm h}^{1/d}\), and
exponent \(\nu+2\); let \(c'_\nu\), \(C'_\nu\), and \(N'_\nu\) be its constants.
Since \(c_\alpha N/4\leq N_\sB\leq N\),
\begin{align}
 h_N(k)\leq h(\sI_\sB;k)\leq(4/c_\alpha)^{1/d}h_N(k),
 \label{eqn:union-scale-comparison}
\end{align}
so \(\radband_N(k)=[C_{\mathrm h}^{-1/d}h_N(k),C_{\mathrm h}^{1/d}h_N(k)]\subseteq[w_1h(\sI_\sB;k),w_2h(\sI_\sB;k)]\).
The conditions of the lemma hold for all these index sets when
\(N\geq4N'_\nu/c_\alpha\) and \eqref{eqn:cell-conditions} holds with
\(c_\nu\coloneqq c'_\nu(c_\alpha/4)^{1/d}\): then
\(\log N_\sB\leq\log N\leq c'_\nu k\) and
\(h(\sI_\sB;k)\leq(4/c_\alpha)^{1/d}h_N(k)\leq c'_\nu\locrad\); decreasing
\(c_\nu\) also gives \(k<c_\alpha N/4\leq N_\sB\), since
\(k/N=(h_N(k)/L)^d\leq c_\nu^d/c_{\mathrm g}\). The error of the lemma is
\(C'_\nu(k/N_\sB)(\log N_\sB/k)^{1/(d+2)}\leq C'_\nu(4/c_\alpha)(k/N)\beta_N(k)\),
and its failure probability is at most
\(N_\sB^{-\nu-2}\leq N^{-\nu-1}\) for all sufficiently large \(N\).

\smallskip
\noindent\emph{Step 2: all unions.} Averaging
over \(\sI_\sB\), adding \(\P(\mathcal E_{\mathrm{bal}}^c)\leq B\mathrm e^{-c_\alpha N/8}\)
for the size condition (Lemma~\ref{lem:class-count-concentration}), and a
union bound over the \(2^B-1\) sets \(\sB\)
prove the claim.
\end{proof}

\subsection{Justification of Remark~\ref{rem:circle-leading-mode-limitation}}
\label{app:eigengap-remark-proof}

\begin{proof}[Justification of Remark~\ref{rem:circle-leading-mode-limitation}]
\smallskip
\noindent\emph{Step 1: the model.}
Consider two disjoint uniform circles of radius \(L\) in a common plane,
with weights \(\alpha_1>\alpha_2>0\), \(\alpha_1+\alpha_2=1\), and the
constant kernel function \(\kf\equiv1\). In coordinates of the plane, take
the centers
\[
 \rvmu_1=(4\alpha_2L,0),\qquad\rvmu_2=(-4\alpha_1L,0).
\]
Then the mixture is centered, the gap between the circles is \(2L\), and
the planar covariance is
\(L^2\operatorname{diag}(1/2+16\alpha_1\alpha_2,\,1/2)\), as the model
requires.

\smallskip
\noindent\emph{Step 2: the spectrum.}
For a local radius \(r<2L\), let
\(\varphi_r=2\arcsin(r/(2L))\asymp r/L\) be the angle subtended by a chord
of length \(r\). On circle \(b\), the operator is \(\alpha_b\) times the
angular convolution with the indicator of an arc of half-angle
\(\varphi_r\). Its eigenfunctions are the Fourier modes, with eigenvalues
\[
 \frac{\alpha_b\varphi_r}{\pi}\quad\text{(constant mode)},
 \qquad
 \frac{\alpha_b\sin(n\varphi_r)}{\pi n}\quad\text{(order }n\geq1\text{, multiplicity two)}.
\]
Since \(|\sin(n\varphi_r)|/n\leq\sin\varphi_r\) for
\(0\leq\varphi_r\leq\pi/2\) and \(n\geq1\), the two largest eigenvalues are
\[
 \gamma_{1,r}=\frac{\alpha_1\varphi_r}{\pi},
 \qquad
 \gamma_{2,r}=\frac{\max\{\alpha_1\sin\varphi_r,\,\alpha_2\varphi_r\}}{\pi}
 =\frac{\alpha_1\sin\varphi_r}{\pi}
 \quad\text{once}\quad\frac{\sin\varphi_r}{\varphi_r}\geq\frac{\alpha_2}{\alpha_1},
\]
which holds for all sufficiently small \(\varphi_r\).

\smallskip
\noindent\emph{Step 3: the leading gap.}
Since \(\varphi_r-\sin\varphi_r\in[\varphi_r^3/12,\varphi_r^3/6]\) for
\(0\leq\varphi_r\leq1\),
\begin{align}
 \operatorname{gap}_1(\IOK_r)=\gamma_{1,r}-\gamma_{2,r}
 =\frac{\alpha_1}{\pi}(\varphi_r-\sin\varphi_r)
 \asymp\Bigl(\frac rL\Bigr)^3,
 \qquad\text{whereas}\qquad
 \gamma_{1,r}\asymp\frac rL.
 \label{eqn:circle-leading-gap}
\end{align}

\smallskip
\noindent\emph{Step 4: the gap condition.}
At radii in \(\radband_N(k)\), \(r\asymp h_N(k)=L\,k/N\) by
\eqref{eqn:local-knn-scale}, with \(d=1\). With \(a_N=b_N=0\), the gap
condition \eqref{eqn:spectral-gap-condition} therefore reads
\[
 \frac kN\Bigl(\frac{\log N}k\Bigr)^{1/3}=o\Bigl(\Bigl(\frac kN\Bigr)^3\Bigr),
 \qquad\text{that is,}\qquad
 k\gg N^{6/7}(\log N)^{1/7}.
\]
\end{proof}

\section{Proofs in Section~\ref{sec:below-separation}}
\label{app:below-separation-proofs}

The conventions on constants and the setting of
Appendix~\ref{app:radius-neighborhood-proofs} apply.
Appendix~\ref{app:mixture-regularity-proof} proves
Lemma~\ref{lem:mixture-regularity}.
Appendix~\ref{app:union-stability-proof} proves
Proposition~\ref{prop:union-stability} from the lemmas of
Appendices~\ref{app:radius-neighborhood-proofs}
and~\ref{app:spectral-convergence-proofs}, which are stated for every
union of communities.

\subsection{Proof of Lemma~\ref{lem:mixture-regularity}}
\label{app:mixture-regularity-proof}

\paragraph{Idea of the proof}
Below half the separation, a ball or a thin shell around a support point
of one community contains no point of another community. On such sets,
the law of a union of communities is therefore the law of that community,
up to its weight in the union.

\begin{proof}[Proof of Lemma~\ref{lem:mixture-regularity}]
The support of \(\P_{\mathrm{s}}^{(\sB)}\) is the union of the supports of
\(\P_{\mathrm{s}}^{(b)}\), \(b\in\sB\). Let \(b\in\sB\), let \(\vs\) be a
support point of \(\P_{\mathrm{s}}^{(b)}\), and let
\(0<r\leq\min\{R_{\mathrm g},\Delta/2\}\) and \(0<u\leq1/2\). When
\(B\geq2\), the ball of radius \(r\) around \(\vs\) and the shell with
outer radius \((1+u)r\leq3\Delta/4<\Delta\) contain no point of
\(\iota_{b'}(\manifold_{b'})\) for \(b'\ne b\), by the definition
\eqref{eqn:manifold-separation} of \(\Delta\). Hence, by
\eqref{eqn:community-clean-law},
\(\P_{\mathrm{s}}^{(\sB)}(\sX)=(\alpha_b/\alpha_\sB)\,\P_{\mathrm{s}}^{(b)}(\sX)\)
when \(\sX\) is this ball or this shell. Since
\(c_\alpha\leq\alpha_b\leq\alpha_b/\alpha_\sB\leq1\), the bounds
\eqref{eqn:uniform-clean-small-ball} and \eqref{eqn:uniform-clean-shell}
for \(\P_{\mathrm{s}}^{(b)}\) give those for \(\P_{\mathrm{s}}^{(\sB)}\) with
the constants \((c_\alpha c_{\mathrm g},C_{\mathrm g})\).
\end{proof}

\subsection{Proof of Proposition~\ref{prop:union-stability}}
\label{app:union-stability-proof}

\begin{proof}[Proof of Proposition~\ref{prop:union-stability}]
Under Assumptions~\ref{ass:local-mass-regularity}
and~\ref{ass:locality-separation}, \(B\geq2\) and
\(\locrad=\Delta/2\), so this is case~(ii) of the setting of
Appendix~\ref{app:radius-neighborhood-proofs}, and the valid radii and the
admissible neighbor counts of Section~\ref{sec:below-separation} are
those of that setting. Fix \(\nu>0\). The event
\(\mathcal E_N(\nu)\cap\mathcal E_{\mathrm D}\) fails with probability at
most \(C_\nu N^{-\nu}\) for all sufficiently large \(N\). On it,
part~(a) is Lemmas~\ref{lem:union-radius-stability}
and~\ref{lem:union-neighborhood-preservation}, and part~(b) is
Lemma~\ref{lem:union-spectral-stability}(a), with Weyl's inequality
(Lemma~\ref{lem:weyl}(a)) for the eigenvalues. Part~(c) is
Lemma~\ref{lem:union-spectral-stability}(b).
\end{proof}

\section{Proofs in Section~\ref{sec:recursive-clustering-guarantee}}
\label{app:recursive-recovery-proofs}

The conventions of Appendix~\ref{app:radius-neighborhood-proofs} apply,
with \(B\geq2\). The deterministic and probabilistic lemmas below are used
by both theorems and proved in Appendix~\ref{app:recovery-lemmas-proofs}.

\paragraph{Active sets}
Use the union notation \(\sI_\sB\) of
Appendix~\ref{app:radius-neighborhood-proofs} and write \(h_N=h_N(k)\).
While a set \(\sB\) of \(|\sB|\geq2\) communities remains, the algorithm is at
stage \(B-|\sB|+1\) and uses the quantile \(\omega_{B-|\sB|+1}\).

\paragraph{Deterministic extraction}
The first lemma identifies the component that a stage extracts, and the
second turns this into a recovery guarantee. For a symmetric matrix
\(\mK\in\R^{\sI\times\sI}\) with nonnegative entries, the
\emph{components} of \(\mK\) are the vertex sets of the connected
components of the graph with edge weights \(\mK_{ij}\), \(i\ne j\), and
the \emph{Perron root} of a component is the largest eigenvalue of the
principal submatrix of \(\mK\) on it.
\begin{lemma}[Selection of one maximizing component]
\label{lem:selected-community-extraction}
Let \(\mK\in\R^{\sI\times \sI}\) be a symmetric matrix with nonnegative
entries and positive leading eigenvalue \(\lambda_{\max}=\lambda_1(\mK)\),
and let \(\vv_1\) be its unit leading eigenvector given by
the convention of Section~\ref{sec:recursive-methodology}. Let
\(\mathfrak C\) be the component that contains the smallest index among
all components whose Perron root equals \(\lambda_{\max}\). Then
\begin{align}
    \Supp(\lambda_{\max}\vv_1)=\mathfrak C.
    \label{eqn:selected-community-weighted-support}
\end{align}
\end{lemma}
By Lemma~\ref{lem:selected-community-extraction}, an implementation may
compute each zero-threshold extraction as a connected component.

\begin{lemma}[Deterministic recovery from connected cores]
\label{lem:core-recovery}
Fix a labeled sample and an input matrix \(\mD\), let
\(\gO_N\subseteq[N]\) be a set of exceptional observations, and put
\(\gC_b^\circ=\gC_b\setminus \gO_N\) with \(|\gC_b^\circ|>k\) for every \(b\).
\begin{enumerate}[label=(\alph*)]
\item Suppose that, for every \(\sB\subseteq[B]\) with \(|\sB|\geq2\) and every
\(E\subseteq \gO_N\), the index set
\(\sI=(\bigcup_{b\in \sB}\gC_b^\circ)\cup E\) has a positive selected radius
\(r(\sI;k,\omega_{B-|\sB|+1})\), and the weighted graph of \(\mK(\sI)\)
satisfies the following:
\begin{enumerate}[label=(\roman*)]
\item each core \(\gC_b^\circ\), \(b\in \sB\), induces a connected graph;
\item different cores lie in different connected components;
\item every connected component with maximal Perron root meets a core.
\end{enumerate}
Then the algorithm
completes and
\begin{align}
 \mismatch_N(\vy,\widehat{\vy})\leq |\gO_N|/N.
 \label{eqn:recursive-core-mismatch}
\end{align}
\item Condition (iii) holds at \(\sI\) if every row sum of every core
submatrix \([|\sI|\mK(\sI)]_{\gC_b^\circ,\gC_b^\circ}\), \(b\in \sB\), is at least \(d_{\min}>0\)
and
\begin{align}
 d_{\min}>M_{\kf}(|\gO_N\cap \sI|-1)_+.
 \label{eqn:recursive-core-root-dominance}
\end{align}
\end{enumerate}
\end{lemma}
With \(\gO_N=\emptyset\), part~(a) gives exact recovery when the
components of every active union of communities are its communities.

\begin{lemma}[Two margins separate the communities]
\label{lem:margin-separation}
Let \(\sI\subseteq[N]\), let \(\mD\) satisfy
\(|D_{ij}-D^{(\mathrm s)}_{ij}|\leq aD^{(\mathrm s)}_{ij}+b\) for all distinct
\(i,j\in\sI\), with \(0\leq a<1\) and \(b\geq0\), and let the selected
radius \(r=r(\sI;k,\omega)\) satisfy
\((1-a)\rho_{\mathrm{in}}-b\leq r\leq(1+a)\rho_{\mathrm{out}}+b\) for some
\(0<\rho_{\mathrm{in}}\leq\rho_{\mathrm{out}}\). If, for some \(R>0\),
\begin{align}
 (1+a)R+2b<(1-a)\rho_{\mathrm{in}},
 \qquad
 (1+a)\rho_{\mathrm{out}}+2b<(1-a)\Delta,
 \label{eqn:two-margins}
\end{align}
then \(r>0\), every pair \(i,j\in\sI\) with \(D^{(\mathrm s)}_{ij}\leq R\)
has \(D_{ij}\leq r\), and every pair \(i,j\in\sI\) from different
communities has \(D_{ij}>r\). Consequently, for a regular kernel function,
the graph of \(\mK(\sI)\) has no edge between different communities, and
it connects every subset of \(\sI\) that has a clean spanning tree with
longest edge at most \(R\).
\end{lemma}
The first margin keeps the edges of a clean spanning tree with edges at
most \(R\) below the selected radius, and the second keeps every pair from
different communities above it.

\paragraph{Probabilistic lemmas}
The first lemma shows that a clean sample has a spanning tree whose edges
are a constant multiple of the sampling resolution
\(h_N(\log N)=L(\log N/N)^{1/d}\).
\begin{lemma}[A clean spanning tree]
\label{lem:clean-spanning-tree-event}
Let \(Q\) be a law with compact connected support \(\sS\) such that
\(Q(\{\vt:\|\vt-\vs\|\leq r\})\geq c_{\mathrm g}(r/L)^d\) for every
\(\vs\in\sS\) and \(0<r\leq R_{\mathrm g}\), and let \(c>0\) and \(\nu>0\).
There is a constant \(C_{\mathrm{tree}}\), depending only on \(\nu\), \(d\),
\(c_{\mathrm g}\), and \(c\), such that the following holds whenever
\(C_{\mathrm{tree}}h_N(\log N)\leq R_{\mathrm g}\) and \(n\geq cN\): with
probability at least \(1-C_\nu N^{-\nu}\), \(n\) iid points from \(Q\) have
a spanning tree with longest edge at most \(C_{\mathrm{tree}}h_N(\log N)\).
\end{lemma}
Under Assumption~\ref{ass:local-mass-regularity}, the lemma applies to
each community law with a connected support, and under
Assumption~\ref{ass:regular-cores}, to each conditional core law.

\begin{lemma}[Admissible neighbor counts above a floor]
\label{lem:admissible-ranks-floor}
Let \(\locrad\leq c_{\mathrm g}^{-1/d}L\), let \(C_{\mathrm h}\) be as in
\eqref{eqn:proof-setting}, and form \(\admset_N(\zeta)\) with these
quantities. Let \(\zeta_N\geq0\) and \(k_0\geq0\) be deterministic
sequences. If
\begin{align}
 \xi_N^2\max\bigl\{\zeta_N,\ h_N(\log N)\bigr\}=o(\locrad)
 \quad\text{and}\quad
 \xi_N^{d+1}k_0=o\Bigl(N\Bigl(\frac{\locrad}{L}\Bigr)^d\Bigr),
 \label{eqn:admissible-floor-existence}
\end{align}
then, for all sufficiently large \(N\), \(\admset_N(\zeta_N)\) contains a
count \(k\geq\xi_Nk_0\). Conversely, if \(\admset_N(\zeta_N)\)
contains such a count, then
\(\xi_N^2\max\{\zeta_N,h_N(\log N)\}\leq\locrad/C_{\mathrm h}^{2/d}\) and
\(\xi_N^{d+1}k_0\leq N(\locrad/L)^d/C_{\mathrm h}\).
\end{lemma}
Under Assumption~\ref{ass:local-mass-regularity} or
Assumption~\ref{ass:regular-cores}, with \(R_{\mathrm g}\geq\Delta\), the
lemma applies with \(\locrad=\Delta/2\), since the lower bound of
\eqref{eqn:uniform-clean-small-ball} at \(r=R_{\mathrm g}\) gives
\(R_{\mathrm g}\leq c_{\mathrm g}^{-1/d}L\).

\subsection{Proof of Theorem~\ref{thm:strong-consistency}}
\label{app:strong-consistency-proof}

\paragraph{Idea of the proof}
On the clean event, every clean selected radius is of order \(h_N\), and
each community has a clean spanning tree with edges of order
\(L(\log N/N)^{1/d}\), which is \(o(h_N)\). Two margins then make every
active graph retain the tree edges and drop every edge between
communities (Lemma~\ref{lem:margin-separation}), so that its components
are the communities, and Lemma~\ref{lem:core-recovery} applies with no
exceptional observations. Proposition~\ref{prop:exact-recovery-fixed-margins}
states this for fixed \(N\), and the theorem is its asymptotic case.

\begin{proposition}[Exact recovery with fixed margins]
\label{prop:exact-recovery-fixed-margins}
Suppose Assumption~\ref{ass:local-mass-regularity} holds with \(B\geq2\)
and connected supports, and write \(h_N=h_N(k)\), with the band constant
\(C_{\mathrm h}\) of \eqref{eqn:band-constants-communities}. Fix
\(\nu>0\), and let \(C_{\mathrm{ev},\nu}\) and \(N_0(\nu)\) be the
constants of Lemma~\ref{lem:clean-event} and \(C_{\mathrm{tree}}\) that of
Lemma~\ref{lem:clean-spanning-tree-event} with \(c=c_\alpha/4\). Let
\(N\geq N_0(\nu)\), let \(k\geq C_{\mathrm{ev},\nu}\log N\) be an integer,
and let \(0\leq a_N<1\) and \(b_N\geq0\). Suppose that the margins
\eqref{eqn:two-margins} hold with \(a=a_N\), \(b=b_N\),
\(R=C_{\mathrm{tree}}h_N(\log N)\),
\(\rho_{\mathrm{in}}=C_{\mathrm h}^{-1/d}h_N\), and
\(\rho_{\mathrm{out}}=C_{\mathrm h}^{1/d}h_N\), and that
\begin{align}
 C_{\mathrm h}^{1/d}h_N\leq\min\{R_{\mathrm g},\Delta/2\},
 \qquad
 C_{\mathrm{tree}}h_N(\log N)\leq R_{\mathrm g}.
 \label{eqn:corrected-recovery-fixed-margins}
\end{align}
Then, outside an event of probability at most \(C'_\nu N^{-\nu}\),
Algorithm~\ref{alg:RGG-clustering}, in the settings of
Section~\ref{sec:recursive-clustering-guarantee} with any quantile levels
in \((0,1)\), completes with zero mismatch on
\(\mathcal E_{\mathrm D}(a_N,b_N)\).
\end{proposition}

\begin{proof}[Proof of Proposition~\ref{prop:exact-recovery-fixed-margins}]
\smallskip
\noindent\emph{Step 1: the events.}
The first condition of \eqref{eqn:corrected-recovery-fixed-margins} gives
\eqref{eqn:finite-local-domain} for the family \(\{k\}\), and, as in
\eqref{eqn:admissible-count-upper-bound}, \(k/N\leq c_\alpha/16\). With
\(k\geq C_{\mathrm{ev},\nu}\log N\), these are the hypotheses of
Lemma~\ref{lem:clean-event}, so \(\mathcal E_N(\nu)\) fails with
probability at most \(B\mathrm e^{-c_\alpha N/8}+2N^{-\nu}\). On it, every
community has at least \(c_\alpha N/4>k\) observations, and every clean
selected radius on a union of communities belongs to \(\radband_N(k)\).
Conditionally on the labels, Lemma~\ref{lem:clean-spanning-tree-event}
applies to each community on \(\mathcal E_{\mathrm{bal}}\), so, except
with probability \(C_\nu N^{-\nu}\), every community has a clean spanning
tree with longest edge at most \(C_{\mathrm{tree}}h_N(\log N)\). Intersect these
two events with the affine event.

\smallskip
\noindent\emph{Step 2: separation on every active set.}
Let \(\sI\) be a union of at least two communities and
\(\omega\in(0,1)\). By Lemma~\ref{lem:affine-order-statistics},
\[
 (1-a_N)C_{\mathrm h}^{-1/d}h_N-b_N\leq r(\sI;k,\omega)
 \leq(1+a_N)C_{\mathrm h}^{1/d}h_N+b_N.
\]
By hypothesis, \eqref{eqn:two-margins} holds with
\(R=C_{\mathrm{tree}}h_N(\log N)\),
\(\rho_{\mathrm{in}}=C_{\mathrm h}^{-1/d}h_N\), and
\(\rho_{\mathrm{out}}=C_{\mathrm h}^{1/d}h_N\). By
Lemma~\ref{lem:margin-separation}, the selected radius is positive and
the connected components of the graph of \(\mK(\sI)\) are the communities
in \(\sI\). Hence conditions \textup{(i)}--\textup{(iii)} of
Lemma~\ref{lem:core-recovery} hold with \(\gO_N=\emptyset\), and part~(a)
of that lemma proves the claim.
\end{proof}

\begin{proof}[Proof of Theorem~\ref{thm:strong-consistency}]
By Proposition~\ref{prop:admissible-ranks} with \(\zeta_N=b_N\) and
\eqref{eqn:separation-condition}, \(\admset_N\) is nonempty for all
sufficiently large \(N\). Let \(k=k_N\in\admset_N\) be a sequence, write
\(h_N=h_N(k)\), fix \(\nu>0\), and write \(a_N\) and \(b_N\) for
\(C_\nu a_N\) and \(C_\nu b_N\), as in
Appendix~\ref{app:radius-neighborhood-proofs}. By
\eqref{eqn:window-bounds} and \eqref{eqn:affine-errors-vanish},
\[
 \frac{\log N}k,\quad\frac kN,\quad\frac{h_N}\Delta,\quad a_N+\frac{b_N}{h_N}
 \longrightarrow0.
\]
Hence \(k\geq C_{\mathrm{ev},\nu}\log N\) eventually, and
\(C_{\mathrm{tree}}h_N(\log N)/h_N=C_{\mathrm{tree}}(\log N/k)^{1/d}\to0\). Since
\(R_{\mathrm g}\geq\Delta\) by Assumption~\ref{ass:locality-separation},
also \(h_N=o(R_{\mathrm g})\) and
\(C_{\mathrm{tree}}h_N(\log N)\leq R_{\mathrm g}\) eventually. The margins
\eqref{eqn:two-margins} with the quantities of the proposition and
\eqref{eqn:corrected-recovery-fixed-margins} therefore hold for all
sufficiently large \(N\), and
Proposition~\ref{prop:exact-recovery-fixed-margins} gives the theorem.
\end{proof}

\subsection{Proof of Theorem~\ref{thm:weak-consistency}}
\label{app:population-core-proof}

\paragraph{Idea of the proof}
The exceptional set \(\gO_N\) consists of the observations outside the
cores, and, with high probability, it has \(o(k)\) elements. On every
possible active set, the core rows therefore still determine the selected
radius (Step~3), the two margins connect each core and separate the
communities (Step~4), and each core has a larger Perron root than any set
of exceptional observations (Step~5). Lemma~\ref{lem:core-recovery} then
bounds the mismatch by \(|\gO_N|/N\).

\begin{proof}[Proof of Theorem~\ref{thm:weak-consistency}]
By Lemma~\ref{lem:admissible-ranks-floor} with \(\locrad=\Delta/2\),
\(\zeta_N=b_N\), and \(k_0=N\alpha_{\mathrm{out}}\), conditions
\eqref{eqn:separation-condition} and \eqref{eqn:core-mass-condition} show
that \(\admset_N^\circ\) is nonempty for all sufficiently large \(N\).
Let \(k=k_N\in\admset_N^\circ\) be a sequence, write \(h_N=h_N(k)\), fix
\(\nu>0\), and write \(a_N\) and \(b_N\) for \(C_\nu a_N\) and
\(C_\nu b_N\), as in Appendix~\ref{app:radius-neighborhood-proofs}. By
\eqref{eqn:window-bounds}, \eqref{eqn:affine-errors-vanish}, and
\(k\geq\xi_NN\alpha_{\mathrm{out}}\),
\[
 \frac{\log N}k,\quad\frac{N\alpha_{\mathrm{out}}}k,\quad\frac kN,\quad
 \frac{h_N}\Delta,\quad a_N+\frac{b_N}{h_N}\longrightarrow0.
\]
The inner and outer radius factors of the clean radii below are
\[
 c_{\mathrm{in}}\coloneqq(2C_{\mathrm h})^{-1/d},
 \qquad
 c_{\mathrm{out}}\coloneqq(2C_{\mathrm h})^{1/d},
\]
with \(C_{\mathrm h}\) from \eqref{eqn:band-constants-communities}.

\smallskip
\noindent\emph{Step 1: the events.}
Write \(\gC_b^\circ=\{i\in\gC_b:\vu_i\in\manifold_b^\circ\}\) for the core
members, so that \(\gO_N=[N]\setminus\bigcup_b\gC_b^\circ\), and put
\[
 q_N\coloneqq(\mathrm e-1)N\alpha_{\mathrm{out}}+(\nu+3)\log N,
 \qquad
 c_{\mathrm{row}}\coloneqq\frac{c_\alpha c_{\mathrm g}}{16}\Bigl(\frac{c_{\mathrm{in}}}4\Bigr)^d.
\]
The proof works on the intersection of the affine event with the
following five events.
\begin{enumerate}[label=(E\arabic*)]
\item \emph{Exceptional count.} \(|\gO_N|\leq q_N\).
\item \emph{Core counts.} \(|\gC_b^\circ|\geq c_\alpha N/4\) for every \(b\).
\item \emph{Core rows.} Every \(i\in\gC_b^\circ\) has fewer than \(k/2\)
other members of \(\gC_b^\circ\) within clean distance
\(c_{\mathrm{in}}h_N\), and at least \(k\) within \(c_{\mathrm{out}}h_N\).
\item \emph{Core degrees.} Every \(i\in\gC_b^\circ\) has at least
\(c_{\mathrm{row}}k\) other members of \(\gC_b^\circ\) within clean
distance \(c_{\mathrm{in}}h_N/4\).
\item \emph{Core trees.} Every core sample \(\gC_b^\circ\) has a clean
spanning tree with longest edge at most \(C_{\mathrm{tree}}h_N(\log N)\).
\end{enumerate}

\smallskip
\noindent\emph{Step 2: probability of the events.}
The cores are deterministic sets, so the core flags are iid, and
\(|\gO_N|\) is binomial with success probability at most
\(\alpha_{\mathrm{out}}\). Each core has probability at least
\(\alpha_b-\alpha_{\mathrm{out}}\geq c_\alpha/2\) eventually. By
\eqref{eqn:binomial-additive-tail} and \eqref{eqn:binomial-half-mean} of
Lemma~\ref{lem:binomial-tails},
\[
 \P\{(\mathrm{E1})^c\}\leq N^{-(\nu+3)},
 \qquad
 \P\{(\mathrm{E2})^c\}\leq B\mathrm e^{-cN},
 \qquad
 q_N=o(k).
\]
For (E3)--(E5), condition on the labels and the core flags on (E2). The
points of \(\gC_b^\circ\) are then iid from the conditional core law,
which satisfies the bounds \eqref{eqn:uniform-clean-small-ball} at every
point of its support \(\iota_b(\manifold_b^\circ)\), up to the radius
\(R_{\mathrm g}\geq\Delta\), and the radii \(c_{\mathrm{out}}h_N\) and
\(C_{\mathrm{tree}}h_N(\log N)\) are below \(\Delta/2\) eventually. Since
\(|\gC_b^\circ|-1\geq c_\alpha N/8\), the conditional mean number of other
points of its core around a core point is
\[
 \leq C_{\mathrm g}c_{\mathrm{in}}^dk\leq\frac k{8\mathrm e}
 \ \text{within }c_{\mathrm{in}}h_N,
 \qquad
 \geq\frac{c_\alpha c_{\mathrm g}}8c_{\mathrm{out}}^dk\geq4k
 \ \text{within }c_{\mathrm{out}}h_N,
 \qquad
 \geq2c_{\mathrm{row}}k
 \ \text{within }c_{\mathrm{in}}h_N/4.
\]
Lemma~\ref{lem:local-binomial-counts} with a union bound over the core
points, and Lemma~\ref{lem:clean-spanning-tree-event} with
\(c=c_\alpha/4\), give, since \(k/\log N\to\infty\),
\[
 \P\{(\mathrm{E3})^c\cup(\mathrm{E4})^c\}\leq CN\mathrm e^{-ck}\leq N^{-\nu},
 \qquad
 \P\{(\mathrm{E5})^c\}\leq C_\nu N^{-\nu}.
\]
The intersection of (E1)--(E5) thus fails with probability at most
\(C_\nu N^{-\nu}\), and the rest of the proof is deterministic on it, for
all sufficiently large \(N\).

\smallskip
\noindent\emph{Step 3: clean radii on every active set.}
Consider every index set of the form
\begin{align}
 \sI=\Bigl(\bigcup_{b\in \sB}\gC_b^\circ\Bigr)\cup E,
 \qquad\emptyset\ne \sB\subseteq[B],\quad E\subseteq \gO_N,
 \label{eqn:core-active-sets}
\end{align}
and every \(\omega\in[\underline\omega,1-\underline\omega]\). We claim
that
\begin{align}
 c_{\mathrm{in}}h_N\leq r_{\mathrm s}(\sI;k,\omega)\leq c_{\mathrm{out}}h_N.
 \label{eqn:core-radius-bracket}
\end{align}
Let \(i\in\gC_b^\circ\subseteq\sI\). Since \(c_{\mathrm{in}}h_N<\Delta\),
the set \(\sI\) has no point of another community within clean distance
\(c_{\mathrm{in}}h_N\) of \(i\), so, by (E3) and (E1),
\[
 \#\{j\in\sI\setminus\{i\}:D^{(\mathrm s)}_{ij}\leq c_{\mathrm{in}}h_N\}
 <\frac k2+|\gO_N|<k,
 \qquad
 \#\{j\in\sI\setminus\{i\}:D^{(\mathrm s)}_{ij}\leq c_{\mathrm{out}}h_N\}
 \geq k.
\]
Hence the clean \(k\)-th nearest-neighbor distance of every core row lies
in \([c_{\mathrm{in}}h_N,c_{\mathrm{out}}h_N]\), whatever the set \(E\) is.
At most \(|\gO_N|\) rows of \(\sI\) are exceptional, and
\(|\sI|\geq c_\alpha N/4\), so \(|\gO_N|+1\leq\underline\omega|\sI|\)
eventually, and the index of the selected quantile satisfies
\[
 |\gO_N|<j\coloneqq\lceil\omega|\sI|\rceil\leq|\sI|-|\gO_N|.
\]
The \(j\)-th smallest row value therefore lies in the same bracket, which
proves \eqref{eqn:core-radius-bracket}.

\smallskip
\noindent\emph{Step 4: conditions (i) and (ii) of
Lemma~\ref{lem:core-recovery}.}
On the affine event, Lemma~\ref{lem:affine-order-statistics} and
\eqref{eqn:core-radius-bracket} give, for all sets
\eqref{eqn:core-active-sets},
\[
 (1-a_N)c_{\mathrm{in}}h_N-b_N\leq r(\sI;k,\omega)
                 \leq(1+a_N)c_{\mathrm{out}}h_N+b_N.
\]
Since
\[
 \frac{C_{\mathrm{tree}}h_N(\log N)}{h_N}=C_{\mathrm{tree}}\Bigl(\frac{\log N}k\Bigr)^{1/d},
 \quad a_N+\frac{b_N}{h_N},\quad\frac{h_N}\Delta\longrightarrow0,
\]
the two margins \eqref{eqn:two-margins} hold eventually with
\(R=C_{\mathrm{tree}}h_N(\log N)\), \(\rho_{\mathrm{in}}=c_{\mathrm{in}}h_N\),
and \(\rho_{\mathrm{out}}=c_{\mathrm{out}}h_N\). By
Lemma~\ref{lem:margin-separation} and (E5), the selected radius is
positive, every core in \(\sI\) induces a connected graph, and the graph
has no edge between different communities, including edges at
exceptional points. Different cores therefore lie in different connected
components.

\smallskip
\noindent\emph{Step 5: condition (iii).}
We verify Lemma~\ref{lem:core-recovery}(b). For large \(N\),
\(a_N\leq1/4\) and \(b_N\leq c_{\mathrm{in}}h_N/16\), so
\[
 (1+a_N)\frac{c_{\mathrm{in}}h_N}4+b_N
 \leq(1-a_N)c_{\mathrm{in}}h_N-b_N\leq r(\sI;k,\omega),
\]
and the core neighbors of (E4) are joined to \(i\) in the graph. By
\eqref{eqn:profile-positive-bounds}, every row sum of each core
submatrix of \(|\sI|\mK(\sI)\) is then at least
\[
 d_{\min}\coloneqq c_{\kf}c_{\mathrm{row}}k,
\]
and, since \(|\gO_N|=o(k)\), \eqref{eqn:recursive-core-root-dominance}
holds eventually.

\smallskip
\noindent\emph{Step 6: conclusion.}
Since \(|\gC_b^\circ|\geq c_\alpha N/4>k\), Lemma~\ref{lem:core-recovery}(a)
shows that the algorithm completes and
\[
 \mismatch_N\leq\frac{|\gO_N|}N\leq\frac{q_N}N
 \leq C_\nu\Bigl(\alpha_{\mathrm{out}}+\frac{\log N}N\Bigr),
\]
which proves \eqref{eqn:recursive-core-probabilistic-mismatch}.

On the affine event minus the five events, which has probability at most
\(C_\nu N^{-\nu}\), the output convention of
Algorithm~\ref{alg:RGG-clustering} defines \(\widehat\vy\in[B]^N\), and
\(\mismatch_N\leq1\). Hence
\[
 \E\bigl[\mismatch_N\mathds{1}_{\mathcal E_{\mathrm D}}\bigr]
 \leq\frac{\E|\gO_N|}N+C_\nu N^{-\nu}
 \leq\alpha_{\mathrm{out}}+C_\nu N^{-\nu}.
\]
\end{proof}

\subsection{Proof of Corollary~\ref{cor:denoised-recovery}}
\label{app:denoised-recovery-proof}

\begin{proof}[Proof of Corollary~\ref{cor:denoised-recovery}]
Fix \(\nu>0\), and let \(C_\nu\) be the constant of
Theorem~\ref{thm:denoising-error-bound}(b), so that, for
\(\mD=\widetilde\mD\), the event
\(\mathcal E_{\mathrm D}\coloneqq\mathcal E_{\mathrm D}(C_\nu\widetilde a_N,C_\nu\widetilde b_N)\)
fails with probability at most \(C_\nu N^{-\nu}\). This is the event of
Section~\ref{sec:stability-of-data-driven-selection} with
\(a_N=\widetilde a_N\), \(b_N=\widetilde b_N\), and \(C=C_\nu\), so that
\(\admset_N=\admset_N(\widetilde b_N)\) in both theorems. The sampling
condition \eqref{eqn:sampling-condition} and the separation-to-noise
condition \(\xi_N^2\widetilde b_N=o(\Delta)\) together are the separation
condition \eqref{eqn:separation-condition} with
\(b_N=\widetilde b_N\). By Theorem~\ref{thm:strong-consistency}, the event
that \(\mathcal E_{\mathrm D}\) holds and the algorithm does not complete
with zero mismatch has probability at most \(C'_\nu N^{-\nu}\). The
algorithm therefore fails with probability at most
\(\P(\mathcal E_{\mathrm D}^c)+C'_\nu N^{-\nu}\leq(C_\nu+C'_\nu)N^{-\nu}\),
which is part~(a). Part~(b) follows from
Theorem~\ref{thm:weak-consistency} in the same way; for the expectation,
\(\E\mismatch_N\leq\E[\mismatch_N\mathds{1}_{\mathcal E_{\mathrm D}}]
+\P(\mathcal E_{\mathrm D}^c)\leq\alpha_{\mathrm{out}}+2C_\nu N^{-\nu}\),
and the constant is renamed.
\end{proof}

\subsection{Proofs of the lemmas}
\label{app:recovery-lemmas-proofs}

\begin{proof}[Proof of Lemma~\ref{lem:selected-community-extraction}]
We identify the eigenspace of \(\lambda_{\max}\) and the columns of its
orthogonal projector, and then read off the vector chosen by the
convention.

\smallskip
\noindent\emph{Step 1: the leading eigenspace.}
After a permutation of the indices, \(\mK\) is block diagonal with one
block per component, so its spectrum is the union of the spectra of the
blocks. Let \(\mathfrak c_1,\ldots,\mathfrak c_q\) be the components whose
Perron root equals \(\lambda_{\max}\). Each block indexed by some
\(\mathfrak c_\ell\) is a symmetric nonnegative matrix whose graph is
connected, that is, an irreducible matrix. By the Perron--Frobenius
theorem, its Perron root \(\lambda_{\max}\) is a simple eigenvalue of the
block with an eigenvector whose entries are all positive. Write
\(\vv_{\mathfrak c_\ell}\in\R^\sI\) for the unit vector equal to this eigenvector on
\(\mathfrak c_\ell\) and to zero off \(\mathfrak c_\ell\). A block whose Perron root is smaller than
\(\lambda_{\max}\) has no eigenvalue equal to \(\lambda_{\max}\), since its
Perron root is its largest eigenvalue. Hence
\[
 \ker(\mK-\lambda_{\max}\mI)
 =\operatorname{span}\{\vv_{\mathfrak c_1},\ldots,\vv_{\mathfrak c_q}\},
\]
and these vectors form an orthonormal basis of it.

\smallskip
\noindent\emph{Step 2: columns of the projector.}
The orthogonal projector onto this eigenspace is
\(\mP_\star=\sum_{\ell=1}^q\vv_{\mathfrak c_\ell}\vv_{\mathfrak c_\ell}^{\sT}\).
Since the supports of the vectors \(\vv_{\mathfrak c_\ell}\) are disjoint,
\begin{align}
    (\mP_\star)_{\cdot i}
       =\begin{cases}
          [\vv_{\mathfrak c_\ell}]_i\vv_{\mathfrak c_\ell},&i\in\mathfrak c_\ell,\quad\ell\in[q],\\
          \bzero,&\text{otherwise}.
        \end{cases}
    \label{eqn:recursive-coordinate-projection}
\end{align}

\smallskip
\noindent\emph{Step 3: the selected vector.}
By \eqref{eqn:recursive-coordinate-projection} and the positivity of
\([\vv_{\mathfrak c_\ell}]_i\) for \(i\in\mathfrak c_\ell\), the column \((\mP_\star)_{\cdot i}\) is nonzero
exactly when \(i\) lies in one of \(\mathfrak c_1,\ldots,\mathfrak c_q\). Hence the
first nonzero column of \(\mP_\star\) has the smallest index \(i_\star\) in
\(\mathfrak c_1\cup\cdots\cup\mathfrak c_q\), which lies in \(\mathfrak C\), and
\[
 \vv_1=\frac{(\mP_\star)_{\cdot i_\star}}{\|(\mP_\star)_{\cdot i_\star}\|}=\vv_{\mathfrak C}.
\]
The entries of \(\vv_{\mathfrak C}\) are positive on \(\mathfrak C\) and zero
elsewhere, and \(\lambda_{\max}>0\), which proves
\eqref{eqn:selected-community-weighted-support}.
\end{proof}

\begin{proof}[Proof of Lemma~\ref{lem:core-recovery}]
\smallskip
\noindent\emph{Part (a), Step 1: the invariant.}
We show by induction that, at every stage, the active set has the form
\[
 \Bigl(\bigcup_{b\in\sB}\gC_b^\circ\Bigr)\cup E,
 \qquad\sB=\text{the remaining cores},\quad E\subseteq\gO_N.
\]
Initially every core is present. Suppose the active set has this form at
stage \(B-|\sB|+1\), where \(\sB\) is the set of remaining cores; it is
then a set \(\sI\) of the hypothesis, so (i)--(iii) hold. Any component meeting a
remaining core contains that entire core by (i), and no other core by
(ii). Since a connected core has more than \(k\geq1\) vertices and its
edges have positive weight, the active matrix has a positive leading
eigenvalue. By (iii) and Lemma~\ref{lem:selected-community-extraction},
the stage extracts a component that contains exactly one whole remaining
core.
It can contain exceptional observations, but no observation of another
core. Every other core remains intact, so the active size exceeds \(k\)
while an extraction stage remains, and the positive radius prevents a
failure.

\smallskip
\noindent\emph{Part (a), Step 2: conclusion.}
Induction gives \(B-1\) extractions containing distinct cores,
and the final remainder contains the last core. Matching each extracted
set to the community of its core gives one common permutation that
labels every observation outside \(\gO_N\) correctly, which proves
\eqref{eqn:recursive-core-mismatch}.

\smallskip
\noindent\emph{Part (b).}
The entries of \(|\sI|\mK(\sI)\) are at most \(M_{\kf}\). By the Rayleigh
principle, a core submatrix with row sums at least \(d_{\min}\), and the
component that contains it, have spectral radius at least \(d_{\min}\). A
component that meets no core consists of exceptional observations only,
and, because the kernel is hollow, its spectral radius is at most
\(M_{\kf}(|\gO_N\cap \sI|-1)_+\). With \eqref{eqn:recursive-core-root-dominance}, for
the kernel \(\mK(\sI)\),
\begin{align}
 \bigl(\text{Perron root of a core component}\bigr)
 \geq &\,\frac{d_{\min}}{|\sI|}
 >\frac{M_{\kf}(|\gO_N\cap\sI|-1)_+}{|\sI|}\\
 \geq &\,\bigl(\text{Perron root of an exceptional component}\bigr).
\end{align}
\end{proof}

\begin{proof}[Proof of Lemma~\ref{lem:margin-separation}]
By the first margin, the bound on \(D_{ij}\), and, for pairs from
different communities, \(D^{(\mathrm s)}_{ij}\geq\Delta\) and the second
margin,
\begin{align}
 r&\geq(1-a)\rho_{\mathrm{in}}-b>(1+a)R+b\geq0,\\
 D_{ij}&\leq(1+a)R+b<r
 &&\text{if }D^{(\mathrm s)}_{ij}\leq R,\\
 D_{ij}&\geq(1-a)\Delta-b>(1+a)\rho_{\mathrm{out}}+b\geq r
 &&\text{if \(i\) and \(j\) lie in different communities}.
\end{align}
A regular
kernel function is positive on \([0,1]\), so the edges of the graph of
\(\mK(\sI)\) are the pairs with \(D_{ij}\leq r\). This gives the last
claim.
\end{proof}

\begin{proof}[Proof of Lemma~\ref{lem:clean-spanning-tree-event}]
\smallskip
\noindent\emph{Step 1: every ball is occupied.}
Put \(\delta=C_{\mathrm{tree}}h_N(\log N)/4\), and choose a maximal
\(\delta\)-separated set of centers in \(\sS\). The balls of radius
\(\delta/3\) around the centers are disjoint and have mass at least
\(c_{\mathrm g}(\delta/(3L))^d\). The number of sample points in the ball
of radius \(\delta/2\) around a center is binomial, so, with
\(C_{\mathrm{tree}}\geq1\) and \eqref{eqn:binomial-half-mean} of
Lemma~\ref{lem:binomial-tails},
\[
 \#\{\text{centers}\}\leq C\Bigl(\frac L\delta\Bigr)^d\leq\frac{CN}{\log N},
 \qquad
 \text{mean}\geq cNc_{\mathrm g}\Bigl(\frac\delta{2L}\Bigr)^d=c'C_{\mathrm{tree}}^d\log N,
 \qquad
 \P\{\text{empty}\}\leq N^{-c'C_{\mathrm{tree}}^d/8}.
\]
Choose \(C_{\mathrm{tree}}\) so that
\(c'C_{\mathrm{tree}}^d/8\geq\nu+1\). A union bound over the centers shows
that, with probability at least \(1-C_\nu N^{-\nu}\), every such ball
contains a sample point. All these radii are at most \(R_{\mathrm g}\).

\smallskip
\noindent\emph{Step 2: a spanning tree.}
The closed balls of radius \(\delta\) around the centers cover the
connected set \(\sS\), so their intersection graph is connected: otherwise
two disjoint nonempty closed unions would disconnect \(\sS\). Choose a
sample point within \(\delta/2\) of each center as its representative.
Then
\begin{align}
 \|\vs'-\vs''\|\leq &\, 3\delta\ \text{for representatives \(\vs',\vs''\) of intersecting balls},\\
 \|\vs-\vs'\|\leq &\, \tfrac32\delta\ \text{for every sample point \(\vs\) and some representative \(\vs'\)},
\end{align}
hence these edges connect the whole sample and contain a spanning tree with
longest edge at most \(3\delta\leq C_{\mathrm{tree}}h_N(\log N)\).
\end{proof}

\begin{proof}[Proof of Lemma~\ref{lem:admissible-ranks-floor}]
Let \(x_N\) and \(y_N\) be the two ends \eqref{eqn:admissible-ends} of
the range of admissible counts, for \(\zeta=\zeta_N\), so that
\(y_N=N(\locrad/L)^d/(C_{\mathrm h}\xi_N^d)\).

\smallskip
\noindent\emph{Necessity.} If \(\admset_N(\zeta_N)\) contains a count
\(k\geq\xi_Nk_0\), then
\[
 x_N\leq k\leq y_N\quad\text{and}\quad\xi_Nk_0\leq k\leq y_N;
\]
the first gives the first bound, as in the proof of
Proposition~\ref{prop:admissible-ranks}, and the second is the second
bound.

\smallskip
\noindent\emph{Sufficiency.} Under \eqref{eqn:admissible-floor-existence},
as in that proof, for all sufficiently large \(N\),
\[
 y_N\geq2x_N\geq2,
 \qquad
 y_N\geq2\xi_Nk_0.
\]
With \(x_N'\coloneqq\max\{x_N,\xi_Nk_0\}\geq1\), the interval
\([x_N',y_N]\) has length at least \(x_N'\geq1\), so it contains an
integer \(k\geq\xi_Nk_0\). Since
\(\locrad\leq c_{\mathrm g}^{-1/d}L\) and
\(c_{\mathrm g}C_{\mathrm h}\geq16\), \(y_N<N-1\), so \(k\) is
admissible.
\end{proof}

\section{Proofs in Section~\ref{sec:denoising-necessity}}
\label{app:raw-necessity}

The conventions on constants of Appendix~\ref{app:denoising} apply.
Proposition~\ref{prop:raw-necessity} below, on the failure of the raw
distances below the noise-fluctuation scale, is used to prove
Proposition~\ref{prop:denoising-necessity} and is proved in
Appendix~\ref{app:raw-necessity-proof}.

\paragraph{Raw distances below the noise-fluctuation scale}
The squared raw distances fluctuate at the scale
\(s_p\coloneqq\sigma^2\sqrt{6p}\), the standard deviation of
\(\|\noise_i-\noise_j\|^2\) up to a constant. The next proposition treats local
scales that are below it by a polynomial factor.
\begin{proposition}[Raw distances below the noise-fluctuation scale]
\label{prop:raw-necessity}
Suppose the noise is isotropic Gaussian, \(\noise_j=\sigma\vg_j\) with
\(\vg_1,\ldots,\vg_N\simiid\Normal(\bzero,\mI_p)\) independent of the
clean signals, and that
Assumptions~\ref{ass:local-mass-regularity} and~\ref{ass:locality-separation} hold,
and fix a regular kernel function, \(\omega\in(0,1)\), and \(c_0>0\).
Suppose that:
\begin{enumerate}[label=(\roman*)]
\item \(\sigma>0\) and \(p\geq\log^{10}N\);
\item \(k=k_N\in\admset_N(0)\) is a deterministic sequence with
\(\log k=o(\sqrt{\log N})\);
\item with \(h_N=h_N(k)\),
\begin{align}
 h_N^2&\leq N^{-c_0}\sigma^2\sqrt p,
 &s_p\log N&\leq R_{\mathrm g}^2/4.
 \label{eqn:raw-necessity-scales}
\end{align}
\end{enumerate}
Let \(r_{\mathrm x}=r([N];k,\omega)\) be the radius selected by
Algorithm~\ref{alg:radius-selection} from \(\mD^{(\mathrm x)}\). Then,
with probability tending to one, there is a set
\(\mathcal Z\subseteq[N]\) such that the following hold.
\begin{enumerate}[label=(\alph*)]
\item \emph{Radius.}
\(|r_{\mathrm x}^2-2p\sigma^2|\leq C\sigma^2\sqrt{p\log N}\), whereas
\(ch_N\leq r_{\mathrm s}([N];k,\omega)\leq Ch_N\).
\item \emph{Isolation.} \(\gN_i([N];r_{\mathrm x})=\emptyset\) for every
\(i\in\mathcal Z\), and
\(|\mathcal Z\cap\gC_b|\geq(1-\omega-o(1))|\gC_b|\) for every \(b\in[B]\).
\item \emph{Neighborhoods.} For the raw neighborhoods,
\(T_N(k,\omega)\leq\omega+o(1)\).
\item \emph{Clustering.} If \(B=2\) and the first stage of
Algorithm~\ref{alg:RGG-clustering} uses \(\omega_1=\omega\), then for
every threshold \(\tau_1\geq0\) every \(i\in\mathcal Z\) receives label
\(2\), and
\(\mismatch_N\geq(1-\omega-o(1))\min_b|\gC_b|/N\).
\end{enumerate}
\end{proposition}
The isolated observations also separate the raw kernel and its embedding
from the clean ones.
\begin{corollary}[Kernels and embeddings from the raw distances]
\label{cor:raw-kernel-embedding}
In the setting of Proposition~\ref{prop:raw-necessity}, write
\(\mK^{(\mathrm x)}\) and \(\mK^{(\mathrm s)}\) for the kernel matrices of
\(\mD^{(\mathrm x)}\) and \(\mD^{(\mathrm s)}\) on \([N]\) at their
selected radii. With probability tending to one, the following hold with
the set \(\mathcal Z\) of the proposition.
\begin{enumerate}[label=(\alph*)]
\item \((N/k)\|\mK^{(\mathrm x)}-\mK^{(\mathrm s)}\|\geq c\) for a
constant \(c>0\).
\item Let \(\sJ\) be a fixed block of eigenvalue indices, for example
\(\sJ=\{1,2\}\), and \(c_{\sJ}>0\) a constant. On the
event that
\(\min_{j\in \sJ}\lambda_j(\mK^{(\mathrm s)})\geq c_{\sJ}k/N\), the rows
of \(\mZ_{\sJ}(\mK^{(\mathrm x)})\) indexed by \(\mathcal Z\) vanish, and
\((N/k)\|\mZ_{\sJ}(\mK^{(\mathrm x)})-\mZ_{\sJ}(\mK^{(\mathrm s)})\mQ\|_{\mathrm F}
\geq c_{\sJ}\{(1-\omega)|\sJ|\}^{1/2}-o(1)\) for every orthogonal \(\mQ\).
\end{enumerate}
\end{corollary}

\subsection{Proof of Proposition~\ref{prop:denoising-necessity}}
\label{app:denoising-necessity-proof}

\paragraph{Idea of the proof}
Lemma~\ref{lem:concentric-circles} verifies, for the concentric circles
of Section~\ref{sec:denoising-necessity} and at every admissible neighbor
count, the hypotheses of Corollary~\ref{cor:denoised-recovery}(a) for the
denoised distances and those of Proposition~\ref{prop:raw-necessity} for
the raw distances. For the latter, every admissible count has
\(h_N(k)\leq\Delta\), and
\(\Delta^2\asymp N^{1/2}\log^6N\leq N^{-1/2}\sigma^2\sqrt p\).

\begin{lemma}[Concentric circles]
\label{lem:concentric-circles}
Consider the model of Section~\ref{sec:denoising-necessity}: two equally
weighted concentric circles of radii \(\varrho_1=L/4\) and
\(\varrho_2=L/4+\Delta\), with isotropic Gaussian noise, \(\sigma=1\), \(p=\lceil N^{5/2}\rceil\),
\(L=\sqrt p\), and \(\Delta=L\log^3N/N\). For all sufficiently large
\(N\), the following hold.
\begin{enumerate}[label=(\alph*)]
\item \emph{Model.} Assumptions~\ref{ass:growing-covariance},
\ref{ass:local-mass-regularity}, and~\ref{ass:locality-separation} hold,
with \(d=1\), \(R_{\mathrm g}=L/8\), and absolute constants, and
\(\SNR_N\to1/16\).
\item \emph{Admissible counts.} For every \(\zeta\leq h_N(\log N)\), a
count lies in \(\admset_N(\zeta)\) if and only if
\(C_{\mathrm h}\xi_N\log N\leq k\leq\log^3N/(2C_{\mathrm h}\xi_N)\). This
range contains \(\lceil\log^2N\rceil\), and every count in it is at most
\(\log^3N\).
\item \emph{Denoised distances.} The hypotheses of
Corollary~\ref{cor:denoised-recovery}(a) and the signal-to-noise
condition \eqref{eqn:snr-condition} hold, for either setting of
\(\mathtt{debias}\), and \(\admset_N(\widetilde b_N)=\admset_N(0)\).
\item \emph{Raw distances.} The hypotheses of
Proposition~\ref{prop:raw-necessity} hold for every deterministic
sequence \(k=k_N\in\admset_N(0)\), with \(c_0=1/2\).
\end{enumerate}
\end{lemma}
\begin{proof}[Proof of Lemma~\ref{lem:concentric-circles}]
\noindent\emph{Part~(a).} Since \(L/4\leq\varrho_b\leq L/2\), for a center
\(\vs\) on circle \(b\) and \(0<r\leq L/8\),
\[
 \P(\|\clean-\vs\|\leq r\mid\rY=b)=\frac2\pi\arcsin\frac r{2\varrho_b}
 \in\Bigl[\frac{2r}{\pi L},\frac{2r}L\Bigr].
\]
For \(u\leq1/2\), the shell probability is at most
\((2/\pi)(8/\sqrt{55})(4ur/L)\), since \(r/(2\varrho_b)\leq1/4\) and
\(\arcsin'\leq8/\sqrt{55}\) on \([0,3/8]\). Hence
Assumption~\ref{ass:local-mass-regularity} holds with \(d=1\),
\(R_{\mathrm g}=L/8\), and absolute constants, so the band constant
\(C_{\mathrm h}\) is absolute. The supports are at distance
\(\varrho_2-\varrho_1=\Delta\leq R_{\mathrm g}\), so
Assumption~\ref{ass:locality-separation} holds. Each circle is centered
at the origin with covariance \((\varrho_b^2/2)\mI_2\), so the mixture is
centered, and \(\mSigma=\theta\mI_2\) with
\[
 \theta=\frac{\varrho_1^2+\varrho_2^2}4=\frac{L^2}{32}(1+o(1)),
 \qquad
 \|\mSigma^{-1/2}\mJ^{\sT}\clean\|^2\leq\frac{\varrho_2^2}\theta\leq2\Bigl(1+\frac{4\Delta}L\Bigr)^2\leq4=2m,
 \qquad
 \SNR_N=\frac{2\theta}p\to\frac1{16},
\]
and \(\theta_{\max}/\theta_{\min}=1\).
\smallskip
\noindent\emph{Part~(b).} Here \(h_N(k)=Lk/N\), so
\(h_N(\log N)=L\log N/N\asymp N^{1/4}\log N\) and
\(\Delta=h_N(\log^3N)\). By \eqref{eqn:admissible-range}, for every
\(\zeta\leq h_N(\log N)\), a count lies in \(\admset_N(\zeta)\) if and only
if \(C_{\mathrm h}\xi_N\log N\leq k\leq\log^3N/(2C_{\mathrm h}\xi_N)\). Since
\(\xi_N\leq\sqrt{\log N}\) for \(N\geq3\), this range contains \(\lceil\log^2N\rceil\),
and every count in it is at most \(\log^3N\).
\smallskip
\noindent\emph{Part~(c).}
Since \(m=2\) and \(N/(p-m)\to0\), for either setting of \(\mathtt{debias}\),
\[
 \widetilde a_N=\frac1\theta\Bigl(1+\sqrt{\frac{p-2}N}\Bigr)\asymp N^{-7/4},
 \qquad
 \rho_N=\frac p{N\theta}\asymp\frac1N,
 \qquad
 \widetilde b_N\asymp\sqrt{\log N}.
\]
Thus \(m\log N=o(N)\) and \(\widetilde a_N\log N\to0\). Isotropic
Gaussian noise satisfies Assumption~\ref{ass:noise-variance-structure}
with \(\mGamma=\mI_p\) and an absolute constant
\(C_{\mathrm{sg}}\), so Theorem~\ref{thm:denoising-error-bound}(b) applies to
\(\widetilde\mD\), with \(a_N=\widetilde a_N\) and
\(b_N=\widetilde b_N\). Since \(\Delta/\sigma\asymp N^{1/4}\log^3N\) and
\((1+\rho_N)(m+\log N)\asymp\log N\), the separation-to-noise condition
\eqref{eqn:denoised-separation-noise} holds, and \(\SNR_N\to1/16\)
satisfies the signal-to-noise condition \eqref{eqn:snr-condition}.
Moreover, \(\widetilde b_N\leq h_N(\log N)\), so
\(\admset_N(\widetilde b_N)=\admset_N(0)\) by~(b), and
\(\xi_N^2h_N(\log N)=\xi_N^2\Delta/\log^2N=o(\Delta)\), so the separation condition
\eqref{eqn:separation-condition} holds with \(b_N=\widetilde b_N\).
These are the hypotheses of Corollary~\ref{cor:denoised-recovery}(a).
\smallskip
\noindent\emph{Part~(d).} Let \(k=k_N\in\admset_N(0)\). By
\eqref{eqn:window-bounds}, \(h_N(k)\leq\Delta/(2C_{\mathrm h}\xi_N)\leq\Delta\), so
\[
 h_N(k)^2\leq\Delta^2\asymp N^{1/2}\log^6N\leq N^{-1/2}\sqrt p.
\]
Moreover,
\(\sqrt{6p}\log N\leq p/256=R_{\mathrm g}^2/4\), \(p\geq\log^{10}N\), and,
by~(b), \(\log k\leq3\log\log N=o(\sqrt{\log N})\), so
the hypotheses of Proposition~\ref{prop:raw-necessity} hold for every
deterministic sequence \(k=k_N\in\admset_N(0)\), with \(c_0=1/2\).
\end{proof}

\begin{proof}[Proof of Proposition~\ref{prop:denoising-necessity}]
The claims about the admissible counts are parts~(b) and~(c) of
Lemma~\ref{lem:concentric-circles}, and part~(a) of the proposition is
its part~(c). By its part~(d), Proposition~\ref{prop:raw-necessity}
applies to \(\mD^{(\mathrm x)}\) at every deterministic sequence
\(k=k_N\in\admset_N(0)\). Its parts~(a)--(d), with \(|\gC_b|/N\to1/2\) in
probability, give part~(b) of the proposition.
\end{proof}

\paragraph{The block \(\sJ=\{1,2\}\)}
In this model, Corollary~\ref{cor:raw-kernel-embedding}(b) applies to
\(\sJ=\{1,2\}\), the block that carries the community structure. On the
clean event \(\mathcal E_N(1)\) of Lemma~\ref{lem:clean-event}, no clean
edge joins the two circles and every clean degree is at least
\(c_{\mathrm{deg}}k\), so the Rayleigh quotients of the normalized
indicators of the two communities and the min--max principle give
\[
 \lambda_2(\mK^{(\mathrm s)})\geq c_{\kf}c_{\mathrm{deg}}\frac kN.
\]

\subsection{Proof of Proposition~\ref{prop:raw-necessity}}
\label{app:raw-necessity-proof}

\paragraph{Idea of the proof}
Write \((D^{(\mathrm x)}_{ij})^2=\|\noise_i\|^2+R_{ij}\). The term
\(\|\noise_i\|^2\) is common to row \(i\) and fluctuates at the scale
\(s_p\) across rows. The \(k\) smallest values of \(R_{ij}\) in a row lie
in a window of width \(o(s_p)\) around a deterministic level
(Lemma~\ref{lem:raw-tails} and Steps~6--9), because the clean squared
distances are below \(s_p\) by a polynomial factor. The \(k\)-th
nearest-neighbor distance of row \(i\) is therefore governed by
\(\|\noise_i\|^2\), the selected radius is its \(\omega\)-quantile, and
the rows with a larger \(\|\noise_i\|^2\), a fraction \(1-\omega\), have
no neighbor at that radius.

\begin{lemma}[Tails of the raw cross term]
\label{lem:raw-tails}
Let \(\noise_i,\noise_j\) be independent \(\Normal(\bzero,\sigma^2\mI_p)\),
suppose \(\sigma>0\) and \(p\geq\log^{10}N\), and put \(s_p=\sigma^2\sqrt{6p}\) and
\(R^0_{ij}=\|\noise_j\|^2-2\langle\noise_i,\noise_j\rangle\). There are an
absolute constant \(c>0\) and a deterministic \(\varepsilon_N\to0\) with the
following property. For all sufficiently large \(N\), uniformly in
\(p\geq\log^{10}N\), and whenever
\(|\|\noise_i\|^2-p\sigma^2|\leq4\sigma^2\sqrt{p\log N}\),
\begin{align}
 \P\{R^0_{ij}\leq p\sigma^2-xs_p\mid\noise_i\}
 &\leq\exp\{-(1-\varepsilon_N/\log N)x^2/2+\varepsilon_N\}
 &&\text{for every }x\geq0,
 \label{eqn:raw-necessity-upper-tail}\\
 \P\{R^0_{ij}\leq p\sigma^2-xs_p\mid\noise_i\}
 &\geq c(1+x)^{-2}\exp\{-x^2/2-\varepsilon_N\}
 &&\text{for every }0\leq x\leq3\sqrt{\log N}.
 \label{eqn:raw-necessity-lower-tail}
\end{align}
\end{lemma}
\begin{proof}[Proof of Lemma~\ref{lem:raw-tails}]
In an orthonormal basis
whose first vector is \(\noise_i/\|\noise_i\|\), write
\(\noise_j=\sigma\vg\) and \(\upsilon=\|\noise_i\|/\sigma\). Then
\begin{align}
 R^0_{ij}-p\sigma^2=\sigma^2(Q_\perp+Q_{\mathrm{lin}}+Q_1),\qquad
 Q_\perp=\sum_{t=2}^p(g_t^2-1),\quad Q_{\mathrm{lin}}=-2\upsilon g_1,\quad Q_1=g_1^2-1,
 \label{eqn:raw-necessity-r0}
\end{align}
with \(Q_\perp\) independent of \((Q_{\mathrm{lin}},Q_1)\). Then
\(\upsilon^2=p+O(\sqrt{p\log N})\).

\smallskip
\noindent\emph{Upper tail.} Let \(\lambda\geq0\) and \(\varkappa_p=2(p-1)+4\upsilon^2\). Since
\(Q_1\geq-1\) and \(\lambda-\frac12\log(1+2\lambda)\leq\lambda^2\),
\[
 \log\E\mathrm e^{-\lambda(Q_\perp+Q_{\mathrm{lin}}+Q_1)}
 \leq(p-1)\lambda^2+2\upsilon^2\lambda^2+\lambda=\lambda^2\varkappa_p/2+\lambda .
\]
The exponential Markov inequality at \(\lambda=x\sqrt{6p}/\varkappa_p\) bounds the
left side of \eqref{eqn:raw-necessity-upper-tail} by
\(\exp\{-3px^2/\varkappa_p+x\sqrt{6p}/\varkappa_p\}\). Since \(6p/\varkappa_p=1+O(\sqrt{\log N/p})\),
\(x\sqrt{6p}/\varkappa_p\leq C(1+x^2)/\sqrt p\), and \((\log N)^{3/2}/\sqrt p\to0\), this
gives \eqref{eqn:raw-necessity-upper-tail}.

\smallskip
\noindent\emph{Lower tail.} Put \(u=2\upsilon x\sqrt{6p}/\varkappa_p\) and
\(z=x\sqrt{6p}\sqrt{2(p-1)}/\varkappa_p\), so that \(u^2+z^2=6px^2/\varkappa_p\) and \(u,z\leq
C\sqrt{\log N}\). On the event
\[
 \Bigl\{u\leq g_1\leq u+\frac1{1+u}\Bigr\}\cap
 \Bigl\{Q_\perp\leq-z\sqrt{2(p-1)}-C(1+x^2)\Bigr\},
\]
we have \(Q_{\mathrm{lin}}\leq-2\upsilon u\), \(Q_1\leq C(1+x^2)\), and
\(Q_\perp+Q_{\mathrm{lin}}+Q_1\leq-x\sqrt{6p}\), because
\[
 2\upsilon u+z\sqrt{2(p-1)}=x\sqrt{6p}.
\]
By Lemma~\ref{lem:gaussian-lower-tails}(a), the first event has
probability at least \(c(1+u)^{-1}\mathrm e^{-u^2/2}\). For the second,
Lemma~\ref{lem:gaussian-lower-tails}(b), applied to the \(p-1\) iid
variables \(g_t^2-1\) at the level
\(z'=z+C(1+x^2)/\sqrt{2(p-1)}\leq C\sqrt{\log N}=o(\sqrt p)\), gives
probability at least
\(c(1+z)^{-1}\exp\{-z^2/2-C(1+z^3)/\sqrt p-C(1+x^2)z/\sqrt p\}\). Since
\(6p/\varkappa_p=1+O(\sqrt{\log N/p})\) and \((\log N)^{3/2}/\sqrt p\to0\), the product of
the two bounds gives \eqref{eqn:raw-necessity-lower-tail}.
\end{proof}
\begin{proof}[Proof of Proposition~\ref{prop:raw-necessity}]
Write \(E_i=\|\noise_i\|^2\). Constants may depend on \(d\), \(c_0\),
\(c_\alpha\), the constants of
Assumption~\ref{ass:local-mass-regularity} and of the kernel function, and
\(\omega\). Recall that \(R_{\mathrm g}\leq c_{\mathrm g}^{-1/d}L\leq CL\).

The proof shows that, with probability tending to one, every row \(i\)
outside a small set \(\mathcal Z_{\mathrm{bad}}\) (Step~9) satisfies
\[
 \bigl|r_i([N];k)^2-(E_i+p\sigma^2-s_px_k)\bigr|\leq s_p(\vartheta_N+\eta_N),
 \qquad
 \min_{j\ne i}(D^{(\mathrm x)}_{ij})^2\geq E_i+p\sigma^2-s_px_k-s_p(\vartheta_N+\eta_N),
\]
with \(x_k\) and \(\vartheta_N\) from Step~4 and \(\eta_N\) from Step~3.
The selected radius is therefore about the \(\omega\)-quantile of the
\(E_i\) plus a common constant (Step~10), and every row whose \(E_i\)
exceeds that quantile by \(2s_p(\vartheta_N+\eta_N)\) has no neighbor
(Step~11).

\smallskip\noindent\emph{Step 1: decomposition.}
For \(i\ne j\),
\begin{align}
 (D^{(\mathrm x)}_{ij})^2=E_i+R_{ij},\qquad
 R_{ij}\coloneqq R^0_{ij}+(D^{(\mathrm s)}_{ij})^2+c_{ij},
 \label{eqn:raw-necessity-decomposition}
\end{align}
where \(R^0_{ij}=E_j-2\langle\noise_i,\noise_j\rangle\) and
\(c_{ij}=2\langle\clean_i-\clean_j,\noise_i-\noise_j\rangle\) is the cross
term. Conditionally on \(\noise_i\) and on the clean sample, the variables
\(R^0_{ij}+(D^{(\mathrm s)}_{ij})^2\), \(j\ne i\), are independent.

\smallskip\noindent\emph{Step 2: global events.}
Let \(\mathcal E_{\mathrm{raw}}\) be the event on which
\begin{enumerate}[label=(G\arabic*)]
\item \(|E_i-p\sigma^2|\leq4\sigma^2\sqrt{p\log N}\) for every \(i\);
\item \(|\langle\vu_{ij},\noise_i\rangle|\leq\sigma\sqrt{6\log N}\) for every
ordered pair \(i\ne j\) with \(D^{(\mathrm s)}_{ij}>0\), where
\(\vu_{ij}=(\clean_i-\clean_j)/D^{(\mathrm s)}_{ij}\);
\item \(R^0_{ij}\geq p\sigma^2-3s_p\sqrt{\log N}\) for every pair \(i\ne j\);
\item the clean event \(\mathcal E_N(1)\) of Lemma~\ref{lem:clean-event}
for the family \(\{k\}\) holds.
\end{enumerate}
By Lemma~\ref{lem:laurent-massart} with a union bound over \(i\); since
each \(\langle\vu_{ij},\noise_i\rangle\) is \(\Normal(0,\sigma^2)\) given the
clean sample; by Lemma~\ref{lem:raw-tails} at \(x=3\sqrt{\log N}\) with a
union bound over the \(N^2\) pairs; and by Lemma~\ref{lem:clean-event},
whose conditions hold by \eqref{eqn:window-bounds} since
\(k\in\admset_N(0)\),
\[
 \P\{(\mathrm{G1})^c\}\leq2N^{-2},\qquad
 \P\{(\mathrm{G2})^c\}\leq2N^2\mathrm e^{-3\log N},\qquad
 \P\{(\mathrm{G3})^c\}\to0,\qquad
 \P\{(\mathrm{G4})^c\}\to0
\]
for large \(N\). Hence \(\P(\mathcal E_{\mathrm{raw}})\to1\).

\smallskip\noindent\emph{Step 3: the cross term.}
Put
\[
 \eta_N\coloneqq\frac{4\sqrt6\,\sigma\log N}{\sqrt{s_p}}\longrightarrow0,
\]
the relative size of the cross term, which tends to zero because
\(\log N=o(p^{1/4})\). On \(\mathcal E_{\mathrm{raw}}\),
\(|c_{ij}|\leq4\sqrt6\,\sigma D^{(\mathrm s)}_{ij}\sqrt{\log N}\), so
\[
 |c_{ij}|\leq s_p\eta_N
 \quad\text{if }(D^{(\mathrm s)}_{ij})^2\leq s_p\log N,
 \qquad
 R_{ij}\geq p\sigma^2+\frac{s_p\log N}4
 \quad\text{if }(D^{(\mathrm s)}_{ij})^2>s_p\log N,
\]
for large \(N\). In the second case,
\(D^{(\mathrm s)}_{ij}\geq8\sqrt6\,\sigma\sqrt{\log N}\), so
\(|c_{ij}|\leq(D^{(\mathrm s)}_{ij})^2/2\), and (G3) gives the bound.

\smallskip\noindent\emph{Step 4: thresholds.} Let
\[
 \Upsilon_N=N\Bigl(\frac{s_p}{L^2}\Bigr)^{d/2}=k\Bigl(\frac{s_p}{h_N^2}\Bigr)^{d/2},
\]
the expected number of clean neighbors within the fluctuation scale, up
to constants; the identity is \eqref{eqn:local-knn-scale}. Let \(x_k\),
the level at which a row has about \(k\) raw neighbors in expectation,
solve
\[
 \Upsilon_Nx_k^{-d/2}\mathrm e^{-x_k^2/2}=k;
\]
the left side decreases in \(x_k\), and
\(x_k^2=2\log(\Upsilon_N/k)-d\log x_k\). By
\eqref{eqn:raw-necessity-scales} and
\(s_p\leq R_{\mathrm g}^2/(4\log N)\leq CL^2\),
\[
 kN^{c_0d/2}\leq\Upsilon_N\leq CN,
 \qquad
 c_1\sqrt{\log N}\leq x_k\leq\sqrt{2\log N+C},
 \qquad c_1=(c_0d/2)^{1/2}.
\]
We define
\[
 \vartheta_N=\frac{4(\log k+\log\log N)}{c_1\sqrt{\log N}}\longrightarrow0,
 \qquad x_\pm=x_k\pm\vartheta_N,
\]
so that \(x_k\vartheta_N\geq4(\log k+\log\log N)\); \(\vartheta_N\) is the
half-width of the window around \(x_k\).

\smallskip\noindent\emph{Step 5: neighbor counts.} Let
\(n_i(t)=|\{j\ne i:(D^{(\mathrm s)}_{ij})^2\leq t\}|\). For \(\sqrt t\leq R_{\mathrm g}/2\),
a ball of radius \(\sqrt t\) around \(\clean_i\) meeting the support of a
community lies in a ball of radius \(2\sqrt t\) around a support point,
so \eqref{eqn:uniform-clean-small-ball} gives
\[
 c_\alpha c_{\mathrm g}\Bigl(\frac{\sqrt t}L\Bigr)^d
 \leq\P\{(D^{(\mathrm s)}_{ij})^2\leq t\mid\clean_i\}
 \leq2^dC_{\mathrm g}\Bigl(\frac{\sqrt t}L\Bigr)^d.
\]
By \eqref{eqn:binomial-additive-tail} and \eqref{eqn:binomial-half-mean} of
Lemma~\ref{lem:binomial-tails} and a union bound over \(i\) and over \(t\) in a dyadic grid,
with probability tending to one, for all \(i\) and all
\(0\leq t\leq s_p\log N\),
\begin{align}
 n_i(t)\leq C\{N(t/L^2)^{d/2}+\log N\},
 \qquad
 n_i(t_0)\geq cN(t_0/L^2)^{d/2}\quad\text{for }t_0\coloneqq s_p/x_-,
 \label{eqn:raw-necessity-counts}
\end{align}
where the lower bound uses
\(N(t_0/L^2)^{d/2}\geq c\Upsilon_N(\log N)^{-d/4}\gg\log N\). Let
\(\mathcal E_{\mathrm{cnt}}\) be this event; it depends on the clean
sample only.

\smallskip\noindent\emph{Step 6: order statistics.} For \(x>0\) let
\(M_i(x)=|\{j\ne i:R^0_{ij}+(D^{(\mathrm s)}_{ij})^2\leq p\sigma^2-xs_p\}|\), a sum of conditionally
independent indicators with conditional mean \(\mu_i(x)\).

\smallskip\noindent\emph{Step 7: the upper bound.} Apply
\eqref{eqn:raw-necessity-upper-tail} with
\(x=x_++(D^{(\mathrm s)}_{ij})^2/s_p\). Since \((x+v)^2\geq x^2+2xv\) and
\(\varepsilon_Nx_+^2/\log N\leq C\varepsilon_N\), each term is at most
\(C\mathrm e^{-x_+^2/2-x'(D^{(\mathrm s)}_{ij})^2/s_p}\) with \(x'=(1-\varepsilon_N/\log N)x_+\).
Summing by parts over the pairs with \((D^{(\mathrm s)}_{ij})^2\leq s_p\log N\), bounding the
others by \(N\mathrm e^{-x'\log N}\), and using
\eqref{eqn:raw-necessity-counts},
\[
 \mu_i(x_+)\leq C\mathrm e^{-x_+^2/2}
 \Bigl\{\int_0^{s_p\log N}\frac{x'}{s_p}\mathrm e^{-x't/s_p}n_i(t)\,\mathrm dt
       +N\mathrm e^{-x'\log N}\Bigr\}
 \leq C\mathrm e^{-x_+^2/2}\bigl(\Upsilon_Nx_+^{-d/2}+\log N+N\mathrm e^{-x_+\log N/2}\bigr).
\]
Since \(\mathrm e^{-x_+^2/2}\leq\mathrm e^{-x_k^2/2}=kx_k^{d/2}/\Upsilon_N\leq
CN^{-c_0d/2}(\log N)^{d/4}\), the last two terms vanish, and the first is at
most \(Ck\,\mathrm e^{-x_k\vartheta_N}\). Hence, uniformly in \(i\) on (G1),
\[
 \P\{M_i(x_+)\geq1\mid\noise_i,\clean\}\leq\mu_i(x_+)\leq Ck^{-3}(\log N)^{-4}\to0.
\]

\smallskip\noindent\emph{Step 8: the lower bound.} The \(n_i(t_0)\) indices with \((D^{(\mathrm s)}_{ij})^2\leq s_p/x_-\)
have \(x=x_-+(D^{(\mathrm s)}_{ij})^2/s_p\leq x_-+1/x_-\leq3\sqrt{\log N}\), so
\eqref{eqn:raw-necessity-lower-tail} and \eqref{eqn:raw-necessity-counts}
give
\[
 \mu_i(x_-)\geq c\Upsilon_Nx_-^{-d/2-2}\mathrm e^{-x_-^2/2}
 \geq c\,k\,(\log N)^{-1}\mathrm e^{(x_k^2-x_-^2)/2}
 \geq c\,k\,(\log N)^{-1}\mathrm e^{x_k\vartheta_N-\vartheta_N^2/2}\geq2k .
\]
The lower tail \eqref{eqn:binomial-half-mean} of Lemma~\ref{lem:binomial-tails}
holds, with the same proof,
for every sum of independent indicators, and gives
\[
 \P\{M_i(x_-)<k\mid\noise_i,\clean\}\leq\mathrm e^{-\mu_i(x_-)/8}\leq
 \mathrm e^{-k/4}\to0.
\]

\smallskip\noindent\emph{Step 9: the window.} Let \(\mathcal Z_{\mathrm{bad}}=\{i:M_i(x_+)\geq1\text{ or }M_i(x_-)<k\}\) be
the rows whose order statistics leave the window. By the two previous
steps, \(\P(i\in\mathcal Z_{\mathrm{bad}}\mid\clean)\leq\pi_N\) for every
\(i\) on \(\mathcal E_{\mathrm{cnt}}\), where \(\pi_N\to0\) is a
deterministic uniform bound on the probability that a row is bad, and
hence, by Markov's inequality, with probability tending to one,
\[
 |\mathcal Z_{\mathrm{bad}}|\leq\pi_N^{1/2}N.
\]
On \(\mathcal E_{\mathrm{raw}}\), Step~3 shows that a pair with
\(R^0_{ij}+(D^{(\mathrm s)}_{ij})^2\leq p\sigma^2\) has \((D^{(\mathrm s)}_{ij})^2\leq s_p\log N\) and hence
\(|R_{ij}-R^0_{ij}-(D^{(\mathrm s)}_{ij})^2|\leq s_p\eta_N\). Therefore, for
\(i\notin\mathcal Z_{\mathrm{bad}}\), with \(R_{i,(k)}\) the \(k\)-th smallest of
\(\{R_{ij}\}_{j\ne i}\),
\begin{align}
 \min_{j\ne i}R_{ij}\geq p\sigma^2-s_px_k-s_p(\vartheta_N+\eta_N),
 \qquad
 R_{i,(k)}\leq p\sigma^2-s_px_k+s_p(\vartheta_N+\eta_N) .
 \label{eqn:raw-order-window}
\end{align}

\smallskip\noindent\emph{Step 10: radius.} By
\eqref{eqn:raw-necessity-decomposition}, \(r_i([N];k)^2=E_i+R_{i,(k)}\). On
\(\mathcal E_{\mathrm{raw}}\), every \(R_{ij}\geq p\sigma^2-3s_p\sqrt{\log N}-s_p\eta_N\) and
every \(E_i\geq p\sigma^2-4\sigma^2\sqrt{p\log N}\), so every \(r_i([N];k)^2\geq
2p\sigma^2-Cs_p\sqrt{\log N}\). Hence
\[
 r_{\mathrm x}^2\geq2p\sigma^2-C\sigma^2\sqrt{p\log N},
\]
and the clean
claim of part~(a) is the event \(\mathcal E_{\mathrm{scale}}\) in (G4). Let \(F_E^{-1}(\cdot)\) be the quantile function of \(E_1\), whose
density is bounded by \(C/s_p\). Let \(\overline\varepsilon_N=\pi_N^{1/2}+N^{-1/3}\), the slack in the
quantile level. By Chebyshev's inequality,
\[
 \#\bigl\{i:E_i\leq F_E^{-1}(\omega+2\overline\varepsilon_N)\bigr\}
 \geq(\omega+\overline\varepsilon_N)N,
\]
and, for \(i\notin\mathcal Z_{\mathrm{bad}}\) with \(E_i\leq F_E^{-1}(\omega+2\overline\varepsilon_N)\),
\(r_i([N];k)^2\leq F_E^{-1}(\omega+2\overline\varepsilon_N)+p\sigma^2-s_px_k+s_p(\vartheta_N+\eta_N)\). Since
\(|\mathcal Z_{\mathrm{bad}}|\leq\pi_N^{1/2}N<\overline\varepsilon_NN\),
the selected squared radius is at most
\[
 r_{\mathrm x}^2\leq v_N\coloneqq F_E^{-1}(\omega+2\overline\varepsilon_N)+p\sigma^2-s_px_k+s_p(\vartheta_N+\eta_N)
 \leq2p\sigma^2+C\sigma^2\sqrt p,
\]
where the last bound uses \(\omega+2\overline\varepsilon_N\leq(1+\omega)/2\)
for large \(N\), so that
\(F_E^{-1}(\omega+2\overline\varepsilon_N)\leq p\sigma^2+C\sigma^2\sqrt p\).
This completes part~(a).

\smallskip\noindent\emph{Step 11: isolation.} Let
\[
 \mathcal Z_0=\bigl\{i:E_i>F_E^{-1}(\omega+2\overline\varepsilon_N)+2s_p(\vartheta_N+\eta_N)\bigr\},
 \qquad
 \mathcal Z=\mathcal Z_0\setminus\mathcal Z_{\mathrm{bad}}.
\]
For \(i\in\mathcal Z\) and \(j\ne i\), \eqref{eqn:raw-order-window} and the
definition of \(\mathcal Z_0\) give
\[
 (D^{(\mathrm x)}_{ij})^2\geq E_i+p\sigma^2-s_px_k-s_p(\vartheta_N+\eta_N)>v_N\geq
 r_{\mathrm x}^2,
\]
so \(\gN_i([N];r_{\mathrm x})=\emptyset\). The threshold
defining \(\mathcal Z_0\) is deterministic, and the variables \(E_i\) are
iid and independent of the labels and of the clean sample. Hence the
indicators of \(\mathcal Z_0\) are iid Bernoulli with parameter
\[
 \bar q_N\geq1-\omega-2\overline\varepsilon_N-2C(\vartheta_N+\eta_N)=1-\omega-o(1),
\]
independent of
the clean sample. Binomial concentration (\eqref{eqn:binomial-multiplicative-tail} of
Lemma~\ref{lem:binomial-tails}) within each community, whose
size is at least \(c_\alpha N/4\) on \(\mathcal E_{\mathrm{bal}}\), and
\(|\mathcal Z_{\mathrm{bad}}|=o(N)\) prove part~(b).

\smallskip\noindent\emph{Step 12: neighborhoods.}
On \(\mathcal E_N(1)\), every clean
neighborhood \(\gN_i^{(\mathrm s)}([N];r_{\mathrm s})\) has between
\(c_{\mathrm{deg}}k\) and \(C_{\mathrm{deg}}k+t_1\leq Ck\) elements. Each \(i\in\mathcal Z\) therefore contributes a
Jaccard dissimilarity of one to
\eqref{eqn:average-jaccard}, and
\(T_N(k,\omega)\leq1-|\mathcal Z|/N\leq\omega+o(1)\), which is part~(c).

\smallskip\noindent\emph{Step 13: clustering.} The first stage computes
\(r_1=r_{\mathrm x}\), \(\mK_1=\mK^{(\mathrm x)}\), and
\(\vw_1=\mK^{(\mathrm x)}\vv_1(\mK^{(\mathrm x)})\), so
\([\vw_1]_i=0\) for \(i\in\mathcal Z\). No such \(i\) satisfies
\(|[\vw_1]_i|>\tau_1\geq0\), so with \(B=2\) all of \(\mathcal Z\)
receives label \(2\). Under either permutation in
\eqref{eqn:mismatch-ratio}, the observations of \(\mathcal Z\) in one of the
two communities are misclassified, so part~(b) gives
\(\mismatch_N\geq\min_b|\mathcal Z\cap\gC_b|/N
\geq(1-\omega-o(1))\min_b|\gC_b|/N\), which is part~(d).
\end{proof}

\begin{proof}[Proof of Corollary~\ref{cor:raw-kernel-embedding}]
We keep the notation and the events of the previous proof.

\smallskip
\noindent\emph{Part (a).}
An isolated vertex has a zero row and column in
\(\mK^{(\mathrm x)}=\mK^{(\mathrm x)}([N])\). Let \(\vf\) be the normalized
indicator of \(\mathcal Z\). Then \(\mK^{(\mathrm x)}\vf=\bzero\), and,
with \(\mK^{(\mathrm s)}=\mK^{(\mathrm s)}([N])\) and
\(\mA^{(\mathrm s)}\) its adjacency,
\[
 \|\mK^{(\mathrm x)}-\mK^{(\mathrm s)}\|
 \geq\vf^{\sT}\mK^{(\mathrm s)}\vf
 \geq\frac{c_{\kf}}{N|\mathcal Z|}
   \Bigl(\sum_{i\ne j}\indi{i,j\in\mathcal Z_0}A^{(\mathrm s)}_{ij}
         -2Ck|\mathcal Z_{\mathrm{bad}}|\Bigr).
\]
Conditionally on the clean sample, the double sum has mean
\(\bar q_N^2\sum_{i\ne j}A^{(\mathrm s)}_{ij}\geq c\bar q_N^2Nk\) and variance at
most \(C\sum_iA^{(\mathrm s)}_{i\cdot}{}^2\leq CNk^2\), where
\(A^{(\mathrm s)}_{i\cdot}\) is the degree of \(i\); Chebyshev's
inequality and \(|\mathcal Z_{\mathrm{bad}}|=o(N)\) give
\(\|\mK^{(\mathrm x)}-\mK^{(\mathrm s)}\|\geq ck/N\), which is part~(a).

\smallskip
\noindent\emph{Part (b).}
Work on the event that
\(\min_{j\in \sJ}\lambda_j(\mK^{(\mathrm s)})\geq c_{\sJ}k/N\). The rows of
\(\mZ_{\sJ}(\mK^{(\mathrm x)})=\mK^{(\mathrm x)}\mV_{\sJ}(\mK^{(\mathrm x)})\) indexed
by \(\mathcal Z\) vanish. Let \(\mV_{\sJ}\) and \(\mLambda_{\sJ}\) be the
eigenvectors and eigenvalues of \(\mK^{(\mathrm s)}\) in \(\sJ\), so
\(\mZ_{\sJ}(\mK^{(\mathrm s)})=\mV_{\sJ}\mLambda_{\sJ}\) and
\(\mV_{\sJ}=\mK^{(\mathrm s)}\mV_{\sJ}\mLambda_{\sJ}^{-1}\). Every row satisfies
\(\|(\mV_{\sJ})_{i\cdot}\|\leq\|\mK^{(\mathrm s)}_{i\cdot}\|/(c_{\sJ}k/N)\leq
M_{\kf}\sqrt{Ck}/(c_{\sJ}k)\leq C/\sqrt k\). Conditionally on the clean sample,
\(\sum_i\indi{i\in\mathcal Z_0}\|(\mV_{\sJ})_{i\cdot}\|^2\) has mean
\(\bar q_N|\sJ|\) and variance at most \(\max_i\|(\mV_{\sJ})_{i\cdot}\|^2|\sJ|\leq
C|\sJ|/k\), and
\(\E\{\sum_{i\in\mathcal Z_{\mathrm{bad}}}\|(\mV_{\sJ})_{i\cdot}\|^2\mid\clean\}\leq
\pi_N|\sJ|\). Hence
\(\|(\mV_{\sJ})_{\mathcal Z\cdot}\|_{\mathrm F}^2\geq(1-\omega)|\sJ|-o(1)\), and
for every orthogonal \(\mQ\),
\[
 \bigl\|\mZ_{\sJ}(\mK^{(\mathrm x)})-\mZ_{\sJ}(\mK^{(\mathrm s)})\mQ\bigr\|_{\mathrm F}
 \geq\|(\mV_{\sJ}\mLambda_{\sJ}\mQ)_{\mathcal Z\cdot}\|_{\mathrm F}
 \geq\frac{c_{\sJ}k}N\|(\mV_{\sJ})_{\mathcal Z\cdot}\|_{\mathrm F}.
\]
\end{proof}

\section{Technical lemmas}\label{app:technical-lemmas}
We collect the probability and linear-algebra tools used in the proofs;
standard results are cited, and the others are proved for completeness.

\subsection{Counting bounds}\label{app:concentration-estimates}

\begin{lemma}[Binomial tail bounds]
\label{lem:binomial-tails}
Let \(Z\sim\operatorname{Bin}(n,\beta)\) with mean \(\mu=n\beta>0\). Then:
\begin{enumerate}[label=(\alph*)]
\item \emph{One-sided count bounds.}
\begin{align}
 \P(Z\geq s)&\leq(\mathrm e\mu/s)^s,
 &&s\geq\mu,\ s>0,
 \label{eqn:binomial-upper-tail}\\
 \P(Z\leq\mu/2)&\leq \mathrm e^{-\mu/8}.
 \label{eqn:binomial-half-mean}
\end{align}
\item \emph{Additive upper tail.}
\begin{align}
 \P\{Z>(\mathrm e-1)\mu+t\}&\leq \mathrm e^{-t},
 &&t>0.
 \label{eqn:binomial-additive-tail}
\end{align}
\item \emph{Relative concentration.}
\begin{align}
 \P\{|Z-\mu|>v\mu\}&\leq2\mathrm e^{-\mu v^2/3},
 &&0<v<1.
 \label{eqn:binomial-multiplicative-tail}
\end{align}
\end{enumerate}
\end{lemma}
\begin{proof}[Proof of Lemma~\ref{lem:binomial-tails}]
By independence and \(1+x\leq \mathrm e^x\), for every \(\lambda\in\R\),
\begin{align}
 \E \mathrm e^{\lambda Z}=(1+\beta(\mathrm e^\lambda-1))^n
 \leq\exp\{\mu(\mathrm e^\lambda-1)\}.
 \label{eqn:binomial-mgf}
\end{align}
The case \(\mu=0\) is deterministic, so assume \(\mu>0\).
By the exponential Markov inequality and \eqref{eqn:binomial-mgf}, for
every \(u\in\R\),
\begin{align}
 \P(Z\geq u)&\leq\exp\{\mu(\mathrm e^\lambda-1)-\lambda u\},
 &&\lambda\geq0,\\
 \P(Z\leq u)&\leq\exp\{\mu(\mathrm e^\lambda-1)-\lambda u\},
 &&\lambda\leq0.
\end{align}
For \eqref{eqn:binomial-upper-tail}, take
\(u=s\) and \(\lambda=\log(s/\mu)\geq0\), obtaining
\[
 \P(Z\geq s)\leq \mathrm e^{-\mu}(\mathrm e\mu/s)^s
 \leq(\mathrm e\mu/s)^s.
\]
For \eqref{eqn:binomial-half-mean}, take \(u=\mu/2\) and
\(\lambda=-\log2\), giving
\(\P(Z\leq\mu/2)\leq
\mathrm e^{-\mu(1-\log2)/2}\leq \mathrm e^{-\mu/8}\).
Taking \(u=(\mathrm e-1)\mu+t\) and \(\lambda=1\) proves
\eqref{eqn:binomial-additive-tail}.

For relative deviations, take \(u=(1+v)\mu\) and
\(\lambda=\log(1+v)\) in the upper-tail estimate, and
\(u=(1-v)\mu\) and \(\lambda=\log(1-v)\) in the lower-tail estimate.
The inequalities
\[
 (1+v)\log(1+v)-v\geq v^2/3,\qquad
 v+(1-v)\log(1-v)\geq v^2/2
 \quad(0<v<1)
\]
bound the two tail probabilities by \(\mathrm e^{-\mu v^2/3}\) and
\(\mathrm e^{-\mu v^2/2}\), respectively.
Adding the two tail bounds yields
\[
 \P\{|Z-\mu|>v\mu\}
 \leq \mathrm e^{-\mu v^2/3}+\mathrm e^{-\mu v^2/2}
 \leq2\mathrm e^{-\mu v^2/3},
\]
which proves \eqref{eqn:binomial-multiplicative-tail}.
\end{proof}

\begin{lemma}[Local binomial counts]
\label{lem:local-binomial-counts}
Let \(\mathcal F\) be a sigma-field, let \(G\) be an
\(\mathcal F\)-measurable finite index set with \(n=|G|\), and suppose
that, conditionally on \(\mathcal F\), the signals \(\clean_j\),
\(j\in G\), are iid with a law \(Q\). Let \(\vc\) and \(r>0\) be
\(\mathcal F\)-measurable, and suppose that
\(c_Q(r/L)^d\leq Q(\{\vs:\|\vs-\vc\|\leq r\})\leq C_Q(r/L)^d\). Then
\(Z=\#\{j\in G:\|\clean_j-\vc\|\leq r\}\) is conditionally binomial with
mean \(\mu\in[c_Qn(r/L)^d,C_Qn(r/L)^d]\), and, almost surely, for every
\(s>0\),
\[
 \P(Z\geq s\mid\mathcal F)\leq4^{-s}\quad\text{if }\mu\leq s/(4\mathrm e),
 \qquad
 \P(Z<s\mid\mathcal F)\leq\mathrm e^{-\mu/8}\quad\text{if }\mu\geq2s.
\]
\end{lemma}
Under Assumption~\ref{ass:local-mass-regularity},
\(Q=\P_{\mathrm s}^{(b)}\) of \eqref{eqn:community-clean-law} satisfies the
hypothesis with \(c_Q=c_{\mathrm g}\) and \(C_Q=C_{\mathrm g}\), and so do the
conditional core laws of Theorem~\ref{thm:weak-consistency} at points of
their cores.
\begin{proof}[Proof of Lemma~\ref{lem:local-binomial-counts}]
Conditionally on \(\mathcal F\), \(Z\) is a sum of \(n\) iid indicators
with success probability \(Q(\{\vs:\|\vs-\vc\|\leq r\})\). If \(\mu=0\),
then \(Z=0\) and both bounds hold. Otherwise apply
\eqref{eqn:binomial-upper-tail} and \eqref{eqn:binomial-half-mean}
conditionally: if \(\mu\leq s/(4\mathrm e)\), then
\((\mathrm e\mu/s)^s\leq4^{-s}\); if \(\mu\geq2s\), then
\(\{Z<s\}\subseteq\{Z\leq\mu/2\}\).
\end{proof}

\begin{lemma}[Uniform radial-interval counts]
\label{lem:uniform-radial-interval-count}
Let \(N\geq4\) and let \(n_0\leq N\) be fixed. For each index
\(i\in[n_0]\), let \(\mathcal F_i\) be a sigma-field and let
\(n_i\in\{0,\ldots,N-1\}\) be \(\mathcal F_i\)-measurable. Suppose
that, conditional on \(\mathcal F_i\), the variables
\(W_{i1},\ldots,W_{in_i}\) are iid with an \(\mathcal F_i\)-measurable
law \(\P_i\). For a closed interval \(\sL\subseteq\R\) and \(t>0\), put
\[
 Z_i(\sL)\coloneqq\sum_{j=1}^{n_i}\indi{W_{ij}\in \sL},
 \qquad
 \mathcal E_{\mathrm{rad}}(t)\coloneqq
 \bigcap_{i=1}^{n_0}\ \bigcap_{\sL\subseteq\R \,\text{closed interval}}
 \left\{Z_i(\sL)\leq(\mathrm e-1)n_i\P_i(\sL)+t+1\right\}.
\]
Then \(\P(\mathcal E_{\mathrm{rad}}(t)^c)\leq N(N^2+1)^2\mathrm e^{-t}\).
\end{lemma}
The bound holds simultaneously for all indices and closed intervals, so
it applies to intervals with sample-selected endpoints; this is how it is
used for the shells around selected radii.

\begin{proof}[Proof of Lemma~\ref{lem:uniform-radial-interval-count}]
\noindent\emph{Transform to uniform variables.}
Let \(F_i(x)=\P_i((-\infty,x])\) and
\(F_i(x-)=\P_i((-\infty,x))\). On an enlarged probability space, let
\((V_{ij})\) be mutually independent uniform random variables on
\([0,1]\), independent of the original probability space. Put
\[
 U_{ij}=F_i(W_{ij}-)
       +V_{ij}\{F_i(W_{ij})-F_i(W_{ij}-)\}.
\]
Randomizing across the jumps of \(F_i\) ensures that, conditional on
\(\mathcal F_i\), the \(U_{ij}\) are iid uniform on \([0,1]\), even when
\(\P_i\) has atoms. For every \(\sL=[l,v]\),
\[
 \{j:W_{ij}\in \sL\}\subseteq
 \{j:F_i(l-)\leq U_{ij}\leq F_i(v)\},
 \qquad F_i(v)-F_i(l-)=\P_i(\sL),
\]
simultaneously for every choice of endpoints.

\smallskip
\noindent\emph{Control all grid intervals.}
Consider the grid \(\{0,\delta,2\delta,\ldots,N^2\delta=1\}\) with
\(\delta=N^{-2}\). For a closed grid interval \(G\) and an index \(i\),
conditional on \(\mathcal F_i\), the count of \(U_{ij}\) in \(G\) is
binomial with parameters \(n_i\) and \(|G|\), so
\eqref{eqn:binomial-additive-tail} gives
\(\P(\#\{j:U_{ij}\in G\}>(\mathrm e-1)n_i|G|+t\mid\mathcal F_i)\leq\mathrm e^{-t}\).
There are at most \((N^2+1)^2\) closed grid intervals, so the event
\(\mathcal G_t\) on which this count bound holds for all indices and grid
intervals has \(\P(\mathcal G_t^c)\leq N(N^2+1)^2\mathrm e^{-t}\).

\smallskip
\noindent\emph{Transfer to arbitrary intervals.}
Fix \(i\) and \(\sL=[l,v]\). Round \(F_i(l-)\) down and \(F_i(v)\) up to grid
points, and let \(G\) be the resulting closed grid interval. Then
\(|G|\leq\P_i(\sL)+2N^{-2}\), and on \(\mathcal G_t\)
\[
 Z_i(\sL)\leq(\mathrm e-1)n_i|G|+t
 \leq(\mathrm e-1)n_i\P_i(\sL)+t+2(\mathrm e-1)n_iN^{-2}
 \leq(\mathrm e-1)n_i\P_i(\sL)+t+1,
\]
since \(n_i\leq N-1\) and \(N\geq4\). Thus
\(\mathcal G_t\subseteq\mathcal E_{\mathrm{rad}}(t)\).
The intersection defining \(\mathcal E_{\mathrm{rad}}(t)\) is measurable:
every closed interval is the decreasing limit of closed intervals with
rational (or infinite) endpoints, and its count and \(\P_i\)-mass are the
limits of the corresponding counts and masses. The event depends only on
the original variables and conditional laws, so averaging over the
auxiliary uniforms proves the bound.
\end{proof}

\begin{lemma}[Concentration of community sizes]
\label{lem:class-count-concentration}
For iid categorical labels with the weights and community sets of
Section~\ref{sec:multi-community-manifolds},
\begin{align}
 \P\left(\min_{b\in[B]}|\gC_b|<\left\lfloor\frac{c_\alpha N}{2}\right\rfloor\right)
 &\leq B \mathrm e^{-c_\alpha N/8}.
 \label{eqn:class-count-floored}
\end{align}
\end{lemma}

\begin{proof}[Proof of Lemma~\ref{lem:class-count-concentration}]
For each \(b\), the community size \(|\gC_b|\) is binomial with mean \(N\alpha_b\).
Since \(\lfloor c_\alpha N/2\rfloor\leq N\alpha_b/2\), the half-mean
bound \eqref{eqn:binomial-half-mean} gives
\[
 \P\left(|\gC_b|<\left\lfloor\frac{c_\alpha N}{2}\right\rfloor\right)
 \leq\P\left(|\gC_b|\leq\frac{N\alpha_b}{2}\right)
 \leq \mathrm e^{-N\alpha_b/8}\leq \mathrm e^{-c_\alpha N/8}.
\]
A union bound over \(b\in[B]\) proves \eqref{eqn:class-count-floored}.
\end{proof}

\subsection{Gaussian and sub-Gaussian bounds}\label{app:gaussian-testing-tools}

\begin{lemma}[Weighted chi-square deviations]
\label{lem:laurent-massart}
Let \(g_1,\ldots,g_k\) be iid \(\Normal(0,1)\), let
\(\lambda=(\lambda_1,\ldots,\lambda_k)\) have nonnegative entries, and
set \(Q=\sum_{\ell=1}^k\lambda_\ell(g_\ell^2-1)\). For every \(t>0\),
\[
 \P\bigl\{Q>2\|\lambda\|_2\sqrt t+2\|\lambda\|_\infty t\bigr\}
 \leq\mathrm e^{-t},
 \qquad
 \P\bigl\{Q<-2\|\lambda\|_2\sqrt t\bigr\}\leq\mathrm e^{-t}.
\]
\end{lemma}
\begin{proof}[Proof of Lemma~\ref{lem:laurent-massart}]
This is \cite[Lemma~1]{laurent2000adaptive}.
\end{proof}

\begin{lemma}[Gaussian lower tails]
\label{lem:gaussian-lower-tails}
\begin{enumerate}[label=(\alph*)]
\item If \(g\sim\Normal(0,1)\) and \(u\geq0\), then
\[
 \P\bigl\{u\leq g\leq u+(1+u)^{-1}\bigr\}
 \geq c_{\mathrm N}(1+u)^{-1}\mathrm e^{-u^2/2},
 \qquad c_{\mathrm N}=\mathrm e^{-3/2}/\sqrt{2\pi}.
\]
\item Let \(Y_1,Y_2,\ldots\) be iid with \(\E Y_1=0\),
\(\Var(Y_1)=s^2>0\), and \(\E\mathrm e^{h|Y_1|}<\infty\) for some \(h>0\).
There is a constant \(C\), depending only on the law of \(Y_1\), with the
following property: for every deterministic sequence
\(\varepsilon_n\to0\) there is \(n_0\) such that, for all \(n\geq n_0\)
and \(0\leq z\leq\varepsilon_n\sqrt n\),
\[
 \P\Bigl\{\sum_{t=1}^nY_t\leq-zs\sqrt n\Bigr\}
 \geq\frac{c_{\mathrm N}}{2(1+z)}\exp\bigl\{-z^2/2-Cz^3/\sqrt n\bigr\}.
\]
\end{enumerate}
\end{lemma}
\begin{proof}[Proof of Lemma~\ref{lem:gaussian-lower-tails}]
(a) On the interval,
\(g^2\leq u^2+2u/(1+u)+(1+u)^{-2}\leq u^2+3\), so the density is at least
\(\mathrm e^{-(u^2+3)/2}/\sqrt{2\pi}\); multiply by the length.

(b) Cram\'er's theorem \cite[Ch.~VIII, Thm.~1]{petrov1975sums}, applied to
\(-Y_1,-Y_2,\ldots\), states that for every sequence \(z_n\geq0\) with
\(z_n=o(\sqrt n)\),
\[
 \P\Bigl\{\sum_{t=1}^nY_t\leq-z_ns\sqrt n\Bigr\}
 =\{1-\ncdf(z_n)\}
  \exp\bigl\{(z_n^3/\sqrt n)\lambda(z_n/\sqrt n)\bigr\}
  \{1+o(1)\},
\]
where \(\ncdf\) is the standard normal distribution function and the
Cram\'er series \(\lambda\) is bounded, say by \(C\), near zero.
If the claimed bound failed for infinitely many \(n\), the violating
values \(z_n\leq\varepsilon_n\sqrt n\) would contradict this display along
that subsequence, because \(1-\ncdf(z)\geq c_{\mathrm N}(1+z)^{-1}\mathrm e^{-z^2/2}\) by part~(a).
\end{proof}

\begin{lemma}[Hanson--Wright inequality and linear forms]
\label{lem:hanson-wright}
Let \(\vw\in\R^n\) have independent coordinates with mean zero and
sub-Gaussian norm at most \(C_{\vw}\), and let \(\mM\in\R^{n\times n}\) and
\(\vu\in\R^n\) be fixed. There is an absolute constant \(C\) such that, for
every \(t>0\),
\[
 \P\bigl\{|\vw^{\sT}\mM\vw-\E\vw^{\sT}\mM\vw|
   >CC_{\vw}^2(\|\mM\|_{\frob}\sqrt t+\|\mM\|t)\bigr\}\leq2\mathrm e^{-t},
 \qquad
 \P\bigl\{|\vu^{\sT}\vw|>CC_{\vw}\|\vu\|\sqrt t\bigr\}\leq2\mathrm e^{-t}.
\]
\end{lemma}
\begin{proof}[Proof of Lemma~\ref{lem:hanson-wright}]
By \cite[Theorem~1.1]{rudelson2013hanson}, the first probability at the
level \(s>0\) is at most
\(2\exp[-c\min\{s^2/(C_{\vw}^4\|\mM\|_{\frob}^2),\,s/(C_{\vw}^2\|\mM\|)\}]\),
and the minimum is at least \(t\) at
\(s=C_{\vw}^2(\|\mM\|_{\frob}\sqrt{t/c}+\|\mM\|t/c)\). For the second,
\(\vu^{\sT}\vw\) has sub-Gaussian norm at most \(CC_{\vw}\|\vu\|\) by
\cite[Proposition~2.6.1]{vershynin2018high}, and the tail bound is
\cite[Proposition~2.5.2]{vershynin2018high}. When the coordinates have
unit variance, \(\E w_kw_l=\indi{k=l}\), so the centering term is
\(\E\vw^{\sT}\mM\vw=\Tr\mM\); the lemmas below use the bound in this form.
\end{proof}

\begin{lemma}[Sub-Gaussian quadratic forms]
\label{lem:gaussian-quadratic-form}
Let \(\vw\in\R^n\) have independent coordinates with mean zero, unit
variance, and sub-Gaussian norm at most \(C_{\vw}\), let
\(\mB\in\R^{k\times n}\) be fixed, and put \(\vh=\mB\vw\) and
\(\mV=\mB\mB^{\sT}\). There is a constant \(C\), depending only on
\(C_{\vw}\), such that, for every \(\vu\in\R^k\) and \(t>0\), except with
probability at most \(4\mathrm e^{-t}\),
\[
 \bigl|\|\vu+\vh\|^2-\Tr\mV-\|\vu\|^2\bigr|
 \leq C\bigl(\sqrt{t\|\mV\|}\,\|\vu\|+\|\mV\|_{\frob}\sqrt t+\|\mV\|t\bigr).
\]
\end{lemma}
\begin{proof}[Proof of Lemma~\ref{lem:gaussian-quadratic-form}]
The left side is at most
\(|2\vu^{\sT}\mB\vw|+|\vw^{\sT}\mB^{\sT}\mB\vw-\Tr\mV|\). Since
\(\|\mB^{\sT}\vu\|\leq\|\mV\|^{1/2}\|\vu\|\), the second bound of
Lemma~\ref{lem:hanson-wright} controls the linear term, except with
probability at most \(2\mathrm e^{-t}\). The matrix \(\mB^{\sT}\mB\) has
trace \(\Tr\mV\), Frobenius norm \(\|\mV\|_{\frob}\), and operator norm
\(\|\mV\|\), so the first bound of Lemma~\ref{lem:hanson-wright} controls
the quadratic term, except with probability at most \(2\mathrm e^{-t}\).
\end{proof}

\begin{lemma}[Matrix Chernoff bound]
\label{lem:isotropic-matrix-chernoff}
Let \(\mX_1,\ldots,\mX_n\) be independent symmetric \(m\times m\) random
matrices with \(\bzero\preceq\mX_i\preceq K\mI_m\) and \(\E\mX_i=\mI_m\).
There is an absolute constant \(c>0\) such that
\[
 \P\left\{\tfrac12n\mI_m\npreceq\sum_{i=1}^n\mX_i
          \ \text{ or }\ \sum_{i=1}^n\mX_i\npreceq\tfrac32n\mI_m\right\}
 \leq2m\,\mathrm e^{-cn/K}.
\]
\end{lemma}
\begin{proof}[Proof of Lemma~\ref{lem:isotropic-matrix-chernoff}]
This is the matrix Chernoff inequality \cite[Theorem~1.1]{tropp2012user}
with \(\mu_{\min}=\mu_{\max}=n\) and relative deviation \(1/2\). The lower
and upper tails are at most
\(m\{\mathrm e^{-1/2}(1/2)^{-1/2}\}^{n/K}\) and
\(m\{\mathrm e^{1/2}(3/2)^{-3/2}\}^{n/K}\), and both bases are less
than one.
\end{proof}

\begin{lemma}[Two-point testing]
\label{lem:two-point-testing}
Let \(P\) and \(Q\) be probability measures on a common space. Write
\(\operatorname{KL}(Q\Vert P)\) for the relative entropy,
\(\chi^2(Q,P)=\int(\mathrm dQ/\mathrm dP)^2\,\mathrm dP-1\) when
\(Q\ll P\), and \(\|P-Q\|_{\rm TV}=\sup_E|P(E)-Q(E)|\).
\begin{enumerate}[label=(\alph*)]
\item \emph{Bretagnolle--Huber.} Every test \(\phi\in\{0,1\}\) satisfies
\(P(\phi=1)+Q(\phi=0)\geq\frac12\exp\{-\operatorname{KL}(Q\Vert P)\}\).
\item \emph{Mixtures.} If \(Q=\sum_l w_l Q_l\) is a finite
mixture, then
\[
 \operatorname{KL}(Q\Vert P)\leq\min\Bigl\{\sum_l w_l
 \operatorname{KL}(Q_l\Vert P),\ \log\bigl(1+\chi^2(Q,P)\bigr)\Bigr\}.
\]
\item \emph{Pinsker and maximal coupling.} There are random variables
\(U\sim P\) and \(V\sim Q\) with
\(\P\{U\ne V\}=\|P-Q\|_{\rm TV}\leq\sqrt{\operatorname{KL}(Q\Vert P)/2}\).
In particular, for \(\sigma>0\), the laws \(\Normal(\vs,\sigma^2\mI_m)\) and
\(\Normal(\va,\sigma^2\mI_m)\) can be coupled, measurably in
\((\vs,\va)\), so that they differ with probability at most
\(\|\vs-\va\|/(2\sigma)\).
\end{enumerate}
\end{lemma}
\begin{proof}[Proof of Lemma~\ref{lem:two-point-testing}]
(a) Since \(P(\phi=1)+Q(\phi=0)\geq\int\min\{\mathrm dP,\mathrm dQ\}\),
this is \cite[Lemma~2.6]{tsybakov2009introduction}.
(b) The first bound is convexity of relative entropy in its first
argument. For the second, assume \(Q\ll P\); by Jensen's inequality,
\(\operatorname{KL}(Q\Vert P)=\E_Q\log(\mathrm dQ/\mathrm dP)
\leq\log\E_Q(\mathrm dQ/\mathrm dP)=\log(1+\chi^2(Q,P))\).
(c) The inequality is Pinsker's
\cite[Lemma~2.5]{tsybakov2009introduction}. For the coupling, let
\(f_P,f_Q\) be densities with respect to \(P+Q\),
\(f_\wedge=\min\{f_P,f_Q\}\), and
\(\varepsilon=1-\int f_\wedge=\|P-Q\|_{\rm TV}\). With probability
\(1-\varepsilon\) draw \(U=V\) from \(f_\wedge/(1-\varepsilon)\); otherwise
draw \(U\) from \((f_P-f_\wedge)/\varepsilon\) and \(V\) from
\((f_Q-f_\wedge)/\varepsilon\), which have disjoint supports. For normal
laws, \(\operatorname{KL}(\Normal(\vs,\sigma^2\mI_m)\Vert\Normal(\va,\sigma^2\mI_m))
=\|\vs-\va\|^2/(2\sigma^2)\), so Pinsker's inequality gives the bound
\(\|\vs-\va\|/(2\sigma)\), and the construction depends measurably on
\((\vs,\va)\).
\end{proof}

\subsection{Noise bounds for the denoising proof}\label{app:fold-noise-tools}

\begin{lemma}[Noise on a fold]
\label{lem:fold-noise-bounds}
Let \(\mXi=\mGamma^{1/2}\mXi^{\circ}\in\R^{p\times n}\), where
\(\mGamma\) is positive semidefinite with
\(\|\mGamma\|=\sigma^2\) and \(\mXi^{\circ}\) has independent entries
with mean zero, unit variance, and sub-Gaussian norm at most
\(C_{\mathrm{sg}}\). Let \([\mJ,\mJ_\perp]\in\R^{p\times p}\) be orthogonal
with \(\mJ\in\R^{p\times m}\), and let
\(\mGamma_\perp=(\mI_p-\mJ\mJ^{\sT})\mGamma(\mI_p-\mJ\mJ^{\sT})\).
There is a constant \(C\), depending only on \(C_{\mathrm{sg}}\), such that,
for every \(t>0\), each of the following holds with probability at least
\(1-2\mathrm e^{-t}\):
\begin{align}
 \|\mJ^{\sT}\mXi\|&\leq C\sigma\bigl(\sqrt n+\sqrt m+\sqrt t\bigr),
 \label{eqn:fold-parallel-noise}\\
 \bigl\|(\mJ_\perp^{\sT}\mXi)^{\sT}\mJ_\perp^{\sT}\mXi-(\Tr\mGamma_\perp)\mI_n\bigr\|
 &\leq C\bigl(\|\mGamma_\perp\|_{\frob}\sqrt{n+t}+\sigma^2(n+t)\bigr).
 \label{eqn:fold-perpendicular-wishart}
\end{align}
\end{lemma}
\begin{proof}[Proof of Lemma~\ref{lem:fold-noise-bounds}]
If \(\sigma=0\), then \(\mXi=\bzero\) and both bounds hold. Let
\(\sigma>0\). For a unit vector \(\vv\in\R^n\), the vector
\(\vw=\mXi^{\circ}\vv\) has independent coordinates with mean zero, unit
variance, and sub-Gaussian norm at most \(CC_{\mathrm{sg}}\), by
\cite[Proposition~2.6.1]{vershynin2018high}.

For \eqref{eqn:fold-parallel-noise}, let
\(\mB=\mJ^{\sT}\mGamma^{1/2}\), so that \(\|\mB\|\leq\sigma\),
\(\|\mB\|_{\frob}^2\leq m\sigma^2\), and
\(\|\mB^{\sT}\mB\|_{\frob}\leq\|\mB\|\|\mB\|_{\frob}\leq\sigma^2\sqrt m\).
By Lemma~\ref{lem:hanson-wright} with \(\mM=\mB^{\sT}\mB\), for every
\(s>0\), except with probability at most \(2\mathrm e^{-s}\),
\(\|\mJ^{\sT}\mXi\vv\|^2=\vw^{\sT}\mM\vw\leq
m\sigma^2+C\sigma^2(\sqrt{ms}+s)\leq C\sigma^2(m+s)\). A \(1/2\)-net
\(\mathcal N\) of the unit sphere of \(\R^n\) with \(|\mathcal N|\leq5^n\)
satisfies \(\|\mJ^{\sT}\mXi\|\leq2\max_{\vv\in\mathcal N}\|\mJ^{\sT}\mXi\vv\|\)
\cite[Lemma~4.4.1]{vershynin2018high}. A union bound over \(\mathcal N\)
with \(s=t+n\log5\) gives \eqref{eqn:fold-parallel-noise}.

For \eqref{eqn:fold-perpendicular-wishart}, let
\(\mM=\mGamma^{1/2}\mJ_\perp\mJ_\perp^{\sT}\mGamma^{1/2}\),
so that the matrix on the left is
\(\mA=(\mXi^{\circ})^{\sT}\mM\mXi^{\circ}-(\Tr\mGamma_\perp)\mI_n\). The matrix
\(\mM\) has the same nonzero eigenvalues as
\(\mJ_\perp^{\sT}\mGamma\mJ_\perp\), and hence as
\(\mGamma_\perp\), so \(\Tr\mM=\Tr\mGamma_\perp\),
\(\|\mM\|_{\frob}=\|\mGamma_\perp\|_{\frob}\), and \(\|\mM\|\leq\sigma^2\).
For a unit vector \(\vv\), \(\vv^{\sT}\mA\vv=\vw^{\sT}\mM\vw-\Tr\mM\), and
Lemma~\ref{lem:hanson-wright} bounds it by
\(C(\|\mGamma_\perp\|_{\frob}\sqrt s+\sigma^2s)\), except with probability
at most \(2\mathrm e^{-s}\). A \(1/4\)-net \(\mathcal N'\) of the unit
sphere with \(|\mathcal N'|\leq9^n\) satisfies
\(\|\mA\|\leq2\max_{\vv\in\mathcal N'}|\vv^{\sT}\mA\vv|\)
\cite[Exercise~4.4.3]{vershynin2018high}. A union bound over
\(\mathcal N'\) with \(s=t+n\log9\) gives
\eqref{eqn:fold-perpendicular-wishart}.
\end{proof}

\begin{lemma}[Weighted noise on a fold]
\label{lem:weighted-fold-noise}
Let \(\mXi^{\circ}\in\R^{p\times n}\) have independent entries with mean zero,
unit variance, and sub-Gaussian norm at most \(C_{\mathrm{sg}}\). There is a
constant \(C\), depending only on \(C_{\mathrm{sg}}\), such that, for every
fixed \(\mB\in\R^{k\times p}\), every fixed symmetric
\(\mW\in\R^{p\times p}\), and every \(t>0\), each of the following holds
with probability at least \(1-2\mathrm e^{-t}\):
\begin{align}
 \|\mB\mXi^{\circ}\|&\leq C\bigl(\|\mB\|_{\frob}+\|\mB\|(\sqrt n+\sqrt t)\bigr),
 \label{eqn:weighted-fold-linear}\\
 \bigl\|(\mXi^{\circ})^{\sT}\mW\mXi^{\circ}-(\Tr\mW)\mI_n\bigr\|
 &\leq C\bigl(\|\mW\|_{\frob}\sqrt{n+t}+\|\mW\|(n+t)\bigr).
 \label{eqn:weighted-fold-quadratic}
\end{align}
\end{lemma}
\begin{proof}[Proof of Lemma~\ref{lem:weighted-fold-noise}]
The proof of Lemma~\ref{lem:fold-noise-bounds} applies with
\(\mM=\mB^{\sT}\mB\) for \eqref{eqn:weighted-fold-linear} and with
\(\mM=\mW\) for \eqref{eqn:weighted-fold-quadratic}. For a unit vector
\(\vv\), Lemma~\ref{lem:hanson-wright} gives
\(\|\mB\mXi^{\circ}\vv\|^2\leq\|\mB\|_{\frob}^2+C(\|\mB\|\|\mB\|_{\frob}\sqrt s+\|\mB\|^2s)
\leq C(\|\mB\|_{\frob}^2+\|\mB\|^2s)\) and
\(|\vv^{\sT}((\mXi^{\circ})^{\sT}\mW\mXi^{\circ}-(\Tr\mW)\mI_n)\vv|\leq C(\|\mW\|_{\frob}\sqrt s+\|\mW\|s)\),
each except with probability at most \(2\mathrm e^{-s}\); Lemma~\ref{lem:hanson-wright}
applies to the indefinite matrix \(\mW\) as well. The same nets and union
bounds, with \(s=t+n\log5\) and \(s=t+n\log9\), give the two bounds.
\end{proof}

\begin{lemma}[Square roots of perturbed squares]
\label{lem:square-root-perturbation}
If \(y,\eta_1,\eta_2\geq0\), \(x\in\R\), and \(|x-y^2|\leq \eta_1 y+\eta_2\), then
\(|\sqrt{[x]_+}-y|\leq \eta_1+\sqrt{\eta_2}\).
\end{lemma}
\begin{proof}[Proof of Lemma~\ref{lem:square-root-perturbation}]
If \(x<0\), then \(y^2\leq \eta_1 y+\eta_2\), which implies
\(y\leq \eta_1+\sqrt{\eta_2}\). If \(x\geq0\) and \(u=\sqrt x-y\geq0\), then
\(2yu+u^2\leq \eta_1 y+\eta_2\), which excludes \(u>\eta_1+\sqrt{\eta_2}\). In
the remaining case put \(u=y-\sqrt x>0\). If \(y<\sqrt{\eta_2}\), then
\(u\leq y<\sqrt{\eta_2}\). Otherwise \(y>0\) and
\(uy\leq u(y+\sqrt x)=y^2-x\leq \eta_1 y+\eta_2\), so
\(u\leq \eta_1+\eta_2/y\leq \eta_1+\sqrt{\eta_2}\).
\end{proof}

\subsection{Linear algebra}\label{sec:linear-algebra}

\begin{lemma}[Schur bounds for matrices and integral operators]\label{lem:schur-test}
\begin{enumerate}[label=(\alph*)]
\item \emph{Matrices.} Let \(\mX\in\R^{n\times m}\), and let \(a_j>0\) and
\(b_l>0\) for \(j\in[n]\), \(l\in[m]\). If \(R_1,R_2>0\) satisfy
\begin{align}
    \sum_{j=1}^n|[\mX]_{jl}|a_j&\leq R_1 b_l,
        \quad\text{for }l\in[m],
        &
    \sum_{l=1}^m|[\mX]_{jl}|b_l&\leq R_2 a_j,
        \quad\text{for }j\in[n],
\end{align}
then \(\|\mX\|\leq\sqrt{R_1R_2}\). In particular, with all weights equal
to one,
\begin{align}
    \|\mX\|\leq
    \left(\max_{j\in[n]}\sum_{l=1}^m|[\mX]_{jl}|\right)^{1/2}
    \left(\max_{l\in[m]}\sum_{j=1}^n|[\mX]_{jl}|\right)^{1/2}.
\end{align}
\item \emph{Integral operators.} Let \(L\) be a measurable kernel on a
probability space \((\Omega,\mathfrak m)\) whose absolute row and column
integrals satisfy, for finite \(R_1,R_2\geq0\),
\[
 \mathop{\mathrm{ess\,sup}}_{s}\int |L(s,t)|\,\mathfrak m(\odif t)\leq R_1,
 \qquad
 \mathop{\mathrm{ess\,sup}}_{t}\int |L(s,t)|\,\mathfrak m(\odif s)\leq R_2.
\]
Then the integral operator \(T_L\) on \(L^2(\mathfrak m)\) satisfies
\(\|T_L\|\leq\sqrt{R_1R_2}\).
\end{enumerate}
\end{lemma}
\begin{proof}[Proof of Lemma~\ref{lem:schur-test}]
Part~(a) is the Schur test~\cite{schur1911bemerkungen}.
For part~(b), Cauchy--Schwarz gives
\[
 |T_Lf(s)|^2\leq R_1\int |L(s,t)|\,|f(t)|^2\,\mathfrak m(\odif t).
\]
Integrating and applying Tonelli's theorem gives
\(\|T_Lf\|_{L^2(\mathfrak m)}^2\leq R_1R_2\|f\|_{L^2(\mathfrak m)}^2\).
\end{proof}

\begin{lemma}[Weyl bounds for matrices and compact operators]
\label{lem:weyl}
\begin{enumerate}[label=(\alph*)]
\item \emph{Matrices.} Let \(\mA,\mB\in\R^{n\times n}\) be real symmetric,
with eigenvalues in non-increasing algebraic order. Then, for every
\(j\in[n]\),
\[
 \lambda_j(\mB)+\lambda_n(\mA-\mB)
 \leq\lambda_j(\mA)
 \leq\lambda_j(\mB)+\lambda_1(\mA-\mB),
\]
and in particular
\begin{align}
 \max_{j\in[n]}|\lambda_j(\mA)-\lambda_j(\mB)|
 \leq\|\mA-\mB\|.
 \label{eqn:finite-weyl-bound}
\end{align}
\item \emph{Compact operators.} If \(S,T\) are compact self-adjoint
operators on a common Hilbert space, their non-increasing positive
spectra, padded by zeros, satisfy
\[
 \sup_{j\geq1}|\lambda_j^+(T)-\lambda_j^+(S)|\leq\|T-S\|.
\]
\end{enumerate}
\end{lemma}
Part~(b) applies to a matrix and an operator together: extending a matrix
by zero on an infinite-dimensional orthogonal summand preserves its
zero-padded positive spectrum.
\begin{proof}[Proof of Lemma~\ref{lem:weyl}]
Part~(a) is Weyl's inequality~\cite{Weyl1912DasAV}.
For part~(b), adjoin an infinite-dimensional zero
summand if necessary. The compact self-adjoint min--max principle gives
\[
 \lambda_j^+(T)=\inf_{\dim F=j-1}
   \sup_{\substack{f\perp F\\\|f\|=1}}\langle Tf,f\rangle.
\]
Since \(|\langle(T-S)f,f\rangle|\leq\|T-S\|\) for unit \(f\),
this formula gives
\(\lambda_j^+(T)\leq\lambda_j^+(S)+\|T-S\|\).
Interchanging \(S,T\) proves the claim, including zero-padded indices.
\end{proof}

\begin{lemma}[Interval form of the Davis--Kahan theorem]
\label{lem:interval-davis-kahan}
Let \(\mA,\mB\in\R^{n\times n}\) be real symmetric, with eigenvalues
in non-increasing algebraic order. Fix a nonempty proper consecutive
block \(\sJ=\{a,\ldots,b\}\subset[n]\). Use the block and projector
notation of Appendix~\ref{app:spectral-convergence-proofs}, evaluated at
\(\mA\) and \(\mB\), with orthonormal eigenbases. Define the intervals
\[
 \sL(\mB)=[\lambda_b(\mB),\lambda_a(\mB)],\qquad
 \mathcal S(\mA)=(-\infty,\lambda_{b+1}(\mA)]
              \cup[\lambda_{a-1}(\mA),+\infty).
\]
The first ray is omitted when \(b=n\) and the second when \(a=1\).
Assume \(\delta\coloneqq\operatorname{dist}(\sL(\mB),\mathcal S(\mA))>0\).
Let \(0\leq\vartheta_1\leq\cdots\leq\vartheta_{|\sJ|}\leq\pi/2\) be the
principal angles between the column spaces of \(\mV_{\sJ}(\mA)\) and
\(\mV_{\sJ}(\mB)\), so that \(\cos\vartheta_1\geq\cdots\geq\cos\vartheta_{|\sJ|}\)
are the singular values of \((\mV_{\sJ}(\mA))^{\sT}\mV_{\sJ}(\mB)\), and let
\[
 \Theta\coloneqq\operatorname{diag}(\vartheta_1,\ldots,\vartheta_{|\sJ|}),
 \qquad
 \sin\Theta\coloneqq\operatorname{diag}(\sin\vartheta_1,\ldots,\sin\vartheta_{|\sJ|}).
\]
Then
\begin{align}
 \|\mP_{\sJ}(\mA)-\mP_{\sJ}(\mB)\|=\|\sin\Theta\|
 &\leq\frac{\|\mA-\mB\|}{\delta},
 &
 \quad
 \|\sin\Theta\|_{\frob}
 &\leq\frac{\|\mA-\mB\|_{\frob}}{\delta}.
\end{align}
\end{lemma}
\begin{proof}[Proof of Lemma~\ref{lem:interval-davis-kahan}]
    These are the interval Davis--Kahan bounds of \cite[Theorem~1 and its operator-norm remark]{yu2015useful}, with the projector formulation given by the principal-angle identity.
\end{proof}

\begin{lemma}[Perturbation of symmetric spectral clusters]
\label{lem:simple-symmetric-spectral-perturbation}
Let \(\mA,\mB\in\R^{n\times n}\) be real symmetric matrices and
\(\sJ=\{a,\ldots,b\}\subset[n]\) a nonempty consecutive block. Use
the eigenvector, projector, and gap notation of
Appendix~\ref{app:spectral-convergence-proofs}, evaluated at \(\mA\)
and \(\mB\), with orthonormal eigenbases. Suppose \(\operatorname{gap}_{\sJ}(\mB)>0\) and
\(\|\mA-\mB\|<\operatorname{gap}_{\sJ}(\mB)/2\). Then there is an orthogonal
matrix \(\mQ\in\R^{|\sJ|\times|\sJ|}\) such that
\begin{align}
 \|\mP_{\sJ}(\mA)-\mP_{\sJ}(\mB)\|&\leq\frac{2\|\mA-\mB\|}{\operatorname{gap}_{\sJ}(\mB)},
 &
 \|\mV_{\sJ}(\mA)-\mV_{\sJ}(\mB)\mQ\|&\leq\frac{2\sqrt2\,\|\mA-\mB\|}{\operatorname{gap}_{\sJ}(\mB)}.
\end{align}
\end{lemma}
\begin{proof}[Proof of Lemma~\ref{lem:simple-symmetric-spectral-perturbation}]
\noindent\emph{Cluster separation.}
For \(\sJ\ne[n]\), since \(\|\mA-\mB\|<\operatorname{gap}_{\sJ}(\mB)/2\), by \eqref{eqn:finite-weyl-bound} in Lemma~\ref{lem:weyl}, every complementary eigenvalue of \(\mA\) lies at least \(\operatorname{gap}_{\sJ}(\mB)-\|\mA-\mB\|\) away from \([\lambda_b(\mB),\lambda_a(\mB)]\). It also bounds each existing boundary gap of the \(\mA\)-cluster below by \(\operatorname{gap}_{\sJ}(\mB)-2\|\mA-\mB\|>0\).

\smallskip
\noindent\emph{Projectors and basis alignment.}
For \(\sJ\ne[n]\), the interval Davis--Kahan bound in
Lemma~\ref{lem:interval-davis-kahan} therefore gives
\[
 \|\mP_{\sJ}(\mA)-\mP_{\sJ}(\mB)\|
 \leq\frac{\|\mA-\mB\|}{\operatorname{gap}_{\sJ}(\mB)-\|\mA-\mB\|}
 \leq\frac{2\|\mA-\mB\|}{\operatorname{gap}_{\sJ}(\mB)}.
\]
For \(\sJ=[n]\), both projectors equal \(\mI_n\), so the same bound
holds under the convention \(\operatorname{gap}_{[n]}(\mB)=+\infty\).
Let \(\vartheta_\ell\in[0,\pi/2]\) be the principal angles.
Consider the singular value decomposition
\((\mV_{\sJ}(\mB))^{\sT}\mV_{\sJ}(\mA)
=\mU_1\operatorname{diag}(\cos\vartheta_\ell)\mU_2^{\sT}\), and let \(\mQ=\mU_1\mU_2^{\sT}\). Then
\[
 \|\mV_{\sJ}(\mA)-\mV_{\sJ}(\mB)\mQ\|^2
 =2\max_\ell(1-\cos\vartheta_\ell)
 \leq2\max_\ell\sin^2\vartheta_\ell
 =2\|\mP_{\sJ}(\mA)-\mP_{\sJ}(\mB)\|^2.
\]
This proves the aligned-basis bound with the same \(\mQ\).
\end{proof}

\begin{corollary}[Simple eigenpairs]
\label{cor:singleton-symmetric-spectral-perturbation}
Use the notation of Lemma~\ref{lem:simple-symmetric-spectral-perturbation}
with \(\sJ=\{j\}\), and suppose \(\operatorname{gap}_j(\mB)>0\) and
\(\|\mA-\mB\|<\operatorname{gap}_j(\mB)/2\). Then there is a sign
\(\chi\in\{-1,1\}\) for which both bounds hold:
\begin{align}
 \|\vv_j(\mA)-\chi\vv_j(\mB)\|_2
 &\leq\frac{2\sqrt2\,\|\mA-\mB\|}{\operatorname{gap}_j(\mB)},
 \label{eqn:finite-davis-kahan-bound}\\
 \|\lambda_j(\mA)\vv_j(\mA)-\chi\lambda_j(\mB)\vv_j(\mB)\|_2
 &\leq\|\mA-\mB\|
 \left(1+\frac{2\sqrt2\,|\lambda_j(\mB)|}{\operatorname{gap}_j(\mB)}\right).
 \label{eqn:finite-weighted-eigenvector-bound}
\end{align}
\end{corollary}
\begin{proof}[Proof of Corollary~\ref{cor:singleton-symmetric-spectral-perturbation}]
The lemma with \(\sJ=\{j\}\) gives a sign \(\mQ=\chi\in\{-1,1\}\) and
\eqref{eqn:finite-davis-kahan-bound}. Since the eigenvectors are unit,
\[
 \|\lambda_j(\mA)\vv_j(\mA)-\chi\lambda_j(\mB)\vv_j(\mB)\|_2
 \leq |\lambda_j(\mA)-\lambda_j(\mB)|
       +|\lambda_j(\mB)|\,\|\vv_j(\mA)-\chi\vv_j(\mB)\|_2.
\]
Combining this with \eqref{eqn:finite-weyl-bound} gives
\eqref{eqn:finite-weighted-eigenvector-bound}.
\end{proof}

\end{document}